\documentclass[preprint,12p]{elsarticle}

\usepackage[margin=2.5cm]{geometry}
\usepackage{amsfonts,amsmath,relsize}
\usepackage{subcaption}
\usepackage{graphicx,times} 
\usepackage{color,soul}
\usepackage{booktabs,multirow}
\usepackage{epsfig,epstopdf,float}
\usepackage{longtable,tabularx}
\usepackage{tikz,pgfplots, pifont}
\usepackage[mathcal]{euscript}
\usepackage{color, colortbl}
\usepackage[hidelinks]{hyperref}
\usepackage[linesnumbered,ruled,vlined]{algorithm2e}

\definecolor{hl}{rgb}{0.75,0.75,0.75}
\sethlcolor{hl}
\hypersetup{
	colorlinks=false,
	linkcolor=blue,
	filecolor=magenta,      
	urlcolor=cyan,
}

\begin{document}
\begin{frontmatter}

\title{AREA: Adaptive Reference-set Based Evolutionary Algorithm for Multiobjective Optimisation}

\author[ad1]{Shouyong Jiang\corref{cor1}}
\cortext[cor1]{Corresponding author}
\ead{math4neu@gmail.com}
\author[ad5]{Hongru~Li}
\author[ad2]{Jinglei~Guo}

\author[ad1]{Mingjun~Zhong}
\author[ad3]{Shengxiang Yang}
\author[ad4]{Marcus~Kaiser}
\author[ad4]{Natalio~Krasnogor}

\address[ad1]{School of Computer Science, University of Lincoln, Lincoln LN6 7TS, UK}
\address[ad5]{College of Information Science and Engineering, Northeastern University, Shenyang, 110819, China}
\address[ad2]{Department of Computer Science, Central China Normal University, Wuhan, China}
\address[ad3]{School of Computer Science and Informatics, De Montfort University, Leicester LE1 9BH, UK}
\address[ad4]{School of Computing, Newcastle University, Newcastle upon Tyne NE4 5TG, UK}

\begin{abstract}
Population-based evolutionary algorithms have great potential to handle multiobjective optimisation problems. However, these algorithms depends largely on problem characteristics, and there is a need to improve their performance for a wider range of problems. References, which are often specified by the decision maker's preference in different forms, are a very effective method to improve the performance of algorithms but have not been fully explored in literature. This paper proposes a novel framework for effective use of references to strengthen algorithms. This framework considers references as search targets which can be adjusted based on the information collected during the search. The proposed framework is combined with new strategies, such as reference adaptation and adaptive local mating, to solve different types of problems. The proposed algorithm is compared with state of the arts on a wide range of problems with diverse characteristics. The comparison and extensive sensitivity analysis demonstrate that the proposed algorithm is competitive and robust across different types of problems studied in this paper.
\end{abstract}


\end{frontmatter}

\section{Introduction}
Evolutionary multiobjective optimisation (EMO) is an important approach to multiobjective optimisation problems (MOPs) that frequently arise from both theoretical studies \cite{DAPM02} and real applications \cite{Jiang2018}. Multiobjectivity means that there is unlikely a single solution that is optimal for all the objective functions of an MOP simultaneously, but instead a set of trade-off solutions that compromise between the objectives. Accordingly, the solution set is called the Pareto-optimal set (PS) and its image in objective space is the  Pareto-optimal front (PF). EMO employs a population of individuals and therefore can search for multiple distinct points on the PF in a single run. 

After several decades of advance, many EMO algorithms are available in literature as the No Free Lunch theorem \cite{Wolpert1997} for optimization proves that no single EMO algorithm solves all MOPs. These algorithms use different techniques to organise population and preserve promising solutions for further evolutionary iterations. For example, Pareto-based algorithms (e.g. NSGA-II \cite{DAPM02} and SPEA2 \cite{ZLT02}) and indicator-based algorithms (e.g. IBEA \cite{ZK04}) have distinct environmental selection strategies, either by Pareto dominance or contribution to indicators for population preservation for the next generation \cite{Liu2018}, whereas decomposition-based algorithms (e.g. MOEA/D \cite{ZL07}) additionally structure a population such that solution collaboration is encouraged within in a specified neighbourhood. 

Decomposition-based EMO algorithms are effective because subproblems derived from decomposition are often easier to solve than the original MOP. MOEA/D uses references, which can be weight vectors \cite{ZL07}, reference points \cite{DJ13,Goulart2016}, reference vectors \cite{CJOS-RVEA}, or reference directions \cite{Jiang17_SPEAR}. While references in MOEA/D are used for problem decomposition to create scalar subproblems, they can be used for objective space decomposition to approximate a small PF segment for each subspace such that the collection of segments form a complete PF \cite{Jiang17_SPEAR,LGZ14}.

Despite their effectiveness, the selection of proper references is a difficult task. The use of references raises two open issues. Firstly, requirements on the uniformity of references vary from one algorithm to another. For example, MOEA/D requires a set of well-diversified weight vectors (ideally evenly-spaced) to obtain a good distribution of solutions on the PF. Other algorithms based on objective space decomposition, however, may not necessarily require uniform references. Instead, the reference set that makes all the PF segments easy to approximate is more desirable. Secondly, optimal reference settings depend on problem properties \cite{ISMN16-PF}. Many existing decomposition-based algorithms use points on a unit simplex as references \cite{DJ13,ZL07}. This reference setting works well for problems with a PF similar to the unit simplex, but has been increasingly reported to be ineffective for problems with other types of PFs, such as degeneracy \cite{ISMN16-PF} and disconnectedness \cite{Li2018}.

Much effort has been devoted to the first issue -- reference uniformity. Qi \emph{et al.} \cite{QMLJ14-AWS} investigated the relationship between references and the resulting subproblems after decomposition, and found that evenly-spaced references do not necessarily lead to uniform distribution of subproblems or solutions. The authors introduced an improved reference setting method, called WS-transformation, to alleviate the problem. 
While optimal reference layout may be hard to obtain, other studies turn to optimising decomposition strategies for subproblem definition since decomposition strategies essentially establish a mapping between references and subproblems \cite{Jiang17_SF}. The basic logics behind this idea are that uniform subproblems can be achieved by a proper mapping, even if a sub-optimal reference set was provided. Decomposition approaches, such as the penalty-based boundary intersection (PBI) \cite{ZL07}, inverted PBI \cite{Sato15-IPBI}, adaptive PBI \cite{Yang2017} and enhanced scalarizing functions \cite{Jiang17_SF}, have shown improved solution distribution. Hybrid use of different decomposition approaches has also been explored in a few recent studies \cite{JY16_TPN}.

In parallel with the above, some improvements have been made to overcome/alleviate the issue that optimal reference layout is geometrically PF dependent. A primary improvement strategy is to evolve a reference set and population together -- the reference set is dynamically adjusted to best fit the PF shape represented by the population. In \cite{JD14}, the reference set is adaptively relocated by first inclusion of new reference points and then removal of non-useful ones, based on their niche count values. A similar idea was used in the MOEA/D with adaptive weight adjustment (MOEA/D-AWA) algorithm \cite{QMLJ14-AWS}, with a difference in reversing the above addition-first-removal-second order. Farias {et al.} \cite{Farias2018} modified MOEA/D-AWA by adopting a uniformly random method to update references every a few generations, and they reported that adaptive weight adjustment with the random method yields improved results in most of their test cases. The reference vector guided evolutionary algorithm (RVEA) \cite{CJOS-RVEA} proposed to maintain two reference sets when handling irregular PF shapes. It maintains a uniformly distributed reference set over time as suggested by NSGA-III \cite{DJ13}, and adaptively updates the other reference set by using randomly generated reference vectors to replace existing ones that are not promising. 

The study of \cite{Cai2018} suggested two types of adjustments for references. The first type is to dynamically increase the number of references as the evolution precedes so that a complete PF can be approximated. The second type of adjustment deals with the effectiveness of the references in use, in which the ineffective ones are re-arranged to better fit the shape of irregular PFs. The use of an indicator to guide the adaptation of references has recently been studied in \cite{Tian2018}. Other adaptive approaches include improved reference set adaptation \cite{Li2018} and sampling \cite{Yan2018}. Note that reference utilisation is similar to preference articulation \cite{Rostami2015,Rostami2017}, where preference information from the decision maker is incorporated into algorithms so as to find only preferred regions on the PF. However, the former is intended to find a good representative of the PF (or a set of well-diversified solutions) whereas the latter often expects solutions in a region of interest, which is a small part of the PF preferred by the decision maker. 


Besides the above methods, the use of machine learning models to select reference sets has been seen in recent years. For example, a linear interpolation model was used in an approach specifically for problems with discontinuous PFs \cite{Zhang2018}. Wu \emph{et al.} introduced a Gaussian process regression model to aid reference sampling \cite{Wu2018}. Similarly, an incremental learning model was employed for reference adaptation \cite{Ge2018}. It should be noted that, however, the use of models for reference setting possibly induces additional computational costs, as there is no free lunch for improvements \cite{Wolpert1997}.  

Despite a variety of success in reference-based evolutionary algorithms, the use of references is not fully explored. Many existing studies use references as a tool to form scalar subproblems \cite{ZL07} or estimate solution densities \cite{Jiang17_SPEAR}. These studies place references in the attainable objective space for space decomposition. Positioning a few reference points outside the attainable objective space has also been studied, showing effectiveness in finding solutions in regions of interest \cite{Said2010}. However, little to none has been discussed if doing so with a large set of references for finding a good representative of the whole PF.

In this paper, we explore efficient use of references and develop a new EMO framework, called AREA. Different from most decomposition-based algorithms where references are used for space decomposition, AREA considers a set of reference points as targets for population members to search. The reference set is adaptively adjusted based on information about the PF estimated by the solution set found so far. This way, the population can be robust to various PF geometries. Extensive sensitivity analysis carried out verifies the importance of each component of AREA. Furthermore, AREA is examined by comparing with state of the arts on 35 test problems with various characteristics, demonstrating its effectiveness and robustness to diverse problem types. 

The rest of the paper is organised as follows. Section 2 introduces the motivation of this research and a detailed description of the proposed AREA framework. Section 3 presents sensitivity analysis of key components of the proposed approach, followed by experimental design in Section 4. Section 5 shows empirical results of comparison between the proposed approach and the state of the arts on a variety of problems. Finally, Section 6 concludes the work.

\section{AREA: The Proposed Approach}
In this section, we will explain the motivation of this work and describe the proposed algorithm as a result of this motivation. 
\subsection{Motivation}
It is widely recognised that a proper reference set can guide the search of population with good diversity toward the PF \cite{ZL07,DJ13}. MOEA/D \cite{ZL07} is a good example to demonstrate this: a good reference set not only generates well-diversified subproblems to enhance solution distribution but also easily structures neighbourhood to encourage collaboration leading to fast approximation to the PF. The importance of good reference sets is also seen in NSGA-III variants \cite{JD14,DJ13} where a reference set helps  to improve environmental selection by facilitating niche counts. Generally, a good reference set can be composed of preferences of the decision maker (DM) represented, in most cases, by weight aggregating objective functions. While the DM's preferences are hardly known ahead, it is important that proper preferences are defined progressively such that a good decision is achieved at the end of optimisation \cite{Rostami2015}. 

\begin{figure*}[t]
	\centering
	\centering
	\begin{tabular}{@{}c@{}c}
		\includegraphics[width=0.5\linewidth]{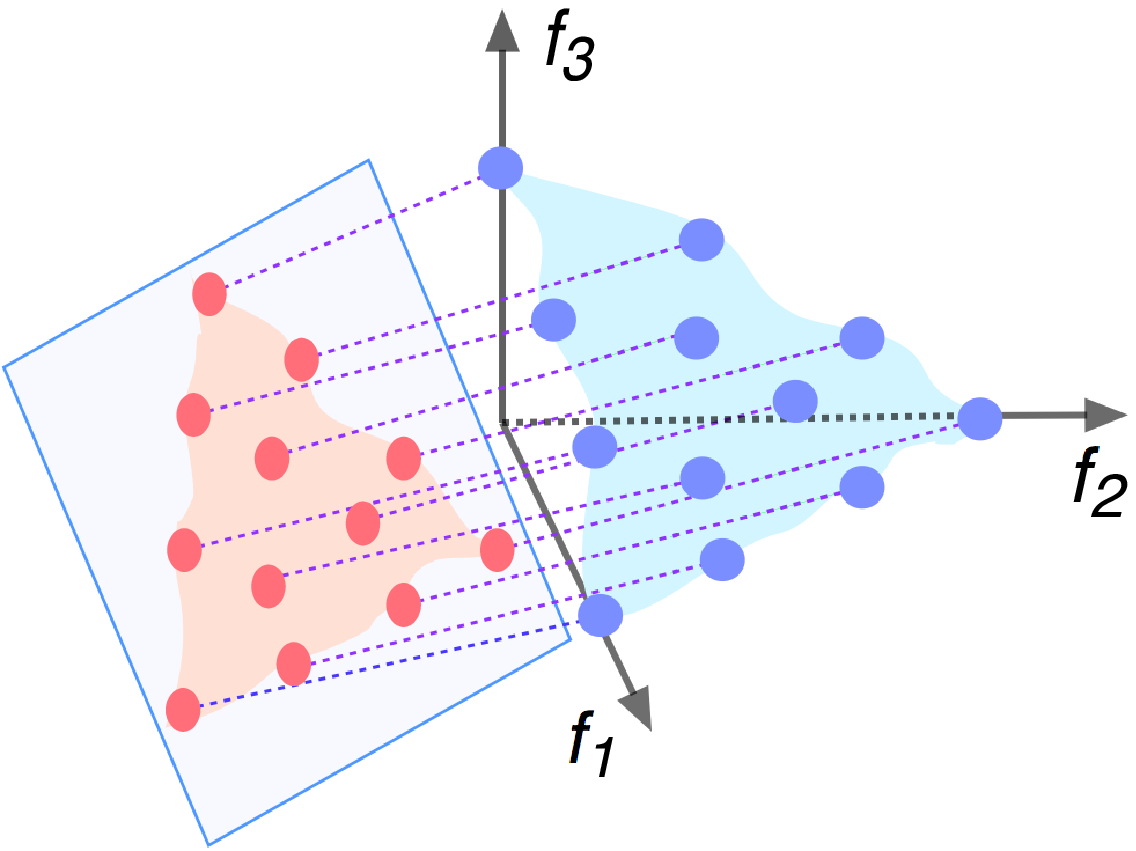}&
		\includegraphics[width=0.5\linewidth]{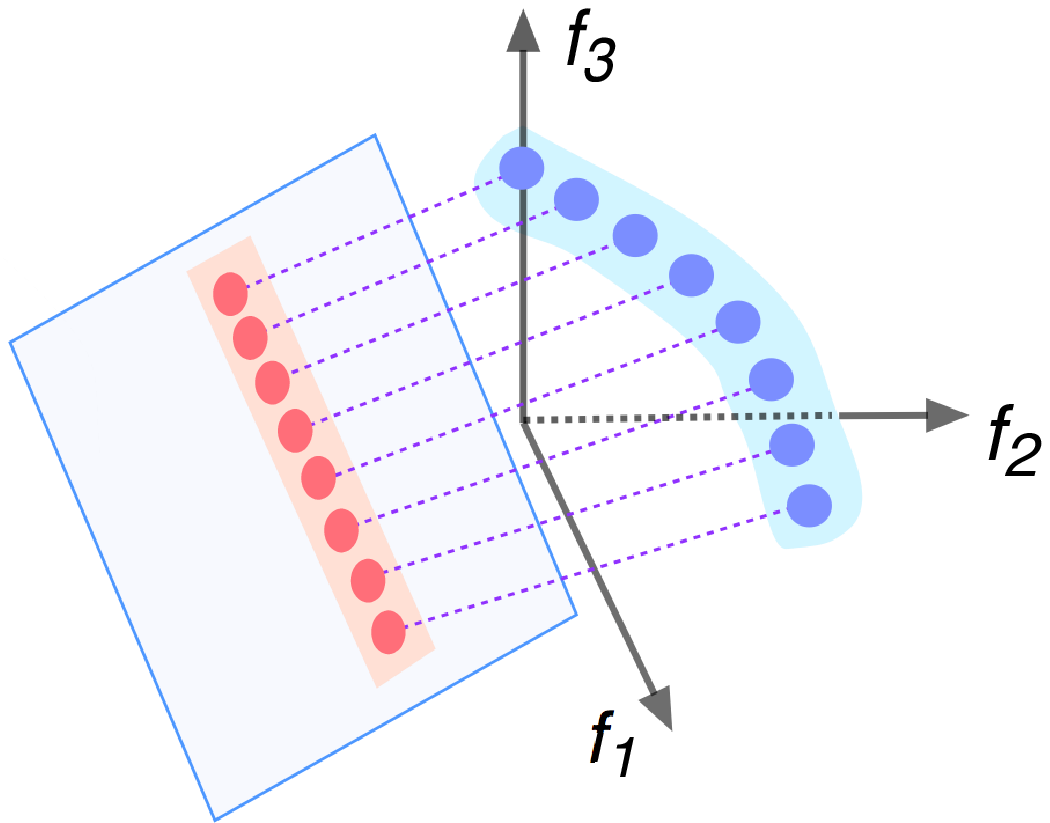}\\
		(a) regular PF & (b) degenerate PF\\[-2mm]
	\end{tabular}	
	\caption{Illustration of EAs guided by reference set for different PF geometries. Orange dots are reference points (targets) and blue ones are solutions on the PF (paleblue) after normalisation, whose projection on a squared plane is in peach.}
	\label{fig:motiv}
	\vspace{-2mm}
\end{figure*}

Many existing studies use reference sets to decompose the first orthant of the objective space in order to form either scalar subproblems \cite{ZL07} or search directions for density estimation \cite{CJOS-RVEA,Jiang17_SPEAR}. We argue that a reference set, if placed properly in the objective space, can be search targets for population. Take Fig.~\ref{fig:motiv}(a) for example, a diversified reference set outside the orthant of the PF can be used as targets for population individuals to approach, and one can expect the final solution set can be as well diversified as the reference set if well converged. In mathematical words, the optimisation task is to minimise the distance between each pair of search targets and individuals.

In irregular cases, e.g., problems with degenerate PF in Fig.~\ref{fig:motiv}(b), a fixed reference set like Fig.~\ref{fig:motiv}(a) is no longer good targets since the reference set is not a good representative of the PF. In this situation, the reference set needs to be properly adjusted to suit the degenerate PF.

The above two examples clearly show (a) reference sets can be used directly as search targets if placed in good position and (b) reference sets need adaptation for different kinds of PFs. The adaptation problem of reference sets has been considered in a few studies \cite{CJOS-RVEA,Li2018}, and we will discuss it in great detail in latter sections. However, how to position reference sets as search targets has not yet been explored in literature. In the following, we focus on (a) and show how to create a good reference set as search targets.  

\begin{figure}[t]
	\centering
	\includegraphics[width=0.5\linewidth]{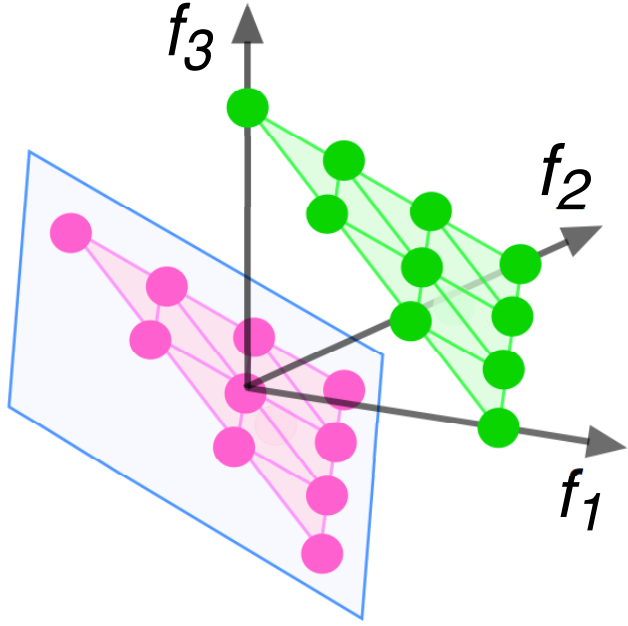}
	\caption{Shifting reference points from a unit simplex to the predefined reference plane (shown in light blue parallelogram). The reference set is in green before shifting and in pink after shifting.}
	\label{fig:shift-ref}
	\vspace{-2mm}
\end{figure}

Considering an $M$-objective problem, $f_i$ ($1\le i \le M$) denotes the $i$-th objective value. We define in the $M$-dimensional objective space a reference (hyper)plane where the reference set is located. Since we consider the reference points as search targets, the reference plane must be outside (or opposite) the attainable objective space. For simplicity, the reference plane is made to pass through the origin, as shown in Fig.~\ref{fig:shift-ref}, and is written as:
\begin{equation}
f_1+f_2+\cdots +f_M=0.
\label{eq:refplane}
\end{equation} 
Ideally, the projection of points sampled uniformly from the PF (after normalisation) on the reference plane is a perfect reference set, as the projection is the minimum euclidean distance from a PF point to the reference plane. However, the PF is often not known in advance, so it is difficult to preset a perfect reference set. However, information about the PF can be gained to help adjust the reference set properly as the search proceeds. This is discussed in following sections in more detail.

Given a reference set, the optimisation task is to minimise the distance between a solution $x$ and a reference point $r$. To reflect proximity of $x$ to $r$, this paper uses the following distance-based measure 
\begin{equation}
D_{<t>}(x,r),
\end{equation}
where $D_{<t>}(x,r)$ measures the distance between $x$ and $r$ using $t$-type distance. Options for $t$ in this paper can be `e' and `c', representing the Euclidean and Chebyshev distance, respectively. The Chebyshev distance is primarily used when discriminating two candidates, as it is found more effective than the Euclidean distance in our preliminary tests. A possible explanation is that an individual $x_1$ with a shorter Euclidean distance to $r$ than another individual $x_2$ does not imply $r$ is more suited to $x_1$ than $x_2$. In contrast, if $x_1$ has a shorter Chebyshev distance than $x_2$ to $r$, then $x_1$ approaches $r$ more directly than $x_2$. In other cases, the Euclidean distance is used.

\subsection{The AREA Framework}
{\bf Algorithm \ref{alg:AREA}} presents the framework of AREA. It consists of four key components: initialisation (lines 1--3), reference set selection (lines 5--9), local mating probability (line 10), and population update (11--17). They are detailed as follows.

\subsubsection{Initialisation}
Initialisation creates a well-distributed reference set $R_0$. This is done by shifting design points of simplex-lattice \cite{DD98} to the predefined reference plane. Let $f_j$ be the $j$-th element of a reference point on an unit simplex, the corresponding element $f'_j$ after shifting is calculated by
\begin{equation}
f'_j=f_j-\frac{1}{M}.
\label{eq:proj}
\end{equation}
This ensures that a unit simplex is shifted so as to pass through the origin. That is, any design point on $\sum\nolimits_{j=1}^{M}f_j=1$ can be shifted to the reference plane defined by Eq.~(\ref{eq:refplane}) since $\sum\nolimits_{j=1}^{M}f'_j=0$. 

Fig.~\ref{fig:shift-ref} shows an example of reference set distribution before and after shifting. After the generation of $R_0$, a neighbourhood structure for each reference point. Meanwhile, an initial population $P$ has to be created and the nondominated set of $P$ fills into an archive $A$. The ideal and anti-ideal points, i.e. $z^l$ and $z^u$, refer to vectors having the best and worst value for each objective, respectively. They are used for objective normalisation when calculating $D_{<t>}(x,r)$ between a solution $x$ and a reference point $r$, although we recognise the identification of these two points are not easy in certain situation \cite{Wang2018}. 

\begin{algorithm}[t]
	\caption{The AREA Framework}
	\label{alg:AREA}
	\KwIn {	stopping criterion, population size ({$N$}), neighborhood size ({$T$}), reference set update frequency ({$f_r$});}
	
	\KwOut {Pareto-optimial approximation $A$;}
	
	Generate initial reference set: $R_0=\{r^1$, $r^2$, $\dots$, $r^N\}$ and then compute for each $r^i$ the $T$ closest neighbours (using Euclidean distance): $B_0(i)=\{i_1, \dots, i_T\}$\;
	
	Generate an initial population $P=\{x^1,\dots,x^N\}$, and store nondominated members in archive $A$ \;
	
	Initialise $ideal$ and $anti\!\!-\!\!ideal$ points, i.e., $z^l$ and $z^u$\;
	
	\While{Stopping criterion not met}{

		\uIf{Reference set update}{
			Update reference set and neighourhood stucture by {\bf Algorithm \ref{alg:refset}};
		}	
		\Else{
			Use reference set $R=R_0$ and neighbourhood structure $B=B_0$;
			
			Match population to $R$ by {\bf Algorithm \ref{alg:fit2ref}};.
		}

		Calculate local mating probability $Prob(i)$ for each individual $x^i$ in $P$ by {\bf Algorithm \ref{alg:mateprob}};
		
		Initialise an empty offspring population $Q$;
		
		\For{ each individual $x^i$ in $P$}{
			
				Select randomly an individual $x^k$ from neighbourhood $B(i)$ with probability $Prob(i)$; Otherwise, select randomly $x^k$ from $P$;
			
			Create offspring $y$ from $x^{i}$ and $x^{k}$ by a genetic operator;
			
			Add $y$ to $Q$ and update $z^l$ with $y$\;
			
			Identify the nearest reference $r^s$ to $y$ in term of Chebyshev distance, and replace individual $x^s$ in $P$ with $y$ if $y$ has a smaller Chebyshev distance to $r^s$ than $x^s$\;
			
		}
		
		Update $z^{u}$ and $A$ using the merged $P \cup Q$\;	
	}	
\end{algorithm}
\setlength{\textfloatsep}{10pt}%
\subsubsection{Reference Set Selection}
Reference set selection occurs with a frequency of $f_r$, i.e. the reference set $R$ is maintained every $f_r$ generations or an equivalent number of function evaluations. The setting of $f_r$ is important and its sensitivity is discussed later in the experiments. The working $R$ alternates between the initial $R_0$ and an updated reference set. The alternation has an advantage over a single mode of reference set: it enables reference set adjustment while stopping the reference set from moving into a local region.

\begin{algorithm}[t]
	\caption{Update reference set}
	\label{alg:refset}
	\KwIn {	population ({$P$}), archive ({$A$}), last updated reference set ($R$), population size ($N$));}
	
	\KwOut {Output $R$, $B$ and $P$;}
	
	Determine the number reference of points to add: $K =\texttt{min }(\sqrt{N}, |A|)$\;
	
	\tcp{Add K reference points}
	
	\While{$|R|-N<K$}{
		Compute the furthest member $\hat{A_i}$ of $A$ to $P$ by the max-min distance \cite{Visalakshi2009}\;
	    Add $\hat{A_i}$ to $P$\;
		Add into $R$ the projection of $\hat{A_i}$ on the reference plane\;
	}
	
	\tcp{Remove K reference points}
	Calculate reference score $\zeta(r_i)$ using Eq.~(\ref{eq:zeta}) for all $r_i \in R$\;
	
	\While{$|R|>N$ \text{and there exists a reference score larger than 0}}{
		\tcp{pick $\hat{i}$ (randomly if a tie exists)}
		
		Identify the reference with the largest score: $\hat{i}=\texttt{argmax}_{1 \le i \le |R|} ~\zeta(r_i)$\; 
		
		Remove the reference point $r_{\hat{i}}$ from $R$ and the solution associating with it from $P$;
		
		Substract one from the reference score for any reference point whose associated solution has shorter Chebyshev distance to $r_{\hat{i}}$ than that reference point;
	}
	\If{$|R|>N$}{
		Remove $|R|-N$ members from $P$, using the $k$-th nearest neighbour \cite{ZLT02}, and their targets from $R$\;
	}	
	Calculate $B$: the neighbourhood structure for $R$\;
\end{algorithm}
If it is the turn to update the reference set $R$, {\bf Algorithm \ref{alg:refset}} is used. This procedure includes adding $K$ promising points to and then removing the same number of unpromising points from the reference set. The points to be added come from archive $A$ as $A$ is the best representative of the PF so far. Thus $K$ should be smaller than $|A|$, which is the size of $A$. In addition, when $K$ reference points are added to $R$, $K$ solutions have to be added to $P$ to guarantee that each reference point is a target for a solution. Associating $K$ reference points with $K$ solutions has a complexity of $O(K^2)$. For computational efficiency, the complexity is restricted to $O(N)$, i.e. $K \propto \sqrt{N}$. Here, $K$ is simply calculated by the line 1 of {\bf Algorithm \ref{alg:refset}}.

During addition (lines 2--5 of {\bf Algorithm \ref{alg:refset}}), each time an archive member $\hat{A_i}$ is selected from $A$ based on the max-min distance in K-means clustering method \cite{Visalakshi2009}. $\hat{A_i}$ is furthest from the population $P$ and its projection on the reference plane can be potentially used as a target for solutions. Thus, $\hat{A_i}$ and its projection (which can be calculated by Eq.~(\ref{eq:proj}) after normalising $\hat{A_i}$ in objective space) are added to $P$ and $R$, respectively. The above addition procedure is repeated $K$ times.

To aid removal (lines 6--12 of {\bf Algorithm \ref{alg:refset}}), we define a score for each reference point $r_i\in R$, which is calculated as:
\begin{equation}
\zeta(r_i)=|\{r_j \in R | {D_{<c>}(x_i, r_j)<D_{<c>}(x_i, r_i)}\}|
\label{eq:zeta}
\end{equation}
where $\zeta(r_i)$ indicates the promise of $r_i$ as a target for its associated solution $x_i$. The larger $\zeta(r_i)$ is, the less likely $r_i$ renders a good target. The Chebyshev distance is used in Eq.~(\ref{eq:zeta}) as it was found generally more effective than the Euclidean distance in our tests.
Each time the worst target $r_{\hat{i}}$ (with the largest score) is identified ($r_{\hat{i}}$ is randomly chosen among the worst if there is a tie) and then removed from $R$, and the solution $x_{\hat{i}}$ for this target is also removed from $P$. This is repeated until the size of $R$ is reduced to $N$ or no more target can have a positive score. If the removal process terminates before $|R|$ reaches $N$, which happens when all the reference points are the best targets for their associated solutions, an additional operation is performed. The operation employs the $k$-th nearest neighbour method \cite{ZLT02} to truncate $P$ to the size of $N$, and the targets for those truncated solutions are removed from $R$. 
After reference set update, the neighbourhood for each $r_i \in R$ has to be recalculated. This assures that the new $R$ and its neighbourhood structure are consistent. 

\begin{figure*}[t]
	\centering
	\begin{tabular}[c]{cccc}
		\begin{subfigure}[t]{0.22\textwidth}
			\centering
			\includegraphics[width=\linewidth]{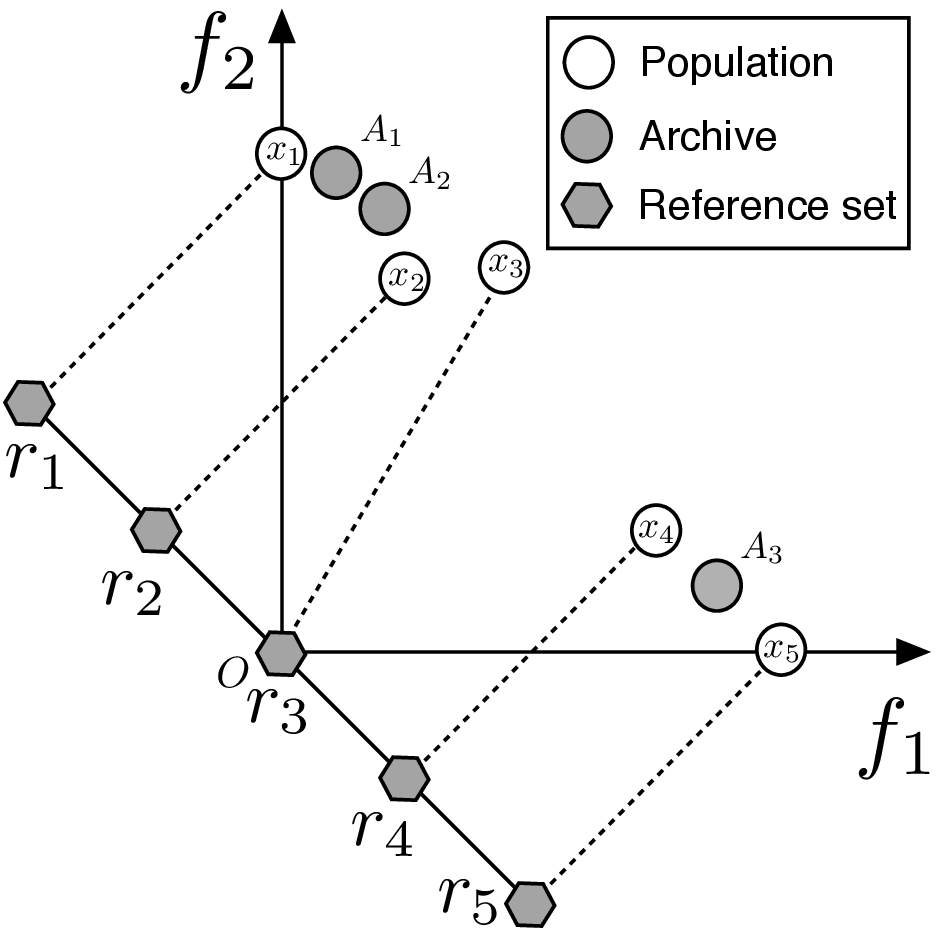}
			\caption{There are five reference points and associated population members and three archived members different from the population.}
		\end{subfigure}&	 
		\begin{subfigure}[t]{0.22\textwidth}
			\centering
			\includegraphics[width=\linewidth]{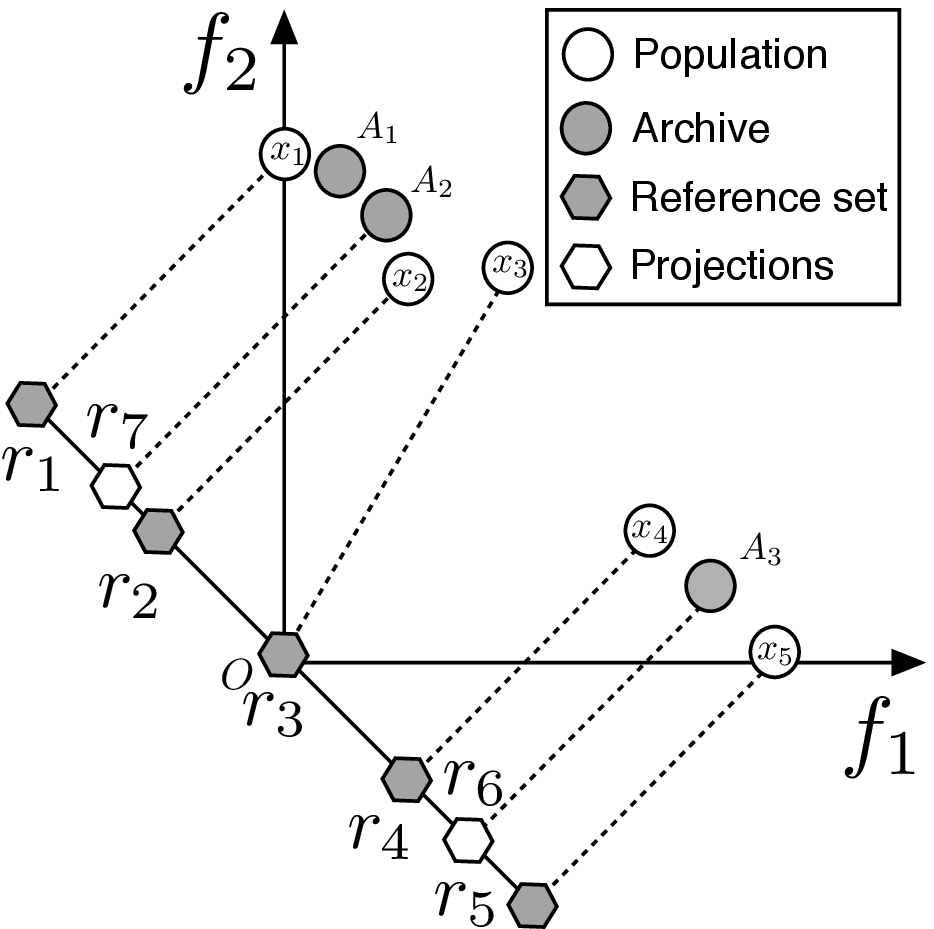}
			\caption{The projections of the top two furthest (to the population) archived members (first $A_3$ and then $A_2$) are added to the reference set.}
		\end{subfigure}
		&	 
		\begin{subfigure}[t]{0.22\textwidth}
			\centering
			\includegraphics[width=\linewidth]{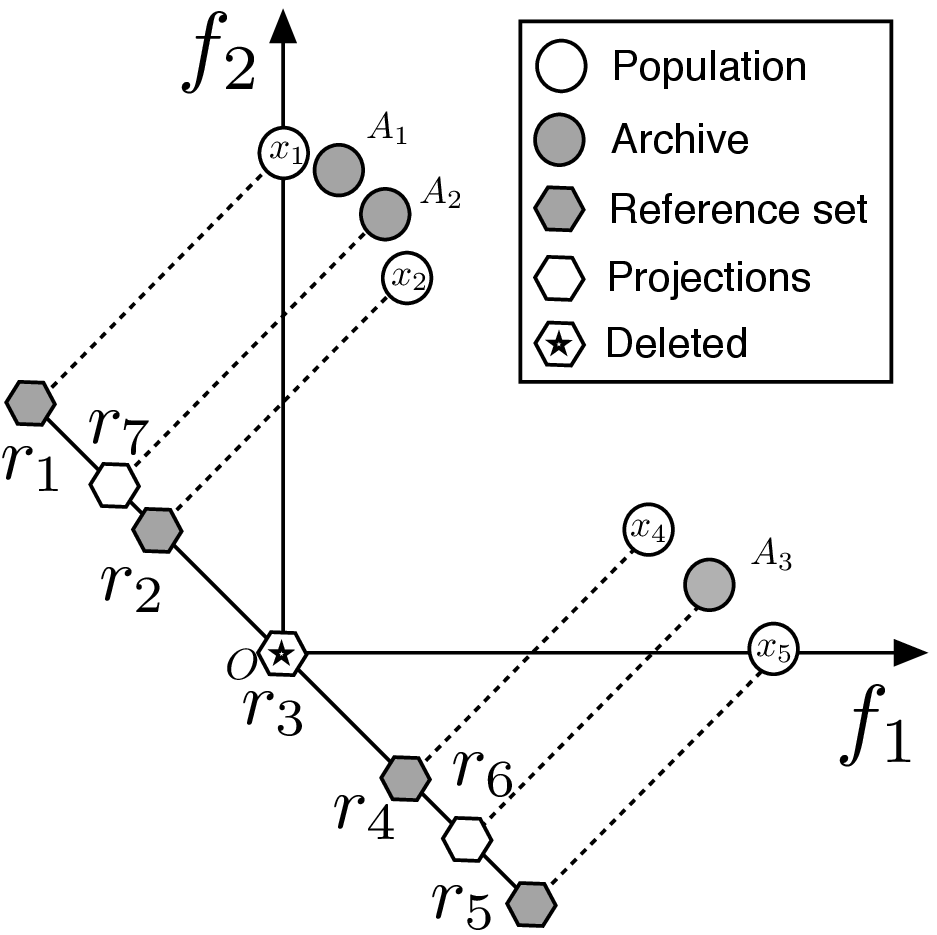}
			\caption{The $\zeta$ score for all the references are calculated, and the reference point $r_3$ is first removed as it has the largest score ($\zeta=1$).}
		\end{subfigure}
		&	
		\begin{subfigure}[t]{0.22\textwidth}
			\centering
			\includegraphics[width=\linewidth]{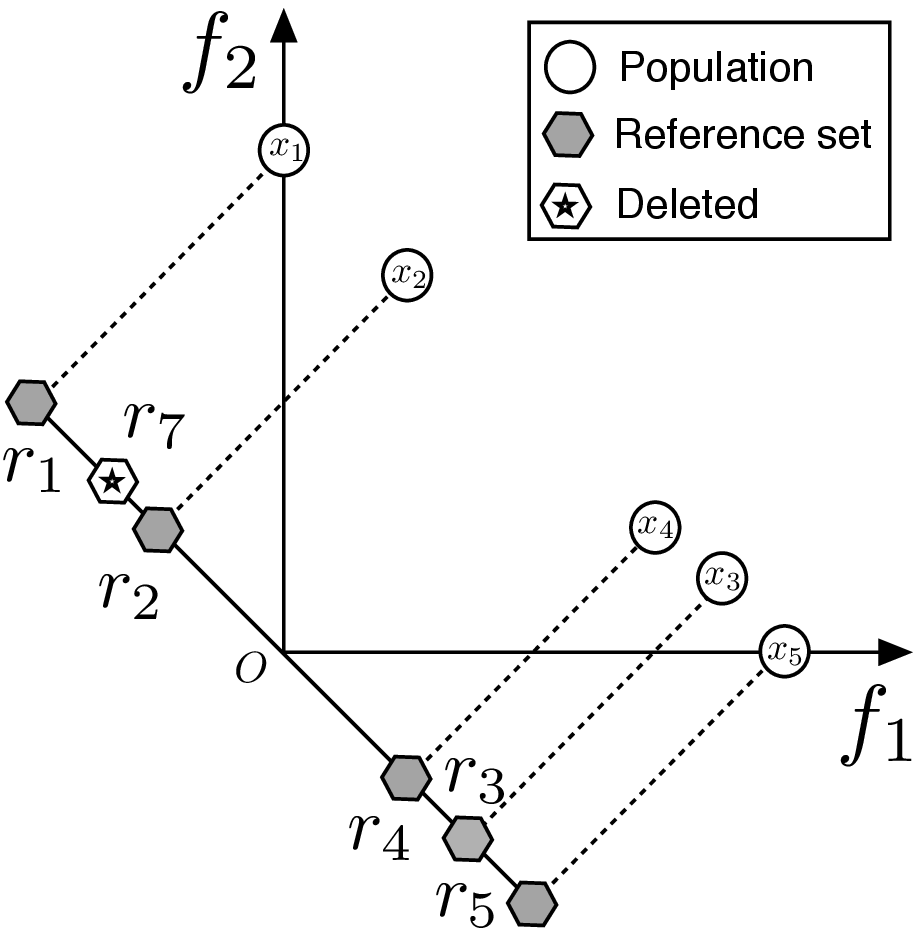}
			\caption{The reference point $r_7$ is removed as a
				result of the $K$-th nearest neighbour method \cite{ZLT02}. The remaining constitutes a new reference set and population.}
		\end{subfigure}
	\end{tabular}
	\caption{Illustration of reference set and population update. Dashed lines show the association relationship between candidates and reference points.}
	\label{fig:refupdate}
\end{figure*}

Fig.~\ref{fig:refupdate} presents an illustrative example of the reference set update procedure of {\bf Algorithm \ref{alg:refset}} (in objective space). As shown in Fig.~\ref{fig:refupdate}(a), the reference set has to be adjusted as it does not render a good population distribution. Suppose $K=2$ reference points are allowed to adjust, two archive members are needed for the adjustment. In Fig.~\ref{fig:refupdate}(b), $A_3$ is first chosen, followed by $A_2$, since they have the maximal Euclidean distance to their nearest population members. The projections of the chosen (e.g., $r_6$ and $r_7$) on the reference plane are added to the current reference set. Then, each reference point is assigned a $\zeta$ score by Eq.~(\ref{eq:zeta}). This leads to $\zeta=0$ for all the reference points except $r_3$, for which $\zeta=1$. So, $r_3$ is removed from the reference set, as shown in Fig.~\ref{fig:refupdate}(c). One more reference point is needed for removal. In Fig.~\ref{fig:refupdate}(d), the k-th nearest neighbour method \cite{ZLT02} identifies $r_7$ as the one having the highest density information. Therefore $r_7$ is removed, and the remaining reference points and their corresponding solutions are used in the next generation. 

If it is the turn to use the initial reference set $R_0$ for $R$, then a match between $P$ and $R$ should be computed such that each solution has a correct target. The procedure in {\bf Algorithm \ref{alg:fit2ref}} is developed to fulfil the task. The population $P$ and archive $A$ are merged to form $S$, and $P$ is emptied at this point. Then a matrix $E$ is created with its element $E_{ij}$ storing the euclidean distance between $r_i \in R$ and $s_j \in S$. To refill $P$ to the size of $N$, each time the nearest reference point to each $s_j \in S$ is calculated and nearest distance is recorded. A solution associated with each nearest reference point is in turn added to $P$, and the corresponding row and column in $E$ are to infinity. In case that some solutions share the same nearest reference point, the one with the shortest distance is added to $P$. The operation is repeated until $P$ has exactly $N$ members. 
\begin{algorithm}[t]
	\caption{Match population to reference set}
	\label{alg:fit2ref}
	\KwIn {	population ({$P$}), archive ({$A$}), reference set ($R$), population size ($N$);}
	
	\KwOut {Updated population;}
	
	Merge $P$ and $A$ to form $S$, then empty $P$;
	
	Compute a distance matrix $E$ between pairs of $R$ and $S$;
	
	\While{$|P|<N$}{
		Identify the nearest reference point for each member of $S$, and the save them in a set $I$;
	
		\For{each reference point $i \in I$}{
			Select the nearest individual $s_{\hat{k}}$ in $S$ to $i$, and associate it with $i$\;
			
			Save $s_{\hat{k}}$ to $P$\;
			
			Set to infinity all the elements of the $i$-th row and $\hat{k}$-th column of the distance matrix $E$;
		}
	}
\end{algorithm}

\begin{figure*}[t]
	\centering
	\begin{tabular}[c]{cccc}
		\begin{subfigure}[t]{0.22\textwidth}
			\centering
			\includegraphics[width=\linewidth]{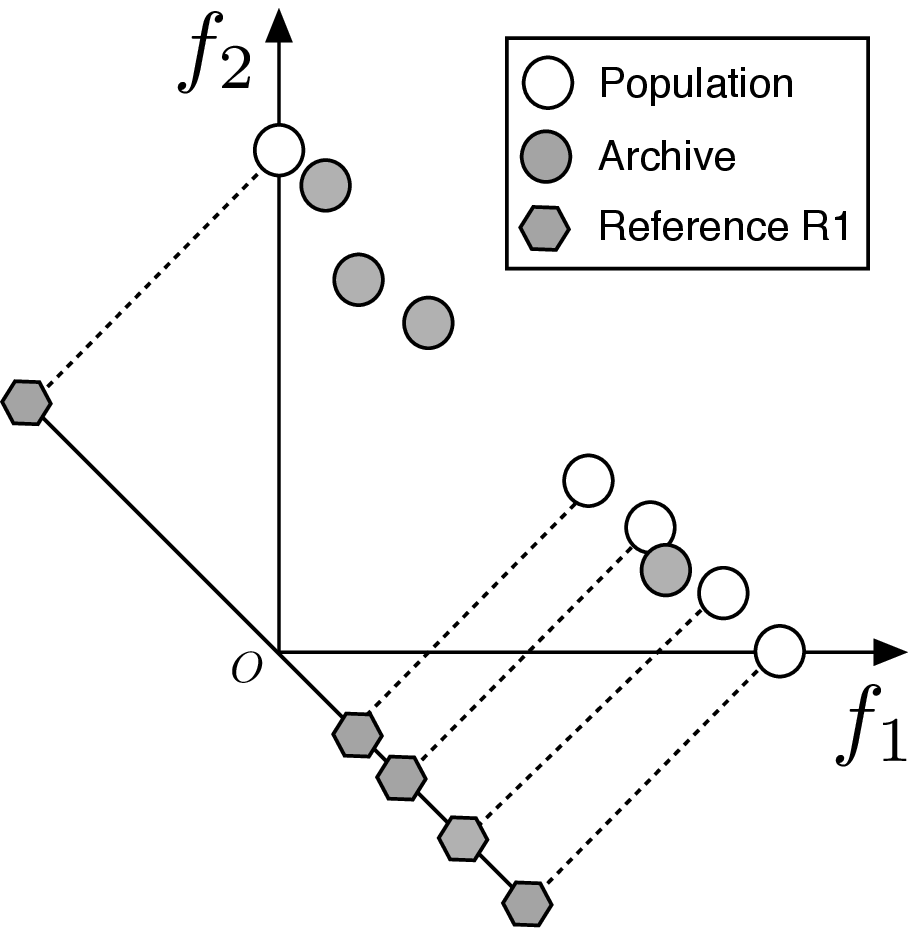}
			\caption{Distribution of solutions from population and archive for reference set $R_1$.}
		\end{subfigure}&	 
		\begin{subfigure}[t]{0.22\textwidth}
			\centering
			\includegraphics[width=\linewidth]{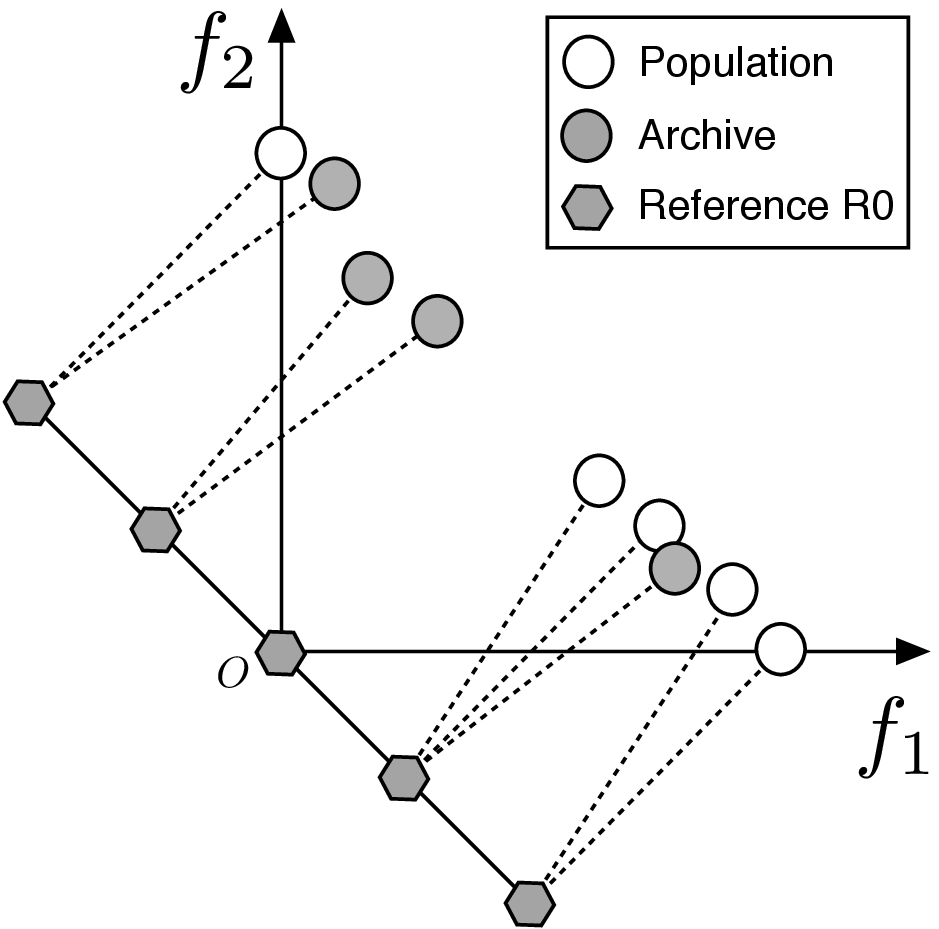}
			\caption{Each solution is associated with the closest member of reference set $R_0$.}
		\end{subfigure}
		&	 
		\begin{subfigure}[t]{0.22\textwidth}
			\centering
			\includegraphics[width=\linewidth]{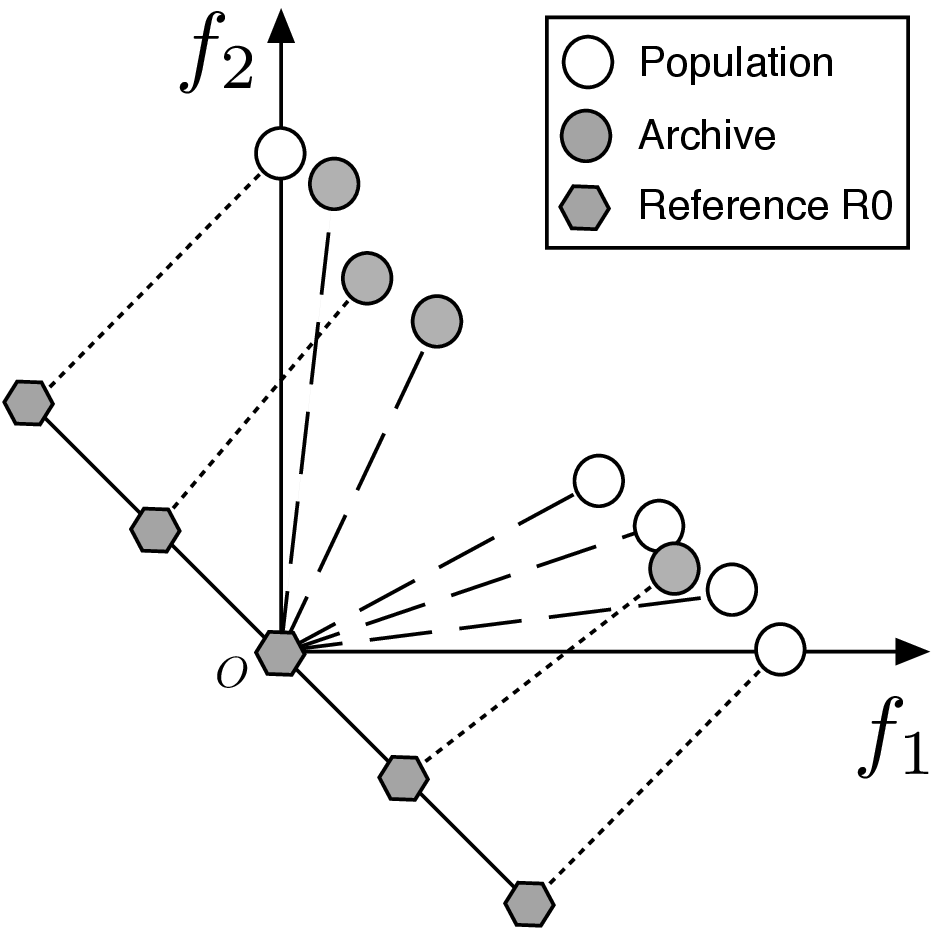}
			\caption{Reference points select the closest solutions and solution association is updated.}
		\end{subfigure}
		&	
		\begin{subfigure}[t]{0.22\textwidth}
			\centering
			\includegraphics[width=\linewidth]{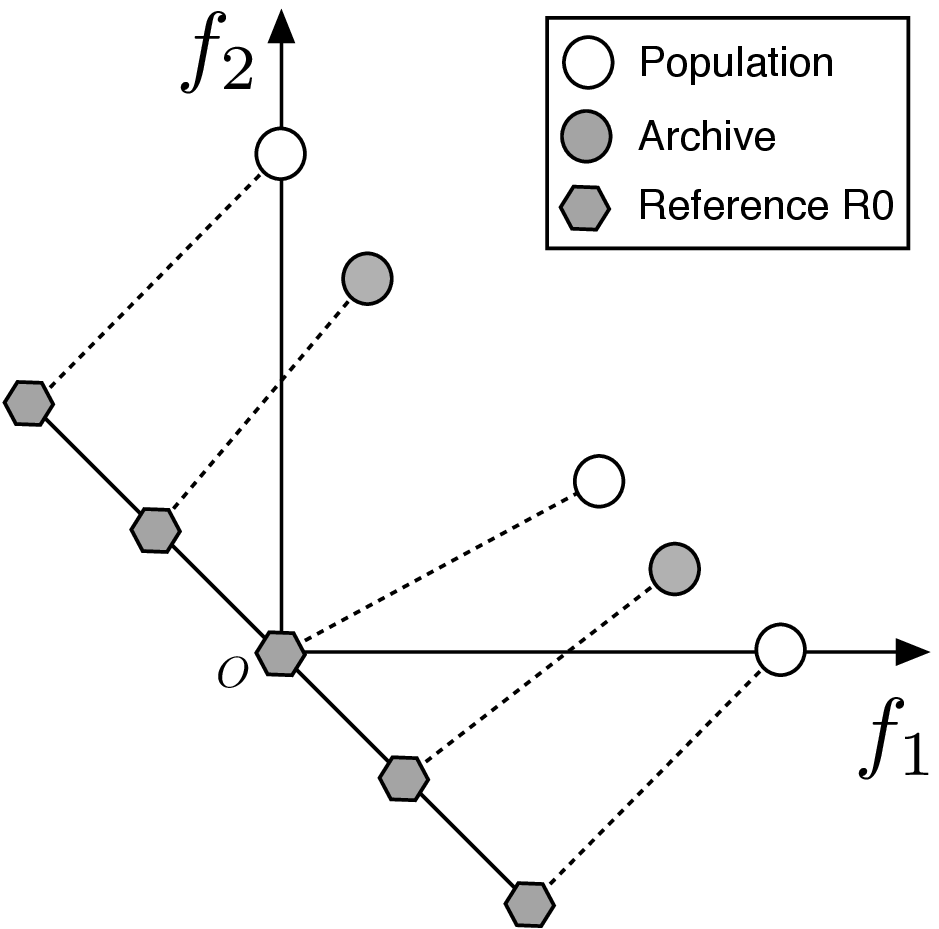}
			\caption{New population is obtained after selecting the closest solution for the reference near $O$.}
		\end{subfigure}
	\end{tabular}
	\caption{Illustration of updating solution association when $R_0$ is used. Dashed lines show the association relationship between candidates and reference points.}
	\label{fig:sol2ref}
\end{figure*}

An illustrative example is provided in Fig.~\ref{fig:sol2ref}. Suppose Fig.~\ref{fig:sol2ref}(a) shows a population and archive distribution resulting from the use of $R_1$, the association relationship between the population and $R_1$ is obviously unsuitable for uniformly-distributed $R_0$. To update solution association, first, every candidate from both the population and archive identifies the nearest reference point in $R_0$, as shown in Fig.~\ref{fig:sol2ref}(b). Then, in Fig.~\ref{fig:sol2ref}(c), for each reference point that is the closest for at least one candidate, the candidate with the shortest Euclidean distance is associated with that reference point. The reference points and candidates that have not been yet assigned association between each other repeat the procedure of Fig.~\ref{fig:sol2ref}(b) until all the reference points in $R_0$ are associated with a candidate, as shown in Fig.~\ref{fig:sol2ref}(d). 

\begin{figure}[t]
	\centering
	\includegraphics[width=0.7\linewidth,height=0.5\linewidth]{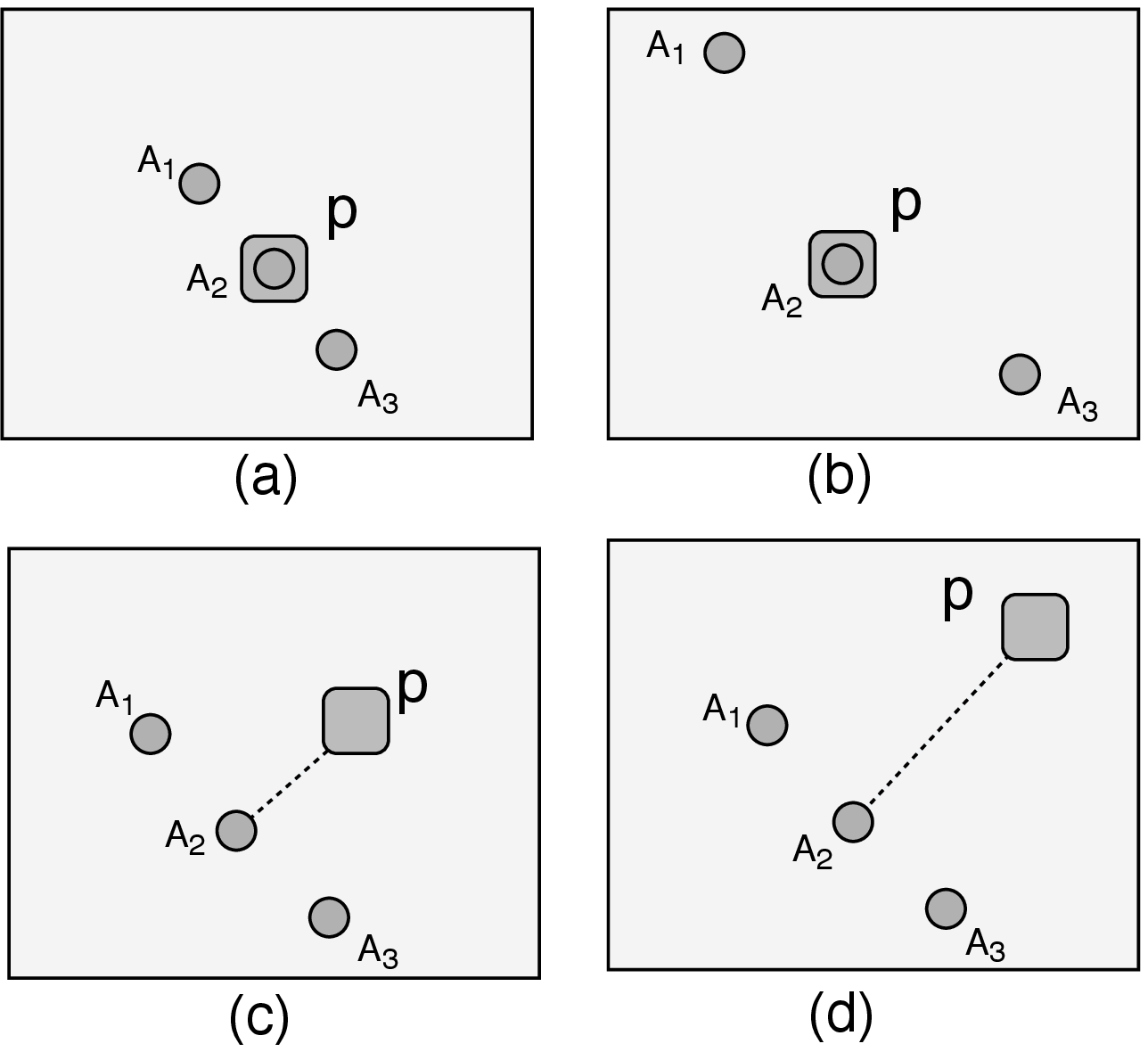}\\[-2mm]	
	\caption{Illustration of the relationship between a solution $p$ (squared) and 
		its neighbours $A_1$--$A_3$ (circled) in the archive. Dashed lines is the 
		distance between $p$ and its closest neighbour. (a) $p$ is identical to $A_2$ 
		and has two close neighbours; (b) $p$ is identical to $A_2$ and has two far neighbours;
		(c) $p$ is close to its nearest neighbour $A_2$; (d) $p$ is far from its nearest neighbour $A_2$.}
	\label{fig:sol_arch}
	\vspace{-2mm}
\end{figure}

\subsubsection{Local Mating Probability}
When a population member $p$ is considered a parent for offspring production, it is important to select carefully its mating partner $q$, either locally or globally, from the population. Here, we calculate the probability of selecting a local partner for $p$ based on closeness to archive $A$, and the neighbours of the nearest in $A$ ($M$ neighbours is considered for $M$-objective problems).  Fig.~\ref{fig:sol_arch} illustrates four possible scenarios of the distribution in objective space. When $p$ is a member of $A$, i.e. $p$ is the same as its nearest neighbour $A_2$, $p$ in Fig.~\ref{fig:sol_arch}(a) is expected to mate with a global partner for global exploration since it has very close neighbours, whereas $p$ in Fig.~\ref{fig:sol_arch}(b) is encouraged to have a local partner for local exploitation to fill the gap between $A_2$ and $A_1$ or $A_3$. When $p$ is different from its nearest neighbour $A_2$,  $p$ in Fig.~\ref{fig:sol_arch}(d) should have a higher probability of selecting a local partner than in Fig.~\ref{fig:sol_arch}(c) for the sake of convergence. For these reasons, we put forward the local mating probability for $p$ as follows: (1) calculating for $p$ the nearest member $\bar{p}$ in $A$ and recording the euclidean distance $d_{p,1}$ between $p$ and $\bar{p}$; (2) calculating $M$ smallest distances of $\bar{p}$ to other members of $A$ and multiplying them to make $d_{p,2}$; (3) summing up $d_{p,1}$ and $d_{p,2}$ to get $d_p$. Then, the local mating probability for $p$ is
\begin{equation}
Prob(p)=\frac{d_p}{max_{v \in P}(d_v)}.
\label{eq:prob}
\end{equation}

\begin{algorithm}[t]
	\caption{Local mating probability}
	\label{alg:mateprob}
	\KwIn {	population ({$P$}), archive ({$A$}), number of objectives ($M$);}
	
	\KwOut {Local mating probability $Prob$;}
	
	\For{each member$i \in P$}{
		Identify the nearest member $A_j$ of $A$ to $i$ and the corresponding nearest distance $d_{i,1}$\;
		Compute the $M$ nearest neighbours around $A_j$ and the corresponding nearest distances: $\bar{d}_1, \dots, \bar{d}_M$ \; 
		Multiply $\bar{d}$ values: $d_{d,2}=\prod\nolimits_{j=1}^{M}{\bar{d}_j}$\;
		Compute $d_i=d_{i, 1}+d_{i, 2}$\; 
	}
	
	Compute $Prob(i)$ for all $x_i \in P$ according to Eq.(\ref{eq:prob})\;	
\end{algorithm}
\noindent It may happen that the minimum $d$ value is very small compared with the maximum value, making $Prob(p)$ close to zero. In practice, a small value (e.g., 0.2) is added to $Prob(p)$ (the probability is capped at one) to enable $p$ a chance of selecting a local mating partner. {\bf Algorithm \ref{alg:mateprob}} presents the procedure for calculating the local mating probability for each member of the population.

\subsubsection{Population Update}
As described in lines 13--16 of {\bf Algorithm \ref{alg:AREA}}, each population member selects a mating partner based the local mating probability mentioned above. Then, the two members create a new solution $y$ using a genetic operator. $y$ updates the ideal point $z^l$ if it is better than $z^l$ in at least one objective, and $y$ is also saved in an offspring population $Q$. The best target $r^{\hat{j}}$ for $y$ is identified by minimising the Chebyshev distance (better than the Euclidean distance in our testing) to all the reference points. Population members whose targets are in the neighbourhood of $r^{\hat{j}}$ undergo update if $y$ has is Chebyshev-closer to $r^{\hat{j}}$ than them. At the end of each generation, the anti-ideal point $z^u$ and nondominated archive $A$ are recalculated based on $P \cup Q$. Note that, if $A$ is over a predefined size, then it is truncated by the $k$-th nearest neighbour method \cite{ZLT02}.

\section{Sensitivity Analysis}
Sensitivity analysis helps to study the robustness of algorithms. This section discusses the impact of different components and parameters on the performance of AREA. The test problems used in this section are from DTLZ1-7 \cite{DTLZ05} and F1-F4 \cite{Yang2017}. The inverted generational distance (IGD) \cite{Jiang17_SPEAR} is the main performance measurement. These will be detailed, together with parameter settings, in Section 4.
\subsection{Influence of Reference Set}
The proposed AREA alternates between a predefined reference set $R_0$ and an evolving one (denoted $R_1$). Here we investigate the importance of each of them to AREA. We turned off the alternation, and allowed AREA to use either $R_0$ or $R_1$ only. The following demonstrates the alteration strategy works better than using one of them only.

\subsubsection{AREA with Fixed Reference Set Only}
\begin{figure*}[t]
	\centering
	\begin{tabular}{@{}c@{}c@{}c}
		\includegraphics[width=0.32\linewidth, height=0.2\linewidth]{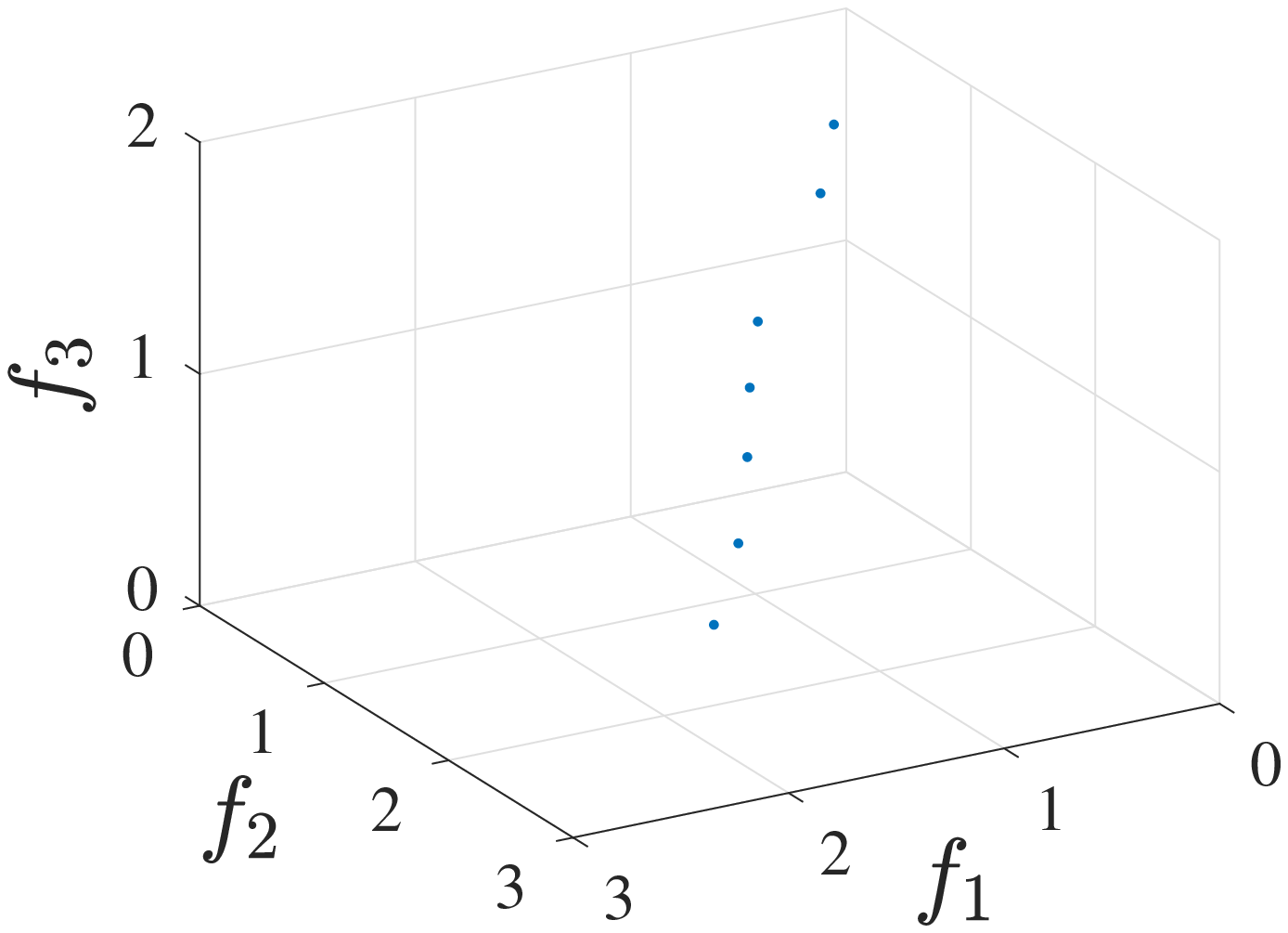}&
		\includegraphics[width=0.32\linewidth, height=0.2\linewidth]{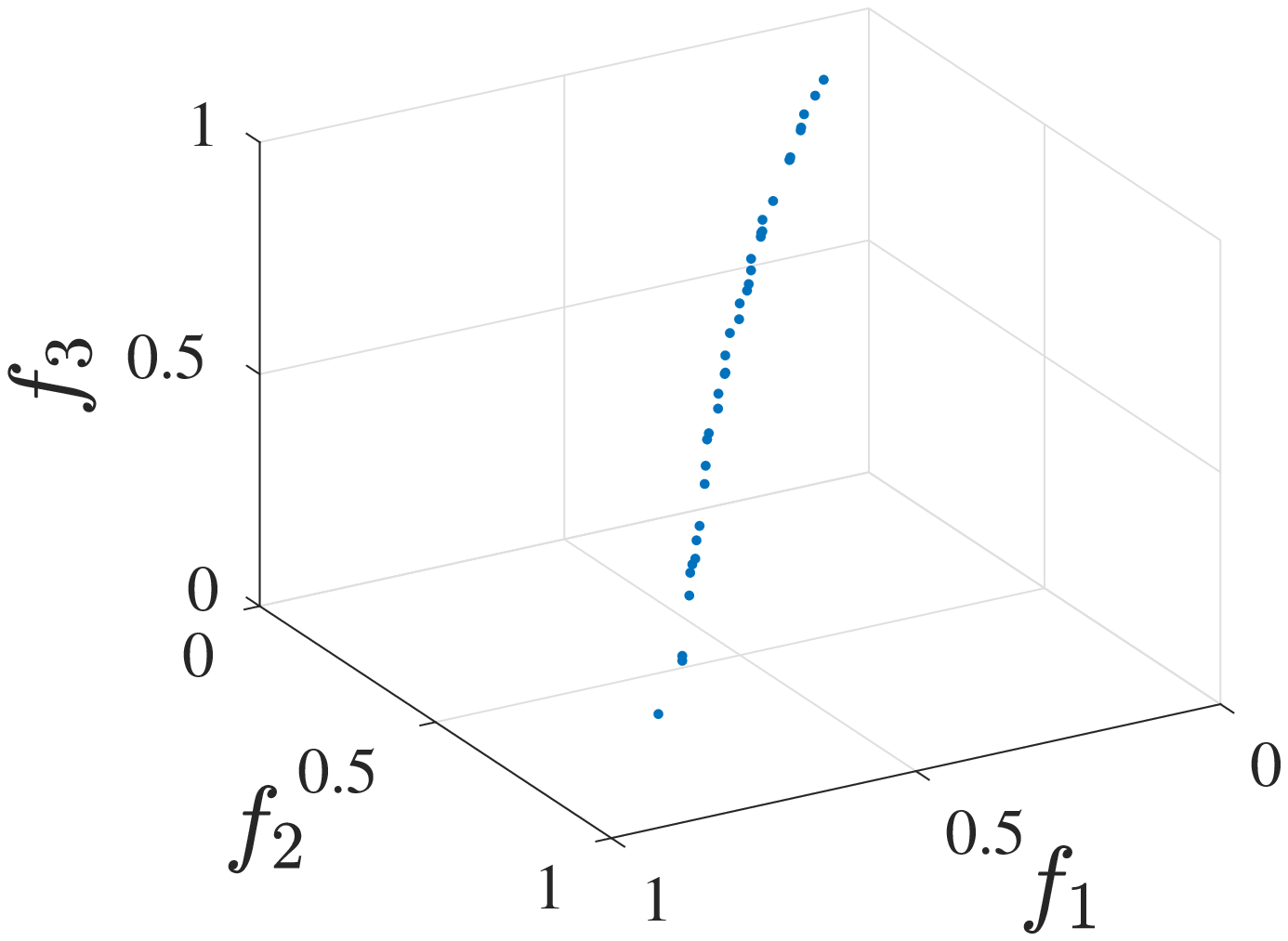}&
		\includegraphics[width=0.32\linewidth, height=0.2\linewidth]{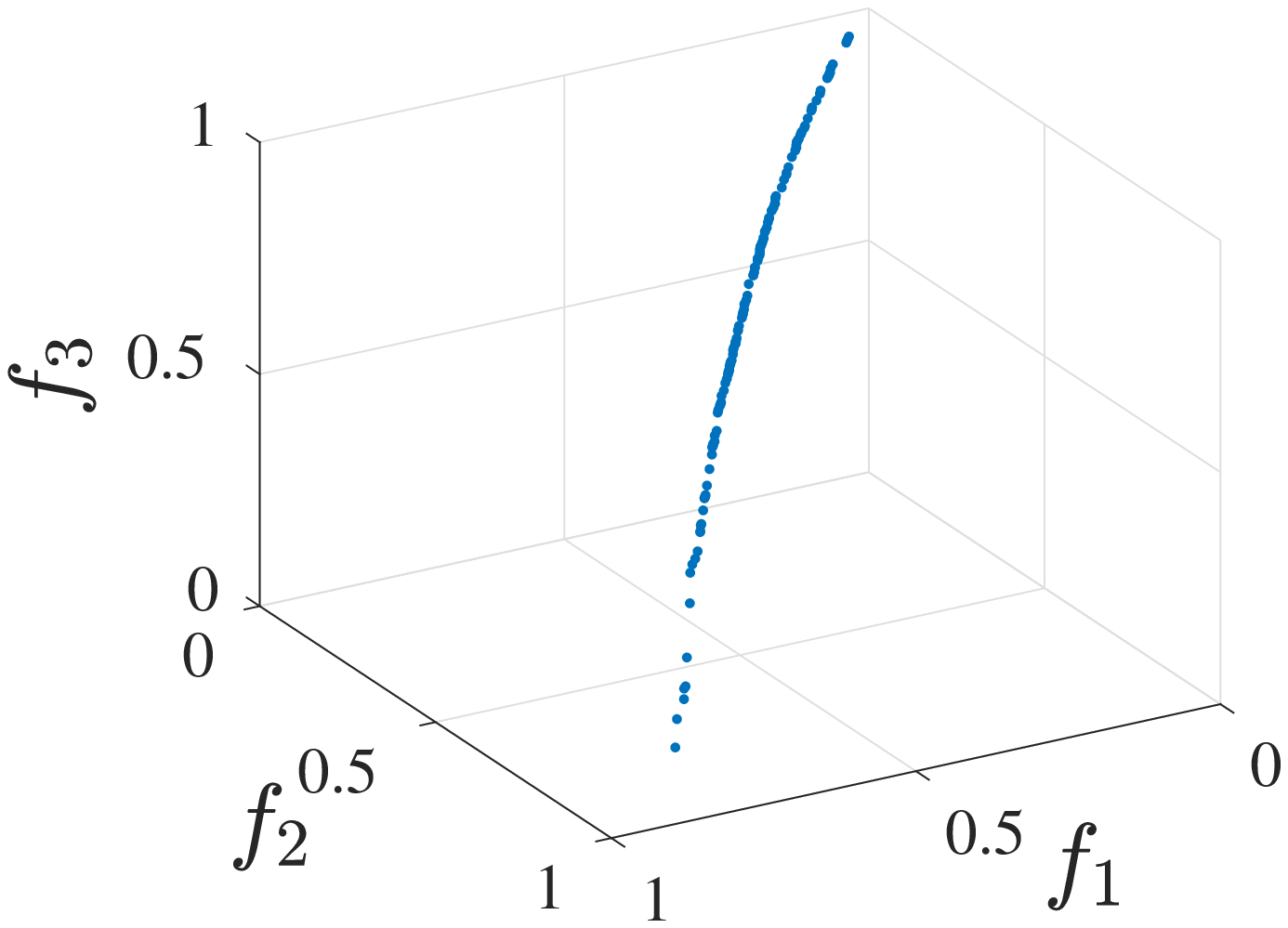}\\
		\includegraphics[width=0.32\linewidth, height=0.2\linewidth]{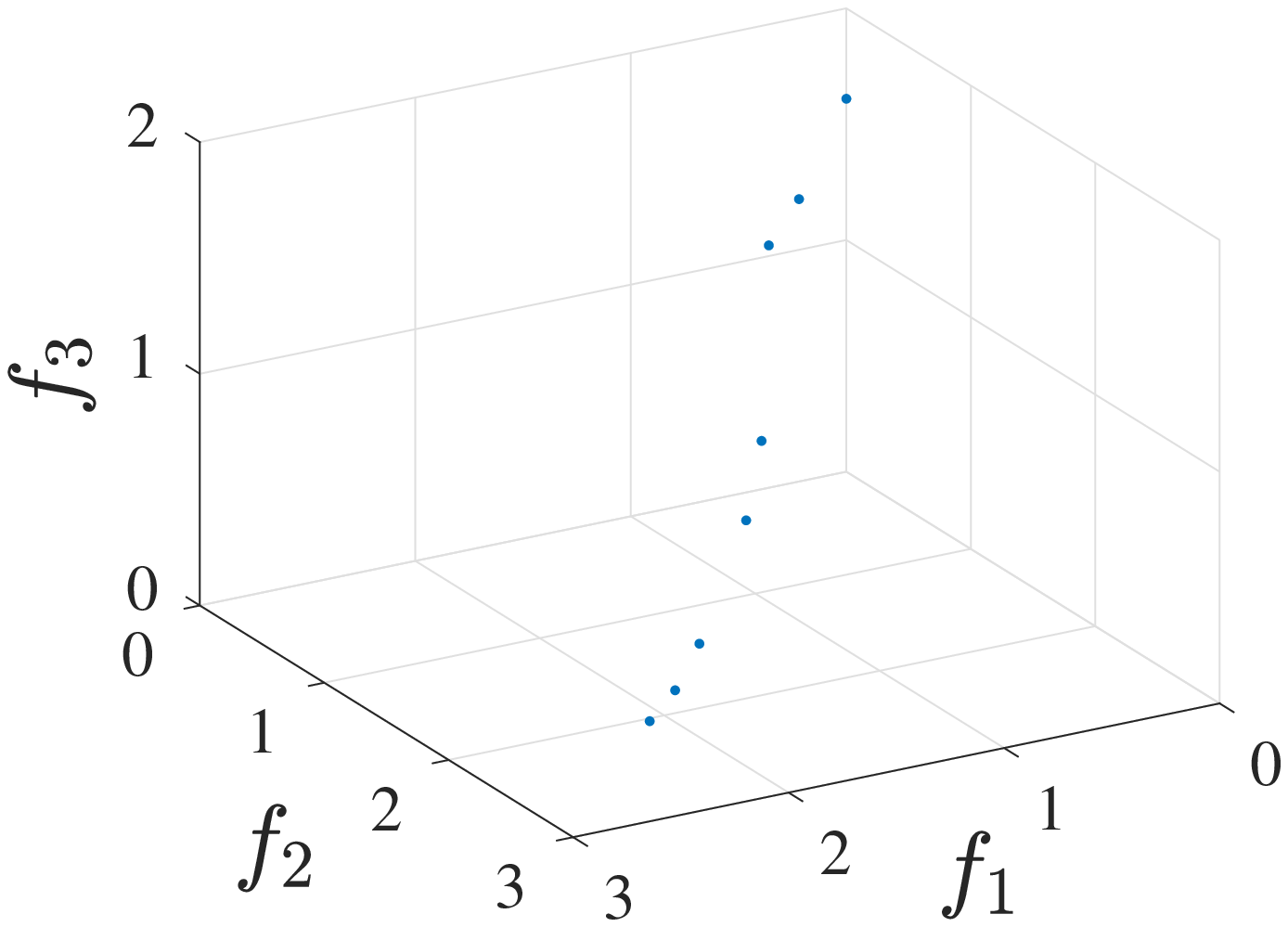}&
		\includegraphics[width=0.32\linewidth, height=0.2\linewidth]{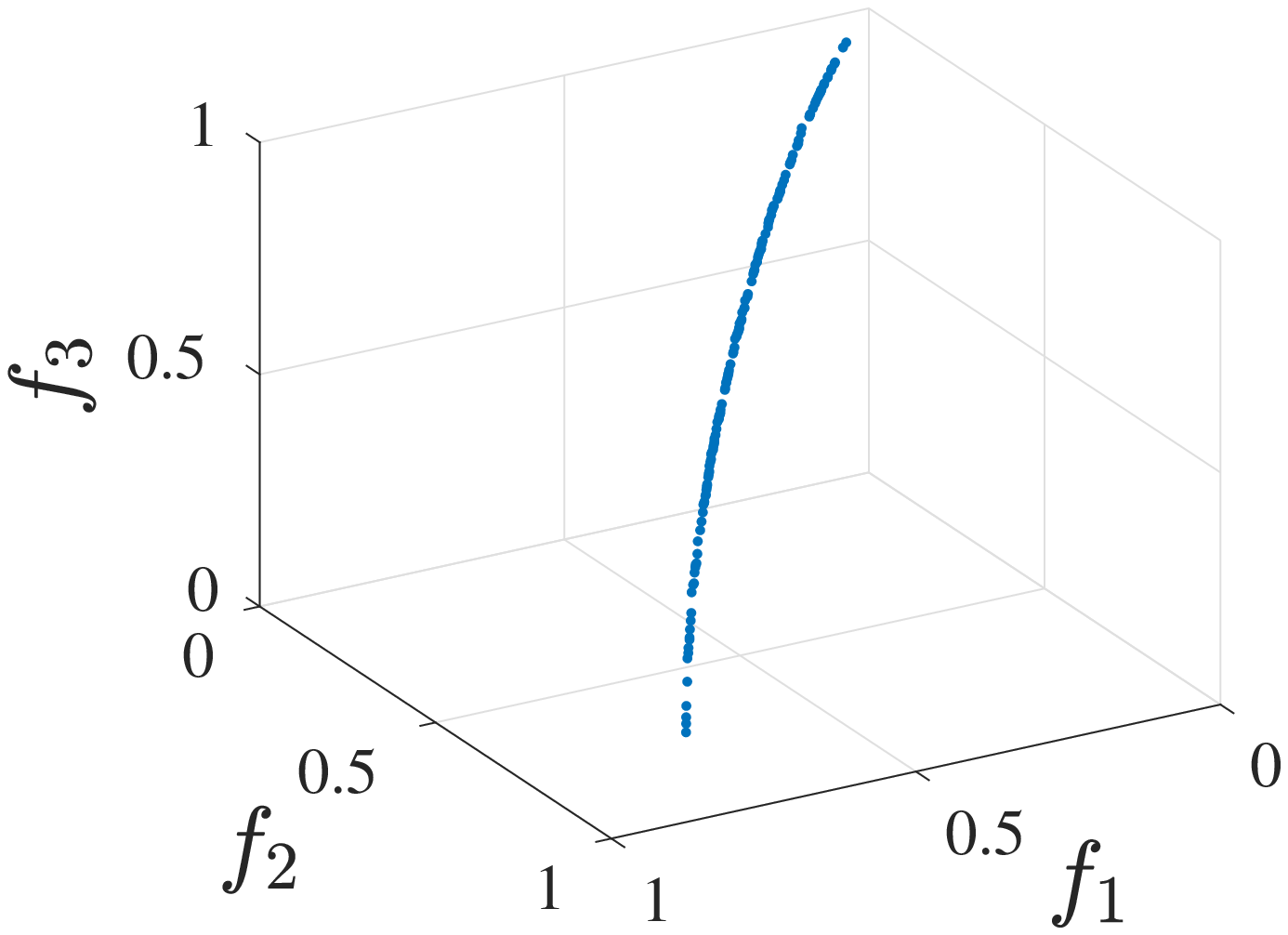}&
		\includegraphics[width=0.32\linewidth, height=0.2\linewidth]{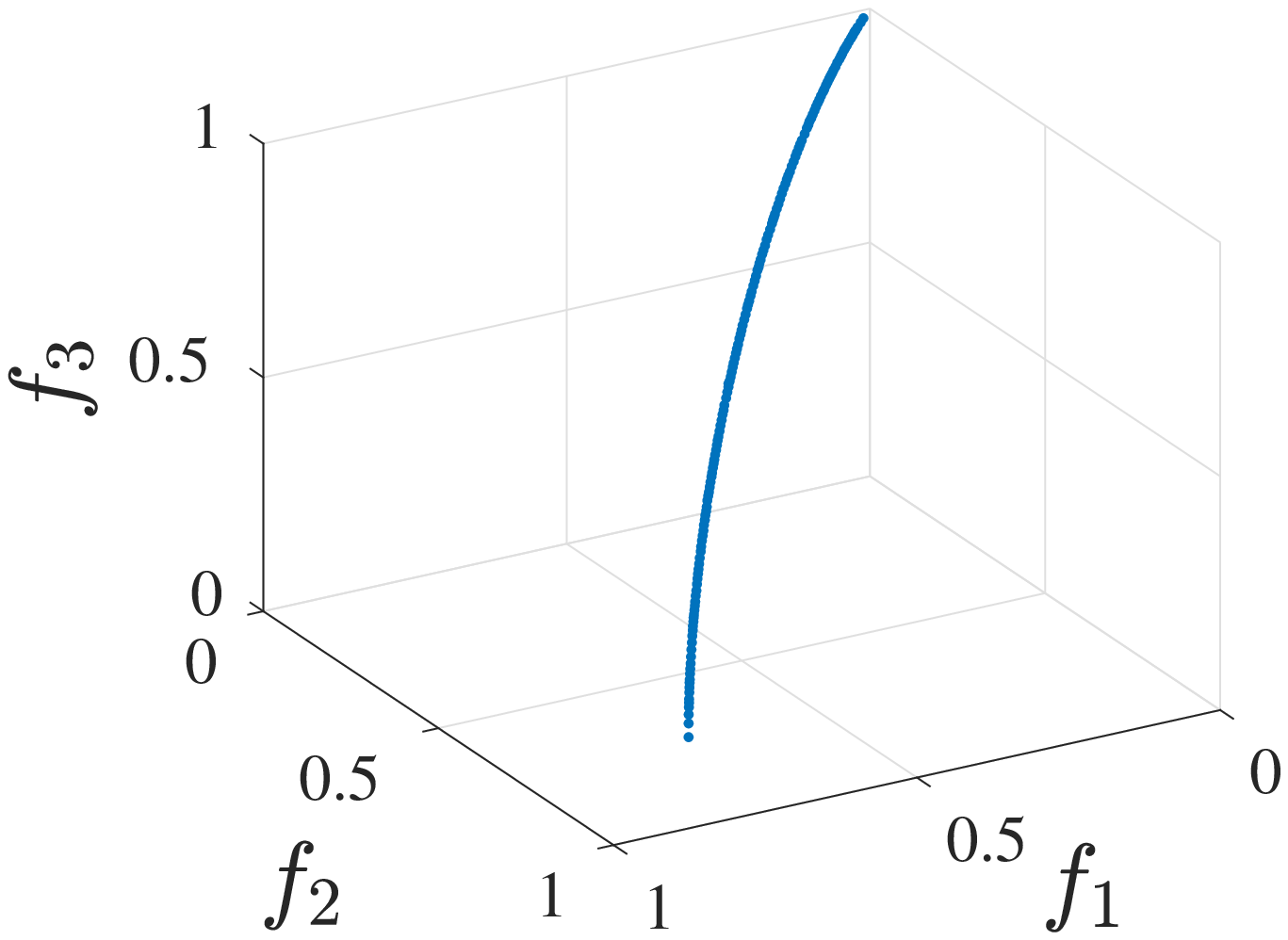}\\
		(a) 5000 FEs & (b) 15000 FEs & (c) 30000 FEs\\[-2mm]
	\end{tabular}
	\caption{PF approximation of DTLZ5 obtained by AREA$\_R_0$ (top) and AREA (bottom) after different number of FEs.}
	\label{fig:area_r0}
	\vspace{-2mm}
\end{figure*}

This AREA variant is denoted AREA$\_R_0$ and compared against the original AREA. We found that no significant difference exists between AREA and AREA$\_R_0$ for regular problems like DTLZ1-2 where the projection of the PF on the $\sum_{i=1}^{M}f_i=0$ is a simplex. However, AREA$\_R_0$ was shown to have coverage limitations for other types of problems. For example, AREA$\_R_0$ takes longer than AREA to find a good coverage of the PF for the degenerate DTLZ5, as illustrated in Fig.~\ref{fig:area_r0}. 

\subsubsection{{AREA with Evolving Reference Set Only}}
\begin{figure*}[t]
	\centering
	\begin{tabular}{@{}c@{}c@{}c}
		\includegraphics[width=0.32\linewidth, height=0.2\linewidth]{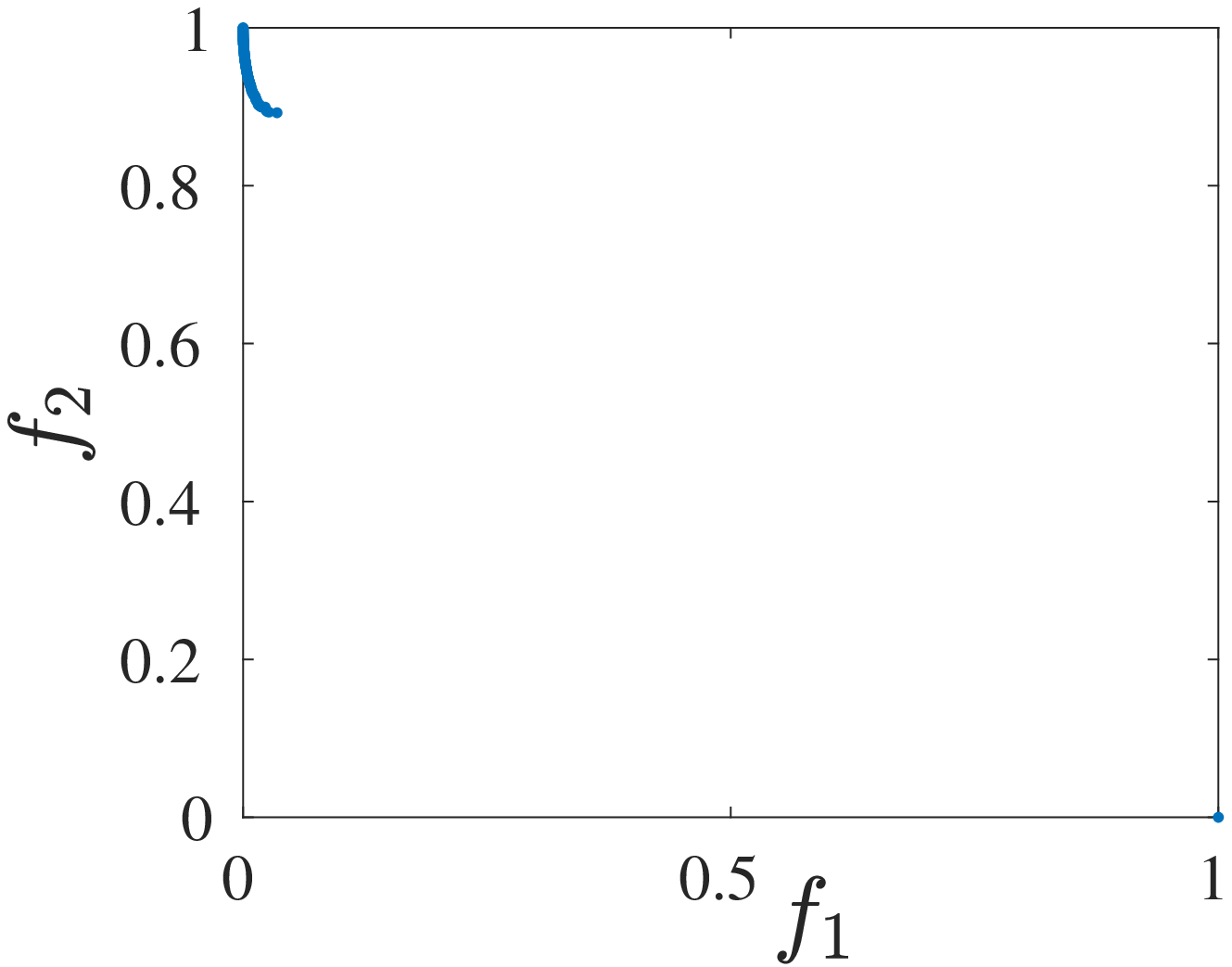}&
		\includegraphics[width=0.32\linewidth, height=0.2\linewidth]{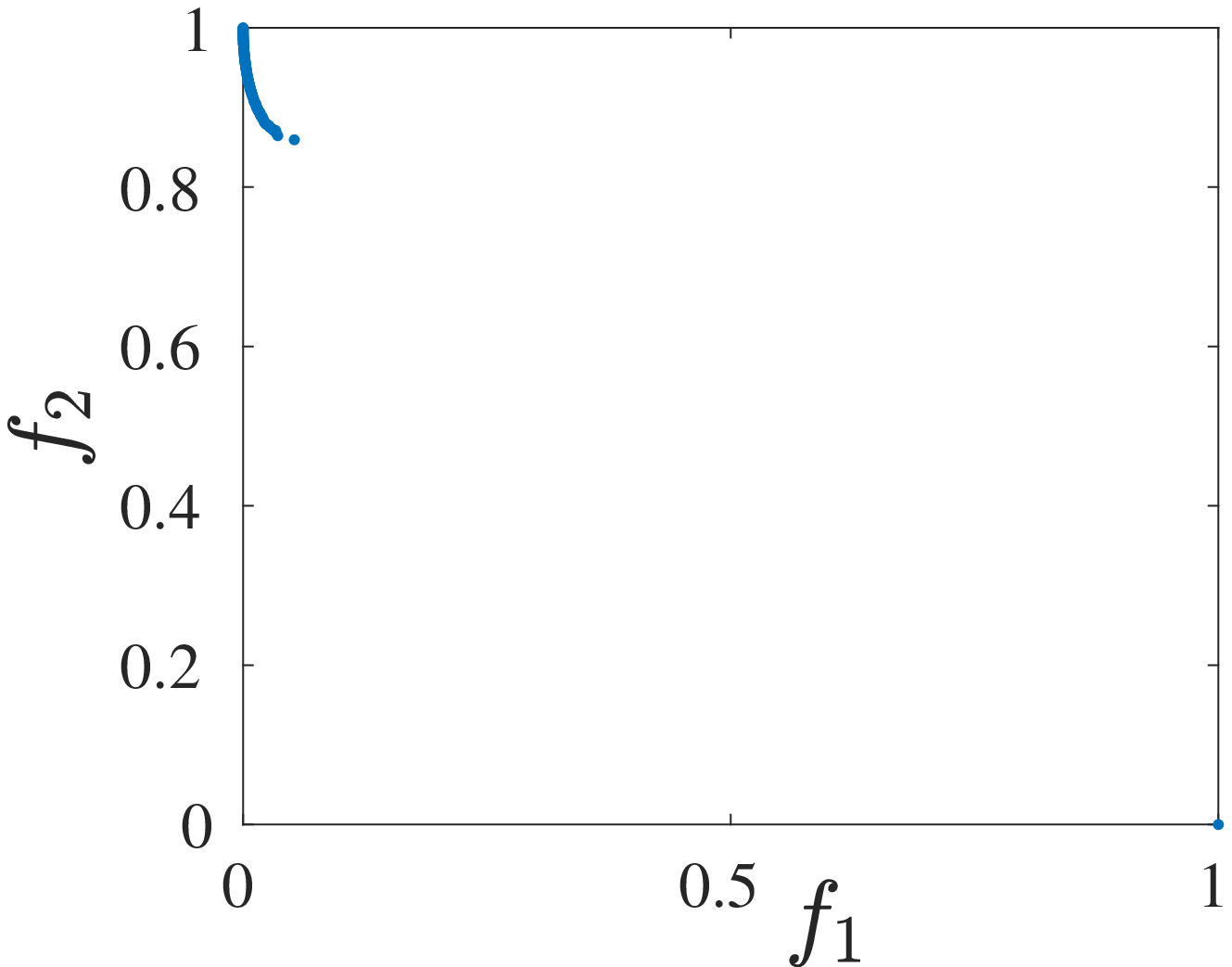}&
		\includegraphics[width=0.32\linewidth, height=0.2\linewidth]{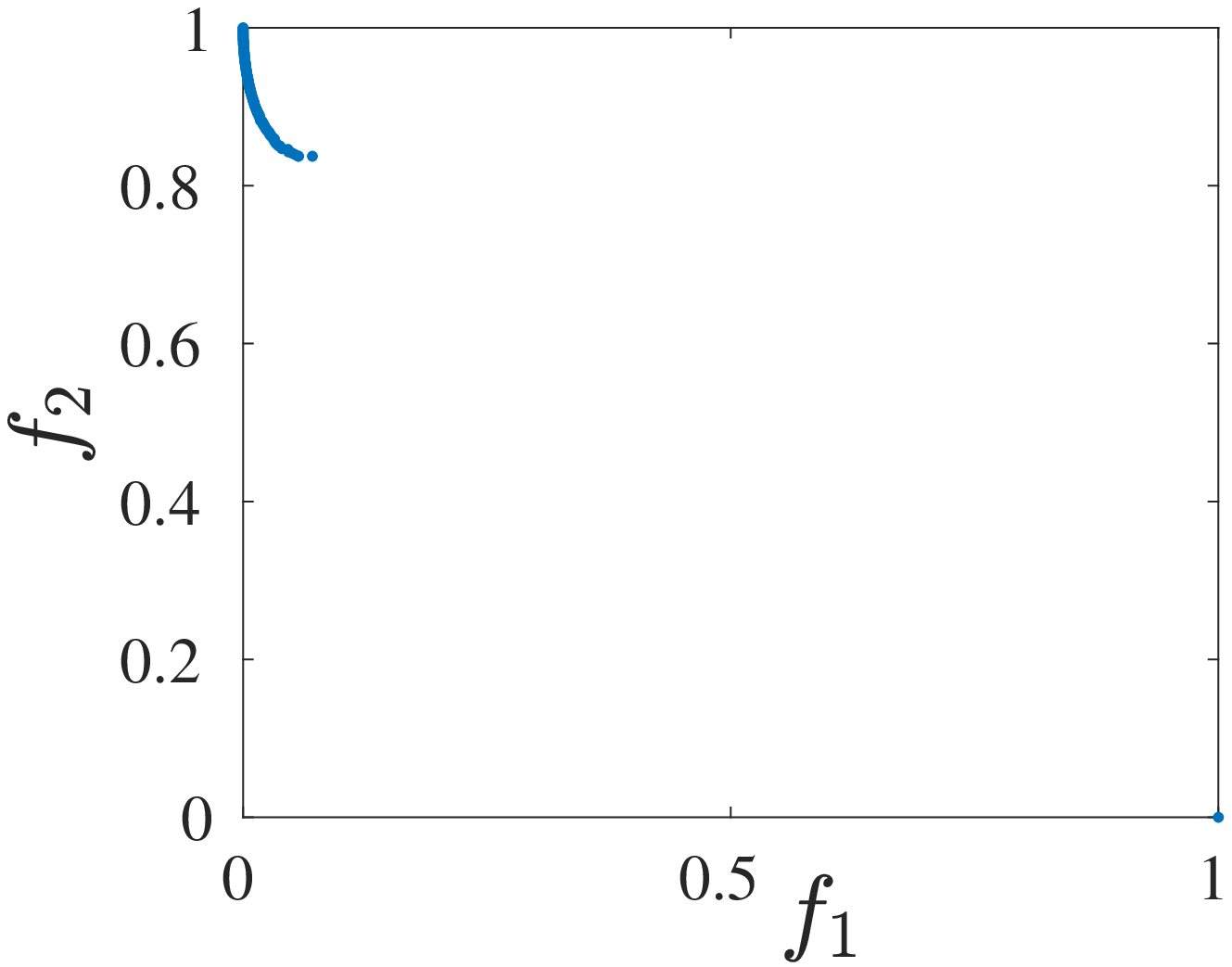}\\
		\includegraphics[width=0.32\linewidth, height=0.2\linewidth]{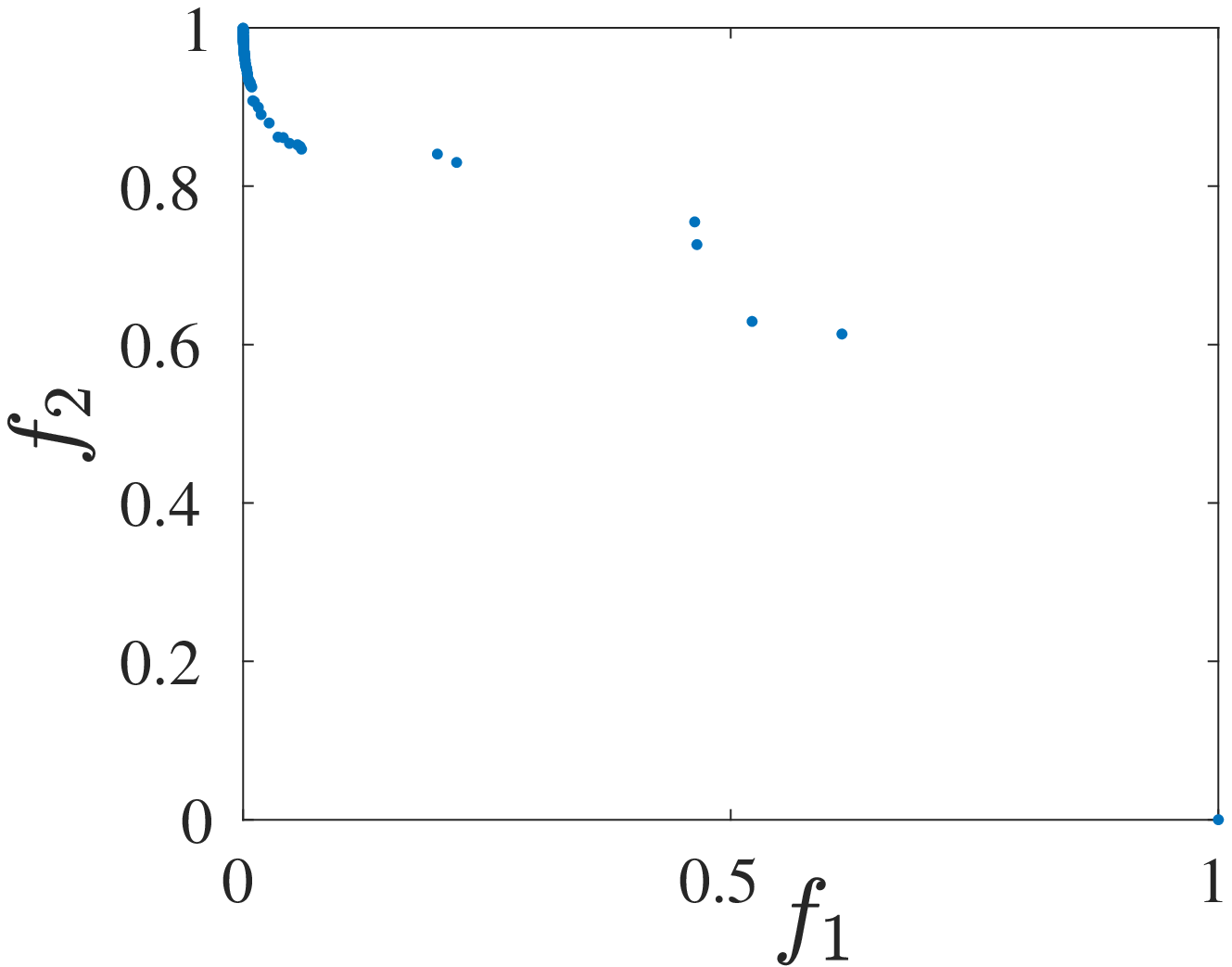}&
		\includegraphics[width=0.32\linewidth, height=0.2\linewidth]{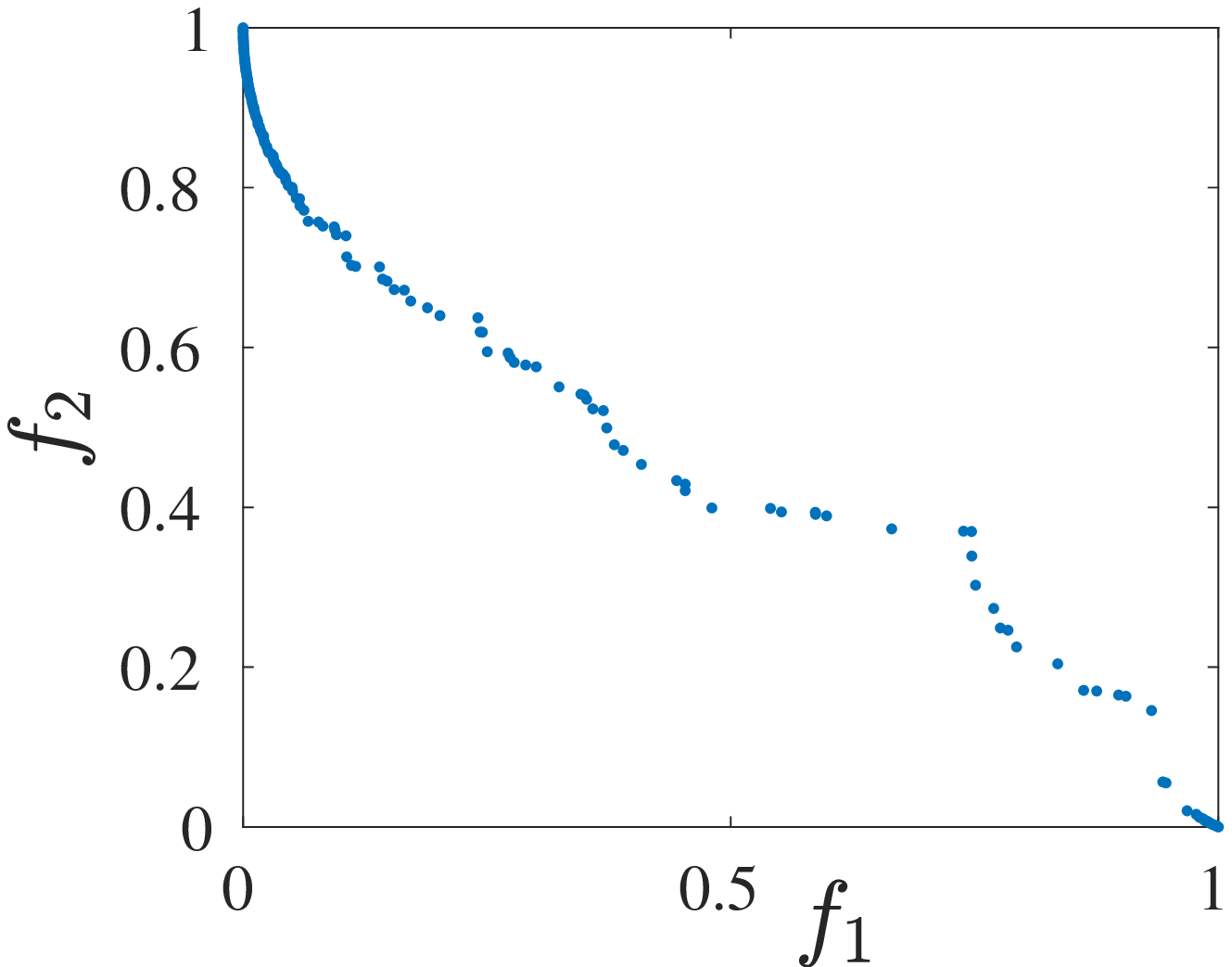}&
		\includegraphics[width=0.32\linewidth, height=0.2\linewidth]{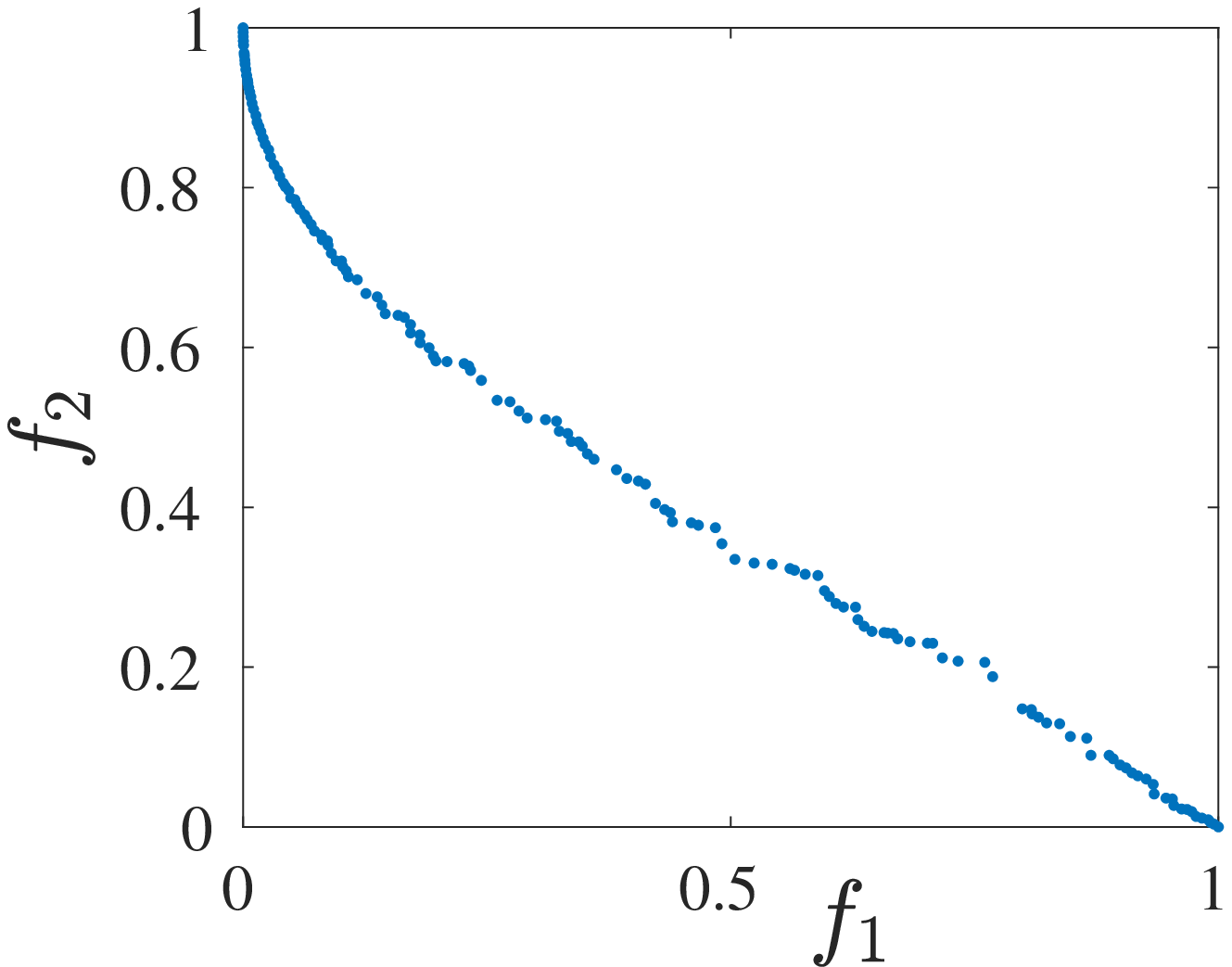}\\
		(a) 5000 FEs & (b) 15000 FEs & (c) 30000 FEs\\[-2mm]
	\end{tabular}
	\caption{PF approximation of MOP1 obtained by AREA$\_R_1$ (top) and AREA (bottom) after different number of FEs.}
	\label{fig:area_r1}
	\vspace{-2mm}
\end{figure*}

\begin{figure*}[!t]
	\centering
	\begin{tabular}{@{}c@{}c@{}c}
		\includegraphics[width=0.32\linewidth, height=0.2\linewidth]{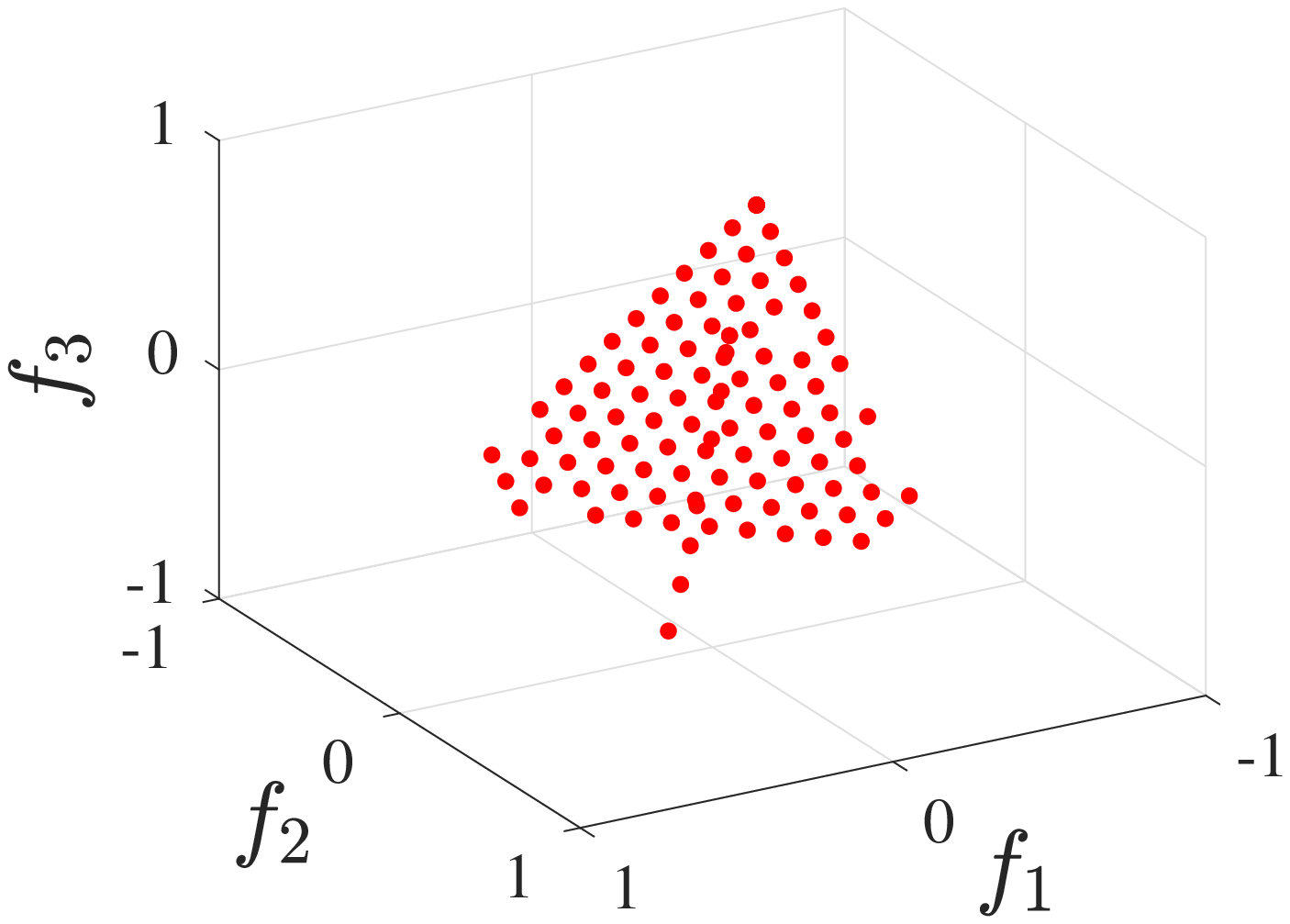}&
		\includegraphics[width=0.32\linewidth, height=0.2\linewidth]{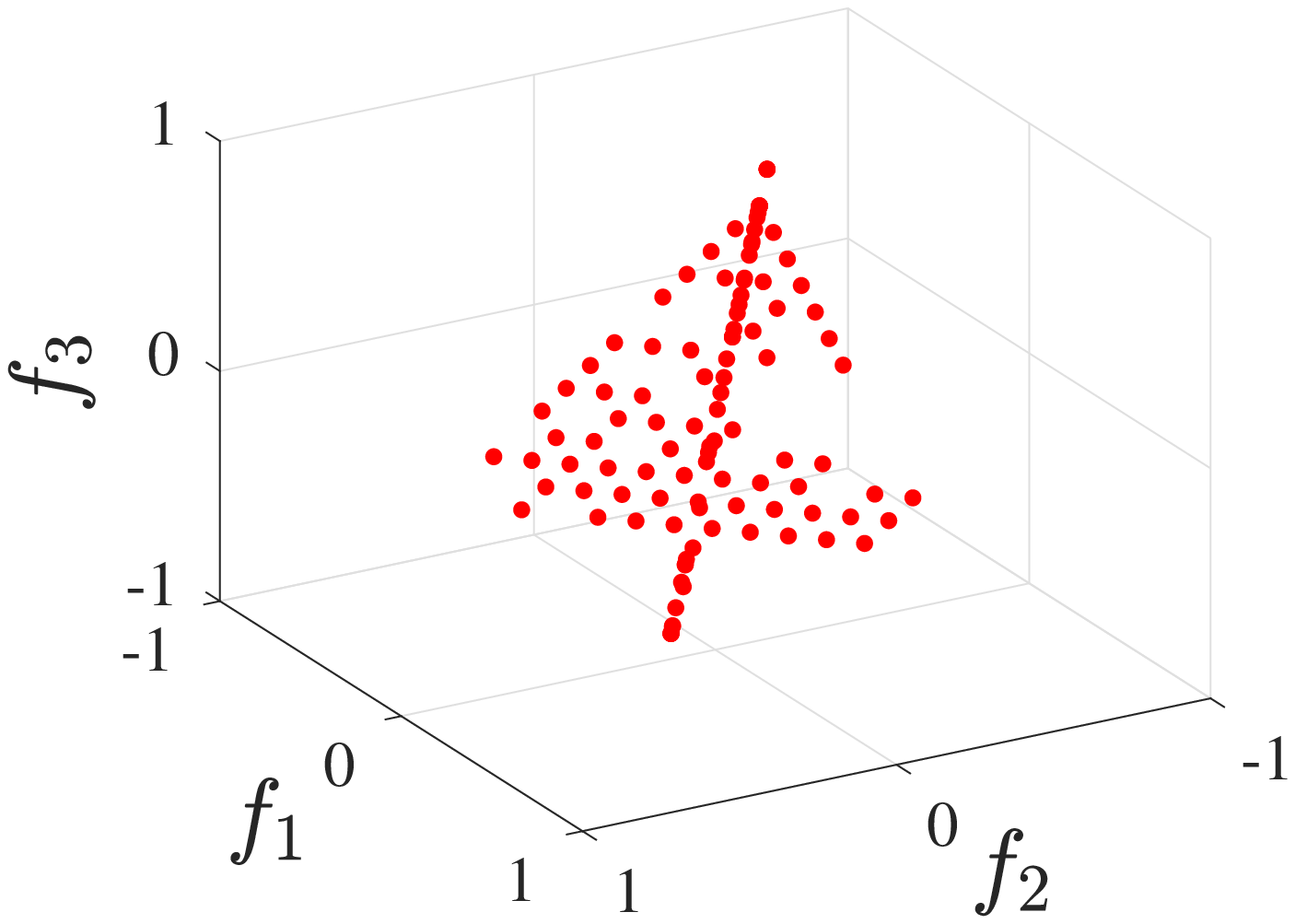}&
		\includegraphics[width=0.32\linewidth, height=0.2\linewidth]{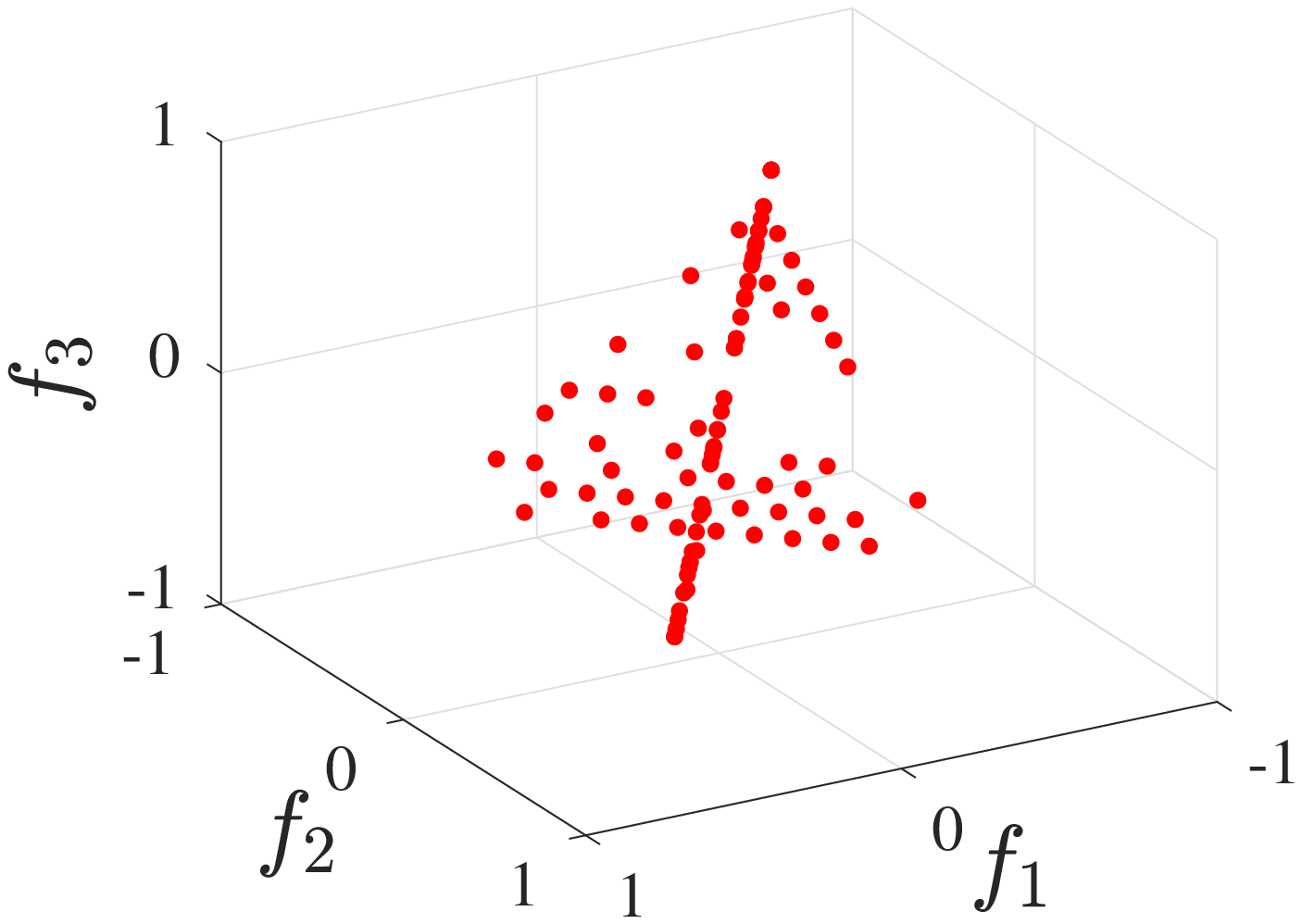}\\
		(a) 5000 FEs & (b) 15000 FEs & (c) 30000 FEs\\[-2mm]
	\end{tabular}
	\caption{Evolved reference set obtained by AREA$\_R_0$ after different number of FEs.}
	\label{fig:area_r0w}
	\vspace{-2mm}
\end{figure*}

AREA with an evolving reference set only is denoted AREA$\_R_1$. We found that using only a reference set evolved with the nondominated solutions found during the search tends to drive the population toward local PF regions, as shown in Fig.~\ref{fig:area_r1} for MOP1. This can be simply explained by the fact that a nondominated set does not guarantee a good distribution across the PF, particularly at the early stage of the search. The use of poorly-distributed nondominated solutions to update the reference set therefore creates dense reference points that are targets for overcrowded nondominated solutions, and eliminates reference points that are desired targets but no nondominated solutions have been found yet for them. In contrast, the original AREA avoids this issue by using a fixed reference set periodically. Fig.~\ref{fig:area_r0w} demonstrates that the evolving reference set is able to adjust itself gradually to fit the PF based on obtained nondominated solutions.
Therefore AREA can solve MOP1 whereas AREA$\_R_1$ cannot.

\begin{table*}[htbp]
	\addtolength{\tabcolsep}{-4.5pt}
	\centering
	\footnotesize
	\caption{Mean and standard deviation of IGD obtained from AREA with different $f_r$ values}
	\begin{tabular}{ccccccc}
		\toprule
		Problem & AREA+$f_r$(5\%)& AREA+$f_r$(10\%) & AREA+$f_r$(20\%) & AREA+$f_r$(30\%) & AREA+$f_r$(40\%) & AREA+$f_r$(50\%)\\
		\midrule
		\multirow{1}{*}{DTLZ1}	& \hl{2.0303e-2(5.59e-4)}	& {2.0968e-2(1.10e-3)}	& {2.0748e-2(1.07e-3)}	& {2.1145e-2(1.60e-3)}	& {2.0693e-2(4.68e-4)}	& {2.0862e-2(8.38e-4)}\\
		\hline                                            	                                                     		                        	                        
		\multirow{1}{*}{DTLZ2}	& \hl{5.2651e-2(4.95e-4)}	& {5.3351e-2(5.86e-4)}	& {5.3541e-2(4.85e-4)}	& {5.3505e-2(4.64e-4)}	& {5.3689e-2(6.15e-4)}	& {5.3717e-2(5.12e-4)}\\
		\hline                                            	                                                     		                        	                        
		\multirow{1}{*}{DTLZ5}	& \hl{4.1568e-3(9.44e-5)}	& {5.1654e-3(3.33e-4)}	& {5.7278e-3(3.60e-4)}	& {5.7360e-3(5.12e-4)}	& {6.5333e-3(7.26e-4)}	& {5.8470e-3(7.84e-4)}\\
		\hline                                            	                                                     		                        	                        
		\multirow{1}{*}{DTLZ7}	& \hl{5.6225e-2(1.47e-3)}	& {1.1494e-1(1.18e-1)}	& {1.4391e-1(1.36e-1)}	& {1.8732e-1(1.48e-1)}	& {1.7493e-1(2.38e-1)}	& {1.4371e-1(1.36e-1)}\\
		\hline                                            	                                                     		                        	                        
		\multirow{1}{*}{F1}		&  {4.6459e-3(3.16e-5)}	& {4.6172e-3(3.46e-5)}	& \hl{4.6103e-3(1.77e-5)}	& {4.6137e-3(3.56e-5)} & {4.6164e-3(4.32e-5)}	& {4.6147e-3(3.89e-5)}\\
		\hline                                            	                                                     		                        	                        
		\multirow{1}{*}{F2}		&  {5.1982e-3(1.04e-3)}	& {4.6484e-3(4.40e-5)}	& {4.6585e-3(1.48e-4)}	& {4.6640e-3(1.36e-4)}	&\hl{4.6149e-3(4.34e-5)}	& {4.7884e-3(1.71e-4)}\\
		\hline                                            	                                                     		                        	                        
		\multirow{1}{*}{F3}		&  {3.1088e-2(1.83e-4)}	& {3.0938e-2(1.71e-4)}	& {3.0803e-2(1.65e-4)}	& {3.0876e-2(1.16e-4)}	& {3.0785e-2(1.60e-4)}	&\hl{3.0774e-2(1.90e-4)}\\
		\hline                                            	                                                     		                        	                        
		\multirow{1}{*}{F4}		&  {3.7011e-3(3.86e-5)}	&\hl{3.7010e-1(3.93e-5)}	& {3.7012e-1(5.67e-5)}	& {3.7010e-1(7.62e-5)}	& {3.7011e-1(4.90e-5)}	& {3.7011e-1(5.78e-5)}\\
		
		\bottomrule
	\end{tabular}
	\label{tab:fr}
\end{table*}

\subsection{Influence of Alternation Frequency of Reference Set}
The alternation frequency $f_r$ of reference set is an important parameter in AREA and should be properly tuned. Frequent alternation increases significantly computational complexity whereas slow alternation may cause untimely update of the evolving reference set.

We investigated the impact of $f_r$ by varying $f_r$  from 5\% to 50\% the total number of fitness evaluations on eight selected test problems. $f_r$  below 5\% is not considered for computational efficiency. The results are shown in Table \ref{tab:fr}. It is clear that $f_r=5\%$ obtains the best results for the DTLZ problems, and no significant improvement is observed for increased $f_r$ on the problems from the F group. $f_r=5\%$ is specially the best setting for DTLZ5 and DTLZ7, as indicated by the fact that an increase in $f_r$ causes drastic performance deterioration.

\subsection{Influence of Archive Size}
The sensitivity of AREA to archive size  is investigated here. The archive size ($|A|$) should neither be  too large nor too small. A too small archive 
stores very limited information about the PF, which renders insufficient to update the evolving reference set. A too large archive is also not good as it increases computational complexity. It is a common practice that the archive is at least as large as the population size ($N$). Therefore, we test $|A| \in \{1, 1.5, 2, 3\}$ times of $N$ over several problems.

\begin{table*}[tbp]
	\addtolength{\tabcolsep}{-1pt}
	\centering
	\caption{Mean and standard deviation of IGD obtained from different archive sizes}
	\begin{tabular}{ccccc}
		\toprule
		Problem & $|A|=1\cdot N$ & $|A|=1.5\cdot N$ & $|A|=2\cdot N$ &$|A|=3\cdot N$ \\
		\midrule
		\multirow{1}{*}{DTLZ1}	&\hl{1.9877e-2 (4.64e-4)}	& {2.0303e-2 (5.59e-4)} & {2.0624e-2 (5.58e-4)}	& {2.0743e-2 (7.10e-4)}\\
		\hline                                                                         	                            
		\multirow{1}{*}{DTLZ2}	& {5.2804e-2 (5.63e-4)}	& \hl{5.2651e-2 (4.95e-4)} & {5.3081e-2 (5.32e-4)}	& {5.2769e-2 (5.38e-4)}\\
		\hline                                                                         	                            
		\multirow{1}{*}{DTLZ5}	&{5.1165e-3 (4.91e-4)}	& \hl{4.1568e-3 (9.44e-5)} &{4.7389e-3 (1.63e-4)}	& {4.6006e-3 (1.70e-4)}\\
		\hline                                                                         	                            
		\multirow{1}{*}{DTLZ7}	&{1.4356e-1 (1.35e-1)}	& \hl{5.6225e-2 (1.47e-3)} &{1.3020e-1 (1.29e-1)}	& {1.4350e-1 (1.35e-1)}\\
		\hline                                                                         	                            
		\multirow{1}{*}{F1}		&\hl{4.5463e-3 (3.11e-5)}	& {4.6459e-3 (3.16e-5)}    &{4.5895e-3 (3.29e-5)}	& {4.6045e-3 (3.19e-5)}\\
		\hline                                                                         	                            
		\multirow{1}{*}{F2}		&\hl{4.4339e-3 (3.23e-5)}	& 5.1982e-3 (1.04e-3) 	   &{4.5831e-3 (9.19e-5)}	& {4.5946e-3 (3.39e-5)}\\
		\bottomrule
	\end{tabular}
	\label{tab:archsize}
\end{table*}

The sensitivity of the archive size to AREA is reported in Table \ref{tab:archsize}. Surprisingly, a large archive size (over $1.5 N$) does not help improve the performance of AREA. We suspect that this is largely due to potential non-uniformity caused by the archive truncation method used in AREA. Although no more than $|A|$ solutions are kept in archive during the search, the archive, if $|A|>N$, is truncated to a size of $N$ for performance assessment. This step likely induces uniformity issues. A larger archive requires more solutions to be removed to keep the size as $N$, and removing more solutions is more likely to reduce the uniformity of solution distribution if solutions in the archive are already well distributed. An archive size of $1.5N$, however, shows to be a wise choice, as it enables remarked performance on some problems, e.g., DTLZ5 and DTLZ7,  while staying robust on the other problems and not increasing computational complexity too much compared with $|A|=N$. $|A|=N$ is not good for some problems like the two irregular DTLZ problems, as it probably eliminates potential high-quality solutions due to the size limit. 

\subsection{Influence of Number of Additions/Removals}
In reference set update, AREA requires to add $K$ potential reference points and then remove the same number of unpromising points. The impact of $K$ on the performance of AREA is investigated through different settings of $K$ values from 0\% to 20\% of population size $N$. 

Table \ref{tab:impactK} shows the performance of AREA with different $K$ values. It is observed that problems with a regular PF (e.g., DTLZ1) prefer no changes in the reference set whereas problems with an irregular PF (e.g., DTLZ5 and DTLZ7) can be better solved by updating around 10\% of reference points. Too large $K$ values may be good for some problems at the expense of high computations. Therefore, The default setting of $K=\sqrt{N}$ ($\sim$ 10\%) is a good choice for AREA.
\begin{table*}[t]
	\addtolength{\tabcolsep}{-2pt}
	\centering
	\caption{Mean and standard deviation of IGD obtained from AREA with different K values}
	\begin{tabular}{ccccc}
		\toprule
		Problem & K=0\% & K=5\% & K=$\sqrt{N}$ ($\sim$ 10\%)  & K=20\% \\
		\midrule
		\multirow{1}{*}{DTLZ1}		&\hl{1.9503e-2 (3.62e-4)}&{2.0423e-2 (2.28e-3)}& {2.0303e-2 (5.59e-4)}    &{2.0948e-2 (3.34e-3)}\\
		\hline                                                  
		\multirow{1}{*}{DTLZ2}		&{5.3006e-2 (4.83e-4)}&{5.2851e-2 (5.66e-4)}& \hl{5.2651e-2 (4.95e-4)} &{5.2740e-2 (6.28e-4)}\\
		\hline                                                  
		\multirow{1}{*}{DTLZ5}		&{7.3860e-3 (6.40e-4)}&{5.5111e-3 (4.86e-4)}& \hl{4.1568e-3 (9.44e-5)} &{4.6164e-3 (4.44e-4)}\\
		\hline                                                  
		\multirow{1}{*}{DTLZ7}		&{2.5149e-1 (2.61e-1)}&{9.5389e-2 (1.02e-1)}& \hl{5.6225e-2 (1.47e-3)} &{1.4561e-1 (2.07e-1)}\\
		\hline                                                  
		\multirow{1}{*}{F1}			&{5.0699e-3 (4.14e-4)}&{4.5311e-3 (3.19e-5)}& {4.6459e-3 (3.16e-5)}    &\hl{4.5284e-3 (3.98e-5)}\\
		\bottomrule
	\end{tabular}
	\label{tab:impactK}
\end{table*}

\subsection{Influence of Local Mating Probability}
It has been widely reported that local mating is an effective strategy to speed up the search toward the PF\cite{LZ09,Jiang17_SPEAR}. AREA adopts this strategy for the same purpose. Dissimilar to the study \cite{LZ09} where the local mating probability ($p_r$) is set manually, AREA defines the probability based on the closeness of population members to the nearest archive members, thereby eliminating a parameter that requires tuning. However, it remains unclear whether such auto-calculation is more beneficial than manual settings. The following compares our method against the $p_r$ of 0.8 and 0.9, which are popular manual settings in a number of studies \cite{LZ09,Wang2016,ZZ16-GRA}. In addition, an AREA variant that allows zero local mating probability (ZLMP) is also investigated.

Table \ref{tab:lpr} reports the comparison between these approaches. While there is no significant difference between the proposed approach and the manual settings for most of the chosen problems, the proposed adaptive local mating does achieve much better performance in some cases, particularly for irregular problems like DTLZ5 and DTLZ7. Specially, the proposed approach helps AREA to obtain IGD values more than twice better than the manual setting method on DTLZ7, which seems to be very sensitive to manual settings of $p_r$ as suggested by the difference between $p_r=0.8$ and $p_r=0.9$ in Table \ref{tab:lpr}. It is also noticed that ZLMP has no significant effect on AREA except in problems with irregular PFs. ZLMP allows all possible parents to have a chance to mate locally, and this seems in particular helpful for the disconnected DTLZ7. 

\begin{table*}[t]
	\addtolength{\tabcolsep}{-2pt}
	\centering
	\caption{Mean and standard deviation of IGD obtained from different local mating probability schemes}
	\begin{tabular}{ccccc}
		\toprule
		Problem & AREA & AREA+$p_r$(0.8) & AREA+$p_r$(0.9) & AREA +ZLMP \\
		\midrule
		\multirow{1}{*}{DTLZ1}		& {2.0303e-2 (5.59e-4)} &{2.0949e-2 (9.14e-4)} & {2.0979e-2 (1.55e-3)} & \hl{1.9752e-2 (3.84e-4)}\\
		\hline                                                  
		\multirow{1}{*}{DTLZ2}		& \hl{5.2651e-2 (4.95e-4)} &{5.3327e-2 (5.76e-4)} & {5.3269e-2 (5.41e-4)} & {5.2820e-2 (3.43e-4)}\\
		\hline                                                  
		\multirow{1}{*}{DTLZ5}		& \hl{4.1568e-3 (9.44e-5)} &{4.6898e-3 (2.29e-4)} & {4.7078e-3 (2.05e-4)} & {4.8886e-3 (3.11e-4)}\\
		\hline                                                  
		\multirow{1}{*}{DTLZ7}		& \hl{5.6225e-2 (1.47e-3)} &{1.4318e-1 (1.36e-1)} & {2.2449e-1 (1.99e-1)} & {9.0849e-2 (9.57e-2)}\\
		\hline                                                  
		\multirow{1}{*}{F1}			& {4.6459e-3 (3.16e-5)} &{4.6417e-3 (4.70e-5)} & {4.6325e-3 (3.30e-5)} & \hl{4.5400e-3 (4.72e-5)}\\
		\hline                                                  
		\multirow{1}{*}{F2}		& 5.1982e-3 (1.04e-3) &{4.7065e-3 (1.14e-4)} & {4.6822e-3 (6.06e-5)} &\hl{4.4528e-3 (2.23e-5)}\\
		\hline                                                  
		\multirow{1}{*}{F3}		& {3.1088e-2 (1.83e-4)} &\hl{3.0951e-2 (1.99e-4)} & {3.0987e-2 (2.35e-4)}&{3.0948e-2 (1.48e-4)}\\
		\hline                                                  
		\multirow{1}{*}{F4}		& {3.7011e-3 (3.86e-5)} &{3.7015e-1 (5.60e-5)} & {3.7010e-1 (3.88e-5)} &\hl{3.7009e-1 (4.57e-5)}\\
		\bottomrule
	\end{tabular}
	\label{tab:lpr}
\end{table*}
It is not surprising that the proposed approach is effective for irregular problems because it can adaptively determine whether local exploitation or global exploration is needed around a parent solution by calculating the nearness of this solution to the archive. Thus, the comparison here demonstrates the effectiveness of this approach.

\section{Experimental Design}
This section describes experimental design for performance assessment. The following subsections provide details of test problems, compared algorithms, simulation setup, and performance measures.   
\subsection{Test Problems}
A total of 35 problems covering a wide range of features, including multimodality, disconnectedness, concavity-convexity, variable linkage, non-simply connectivity, and degeneracy, are selected for experimental studies from literature. Table \ref{tab:exist} presents a brief summary of problems used for study. To be specific, the test problems have four DTLZ \cite{DTLZ05} problems (three objectives considered here) and some variants whose importance has been recently recognised \cite{Cheng2017,ISMN16-PF}. Seven problems (MOP1-7) from the MOP test suite \cite{LGZ14} are chosen due to their ability to analyse population diversity. UF1-9 \cite{Zhang2009} involve strong variable linkages and thus are difficult to solve. WFG \cite{HHBW06} problems are scalable in terms of objectives and can be used to test the potential of algorithms on many-objective optimisation \cite{CJOS-RVEA}. Some problems ($F1-F4$) with irregularly shaped PF are also selected from \cite{Yang2017} to test where algorithms are able to find a uniform distribution of solutions on the PF or not.  We take into account additionally four test examples, i.e., F5--F8, which have new problem features different from the above ones. $F5$ and $F6$ are non-simply disconnected, and the PF of $F6$ consists of 3D curves. $F7$ has two PF regions, one of which is close to the origin and the other not. $F8$ is continuous but its PF has four corners and three of which are not on the axis. These new features are largely missing from the literature and their inclusion should provide a more comprehensive analysis on the robustness of algorithms.   

\subsection{Compared Algorithms}
AREA \footnote{MATLAB implementation is freely available at \url{https://github.com/chang88ye/AREA} or \url{https://sites.google.com/view/shouyongjiang/resources}.} is compared with a range of algorithms: MOEA based on decomposition (MOEA/D) \cite{ZL07}, adaptive nondominated sorting genetic algrithm III (ANSGA-III) \cite{JD14}, streng Pareto evolutionary algorithm based reference directions (SPEA/R) \cite{Jiang17_SPEAR}, and reference vector based evolutionary algorithm (RVEA) \cite{CJOS-RVEA}. MOEA/D is used as a baseline algorithm, and ANSGA-III is an upgraded version of NSGA-III for tackling complicated problems. SPEA/R and RVEA are recently developed algorithms and have shown good ability to solve a wide range of problems. 


The settings of other parameters in the compared algorithms are shown in Table \ref{tab:algos}, as suggested in their original papers. The parameters in AREA are configured based on sensitivity analysis in the previous section. All the algorithms use the same crossover and mutation operators as in MOEA/D \cite{ZL07} for most of the problems except the MOP group. For the MOP group, an adaptive operator \cite{LL09} is used to speed up the convergence of population.

\begin{table}[t]
	\newcommand{\tabincell}[2]{\begin{tabular}{@{}#1@{}}#2\end{tabular}}
	\centering
	\footnotesize
	\addtolength{\tabcolsep}{0pt}
	\caption{A summary of test problems. $M$ denotes the number of objectives and $n$ the number of variables which is the same as as DTLZ2. }
	\begin{tabular}{l|l|l}
		\hline
		Category & Problem & Description \\
		\hline
		\multirow{16}{*}{Old}&DTLZ1 \cite{DTLZ05} & linear PF, multimodality \\
		&DTLZ2 \cite{DTLZ05}& concave PF that is a part of hypersphere in the positive orthant\\
		&DTLZ5 \cite{DTLZ05}& degeneracy with an 1-D PF \\
		&DTLZ7 \cite{DTLZ05}& disconnectivity with four disconnected PF segments\\
		&IDTLZ1-2 \cite{Cheng2017} & inverted DTLZ1 and DTLZ2, corner points not on objective axis\\
		&SDTLZ2 \cite{DJ13}& scaled DTLZ2, $f_i$ multiplied by $2^{i-1}$ for $1 \leq i \leq M$ \\
		&CDTLZ2 \cite{DJ13}& convex DTLZ2, $f_i \leftarrow f_i^4$ for $1 \leq i \leq M-1$ and $f_M \leftarrow f_M^2$\\
		&MOP1-7 \cite{LGZ14} & variable linkage, multimodality \\
		&UF1-9 \cite{Zhang2009}  & variable linkage, multimodality, disconnectivity\\
		&WFG2 \cite{HHBW06}  & convexity, nonseparability, disconnectivity\\
		&WFG4 \cite{HHBW06}  & concavity, nonseparability\\
		&WFG6 \cite{HHBW06}  & concavity, nonseparability\\
		&F1 \cite{Yang2017} & extremely convex PF, intermediate PF points close to the origin \\
		&F2 \cite{Yang2017}   & multiple disconnected PF segments  \\
		&F3 \cite{Yang2017}   & disconnectivity and extremely convex geometry \\
		&F4 \cite{Yang2017}   & a mix of concavity and convexity \\	\hline
		\multirow{4}{*}{New}&
		{F5} & 
		\tabincell{l}{objective functions are the same as DTLZ2, except\\
			$g(x)\!=\!\sum_{i=M}^{n}(x_i-0.5)^2+\prod_{i=1}^{M-1}|\sin(0.5\pi\lfloor 4x_i+1.6 \rfloor )|$}\\		
		
		&{F6} &  
		\tabincell{l}{objective functions are the same as DTLZ2, except\\
			$g(x)\!=\!(\prod_{i=1}^{M-1}(x_i))^{0.1}\sum_{i=M}^{n}(x_i\!-\!0.5)^2\!+\!\prod_{i=1}^{M\!-\!1}\!|\sin(2\pi x_i)|$} \\		
		
		&{F7} &  
		\tabincell{l}{objective functions are the same as DTLZ2, except\\
			$g(x)\!=\!\sum_{i=M}^{n}(x_i\!-\!\prod_{j=1}^{M\!-\!1}\!\!x_j)^2\!+\!|\sin(0.5\pi\lfloor 4\prod_{i=1}^{M\!-\!1}\!\!(x_i\!-\!2)\rfloor )|$}\\
		
		&{F8} &  
		\tabincell{l}{$f_{i\in\{1,M-1\}}=1+g(x)((1+\sum_{j=1}^{M}x_j)/x_i-1)$\\
			$f_M=\sum_{j=1}^{M}x_j$, $g(x)\!=\!\sum_{i=M}^{n}((x_i-x_1)/n)^2$, $x \in [1,4]^n$}\\	
		\hline
	\end{tabular}%
	\label{tab:exist}%
	\vspace{-2mm}
\end{table}%

\begin{table}[t]
	\newcommand{\tabincell}[2]{\begin{tabular}{@{}#1@{}}#2\end{tabular}}
	\centering
	\footnotesize
	\addtolength{\tabcolsep}{0pt}
	\caption{Configuration of algorithms for comparison. $M$ denotes the number of objectives and $N$ population size. FE is short for function evaluation.}
	\begin{tabular}{l|l}
		\hline
		Algorithm & Parameter configuration \\
		\hline
		MOEA/D & neighbourhood size ($T=20$) \\
		ANSGA-III & $M$ new points added for a reference point\\
		SPEA/R & restricted mating ($K=20\%N$) \\
		RVEA & rate of change ($\alpha=2$), frequency of adaption ($fr=0.1$)\\		
		AREA & neighbourhood size ($T=20$), update frequency ($f_r$=5\% \#FEs), archive size ($1.5N$)\\
		\hline
	\end{tabular}%
	\label{tab:algos}%
	\vspace{-2mm}
\end{table}

\subsection{Simulation Setup}
The setting of population size ($N$) depends on the number of objectives. A population size of 100 is used for 2 objectives, as recommended by \cite{DAPM02}. A population size of 105, and 200 is used for 3 objectives, according to the suggestion in \cite{Cheng2017,Li2018,Yan2018}. For scalable WFG problems, the population size is set according to \cite{Cheng2017}.

\begin{figure*}[t]
	\centering
	\begin{tabular}[c]{cccc}
		\begin{subfigure}[t]{0.48\textwidth}
			\centering
			\includegraphics[width=\linewidth]{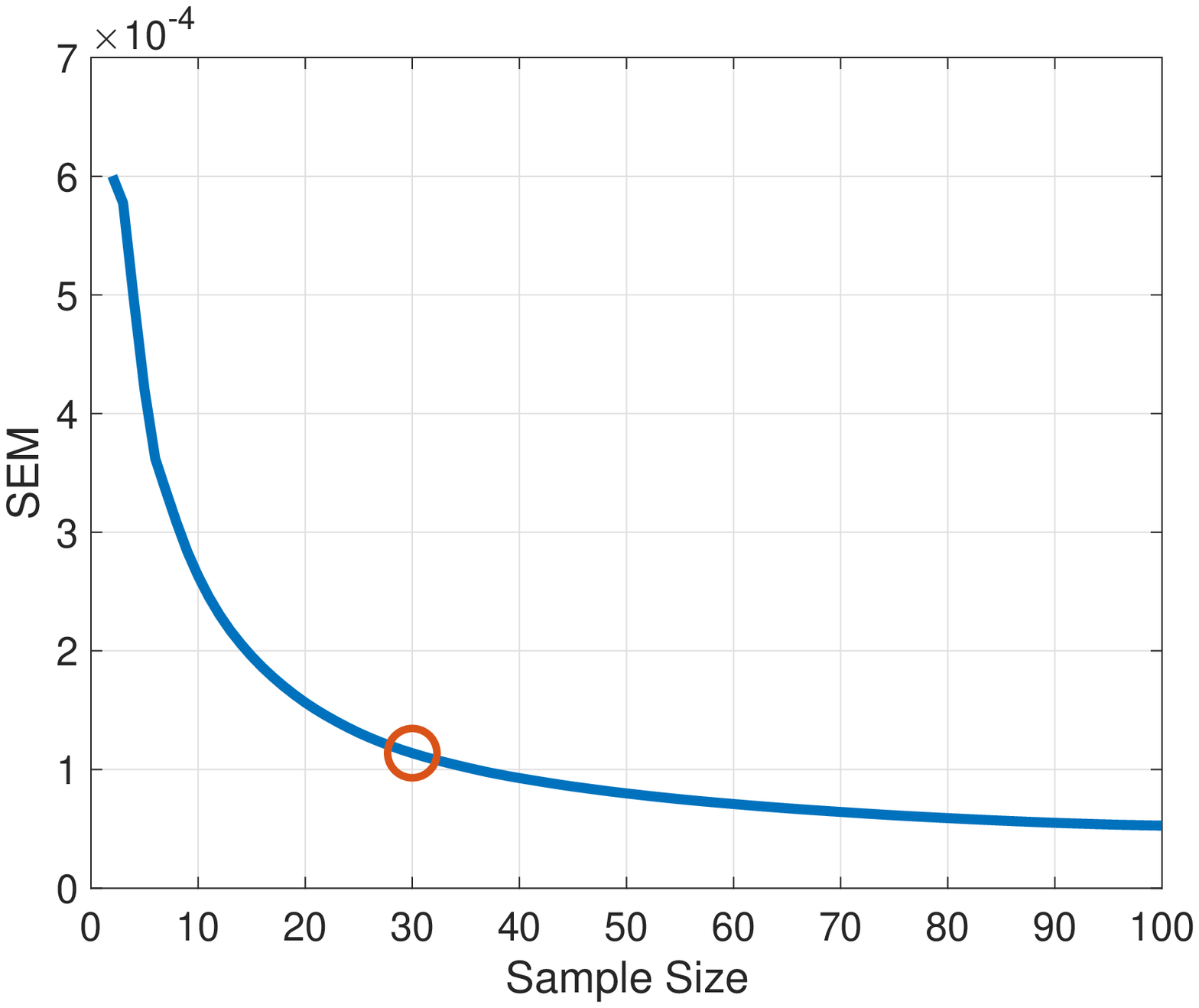}
			\caption{Impact of sample size on SEM}
		\end{subfigure}&	
		\begin{subfigure}[t]{0.48\textwidth}
			\centering
			\includegraphics[width=\linewidth]{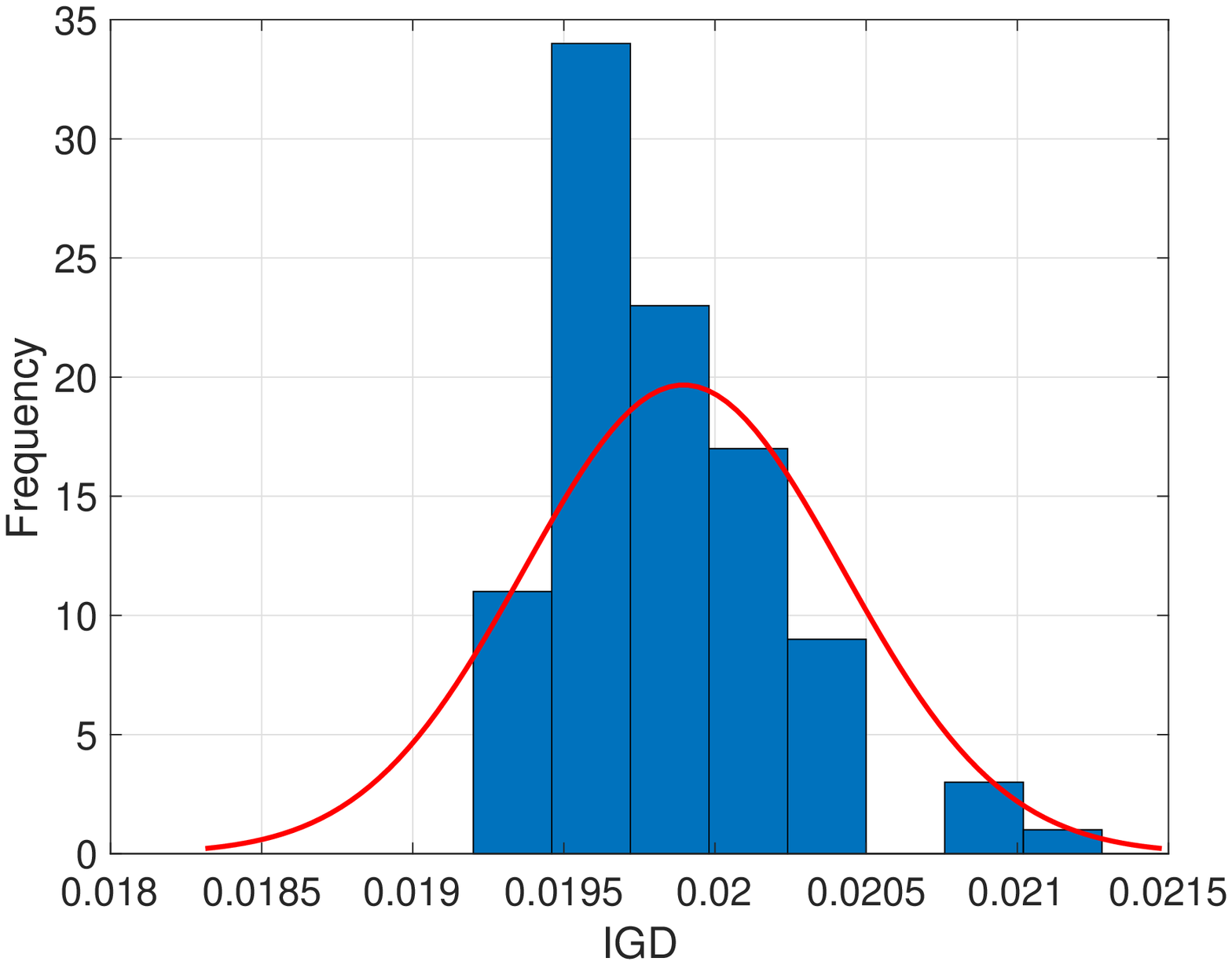}
			\caption{Sample distribution and normal distribution (in red)}
		\end{subfigure}
	\end{tabular}
	\caption{Relationship between sample size and the SEM of IGD and distribution of the IGD values from 100 runs of AREA on DTLZ1.}
	\label{fig:design}
	\vspace{-2mm}
\end{figure*}

All the algorithms are run 30 times for each test problem. This sample size meets the requirement for a non-parametric test like the Wilcoxon rank-sum test \cite{Wilcoxon45}. We further justify the sufficiency of this sample size by the method used in \cite{Rostami2015}. That is, AREA is performed 100 times on DTLZ1. The standard error of the mean (SEM) \cite{Rostami2015} of IGD is plotted against the sample size in Fig. \ref{fig:design}(a). The figure shows that SEM decreases slowly after a sample size of 30. The distribution of the 100 IGD values is shown in Fig. \ref{fig:design}(b), suggesting that the samples do not conform well to the normal distribution. Therefore, the non-parametric Wilcoxon rank-sum test \cite{Wilcoxon45} is used in this work to indicate significant differences between compared algorithms.

For fair comparison, all the algorithms should use the same termination condition. The termination condition is the maximum number of function evaluations (FEs). 20,000 FEs are used for all the problems except MOP1-5, where the number is set to 200,000 due to high optimisation difficulty in these problems. MOP6-7 use 300,000 FEs as it takes longer to get population converged toward their PFs. These termination conditions are based on existing studies \cite{DAPM02,LGZ14} and also take into account practical demands on runtime. In addition, WFG problems are given the same termination condition as in \cite{Cheng2017}.

\subsection{Performance Measures}
The following three measures are used for performance assessment:

\paragraph{Inverted Generational Distance (IGD)}
IGD \cite{LGZ14} accounts for both diversity and convergence of approximations to the PF. Let $PF$ be a set of solutions uniformly sampled from the true PF, and $PF^*$ be an approximate solution set in the objective space, the metric measures the gap 
between $PF^*$ and $PF$, calculated as follows:
\begin{equation}
IGD(PF^*,PF)=\frac{\sum_{p\in PF}{d(p,PF^*)}}{|PF|}
\label{eq:igd}
\end{equation}
where $d(p,PF^*)$ is the distance between the member $p$ of $PF$ and the nearest member in $PF^*$. The calculation of IGD in this paper uses a $PF$ of around 1000 points uniformly sampled from the true PF.

\paragraph{Hypervolume (HV)}
HV \cite{ZT99} calculates the size of the objective space dominated by the approximation set $S$ and bounded by a reference point $R=(R_1,\dots,R_M)^{T}$ that is dominated by all points on the PF. In this paper, $R$ is slightly worse than the nadir point by adding 0.1 to each dimension of it. The HV values presented in this paper are all normalised to $\left[0, 1\right]$ by dividing $\prod_{i=1}^{M}{R_i}$. 

\paragraph{Spacing (S)}
The S measure proposed by Schott \cite{Schott1995} is a performance indicator of solution uniformity. It is calculated by 
\begin{equation}
S=\sqrt{\frac{1}{|PF^*|}\sum_{p\in PF^*}{[d(p,PF^*)-\bar{d}]^2}}
\end{equation} 
where $d(p,PF^*)$ has the same definition as in Eq.~(\ref{eq:igd}), and $\bar{d}$ is the mean of $d(p,PF^*)$ for all $p$ in $PF^*$.

\section{Experimental Results}
This section is devoted to reporting the results of comparison between AREA and other algorithms on a variety of test problems. All the test problems except UF and WFG are somewhat ZDT/DTLZ variants. UF and WFG have different problem characteristics. For example, UF problems have strong variable linkages, and WFG problems involve nonseparability and scalability. Therefore, The following discusses them separately.

\begin{table*}[t]
	\footnotesize
	\addtolength{\tabcolsep}{-3.5pt}
	\centering
	\caption{Mean (and standard deviation) values of IGD obtained by five algorithms. Best (smallest mean) values are highlighted in boldface. Algorithms that are better than, worse than, or equivalent to AREA on ranksum test with a significance level of 0.05 are indicated by $+$, $-$, or $\approx$, respectively.}
	\begin{tabular}{ccccccc}
		\toprule
		Problem&$M$&MOEA/D&ANSGA-III&SPEA/R&RVEA&AREA\\
		\midrule
		\multirow{1}{*}{DTLZ1}&3&\hl{1.9455e-2 (6.26e-4) $+$}&2.4140e-2 (6.56e-3) $-$&2.7387e-2 (1.47e-2) $-$&1.9589e-2 (7.57e-4) $+$&2.0303e-2 (5.59e-4)\\
		\hline
		\multirow{1}{*}{DTLZ2}&3&5.0856e-2 (3.23e-4) $+$&5.2590e-2 (1.52e-3) $\approx$&\hl{5.0334e-2 (2.19e-5) $+$}&5.0424e-2 (2.18e-4) $+$&5.2651e-2 (4.95e-4)\\
		\hline
		\multirow{1}{*}{DTLZ5}&3&1.8581e-2 (9.82e-6) $-$&9.3907e-3 (8.52e-4) $-$&3.0361e-2 (1.59e-3) $-$&7.1990e-2 (1.03e-2) $-$&\hl{4.1568e-3 (9.44e-5)}\\
		\hline
		\multirow{1}{*}{DTLZ7}&3&2.2484e-1 (1.53e-1) $-$&7.0095e-2 (2.89e-3) $-$&8.5436e-2 (2.05e-3) $-$&1.0412e-1 (4.60e-3) $-$&\hl{5.6225e-2 (1.47e-3)}\\
		\hline
		\multirow{1}{*}{IDTLZ1}&3&3.2832e-2 (3.09e-4) $-$&2.1707e-2 (2.45e-3) $\approx$&1.1479e-1 (7.27e-2) $-$&4.8517e-2 (1.85e-2) $-$&\hl{2.1485e-2 (2.05e-3)}\\
		\hline
		\multirow{1}{*}{IDTLZ2}&3&9.7814e-2 (4.02e-4) $-$&7.0362e-2 (4.16e-3) $-$&7.5515e-2 (7.49e-4) $-$&7.5553e-2 (1.12e-3) $-$&\hl{5.2069e-2 (5.01e-4)}\\
		\hline
		\multirow{1}{*}{SDTLZ2}&3&1.3968e-1 (4.25e-3) $-$&1.3168e-1 (2.67e-3) $-$&1.2007e-1 (1.44e-4) $-$&1.2021e-1 (2.88e-4) $-$&\hl{1.1792e-1 (1.25e-3)}\\
		\hline
		\multirow{1}{*}{CDTLZ2}&3&4.3339e-2 (4.25e-5) $-$&4.8019e-2 (3.63e-3) $-$&4.3005e-2 (1.03e-3) $-$&3.8184e-2 (6.84e-4) $-$&\hl{3.3358e-2 (7.83e-4)}\\
		\hline
		\multirow{1}{*}{MOP1}&2&3.4925e-1 (5.92e-3) $-$&1.6972e-1 (1.07e-1) $-$&\hl{9.6154e-3 (1.20e-3) $+$}&4.6443e-2 (3.08e-3) $-$&1.0964e-2 (2.65e-4)\\
		\hline
		\multirow{1}{*}{MOP2}&2&3.5375e-1 (6.52e-3) $-$&2.5475e-1 (3.92e-2) $-$&\hl{2.3403e-2 (4.40e-2) $+$}&8.6316e-2 (2.38e-2) $\approx$&1.1691e-1 (8.64e-2)\\
		\hline
		\multirow{1}{*}{MOP3}&2&5.3701e-1 (8.60e-2) $-$&4.0667e-1 (4.28e-2) $-$&3.0235e-2 (4.61e-2) $\approx$&1.3239e-1 (5.95e-2) $-$&\hl{1.3660e-2 (3.33e-2)}\\
		\hline
		\multirow{1}{*}{MOP4}&2&2.7318e-1 (1.39e-2) $-$&2.4355e-1 (2.55e-2) $-$&\hl{7.9912e-3 (3.13e-3) $+$}&3.4180e-2 (1.72e-2) $-$&2.1709e-2 (3.37e-2)\\
		\hline
		\multirow{1}{*}{MOP5}&2&3.1330e-1 (1.10e-2) $-$&2.1310e-1 (6.82e-2) $-$&2.2083e-2 (1.78e-3) $-$&7.8166e-1 (3.60e-2) $-$&\hl{1.5858e-2 (7.78e-4)}\\
		\hline
		\multirow{1}{*}{MOP6}&3&3.0983e-1 (1.88e-8) $-$&2.7121e-1 (3.25e-2) $-$&2.5371e-1 (3.88e-2) $-$&\hl{1.3391e-1 (1.02e-2) $+$}&2.0143e-1 (4.42e-2)\\
		\hline
		\multirow{1}{*}{MOP7}&3&3.5765e-1 (1.84e-8) $-$&3.5410e-1 (1.58e-2) $-$&3.3627e-1 (2.49e-2) $\approx$&\hl{2.1851e-1 (1.08e-2) $+$}&3.3486e-1 (2.72e-2)\\
		\hline
		\multirow{1}{*}{F1}&2&2.4318e-2 (2.35e-3) $-$&4.1538e-2 (1.60e-2) $-$&2.5016e-2 (2.26e-3) $-$&2.1641e-2 (1.29e-3) $-$&\hl{4.6459e-3 (3.16e-5)}\\
		\hline
		\multirow{1}{*}{F2}&2&7.5871e-1 (1.38e-1) $-$&2.0699e-1 (2.18e-1) $-$&\hl{4.7002e-3 (1.11e-6) $\approx$}&2.0577e-2 (4.57e-3) $-$&5.1982e-3 (1.04e-3)\\
		\hline
		\multirow{1}{*}{F3}&2&9.1038e-2 (1.30e-1) $-$&8.0237e-2 (8.03e-2) $-$&4.0067e-2 (4.49e-4) $-$&3.8267e-2 (1.02e-3) $-$&\hl{3.1088e-2 (1.83e-4)}\\
		\hline
		\multirow{1}{*}{F4}&2&5.2306e-1 (1.91e-1) $-$&4.1481e-1 (4.56e-2) $-$&3.7015e-1 (2.14e-4) $\approx$&3.7219e-1 (1.32e-3) $-$&\hl{3.7011e-3 (3.86e-5)}\\
		\hline
		\multirow{1}{*}{F5}&3&5.1104e-2 (8.57e-4) $-$&4.8516e-2 (9.13e-4) $+$&\hl{4.8215e-2 (7.88e-5) $+$}&5.6043e-2 (2.05e-3) $-$&4.9318e-2 (1.09e-3)\\
		\hline
		\multirow{1}{*}{F6}&3&1.0757e-1 (7.50e-2) $-$&2.7622e-2 (2.09e-3) $\approx$&3.4320e-2 (3.44e-3) $-$&6.7625e-2 (7.76e-3) $-$&\hl{2.7028e-2 (2.86e-3)}\\
		\hline
		\multirow{1}{*}{F7}&3&8.9180e-2 (9.47e-3) $\approx$&9.8225e-2 (1.33e-2) $-$&\hl{8.4542e-2 (6.63e-3) $\approx$}&1.0226e-1 (1.54e-2) $-$&8.5116e-2 (5.93e-3)\\
		\hline
		\multirow{1}{*}{F8}&3&3.9632e-1 (1.09e-2) $-$&2.6829e-1 (2.06e-2) $-$&2.9981e-1 (1.86e-2) $-$&5.1124e-1 (5.96e-2) $-$&\hl{1.9482e-1 (4.43e-3)}\\
		\hline
		\multicolumn{2}{c}{$+/-/\approx$}&2/20/1&1/19/3&5/13/5&4/18/1&\\
		\bottomrule
	\end{tabular}
	\label{tab:all_igd}
	\vspace{-2mm}
\end{table*}

\subsection{Results on ZDT/DTLZ Variants}
\begin{figure*}[t]
	\centering
	\begin{tabular}{@{}c@{}c@{}c@{}c@{}c}
		\includegraphics[width=0.2\linewidth]{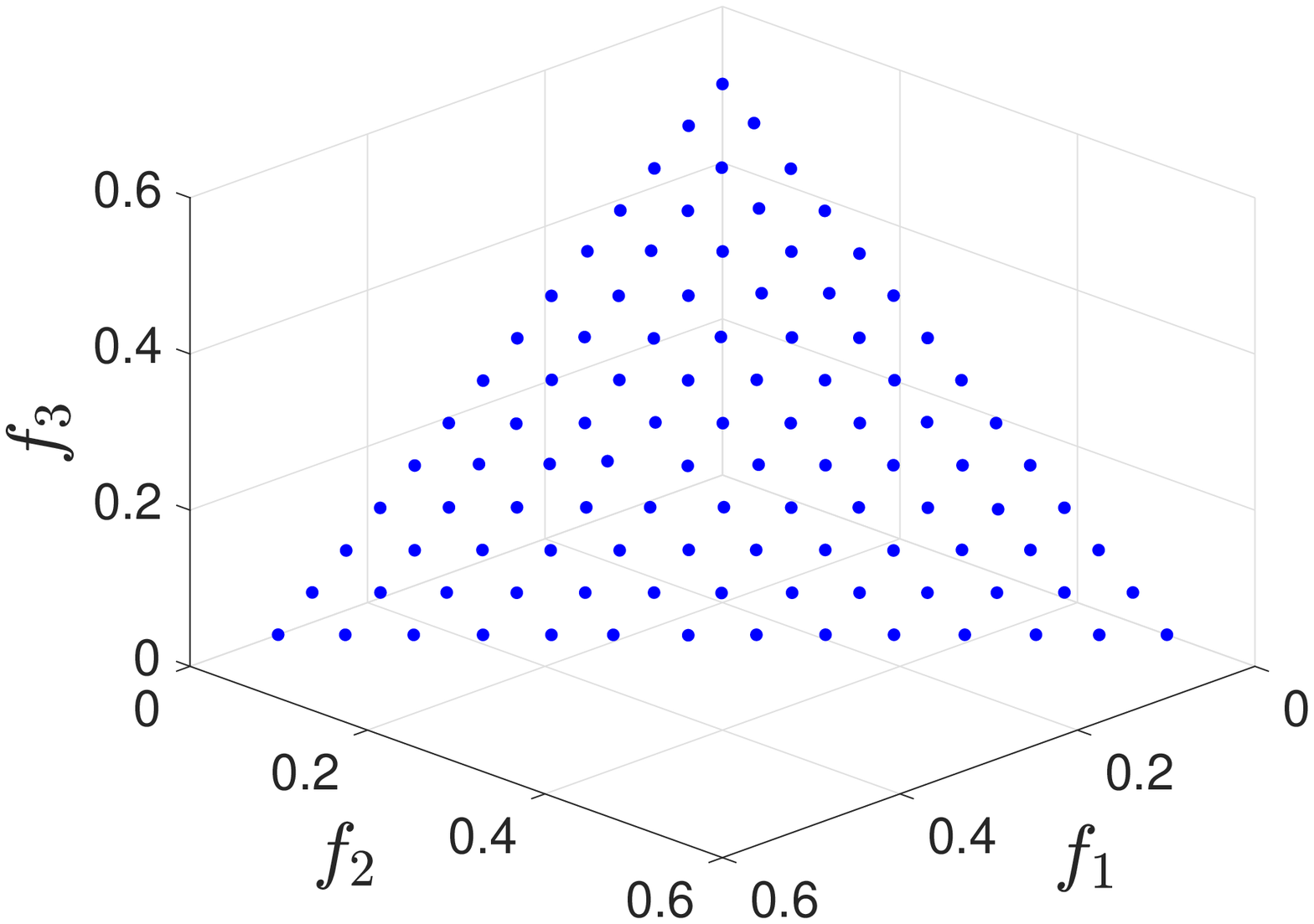}&
		\includegraphics[width=0.2\linewidth]{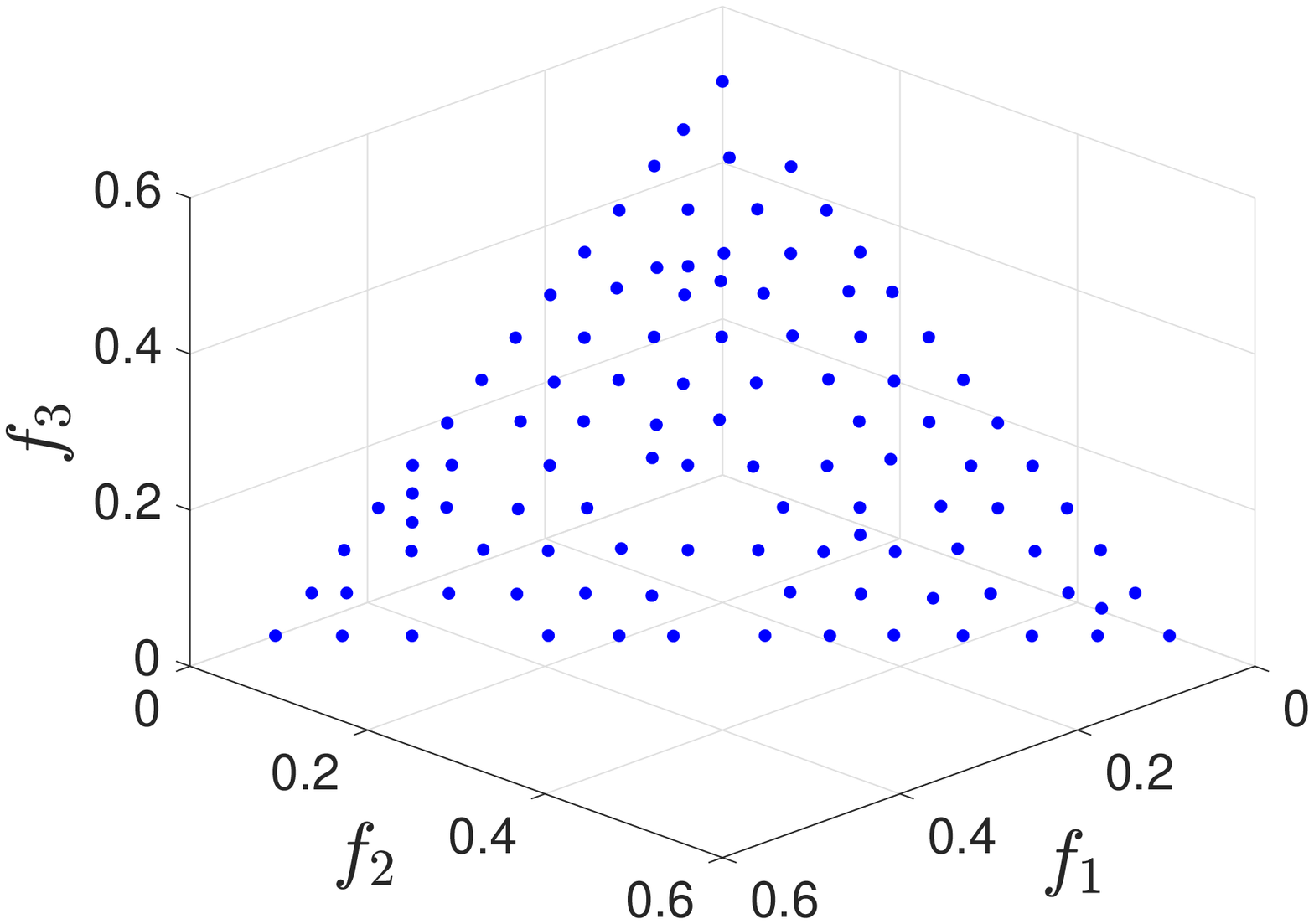}&
		\includegraphics[width=0.2\linewidth]{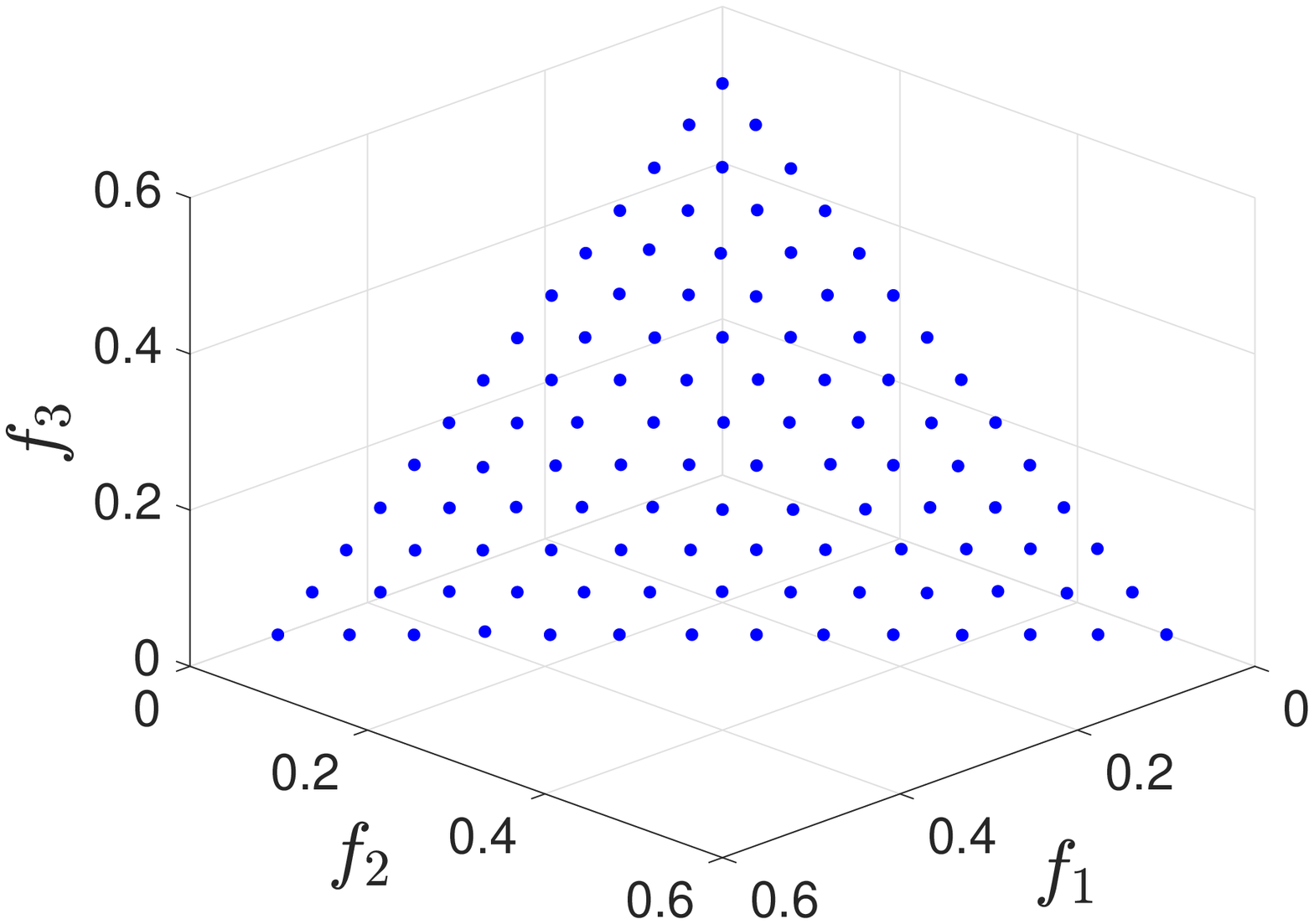}&
		\includegraphics[width=0.2\linewidth]{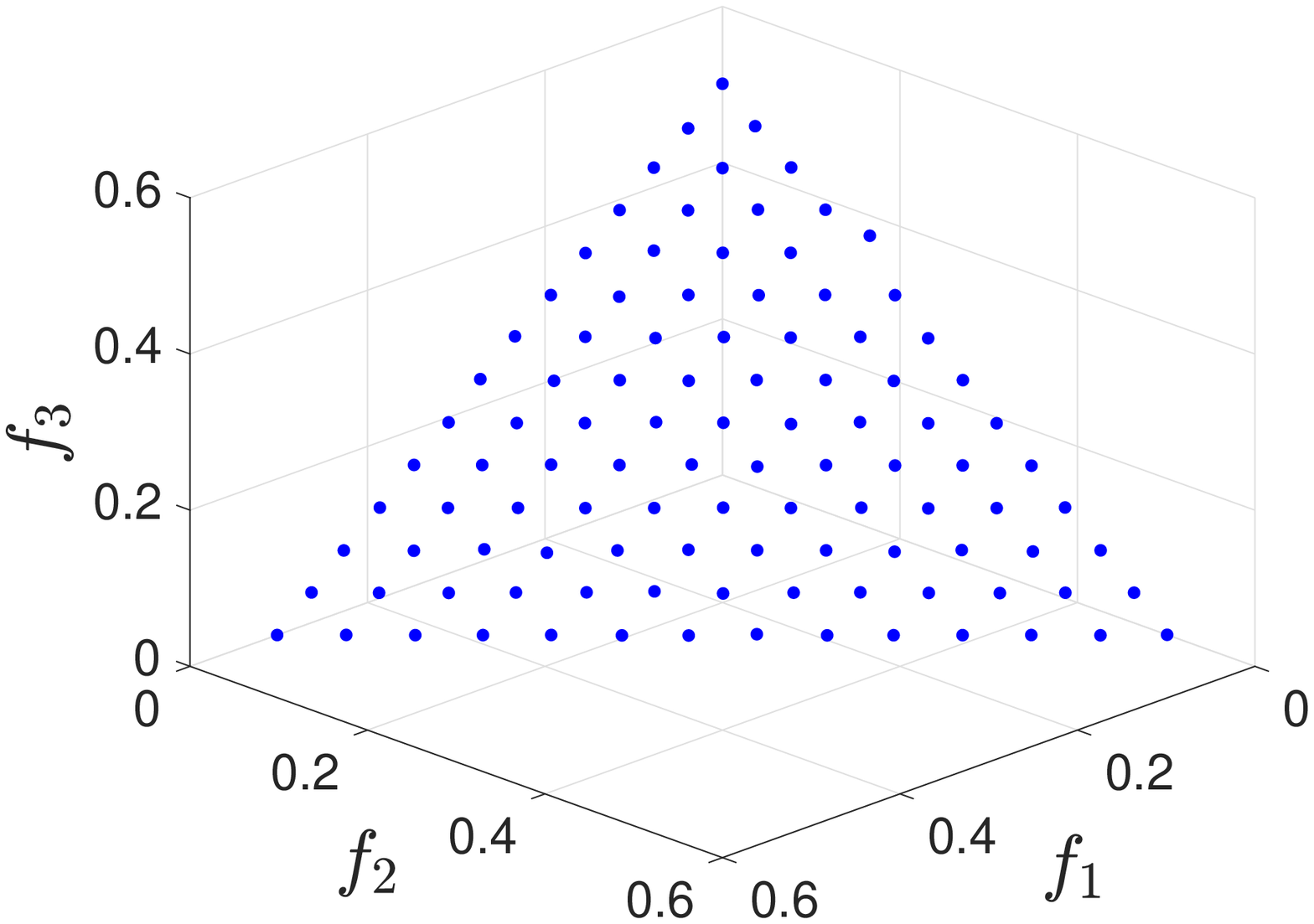}&
		\includegraphics[width=0.2\linewidth]{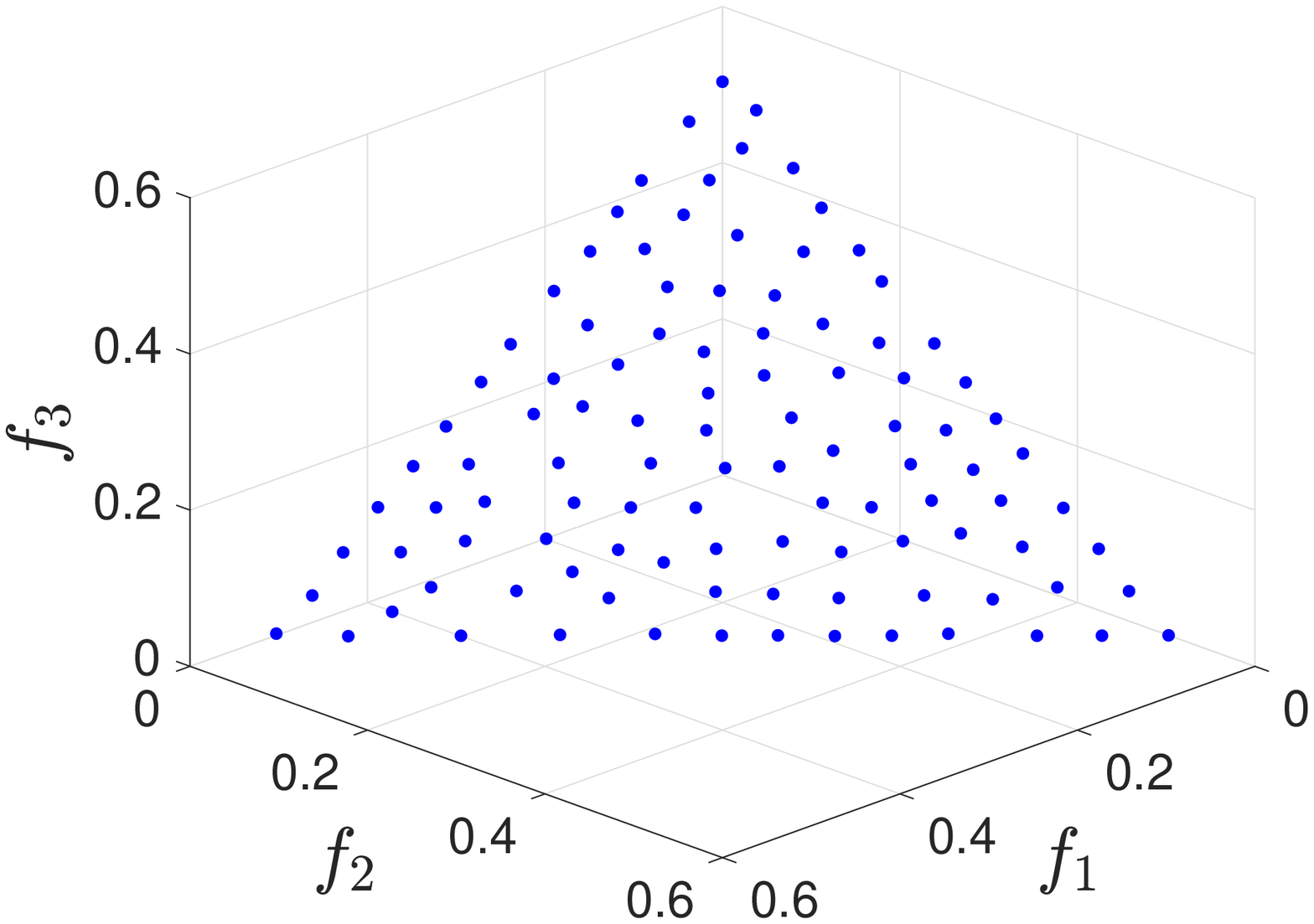}\\
		(a) MOEA/D & (b) ANSGA-III & (c) SPEA/R &(d) RVEA & (e) AREA\\[-2mm]
	\end{tabular}
	\caption{PF approximation of DTLZ1 obtained by different algorithms.}
	\label{fig:dtlz1_pf}
	\vspace{-2mm}
\end{figure*}

\begin{figure*}[!tp]
	\centering
	\begin{tabular}{@{}c@{}c@{}c@{}c@{}c}
		\includegraphics[width=0.2\linewidth]{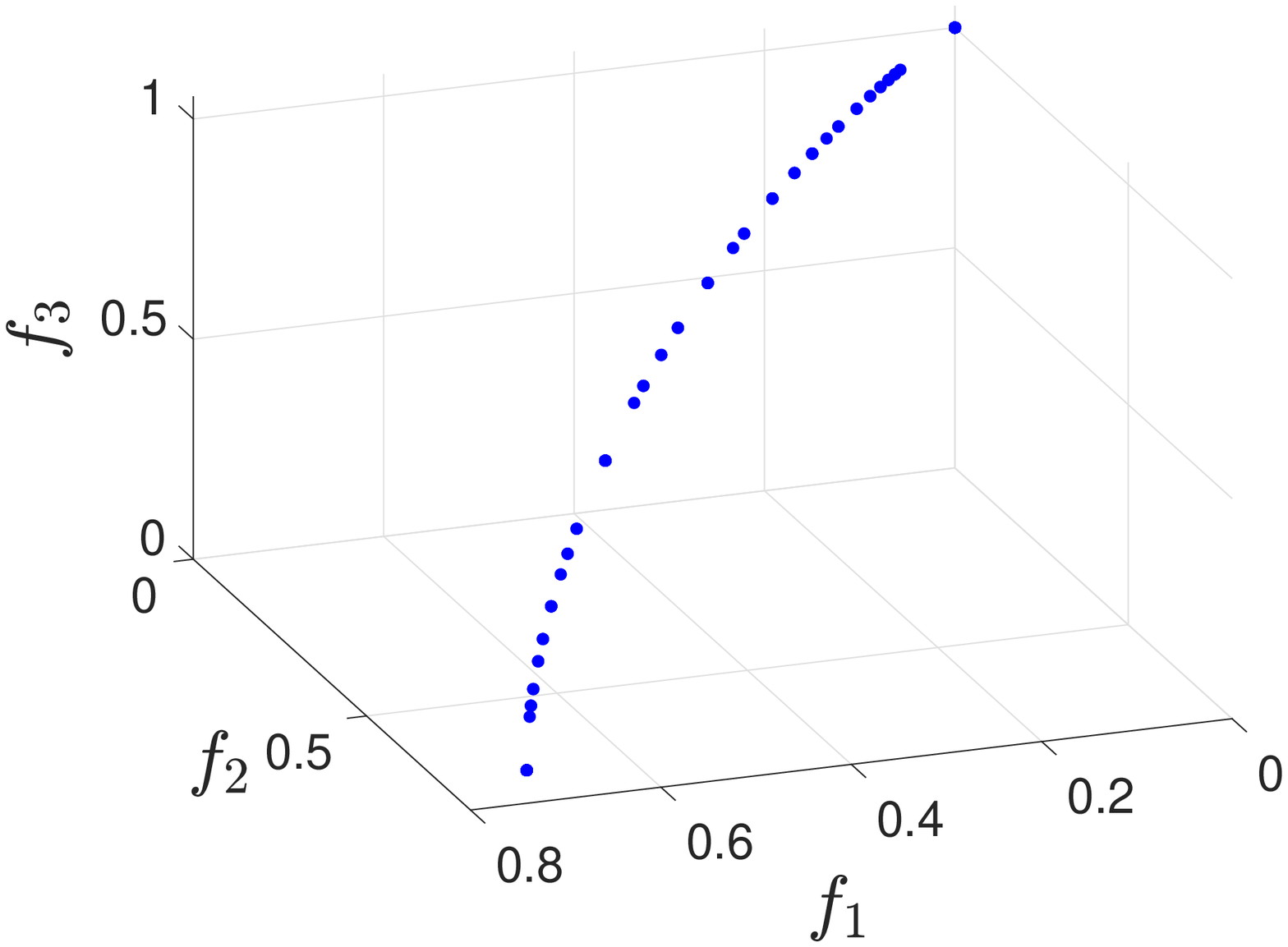}&
		\includegraphics[width=0.2\linewidth]{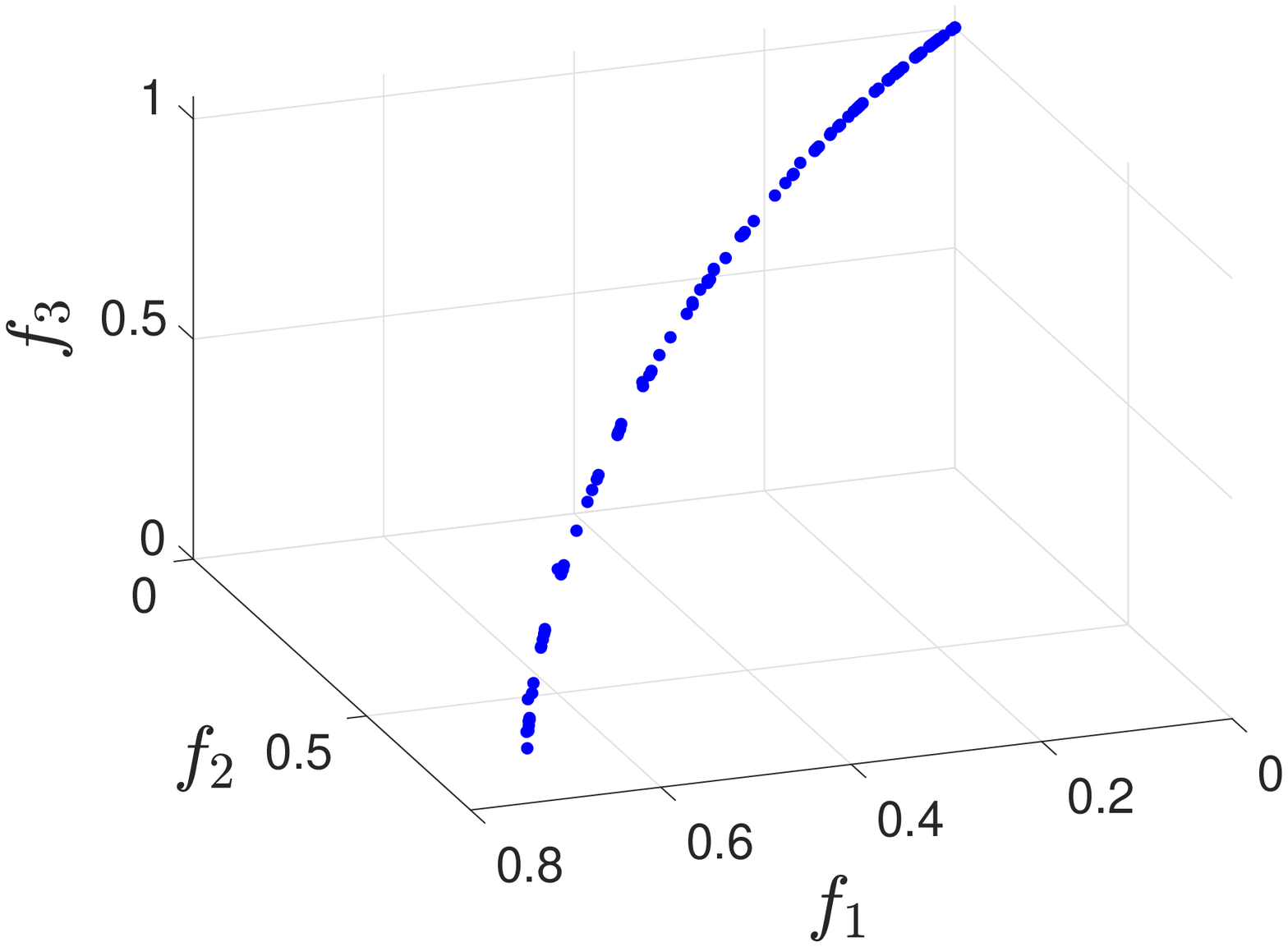}&
		\includegraphics[width=0.2\linewidth]{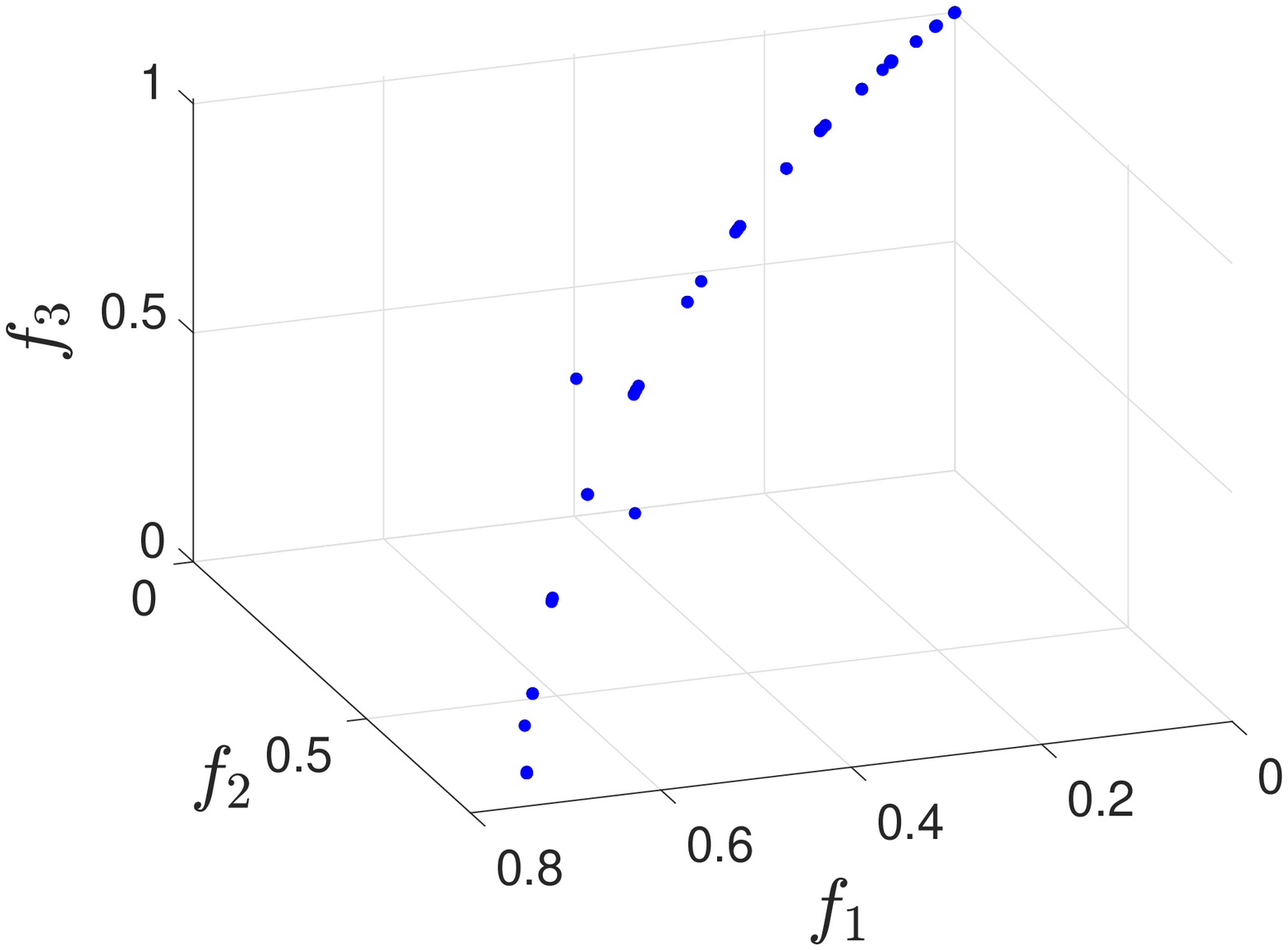}&
		\includegraphics[width=0.2\linewidth]{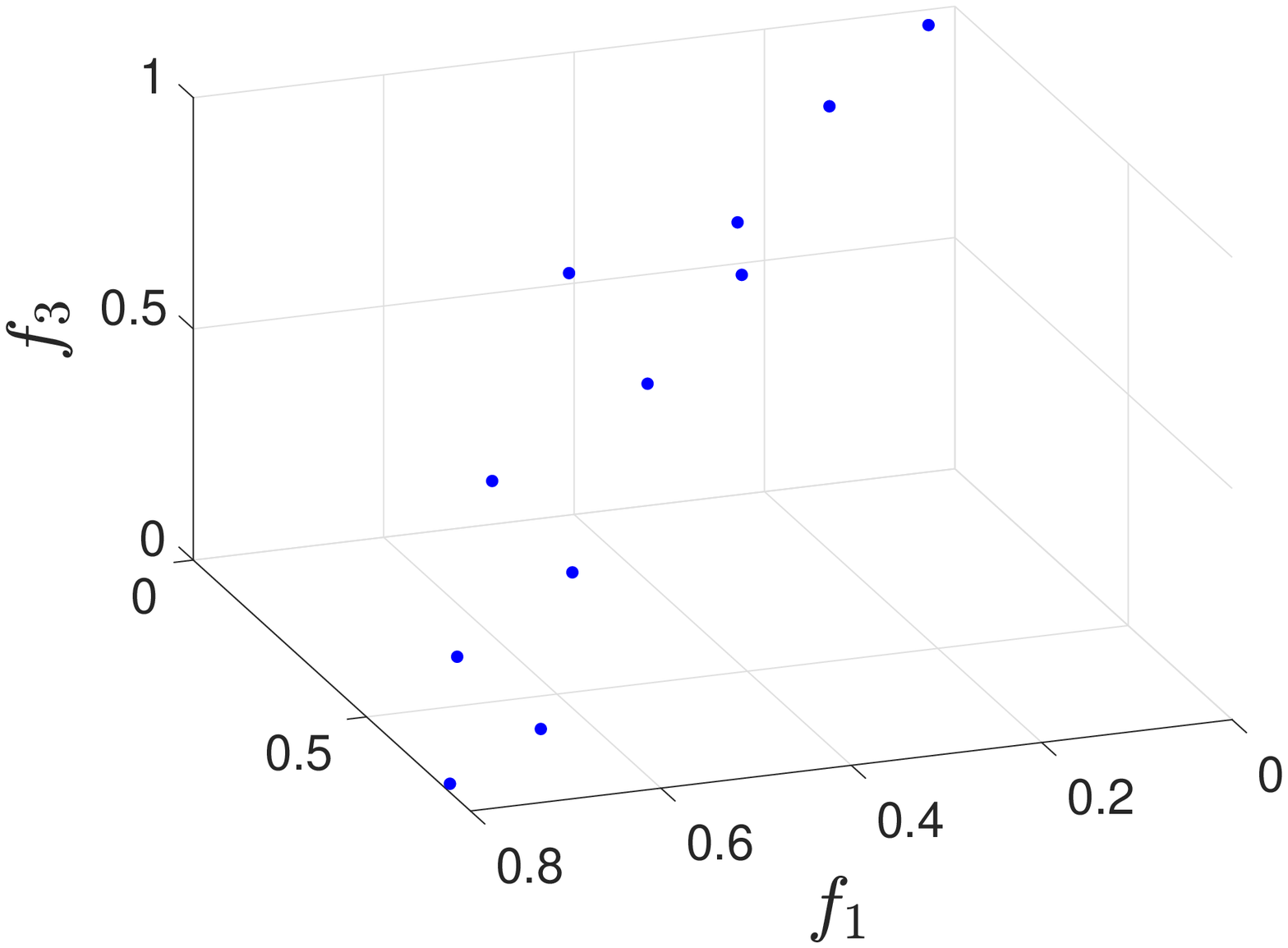}&
		\includegraphics[width=0.2\linewidth]{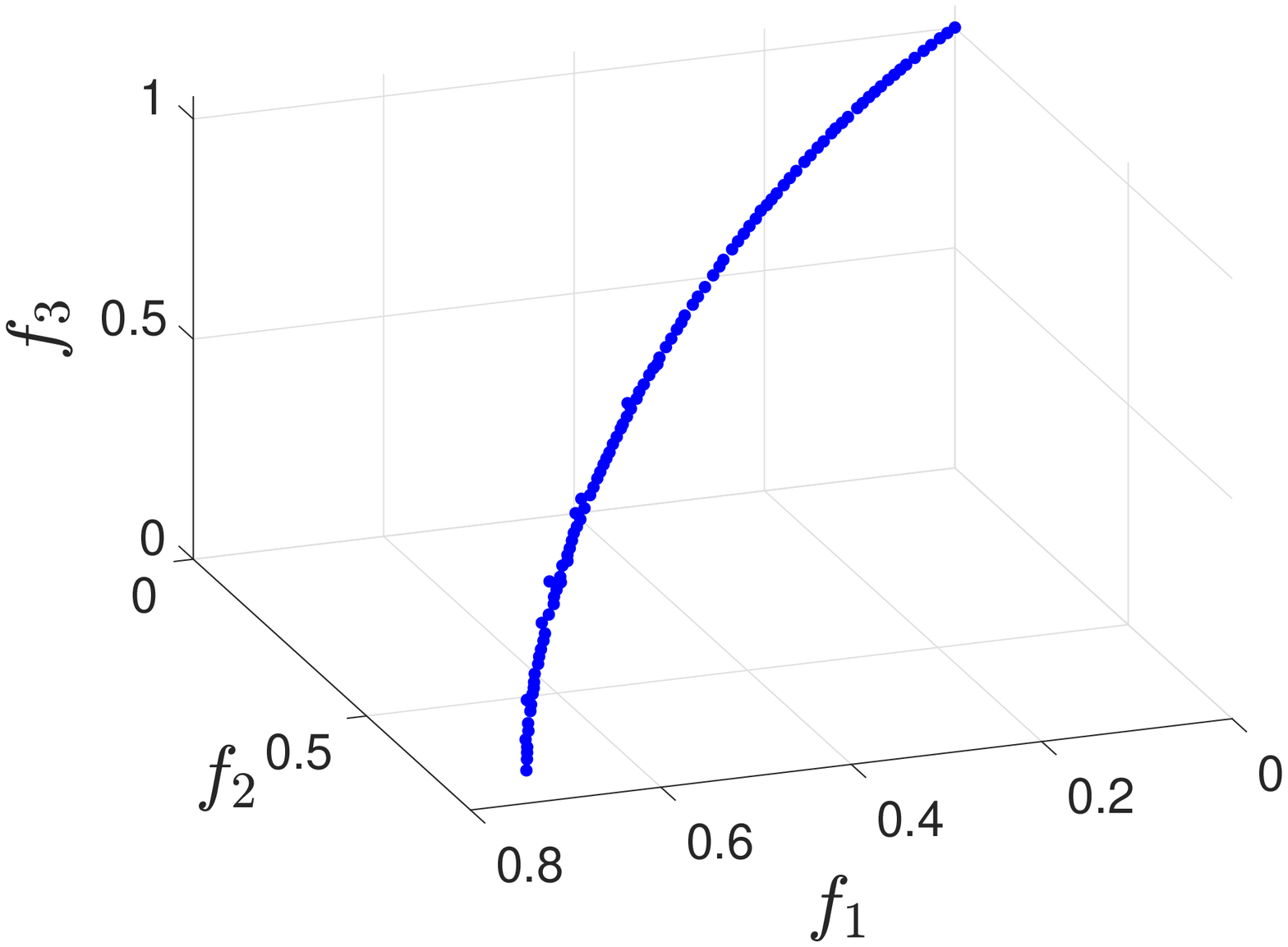}\\
		(a) MOEA/D & (b) ANSGA-III & (c) SPEA/R & (d) RVEA & (e) AREA\\
	\end{tabular}
	\caption{PF approximation of DTLZ5 obtained by different algorithms.}
	\label{fig:dtlz5_pf}
	\vspace{-2mm}
\end{figure*}

\begin{figure*}[!tp]
	\centering
	\begin{tabular}{@{}c@{}c@{}c@{}c@{}c}
		\includegraphics[width=0.2\linewidth]{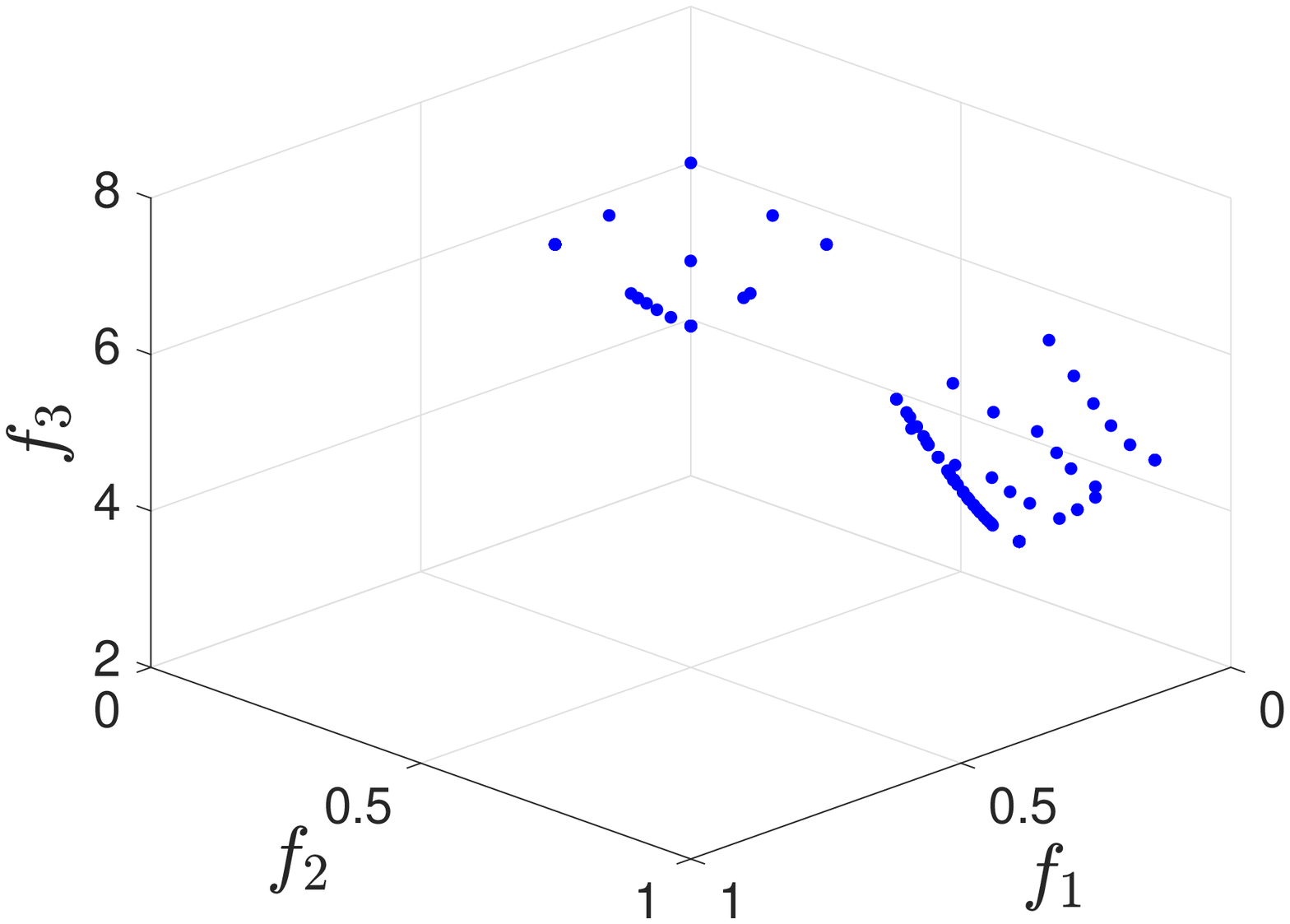}&
		\includegraphics[width=0.2\linewidth]{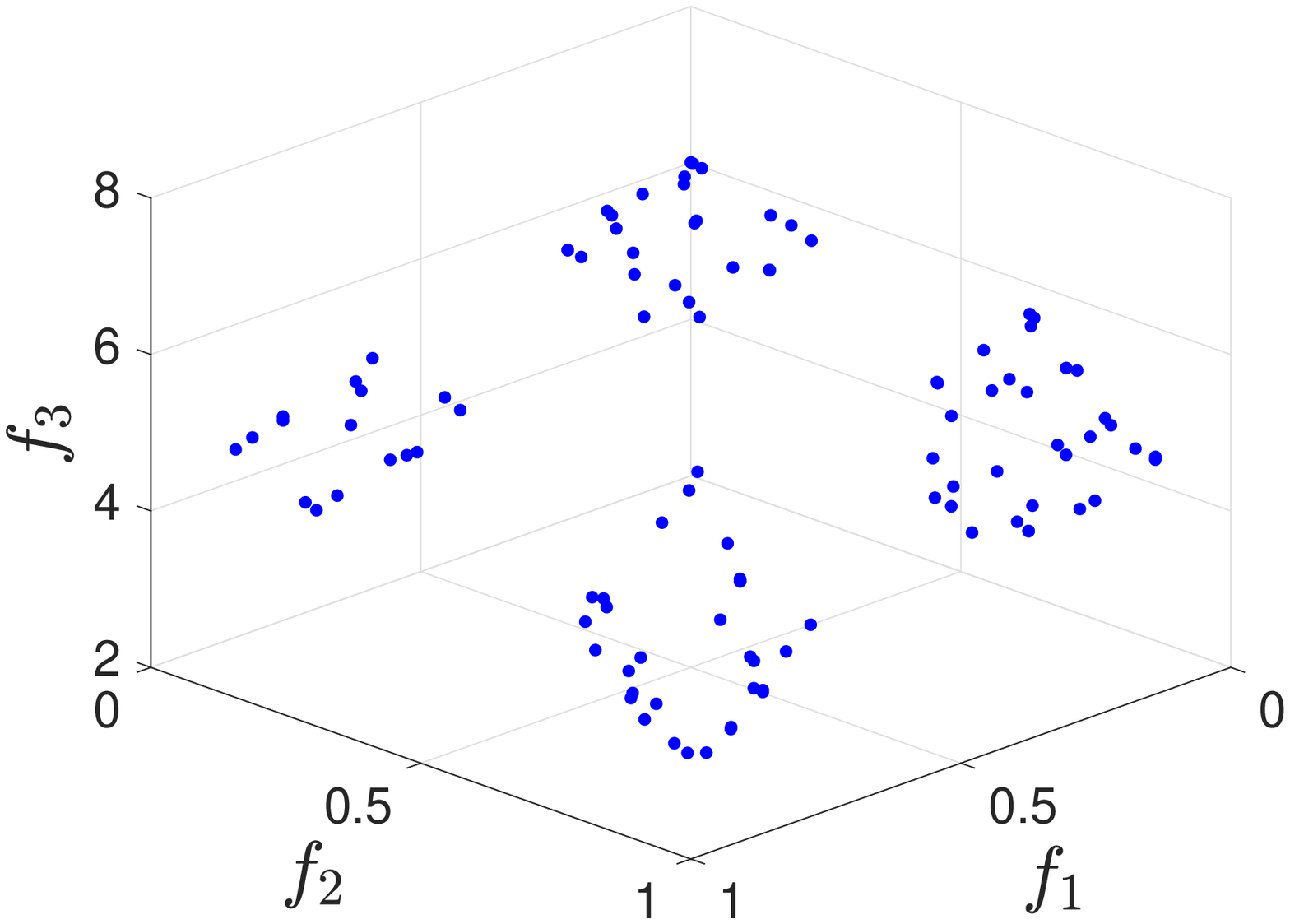}&
		\includegraphics[width=0.2\linewidth]{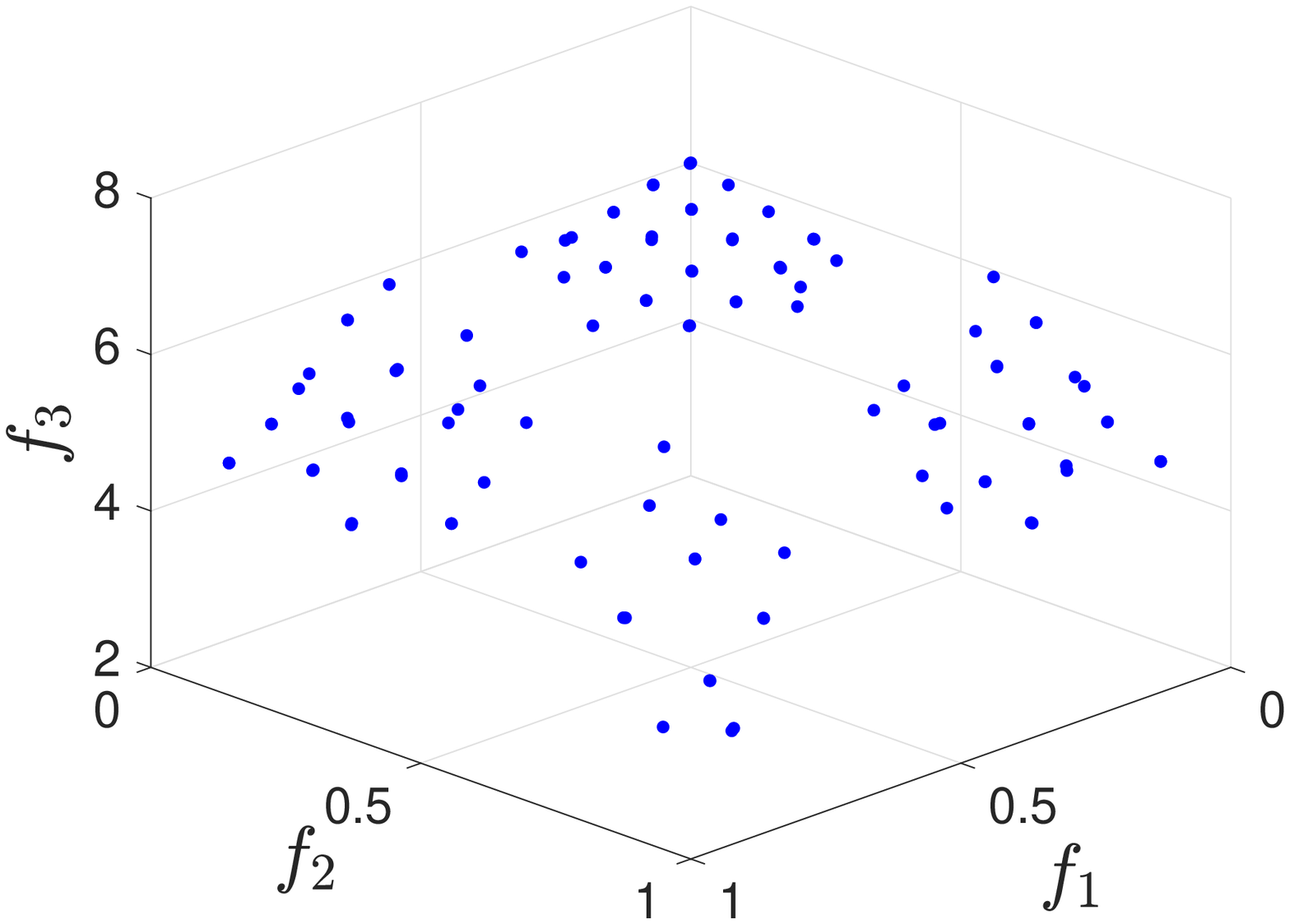}&
		\includegraphics[width=0.2\linewidth]{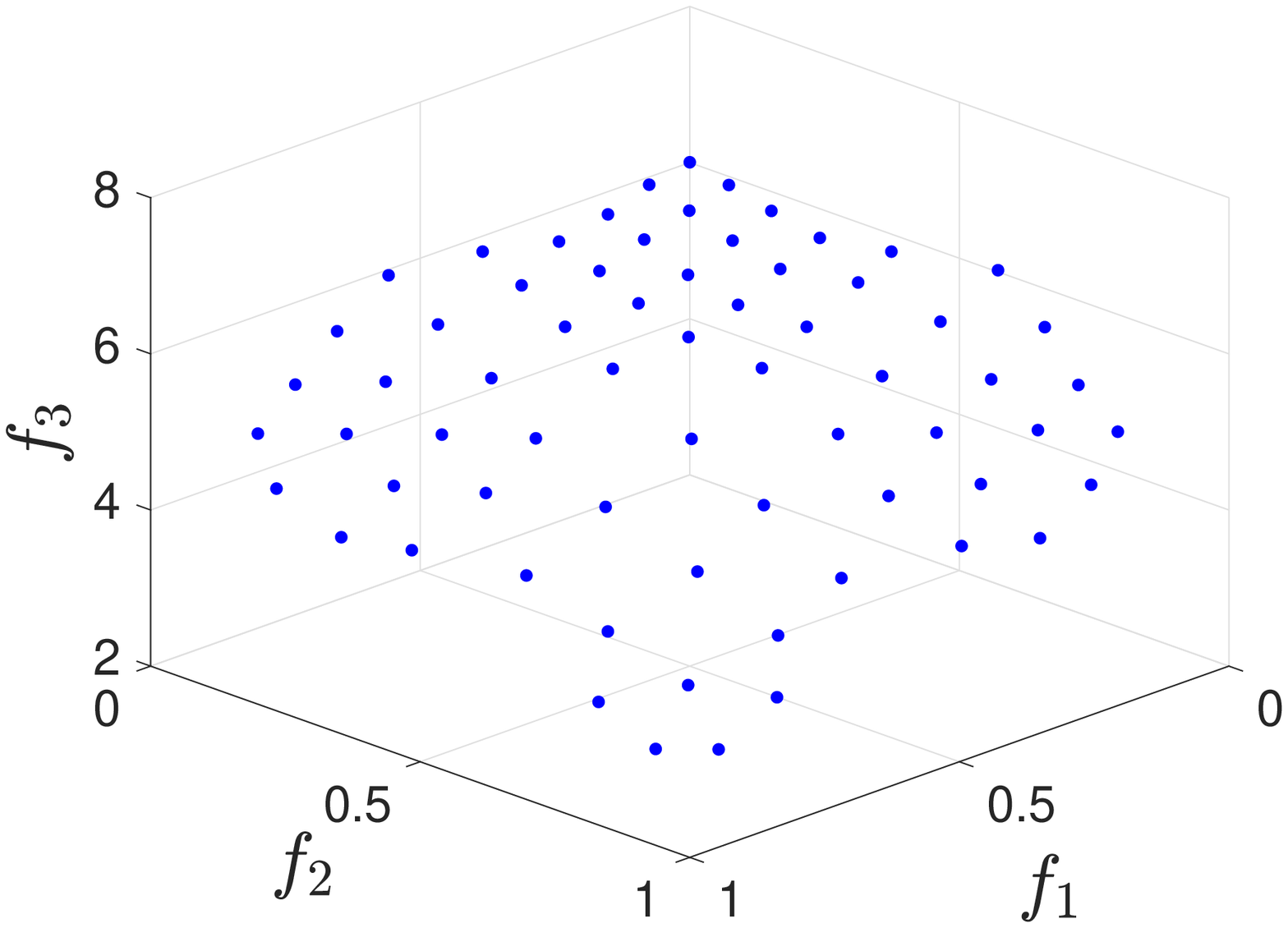}&
		\includegraphics[width=0.2\linewidth]{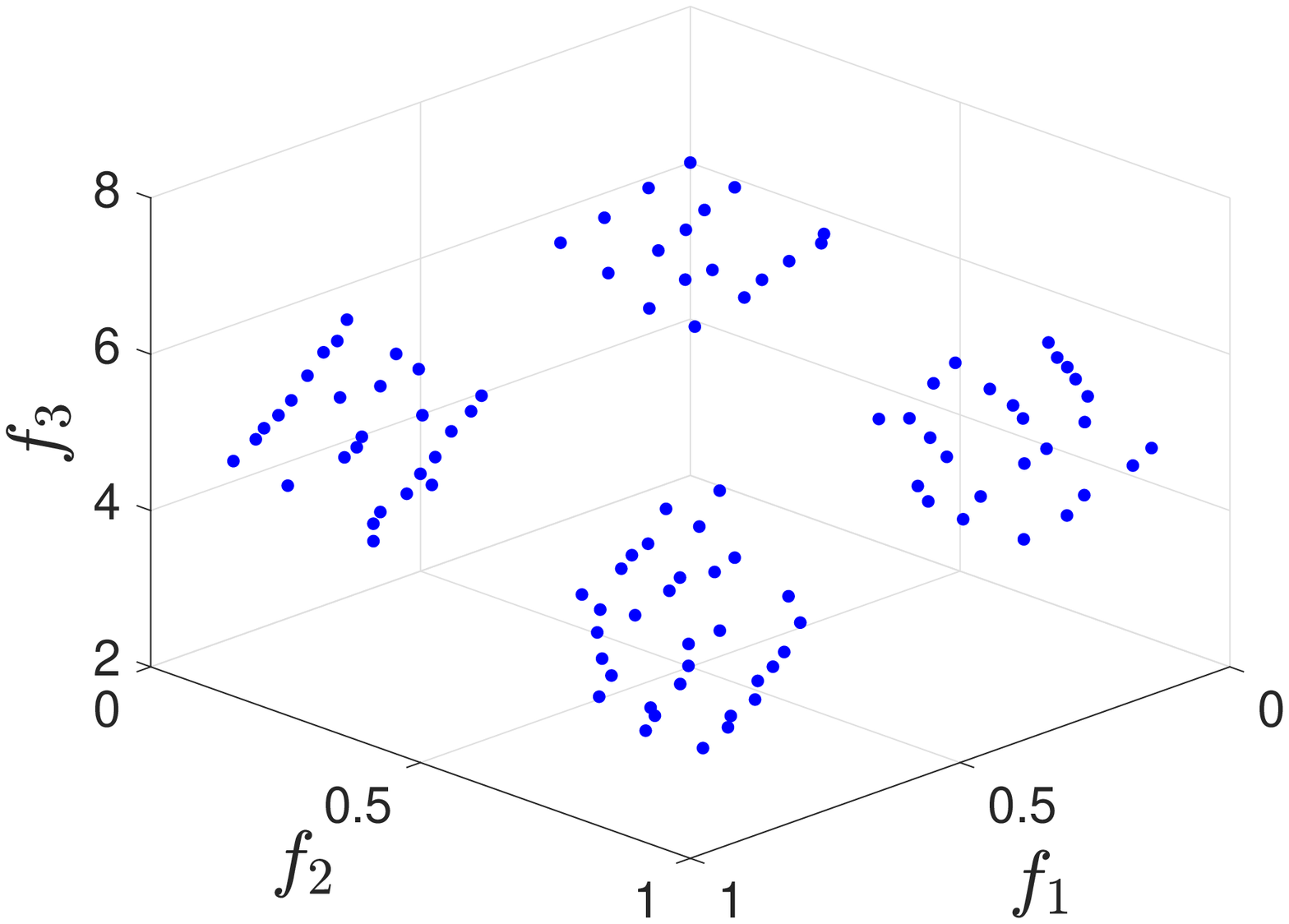}\\
		(a) MOEA/D & (b) ANSGA-III & (c) SPEA/R & (d) RVEA & (e) AREA\\
	\end{tabular}
	\caption{PF approximation of DTLZ7 obtained by different algorithms.}
	\label{fig:dtlz7_pf}
	\vspace{-2mm}
\end{figure*}

\begin{figure*}[!tp]
	\centering
	\begin{tabular}{@{}c@{}c@{}c@{}c@{}c}
		\includegraphics[width=0.2\linewidth]{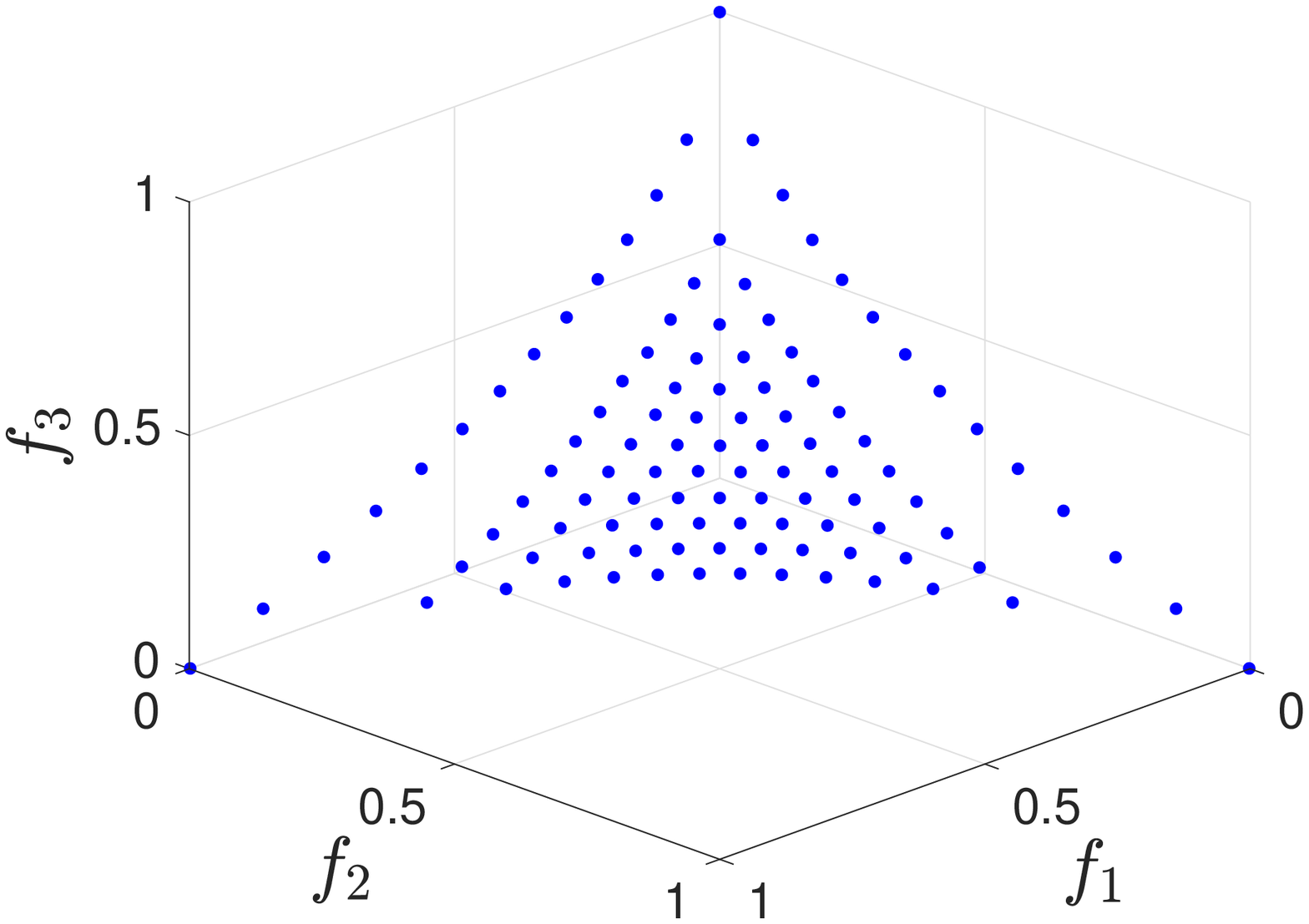}&
		\includegraphics[width=0.2\linewidth]{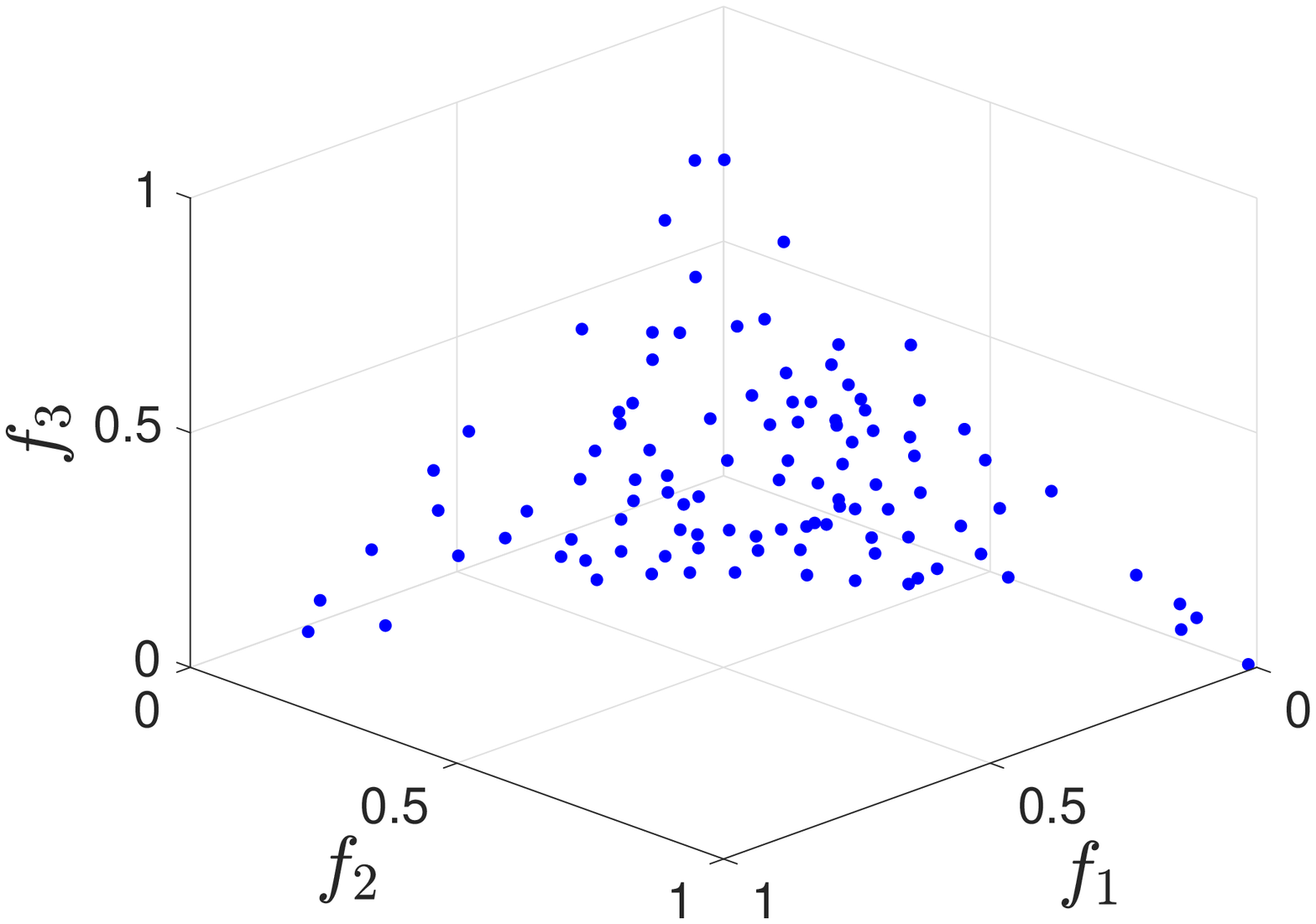}&
		\includegraphics[width=0.2\linewidth]{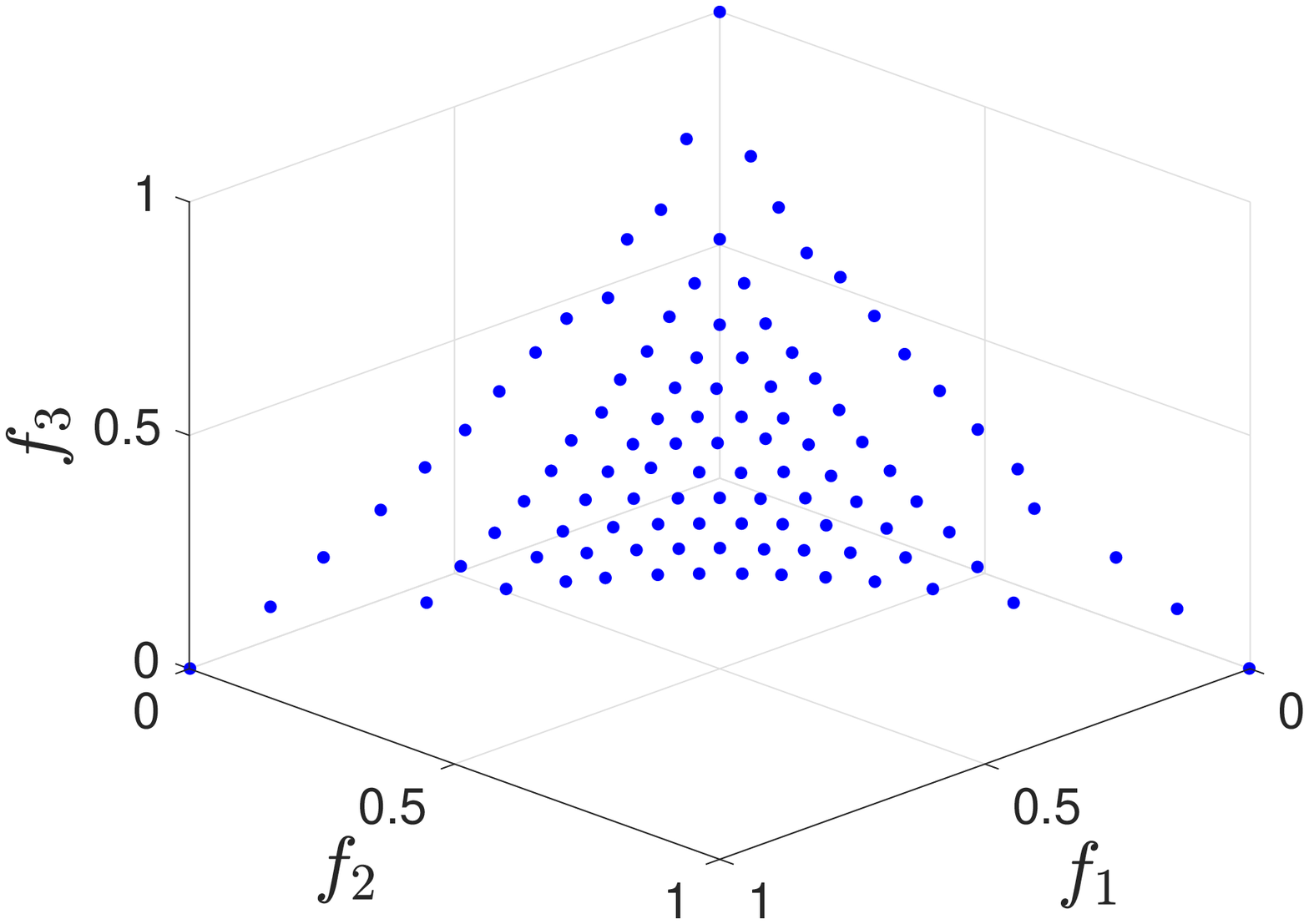}&
		\includegraphics[width=0.2\linewidth]{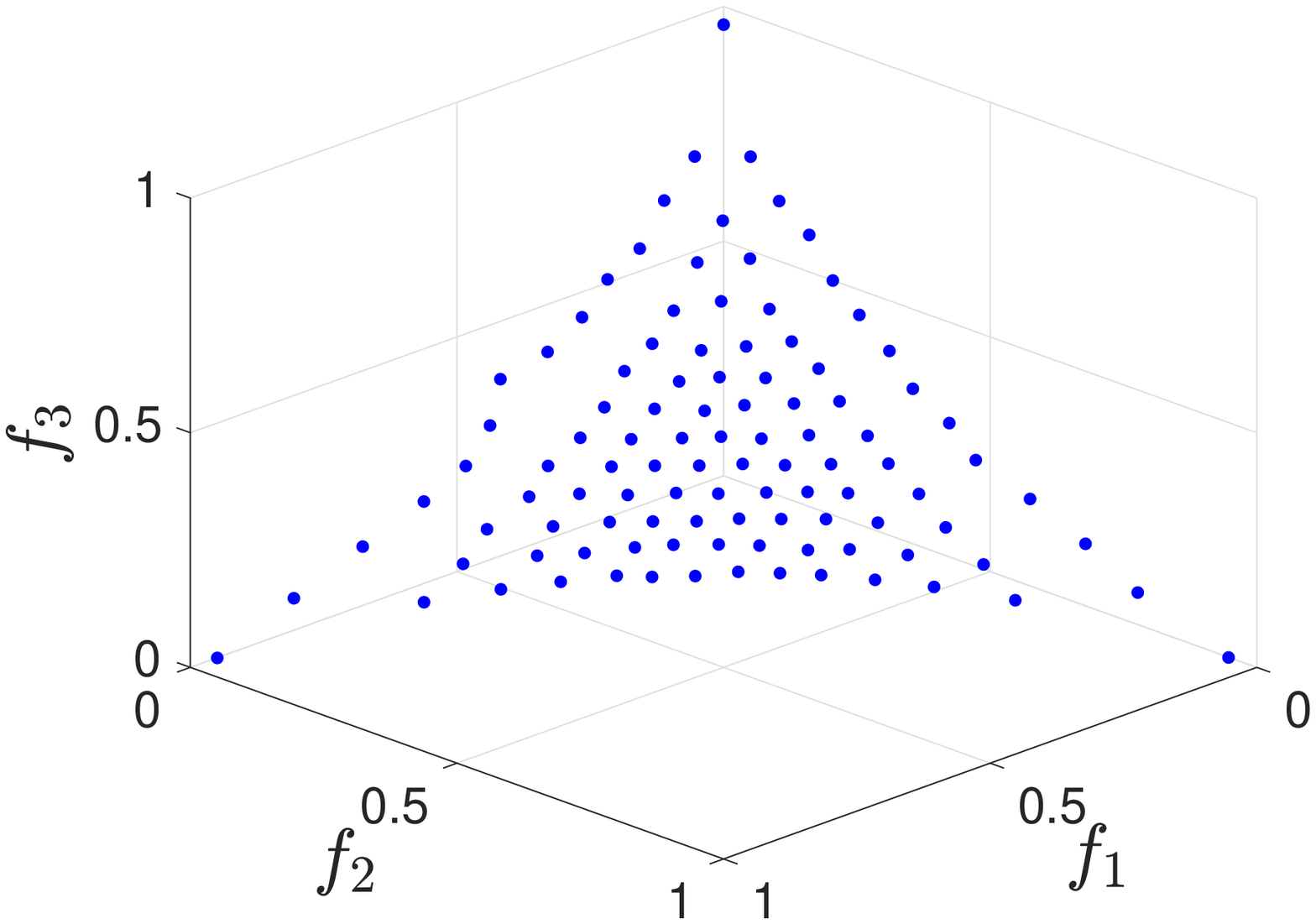}&
		\includegraphics[width=0.2\linewidth]{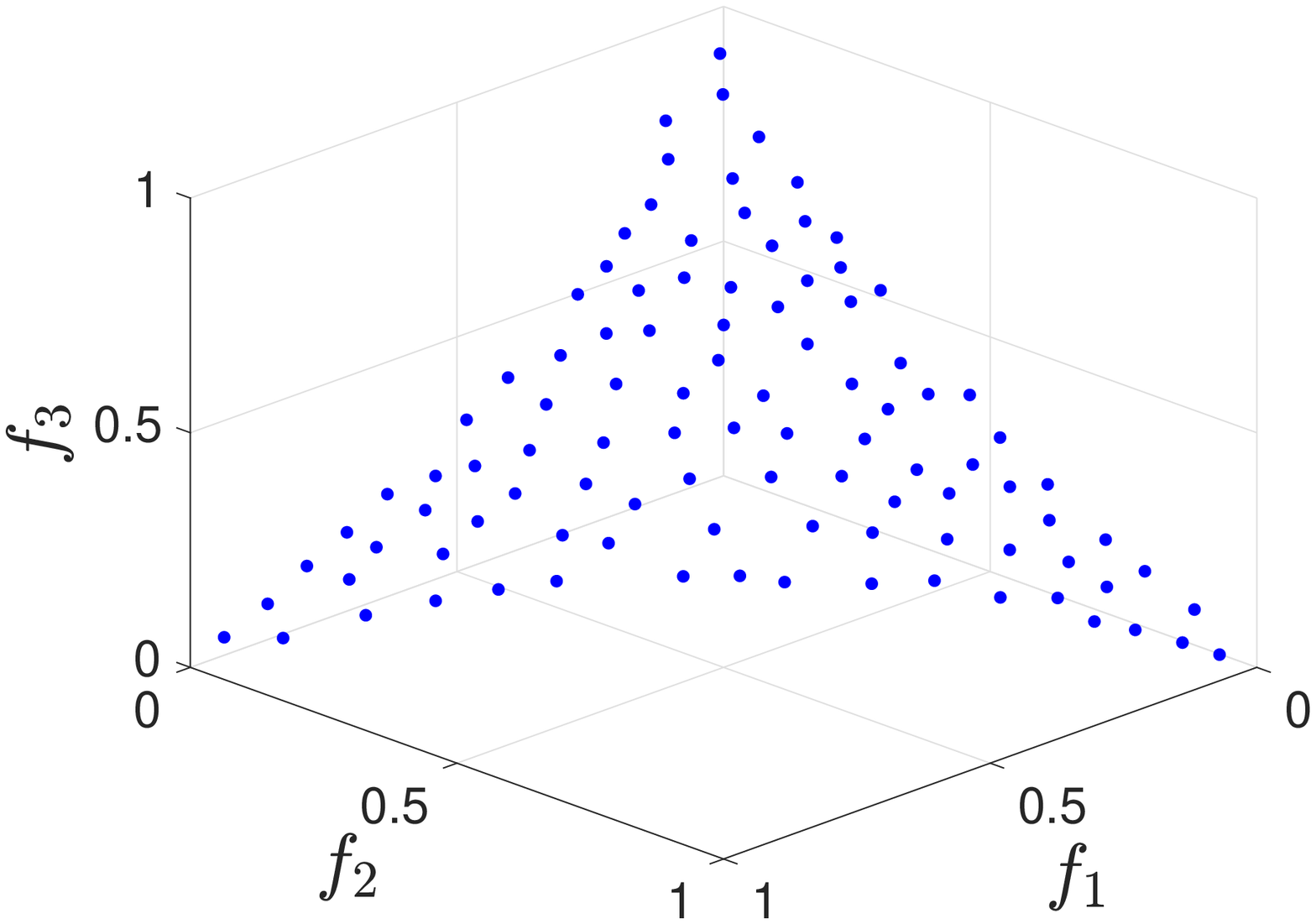}\\
		(a) MOEA/D & (b) ANSGA-III & (c) SPEA/R &(d) RVEA & (e) AREA\\
	\end{tabular}
	\caption{PF approximation of CDTLZ2 obtained by different algorithms.}
	\label{fig:cdtlz2_pf}
	\vspace{-2mm}
\end{figure*}

\begin{figure*}[!tp]
	\centering
	\begin{tabular}{@{}c@{}c@{}c@{}c@{}c}
		\includegraphics[width=0.2\linewidth]{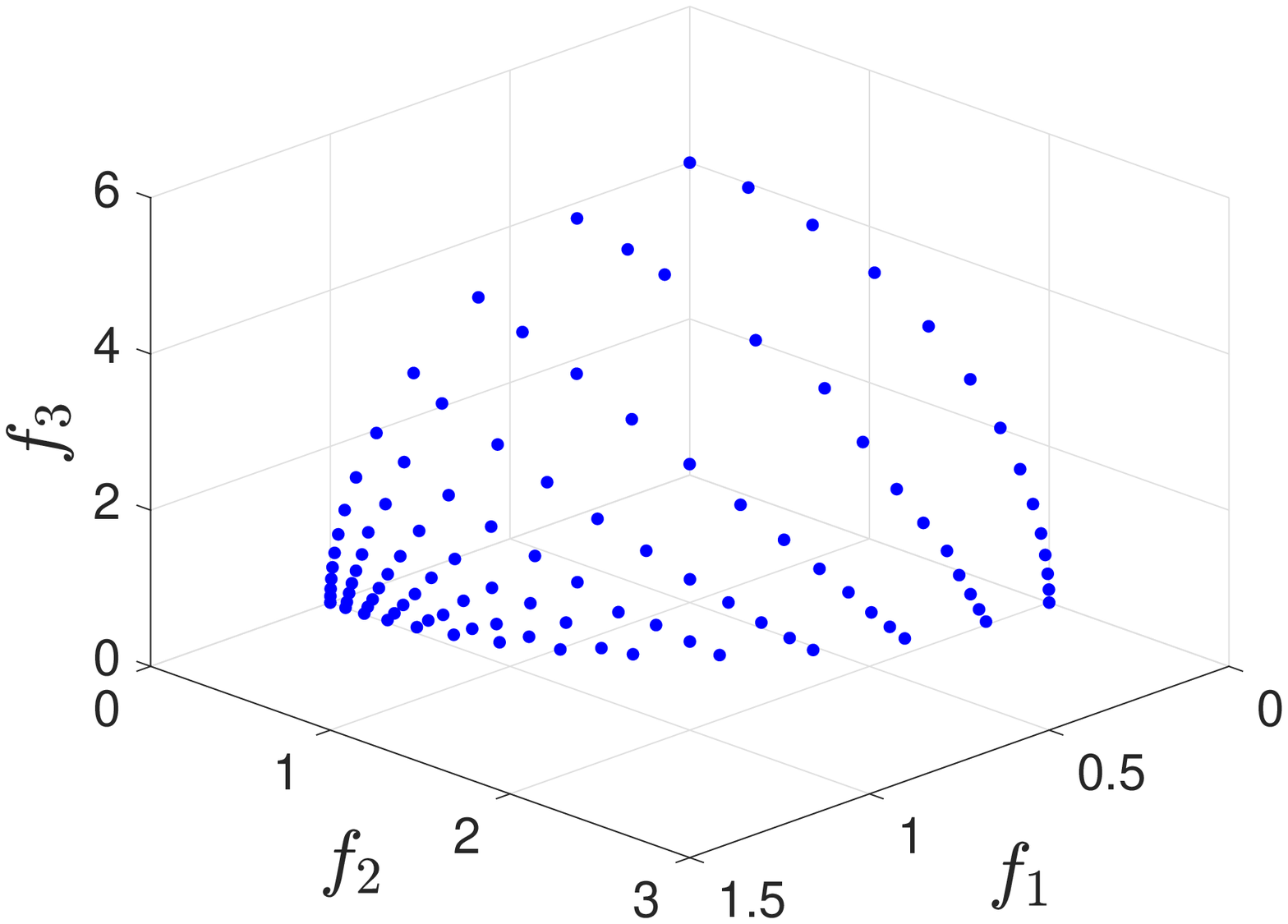}&
		\includegraphics[width=0.2\linewidth]{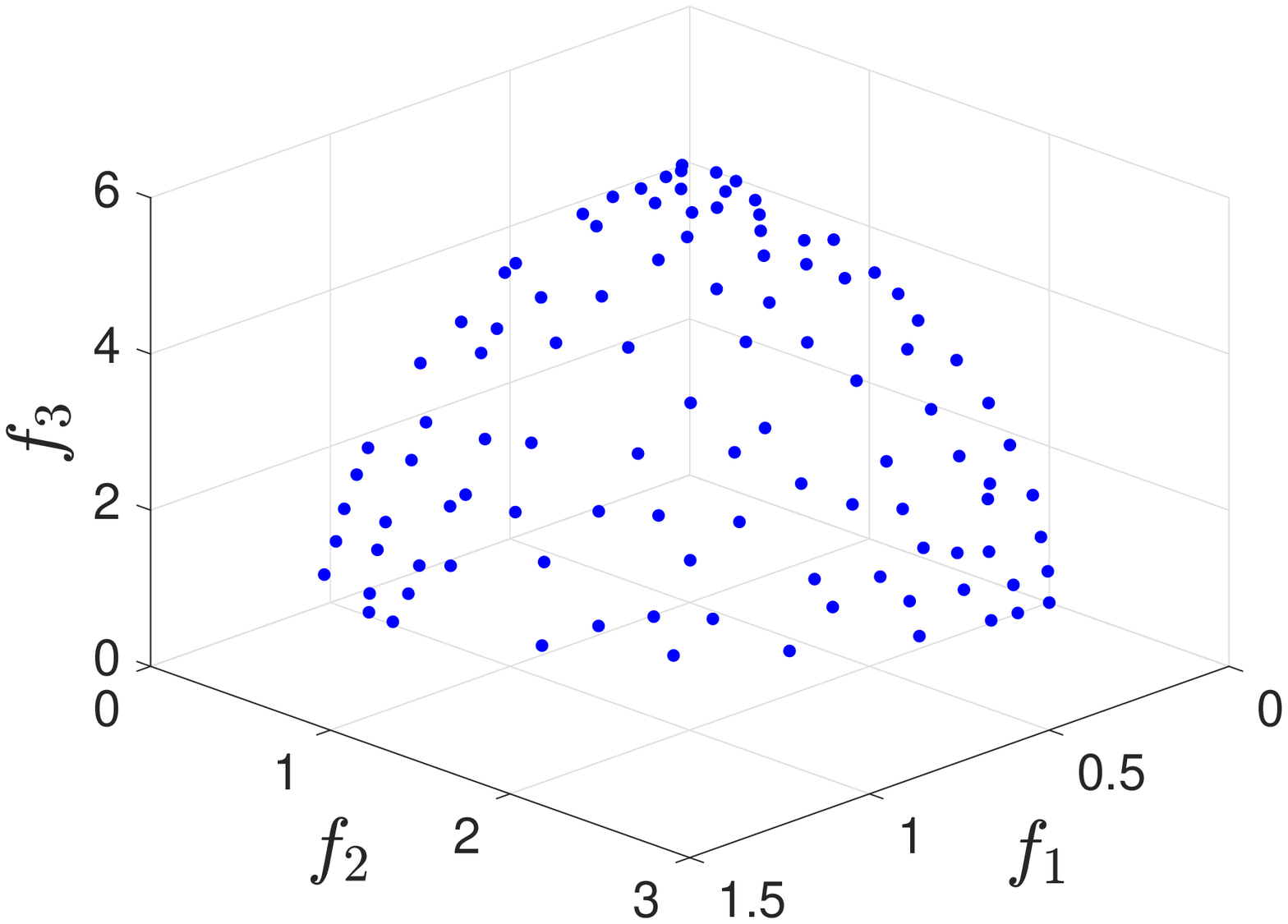}&
		\includegraphics[width=0.2\linewidth]{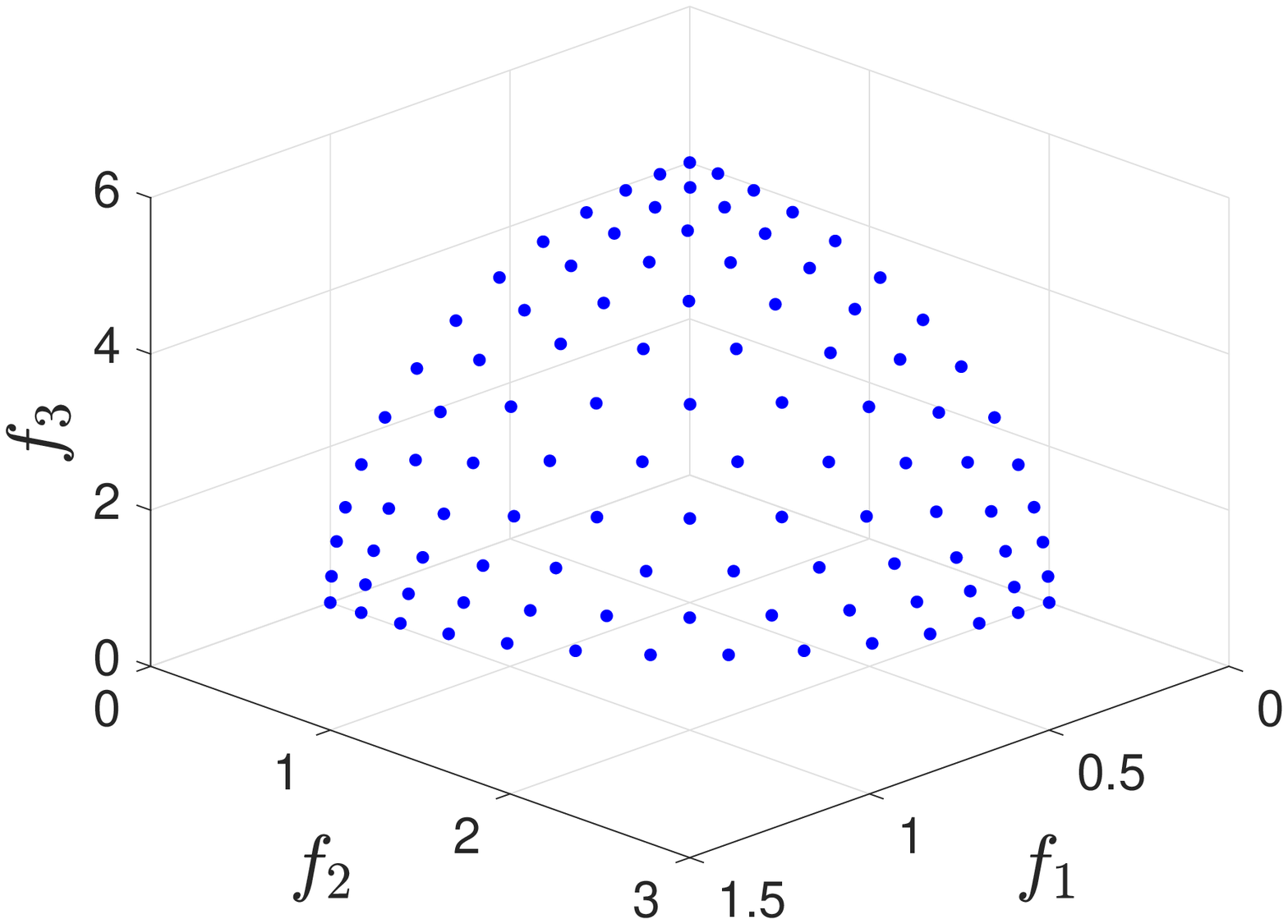}&
		\includegraphics[width=0.2\linewidth]{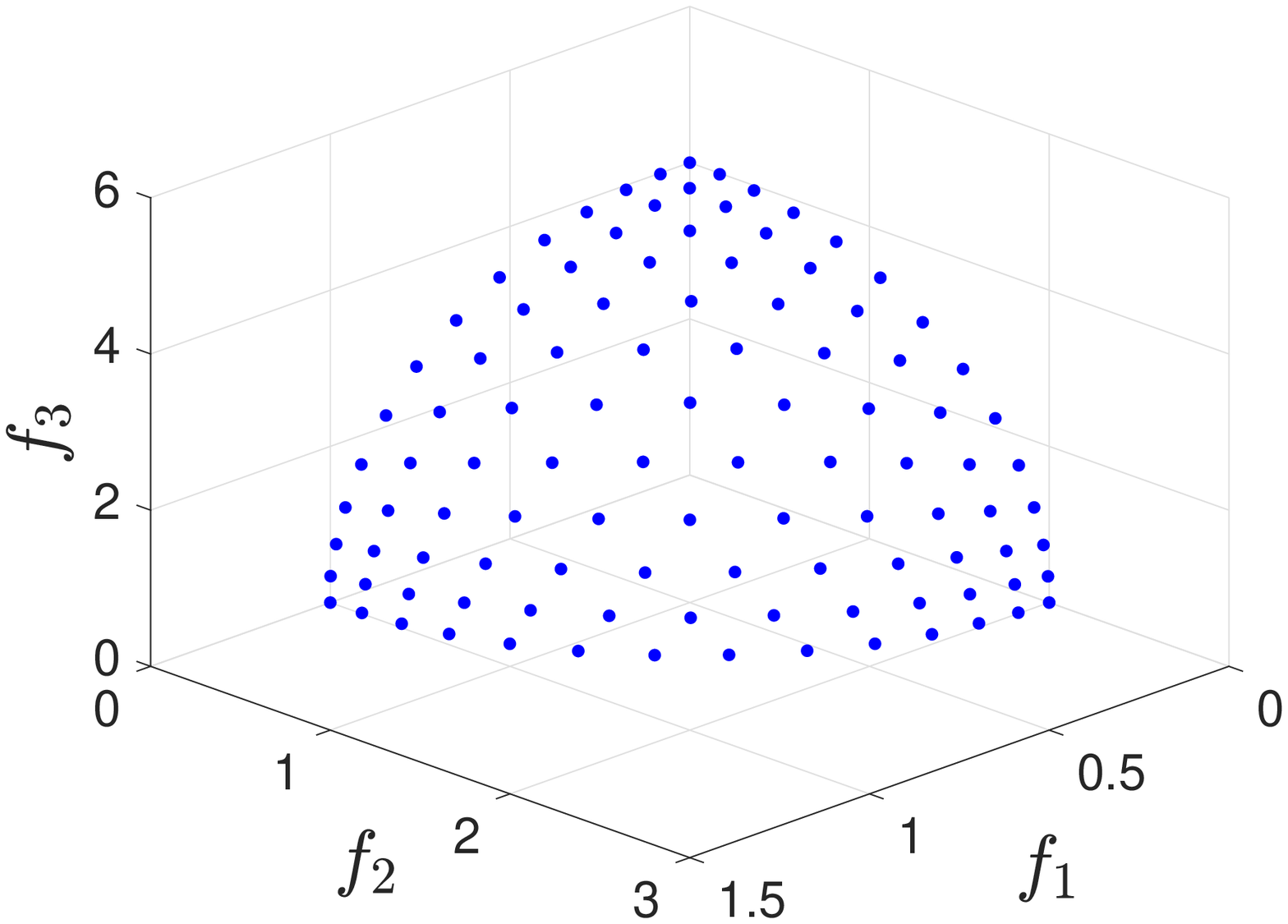}&
		\includegraphics[width=0.2\linewidth]{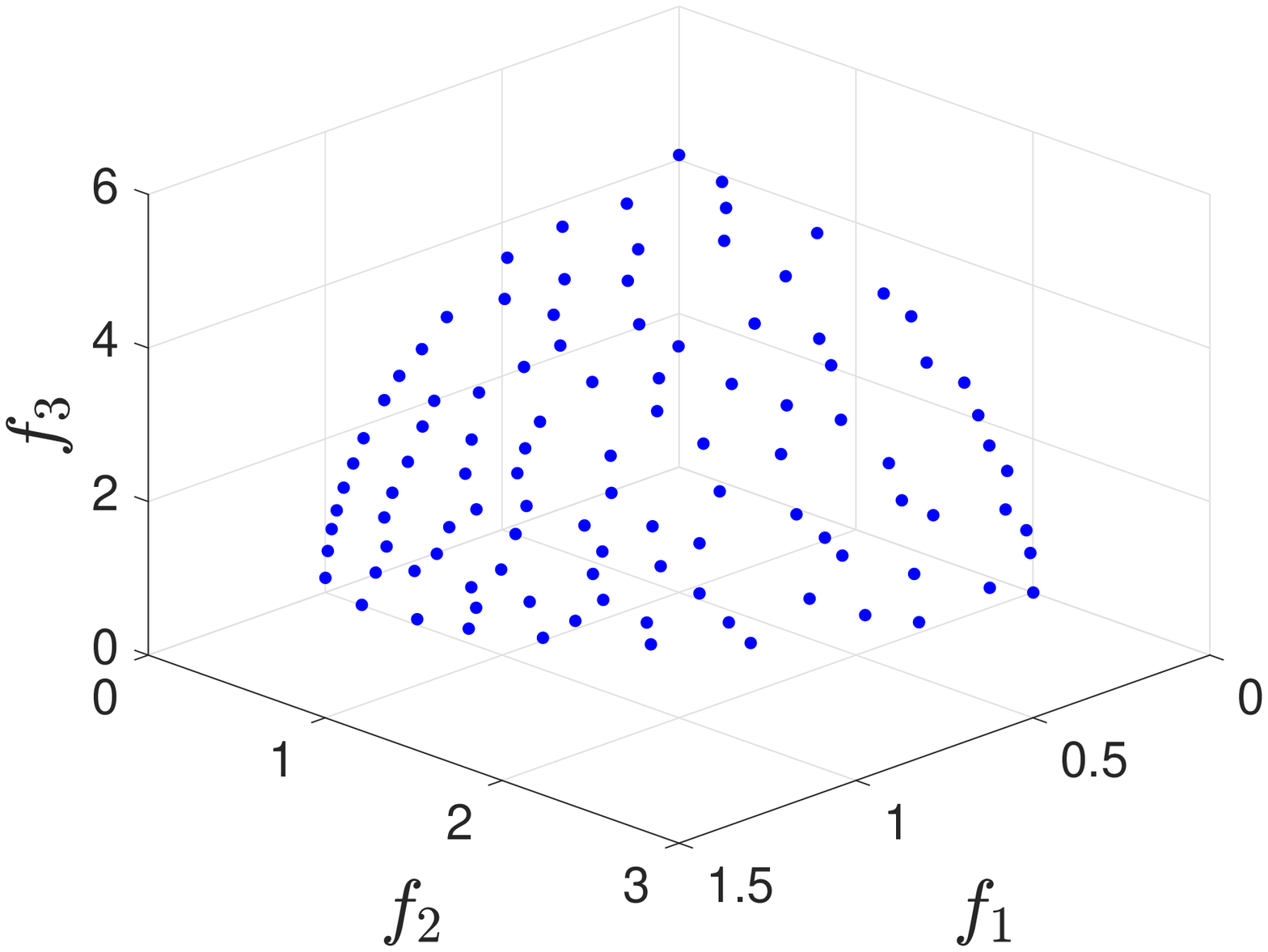}\\
		(a) MOEA/D & (b) ANSGA-III & (c) SPEA/R & (d) RVEA & (e) AREA\\[-2mm]
	\end{tabular}
	\caption{PF approximation of SDTLZ2 obtained by different algorithms.}
	\label{fig:sdtlz2_pf}
	\vspace{-2mm}
\end{figure*}

\begin{figure*}[!tp]
	\centering
	\begin{tabular}{@{}c@{}c@{}c@{}c@{}c}
		\includegraphics[width=0.2\linewidth]{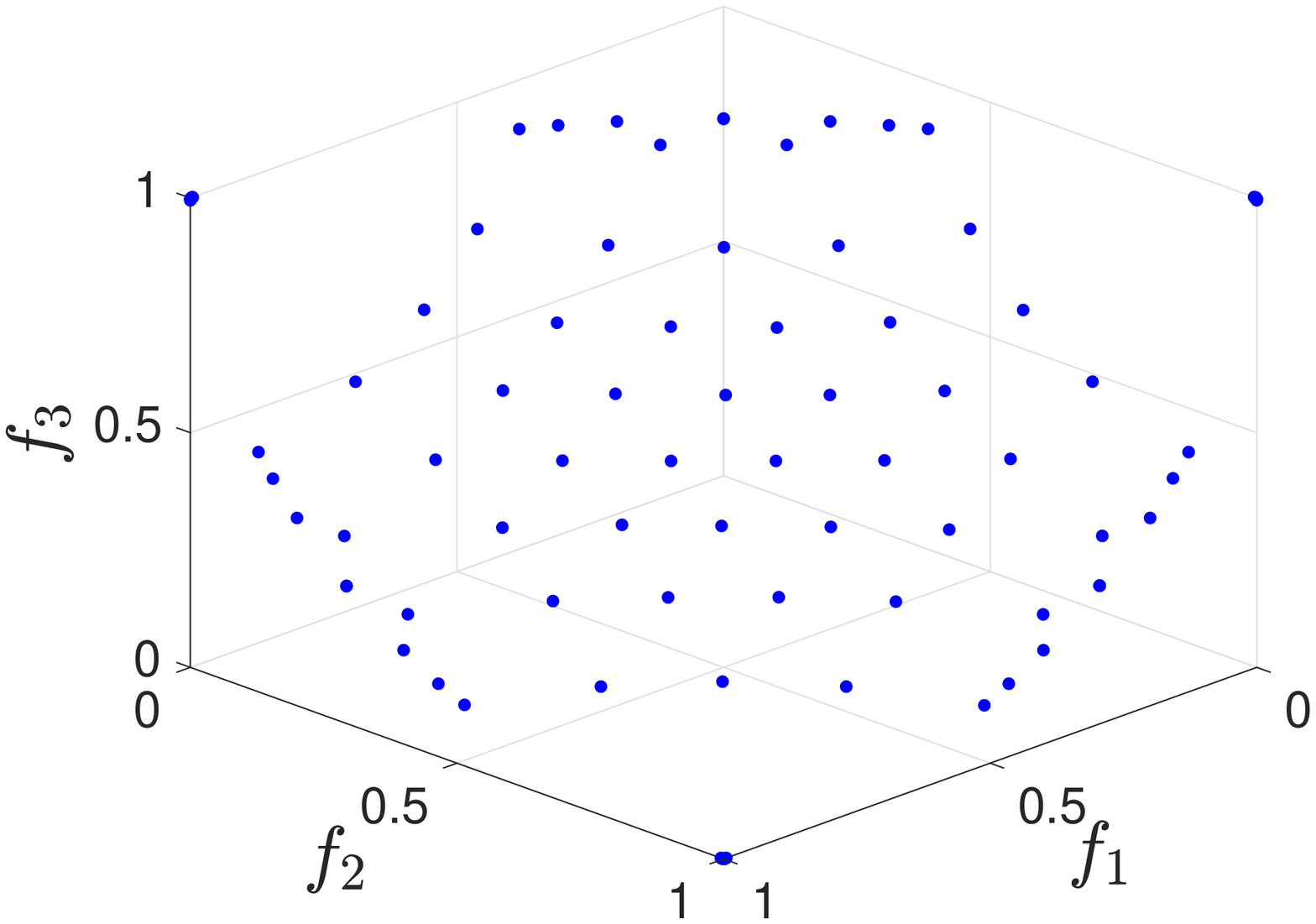}&
		\includegraphics[width=0.2\linewidth]{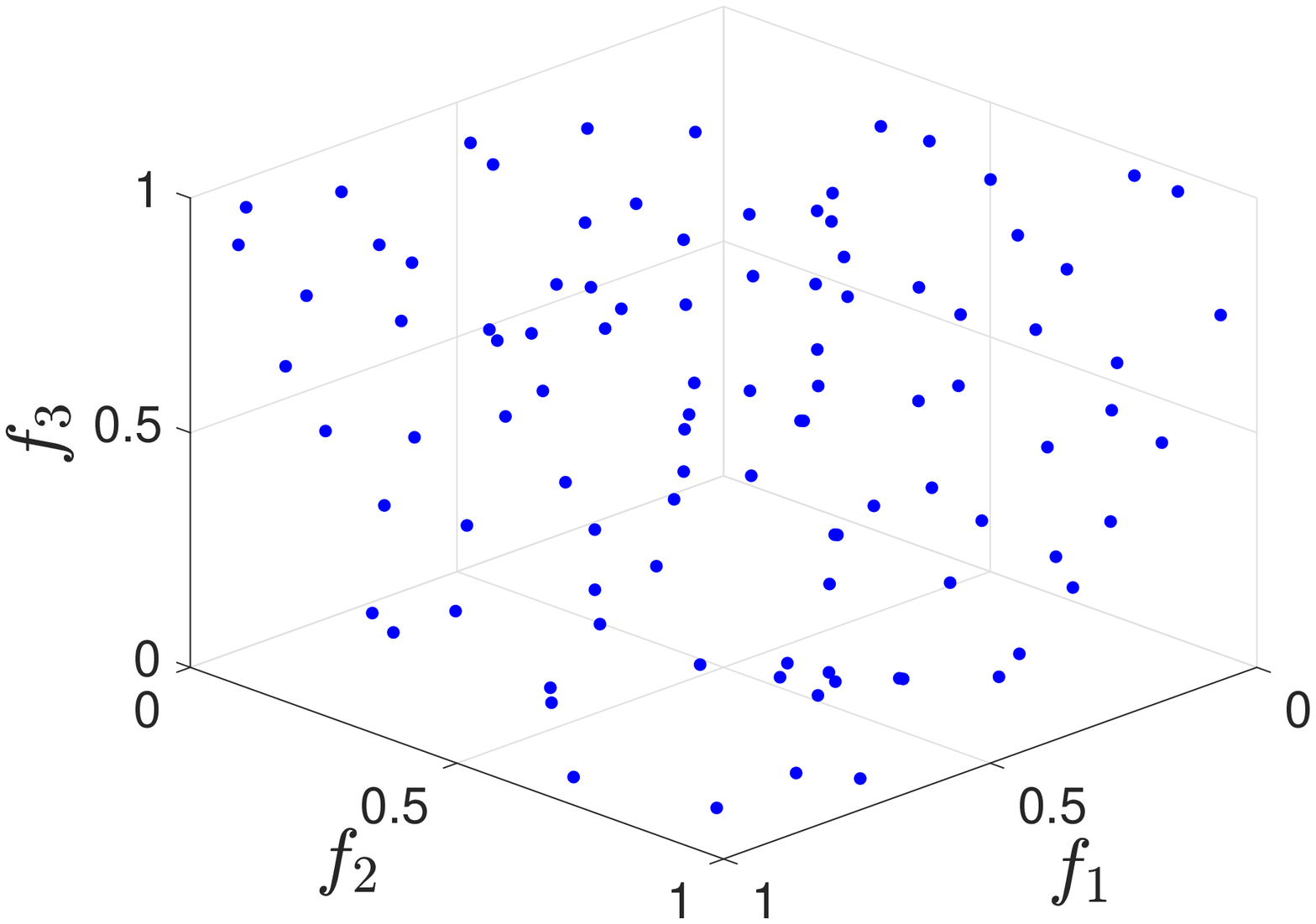}&
		\includegraphics[width=0.2\linewidth]{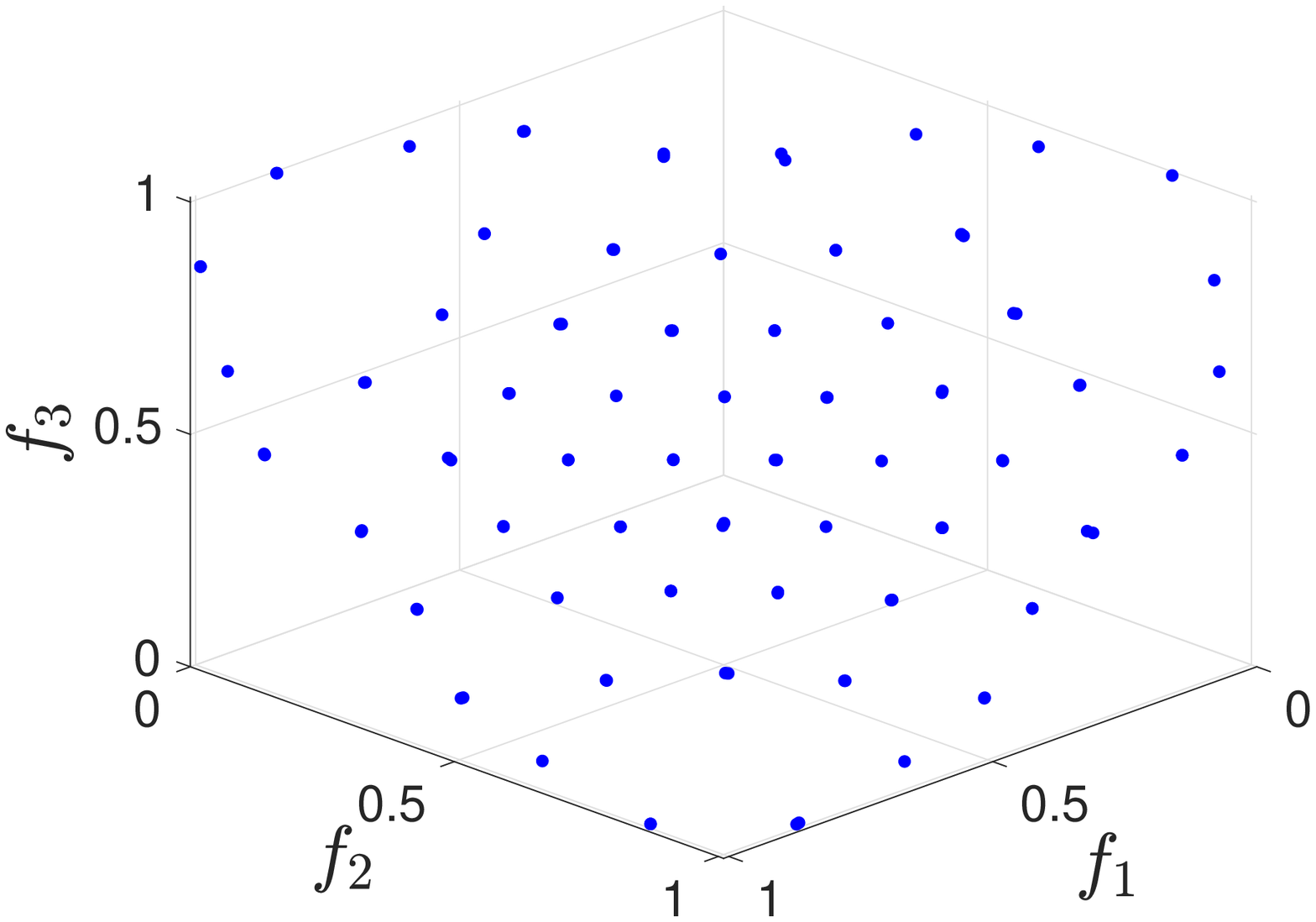}&
		\includegraphics[width=0.2\linewidth]{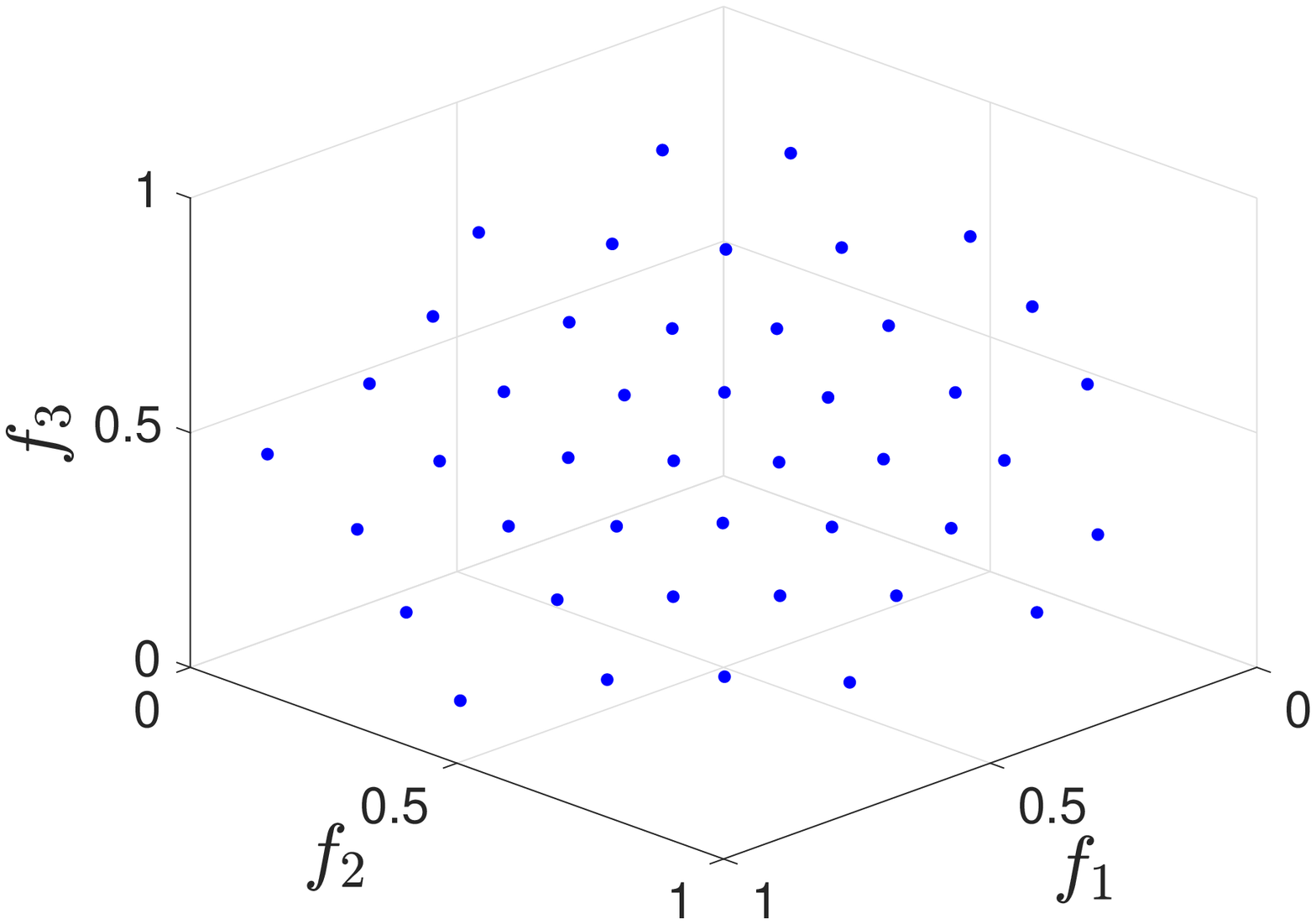}&
		\includegraphics[width=0.2\linewidth]{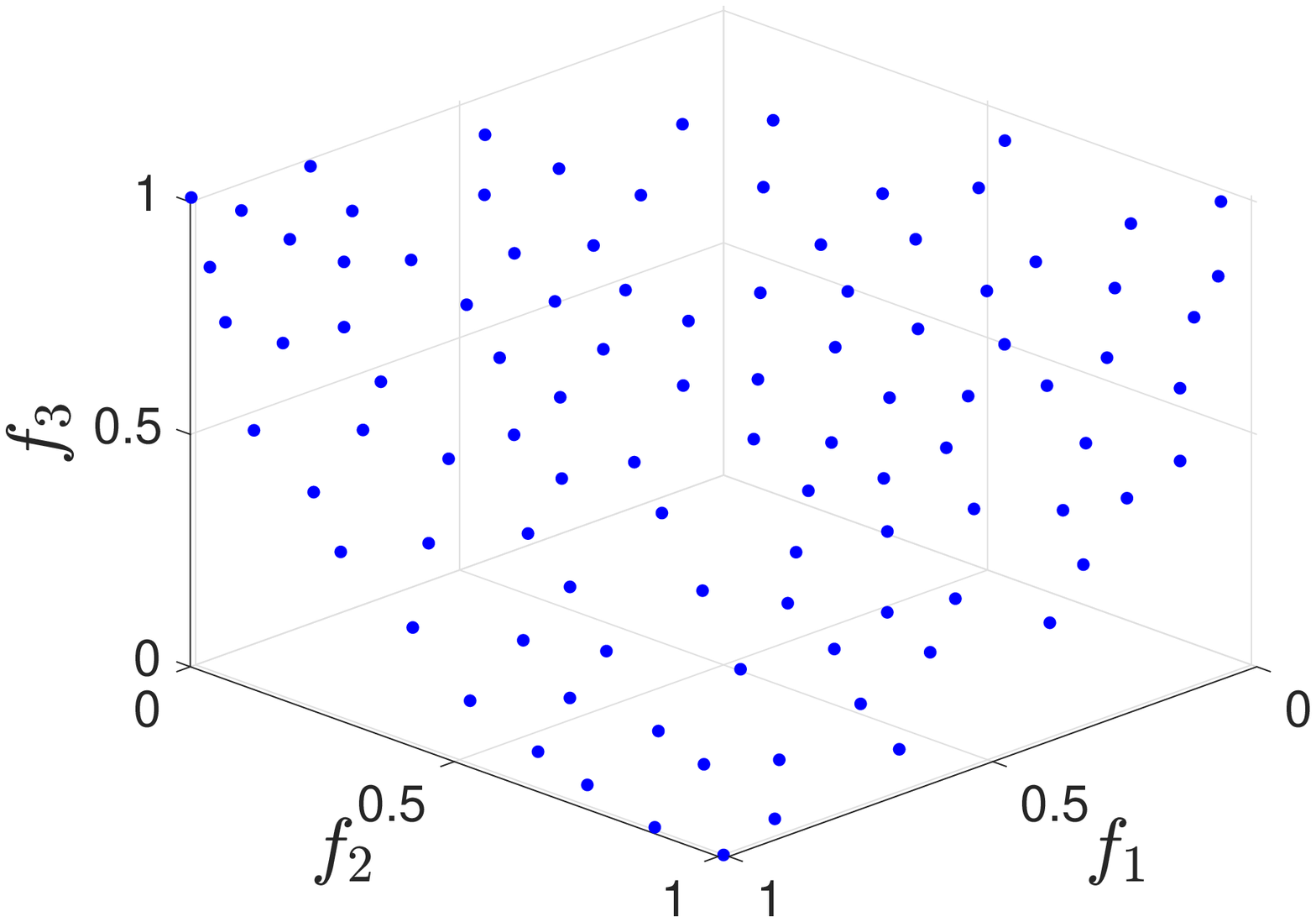}\\
		(a) MOEA/D & (b) ANSGA-III & (c) SPEA/R  & (d) RVEA & (e) AREA\\[-2mm]
	\end{tabular}
	\caption{PF approximation of IDTLZ2 obtained by different algorithms.}
	\label{fig:idtlz2_pf}
	\vspace{-2mm}
\end{figure*}

The comparison between AREA and the other algorithms are presented in Table \ref{tab:all_igd}, Table \ref{tab:all_nhv}, and Table \ref{tab:spacing} with respect to IGD, HV, and S, respectively. As mentioned previously, the Wilcoxon rank-sum test \cite{Wilcoxon45} is applied to determine the significance between AREA and the compared algorithms. Therefore, ``+'', ``-'', or ``$\approx$'' indicates the corresponding algorithm is significantly better than, worse than, or equivalent to AREA, respectively, with a significance level of 0.05. The tables show that the algorithms' IGD values are in good agreement with their HV. The S measure shows that solutions obtained by AREA for most of the problems is more uniform than by the other algorithms. MOEA/D has some good S values mainly on the MOP group. As will be shown later, small S values by MOEA/D is due to the loss of diversity in the solution set. In what follows, detailed performance analysis of AREA relying on the IGD measure is provided for each type of test problems.
For the first two DTLZ problems, AREA has no much advantage over the other algorithms, and even slightly worse than MOEA/D or RVEA, indicating AREA is not the best choice for problems with simple PF geometries. A close visual inspection of the PF approximation for these two problems shows that AREA has a good distribution of solutions, but the distribution is just not as well-organised as those of the other algorithms (see Fig.~\ref{fig:dtlz1_pf}). It is actually the solution organisation that makes the tiny difference between these algorithms, and thus AREA is indeed a good performer on these two problems. For the other DTLZ variants, AREA is clearly the winner, which demonstrates the proposed approach has the ability to handle complex PF shapes effectively. The second best performer for these irregular DTLZ variants is ANSGA-III. The two top performers suggest that adjusting reference set to guide the search is indeed significantly better than doing nothing for irregular PF shapes. Figs.~\ref{fig:dtlz5_pf}--\ref{fig:idtlz2_pf} graphically support the above conclusion and illustrate that AREA has the advantage of finding more well-diversified solutions covering the whole PF than the other algorithms.

\begin{table*}[t]
	\footnotesize
	\addtolength{\tabcolsep}{-3.5pt}
	\centering
	\caption{Mean (and standard deviation) values of normalised HV obtained by five algorithms. Best (highest mean) values are highlighted in boldface. Algorithms that are better than, worse than, or equivalent to AREA on ranksum test with a significance level of 0.05 are indicated by $+$, $-$, or $\approx$, respectively.}
	\begin{tabular}{ccccccc}
		\toprule
		Problem&$M$&MOEA/D&ANSGA-III&SPEA/R&RVEA&AREA\\
		\midrule
		\multirow{1}{*}{DTLZ1}&3&8.4080e-1 (5.41e-3) $+$&8.3142e-1 (1.32e-2) $-$&8.1931e-1 (3.62e-2) $-$&\hl{8.4090e-1 (2.40e-3) $+$}&8.3905e-1 (2.91e-3)\\
		\hline
		\multirow{1}{*}{DTLZ2}&3&\hl{5.6284e-1 (7.32e-5) $+$}&5.5848e-1 (2.98e-3) $\approx$&5.6265e-1 (1.40e-4) $+$&5.6247e-1 (4.33e-4) $+$&5.5970e-1 (1.04e-3)\\
		\hline
		\multirow{1}{*}{DTLZ5}&3&1.9270e-1 (8.98e-6) $-$&1.9620e-1 (5.42e-4) $-$&1.8474e-1 (7.40e-4) $-$&1.5229e-1 (8.00e-3) $-$&\hl{1.9980e-1 (1.19e-4)}\\
		\hline
		\multirow{1}{*}{DTLZ7}&3&2.5602e-1 (1.55e-2) $-$&2.7122e-1 (1.33e-3) $-$&2.7224e-1 (7.37e-4) $-$&2.5954e-1 (4.19e-3) $-$&\hl{2.7741e-1 (1.06e-3)}\\
		\hline
		\multirow{1}{*}{IDTLZ1}&3&2.0107e-1 (2.01e-3) $-$&2.1427e-1 (6.58e-3) $\approx$&1.1620e-1 (5.08e-2) $-$&1.7373e-1 (2.24e-2) $-$&\hl{2.1434e-1 (6.44e-3)}\\
		\hline
		\multirow{1}{*}{IDTLZ2}&3&5.0858e-1 (2.26e-4) $-$&5.2272e-1 (3.95e-3) $-$&5.1722e-1 (7.68e-4) $-$&5.1456e-1 (1.30e-3) $-$&\hl{5.3789e-1 (7.25e-4)}\\
		\hline
		\multirow{1}{*}{SDTLZ2}&3&5.4925e-1 (3.69e-3) $-$&5.5087e-1 (1.65e-3) $-$&\hl{5.6261e-1 (1.39e-4) $+$}&5.6230e-1 (2.74e-4) $+$&5.5887e-1 (1.15e-3)\\
		\hline
		\multirow{1}{*}{CDTLZ2}&3&9.5974e-1 (5.15e-5) $-$&9.5725e-1 (1.57e-3) $-$&9.6013e-1 (4.86e-4) $-$&9.6001e-1 (7.48e-4) $-$&\hl{9.6224e-1 (5.38e-4)}\\
		\hline
		\multirow{1}{*}{MOP1}&2&2.6452e-1 (1.20e-2) $-$&5.1353e-1 (1.37e-1) $-$&\hl{7.1162e-1 (1.26e-3) $+$}&5.6704e-1 (3.42e-3) $-$&7.1006e-1 (3.36e-4)\\
		\hline
		\multirow{1}{*}{MOP2}&2&1.7408e-1 (2.88e-3) $-$&2.5167e-1 (3.60e-2) $-$&\hl{4.2335e-1 (3.94e-2) $+$}&3.3901e-1 (3.20e-2) $-$&3.4022e-1 (7.80e-2)\\
		\hline
		\multirow{1}{*}{MOP3}&2&9.0909e-2 (7.06e-17) $-$&1.5771e-1 (3.01e-2) $-$&3.1959e-1 (4.56e-2) $\approx$&2.4461e-1 (4.60e-2) $-$&\hl{3.3640e-1 (3.55e-2)}\\
		\hline
		\multirow{1}{*}{MOP4}&2&2.9434e-1 (7.88e-3) $-$&3.3744e-1 (2.19e-2) $-$&\hl{5.9251e-1 (6.16e-3) $+$}&5.5592e-1 (2.42e-2) $-$&5.7784e-1 (3.66e-2)\\
		\hline
		\multirow{1}{*}{MOP5}&2&4.0194e-1 (2.51e-3) $-$&4.6204e-1 (8.45e-2) $-$&6.9323e-1 (2.48e-3) $-$&6.1629e-1 (5.29e-3) $-$&\hl{7.0285e-1 (9.59e-4)}\\
		\hline
		\multirow{1}{*}{MOP6}&3&6.1397e-1 (1.76e-7) $-$&6.5164e-1 (3.53e-2) $-$&6.3036e-1 (2.72e-2) $-$&\hl{7.4727e-1 (1.26e-2) $+$}&6.9172e-1 (4.12e-2)\\
		\hline
		\multirow{1}{*}{MOP7}&3&4.0220e-1 (1.64e-7) $-$&3.9730e-1 (3.56e-3) $-$&3.9306e-1 (8.16e-3) $-$&\hl{4.6164e-1 (4.84e-3) $+$}&4.1530e-1 (2.18e-2)\\
		\hline
		\multirow{1}{*}{F1}&2&9.1326e-1 (2.67e-4) $-$&9.1057e-1 (3.07e-3) $-$&9.1275e-1 (6.06e-4) $-$&9.1259e-1 (5.09e-4) $-$&\hl{9.1470e-1 (8.86e-5)}\\
		\hline
		\multirow{1}{*}{F2}&2&9.6426e-2 (3.02e-2) $-$&2.0019e-1 (6.28e-2) $-$&2.5698e-1 (1.52e-7) $-$&2.5043e-1 (2.11e-3) $-$&\hl{2.5744e-1 (9.82e-5)}\\
		\hline
		\multirow{1}{*}{F3}&2&8.5602e-1 (2.33e-2) $-$&8.5954e-1 (1.26e-2) $-$&8.6532e-1 (2.07e-4) $-$&8.6393e-1 (6.56e-4) $-$&\hl{8.6594e-1 (1.58e-4)}\\
		\hline
		\multirow{1}{*}{F4}&2&2.4826e-1 (1.31e-1) $-$&2.2656e-1 (8.85e-2) $-$&3.5264e-1 (4.43e-4) $-$&3.4757e-1 (2.31e-3) $-$&\hl{3.5336e-1 (7.18e-5)}\\
		\hline
		\multirow{1}{*}{F5}&3&5.6127e-1 (5.17e-4) $+$&5.6059e-1 (1.78e-3) $+$&\hl{5.6207e-1 (2.11e-4) $+$}&5.5648e-1 (2.27e-3) $-$&5.5970e-1 (1.02e-3)\\
		\hline
		\multirow{1}{*}{F6}&3&5.3669e-1 (3.01e-2) $-$&5.5286e-1 (1.07e-3) $-$&5.4505e-1 (3.30e-3) $-$&5.2716e-1 (2.50e-3) $-$&\hl{5.5672e-1 (8.05e-4)}\\
		\hline
		\multirow{1}{*}{F7}&3&\hl{8.4265e-1 (1.37e-3) $+$}&8.2664e-1 (6.60e-3) $-$&8.2016e-1 (6.78e-3) $-$&8.2307e-1 (4.91e-3) $-$&8.3293e-1 (2.34e-3)\\
		\hline
		\multirow{1}{*}{F8}&3&5.0984e-1 (2.33e-4) $-$&5.0957e-1 (3.25e-3) $-$&5.1518e-1 (2.38e-3) $-$&4.9933e-1 (2.09e-3) $-$&\hl{5.2316e-1 (2.01e-3)}\\
		\hline
		\multicolumn{2}{c}{$+/-/\approx$}&4/19/0&1/20/2&6/16/1&5/18/0&\\
		\bottomrule
	\end{tabular}
	\label{tab:all_nhv}
	\vspace{-2mm}
\end{table*}

\begin{table*}[htbp]
	\footnotesize
	\addtolength{\tabcolsep}{-3.5pt}
	\centering
	\caption{Mean (and standard deviation) values of S obtained by five algorithms. Best (smallest mean) values are highlighted in boldface. Algorithms that are better than, worse than, or equivalent to AREA on ranksum test with a significance level of 0.05 are indicated by $+$, $-$, or $\approx$, respectively.}
	\begin{tabular}{ccccccc}
		\toprule
		Problem&$M$&MOEA/D&ANSGA-III&SPEA/R&RVEA&AREA\\
		\midrule
		\multirow{1}{*}{DTLZ1}&3&4.5066e-3 (2.63e-3) $+$&1.8173e-1 (8.33e-1) $-$&3.4367e-1 (1.74e+0) $-$&\hl{3.0358e-3 (2.35e-3) $+$}&9.9764e-3 (1.67e-2)\\
		\hline
		\multirow{1}{*}{DTLZ2}&3&5.5332e-2 (2.16e-3) $-$&4.9465e-2 (2.68e-3) $-$&5.2473e-2 (2.71e-4) $-$&5.2531e-2 (1.68e-4) $-$&\hl{2.1922e-2 (2.01e-3)}\\
		\hline
		\multirow{1}{*}{DTLZ5}&3&1.5355e-2 (2.07e-3) $-$&1.2224e-2 (1.60e-3) $-$&1.3130e-2 (1.19e-2) $-$&1.0213e-1 (2.98e-2) $-$&\hl{4.7130e-3 (4.21e-4)}\\
		\hline
		\multirow{1}{*}{DTLZ7}&3&1.1663e-1 (2.41e-2) $-$&7.0257e-2 (6.08e-3) $-$&8.1876e-2 (1.24e-2) $-$&9.7947e-2 (8.88e-3) $-$&\hl{2.8537e-2 (7.56e-3)}\\
		\hline
		\multirow{1}{*}{IDTLZ1}&3&4.3653e-2 (1.32e-3) $-$&2.3937e+0 (7.39e+0) $-$&5.0843e+0 (1.18e+1) $-$&5.4957e-2 (4.09e-2) $-$&\hl{7.7326e-3 (7.12e-4)}\\
		\hline
		\multirow{1}{*}{IDTLZ2}&3&9.5006e-2 (9.60e-4) $-$&5.7135e-2 (8.95e-3) $-$&7.5360e-2 (2.77e-3) $-$&2.4001e-2 (2.67e-3) $-$&\hl{2.2644e-2 (1.58e-3)}\\
		\hline
		\multirow{1}{*}{SDTLZ2}&3&1.6470e-1 (4.33e-3) $-$&1.2234e-1 (8.49e-3) $-$&1.0775e-1 (1.15e-3) $-$&1.0787e-1 (1.12e-3) $-$&\hl{5.3431e-2 (5.48e-3)}\\
		\hline
		\multirow{1}{*}{CDTLZ2}&3&3.3489e-2 (3.42e-4) $-$&4.0234e-2 (7.09e-3) $-$&3.6268e-2 (3.70e-3) $-$&3.2319e-2 (2.01e-3) $-$&\hl{1.5687e-2 (1.63e-3)}\\
		\hline
		\multirow{1}{*}{MOP1}&2&9.6621e-3 (7.38e-4) $-$&1.3169e-2 (2.37e-2) $\approx$&9.8662e-3 (1.61e-4) $-$&4.4518e-2 (1.41e-2) $-$&\hl{3.5248e-3 (5.51e-4)}\\
		\hline
		\multirow{1}{*}{MOP2}&2&\hl{1.2148e-4 (6.65e-4) $+$}&1.1699e-2 (3.84e-2) $+$&1.5014e-2 (2.98e-2) $+$&1.7734e-1 (1.07e-1) $-$&7.2180e-2 (1.85e-2)\\
		\hline
		\multirow{1}{*}{MOP3}&2&\hl{3.8025e-5 (1.74e-4) $+$}&5.8225e-5 (3.02e-4) $+$&2.4566e-2 (2.84e-2) $+$&1.7402e-1 (1.32e-1) $-$&4.0200e-2 (1.21e-1)\\
		\hline
		\multirow{1}{*}{MOP4}&2&1.1218e-2 (7.54e-4) $-$&5.2088e-3 (2.38e-3) $-$&1.1445e-2 (1.63e-3) $-$&2.8181e-2 (1.41e-2) $-$&\hl{2.6247e-3 (1.43e-3)}\\
		\hline
		\multirow{1}{*}{MOP5}&2&\hl{7.9106e-3 (6.45e-3) $\approx$}&8.4370e-2 (3.48e-2) $-$&1.5577e-2 (4.13e-3) $-$&5.7877e-2 (4.11e-2) $-$&9.5701e-3 (1.03e-2)\\
		\hline
		\multirow{1}{*}{MOP6}&3&\hl{1.1480e-2 (1.61e-4) $+$}&1.3729e-2 (4.16e-3) $+$&7.7551e-2 (3.59e-2) $-$&4.3294e-2 (2.09e-2) $-$&4.1046e-2 (4.39e-2)\\
		\hline
		\multirow{1}{*}{MOP7}&3&\hl{1.3615e-2 (1.03e-3) $+$}&1.4848e-2 (3.08e-2) $+$&1.1384e-1 (7.82e-2) $\approx$&1.1523e-1 (6.26e-2) $\approx$&9.3868e-2 (7.62e-2)\\
		\hline
		\multirow{1}{*}{F1}&2&3.7998e-2 (4.81e-3) $-$&2.2475e-2 (7.33e-3) $-$&1.8889e-2 (1.25e-2) $-$&3.2146e-2 (2.53e-3) $-$&\hl{3.1936e-3 (4.50e-4)}\\
		\hline
		\multirow{1}{*}{F2}&2&\hl{4.0123e-4 (2.20e-3) $+$}&5.7417e-2 (5.82e-2) $-$&7.4123e-3 (1.70e-5) $+$&3.7215e-2 (1.01e-2) $-$&1.4676e-2 (5.93e-3)\\
		\hline
		\multirow{1}{*}{F3}&2&8.8891e-2 (1.98e-3) $-$&5.9362e-2 (1.79e-2) $-$&7.4343e-2 (2.17e-2) $-$&4.4087e-2 (3.34e-3) $-$&\hl{2.1394e-2 (5.25e-3)}\\
		\hline
		\multirow{1}{*}{F4}&2&5.3214e-3 (4.50e-3) $\approx$&5.2218e-2 (7.32e-2) $-$&1.3203e-2 (4.62e-3) $-$&1.3208e-2 (2.83e-3) $-$&\hl{3.3803e-3 (3.92e-4)}\\
		\hline
		\multirow{1}{*}{F5}&3&5.9764e-2 (2.06e-3) $-$&4.7852e-2 (2.19e-3) $-$&4.8743e-2 (1.02e-3) $-$&5.8263e-2 (6.90e-3) $-$&\hl{2.1816e-2 (2.12e-3)}\\
		\hline
		\multirow{1}{*}{F6}&3&4.5513e-2 (1.79e-2) $-$&5.6561e-2 (5.43e-3) $-$&6.8635e-2 (1.36e-2) $-$&6.9962e-2 (5.97e-3) $-$&\hl{2.1303e-2 (1.75e-3)}\\
		\hline
		\multirow{1}{*}{F7}&3&1.2604e-1 (2.97e-3) $-$&1.4898e-1 (1.25e-1) $-$&2.1306e-1 (1.25e-1) $-$&1.1459e-1 (2.93e-2) $-$&\hl{9.9732e-2 (5.38e-2)}\\
		\hline
		\multirow{1}{*}{F8}&3&3.1472e-1 (6.33e-2) $-$&2.3541e-1 (2.45e-2) $-$&2.1309e-1 (4.25e-2) $-$&1.8645e-1 (1.77e-2) $-$&\hl{8.3958e-2 (9.20e-3)}\\
		\hline
		\multicolumn{2}{c}{$+/-/\approx$}&6/15/2&4/18/1&3/19/1&1/21/1&\\
		\bottomrule
	\end{tabular}
	\label{tab:spacing}
\end{table*}


\begin{figure*}[!tp]
	\centering
	\begin{tabular}{@{}c@{}c@{}c@{}c@{}c}
		\includegraphics[width=0.2\linewidth]{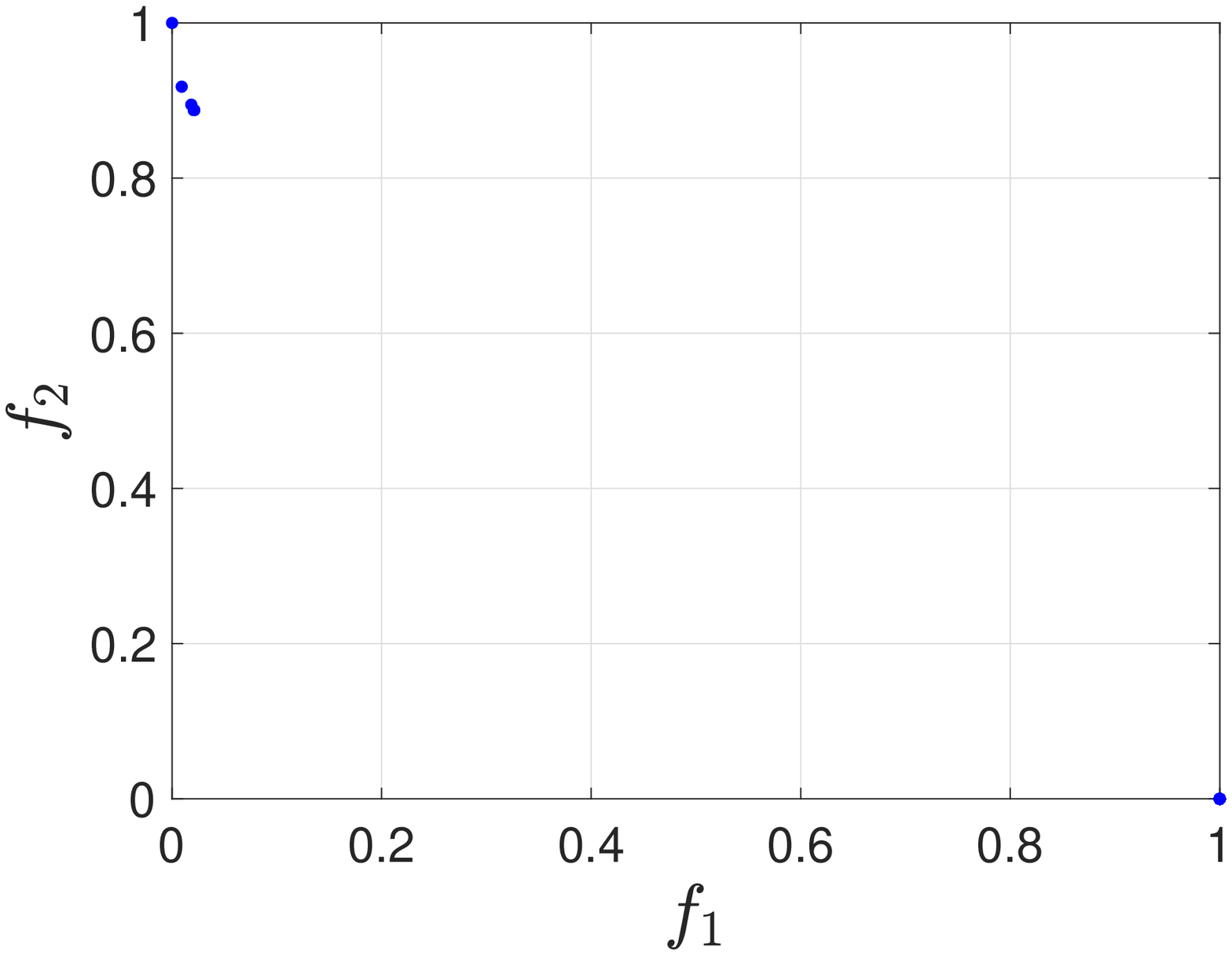}&
		\includegraphics[width=0.2\linewidth]{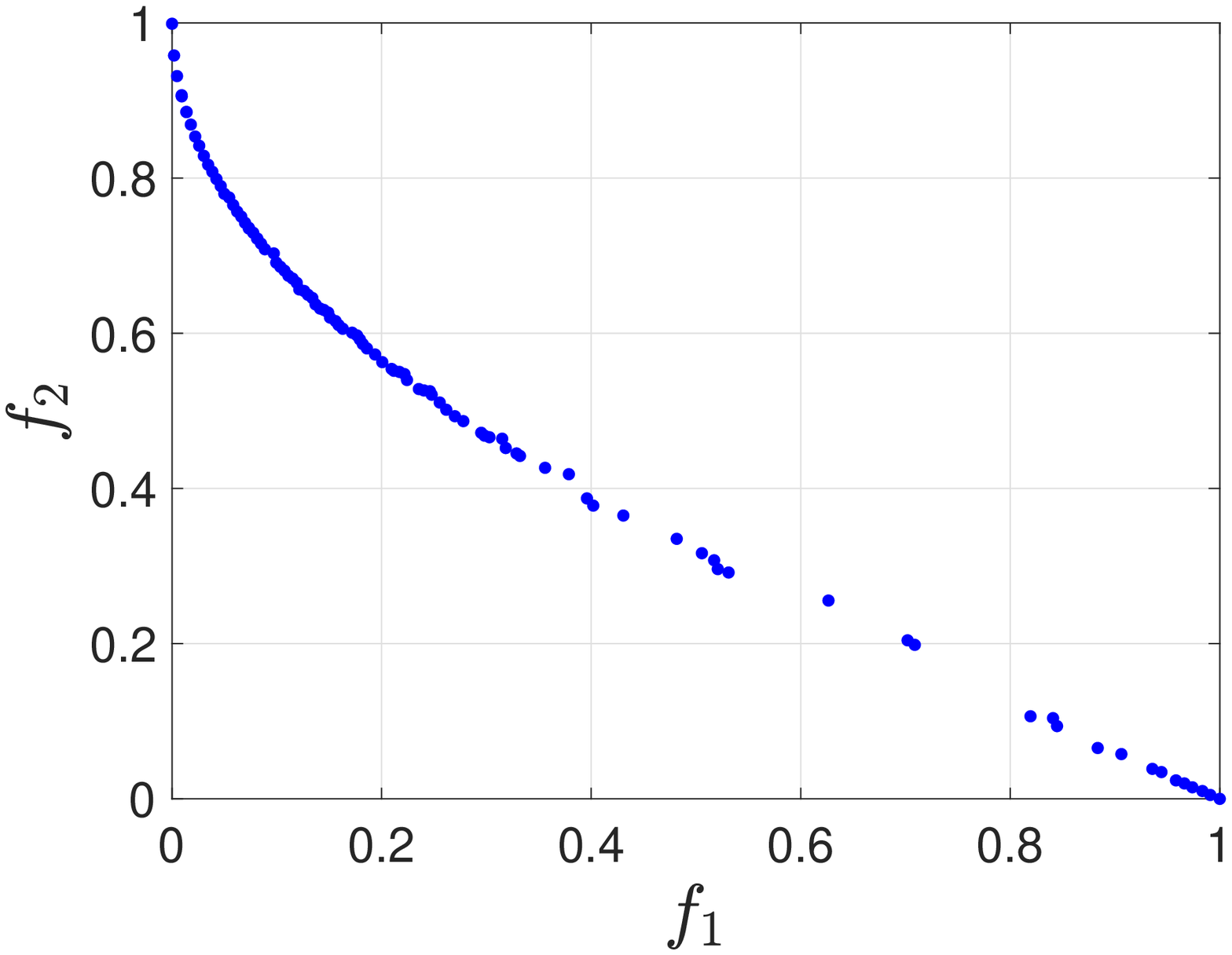}&
		\includegraphics[width=0.2\linewidth]{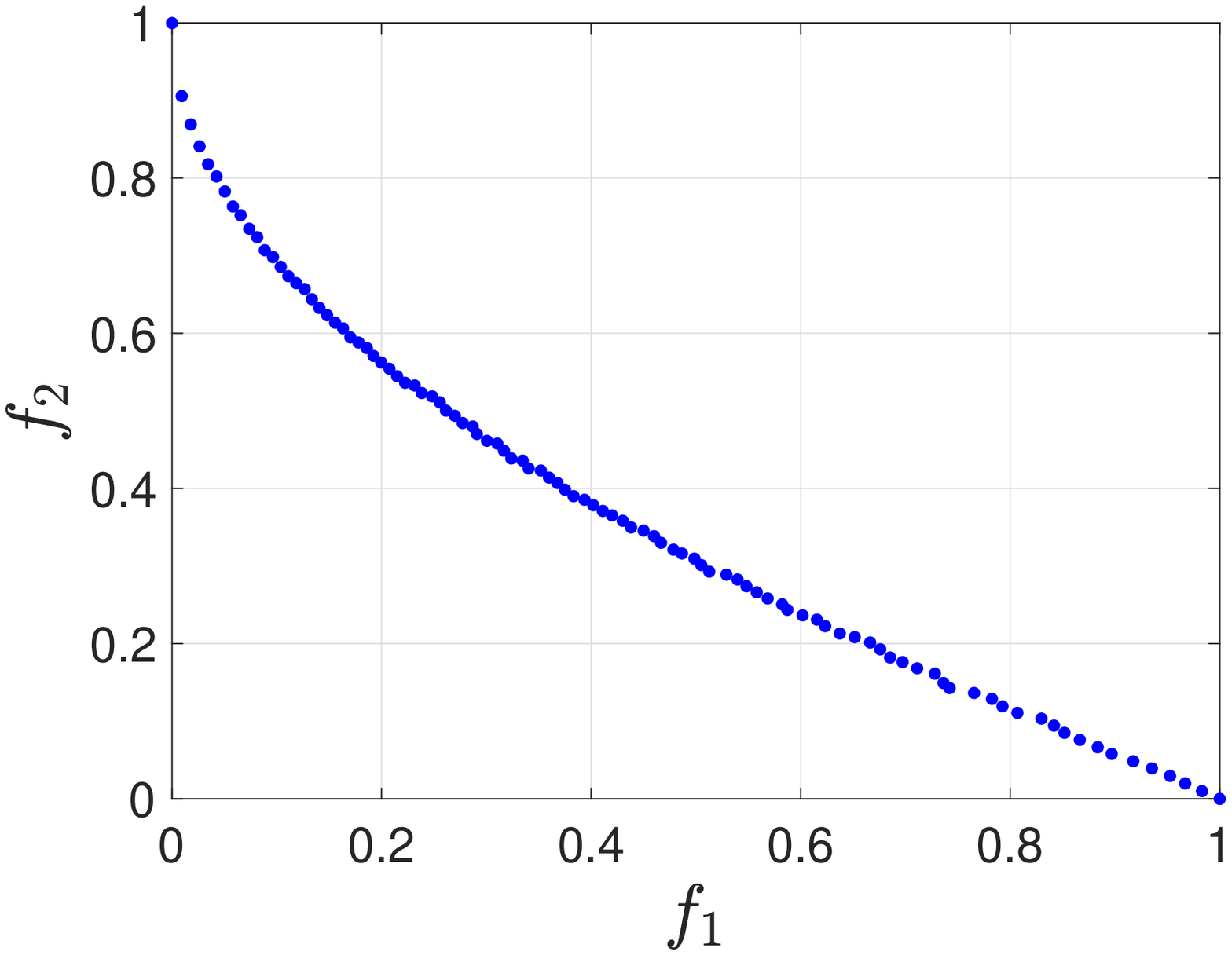}&
		\includegraphics[width=0.2\linewidth]{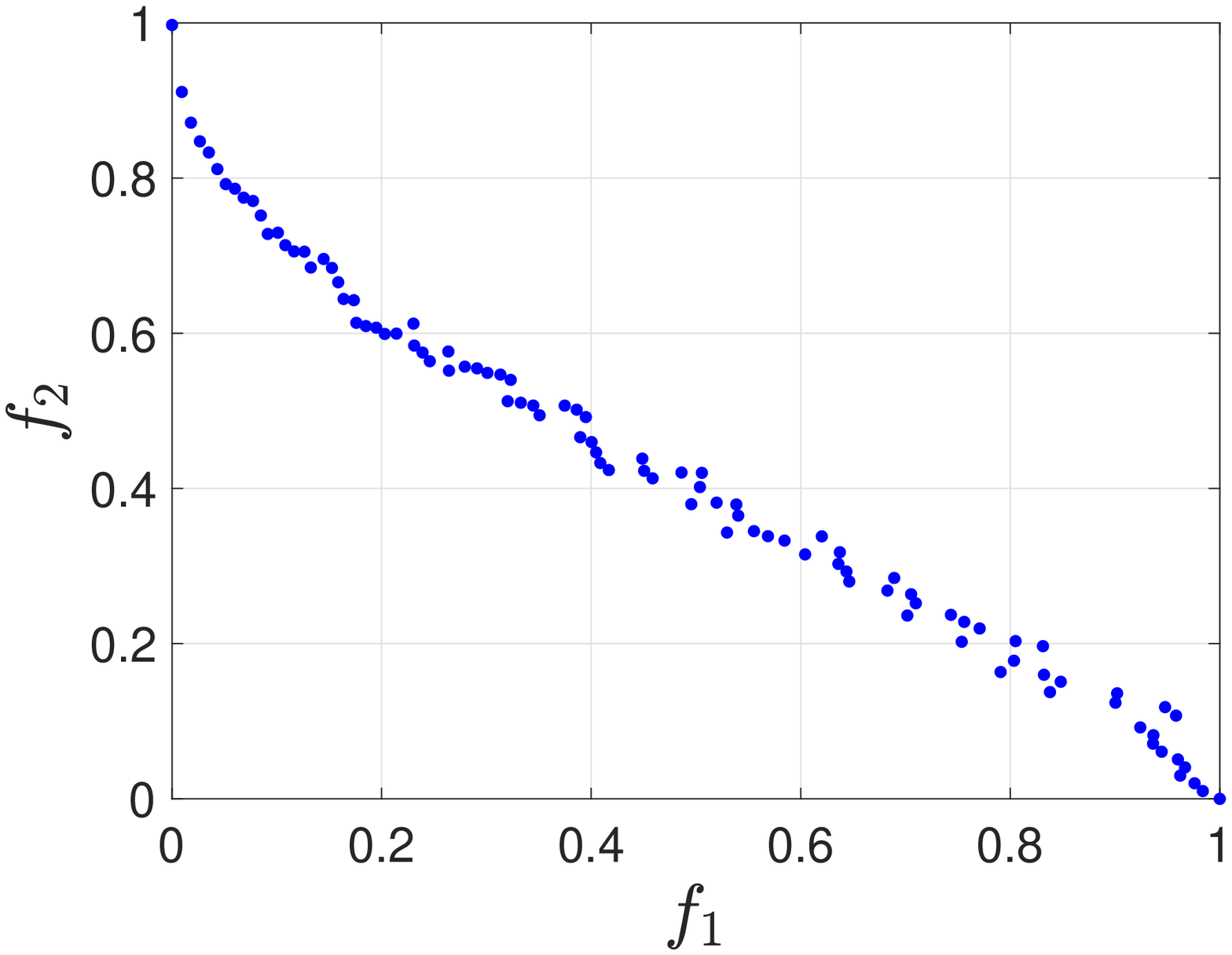}&
		\includegraphics[width=0.2\linewidth]{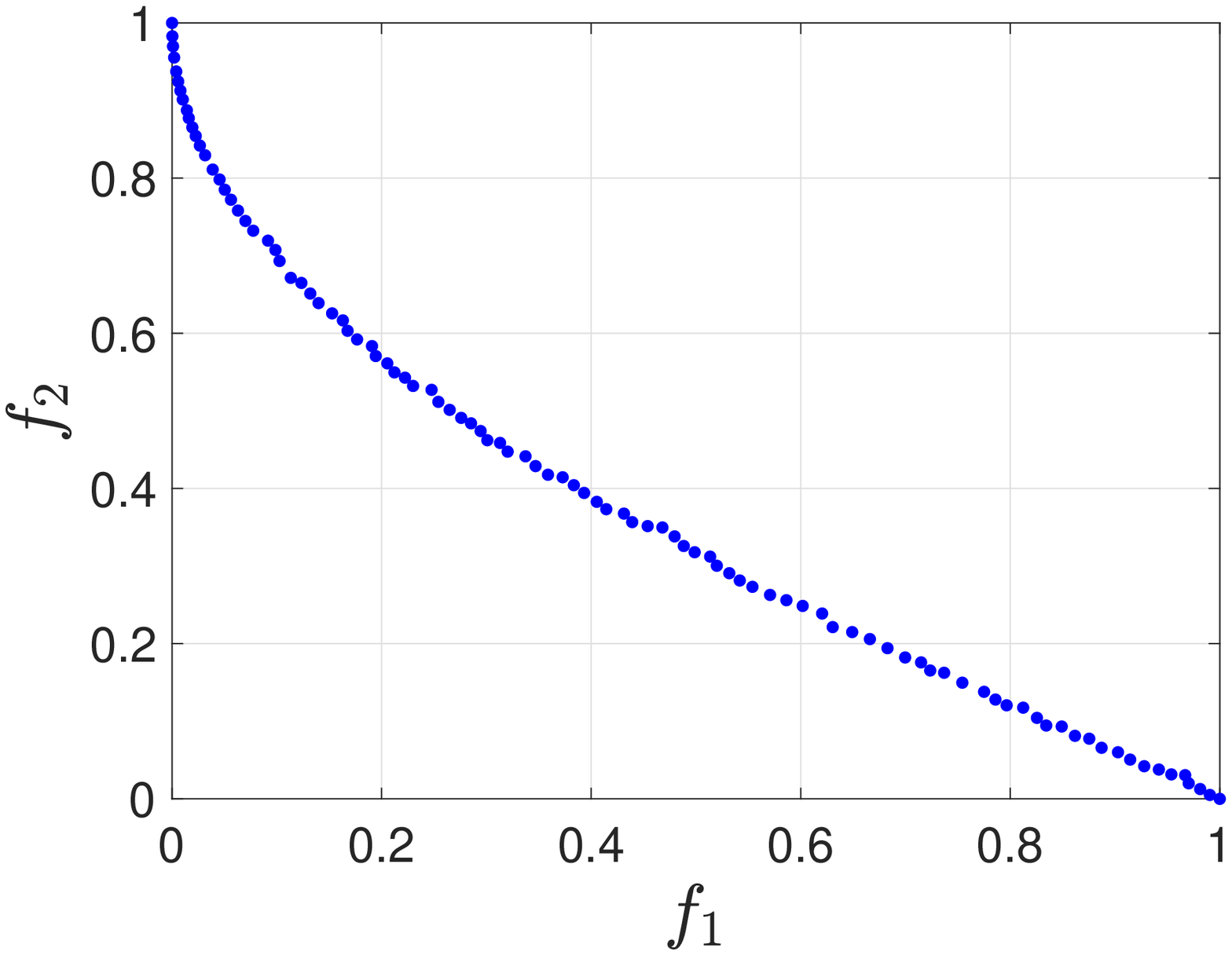}\\
		(a) MOEA/D & (b) ANSGA-III & (c) SPEA/R  & (d) RVEA & (e) AREA\\[-2mm]
	\end{tabular}
	\caption{PF approximation of MOP1 obtained by different algorithms.}
	\label{fig:mop1_pf}
	\vspace{-2mm}
\end{figure*}

\begin{figure*}[!tp]
	\centering
	\begin{tabular}{@{}c@{}c@{}c@{}c@{}c}
		\includegraphics[width=0.2\linewidth]{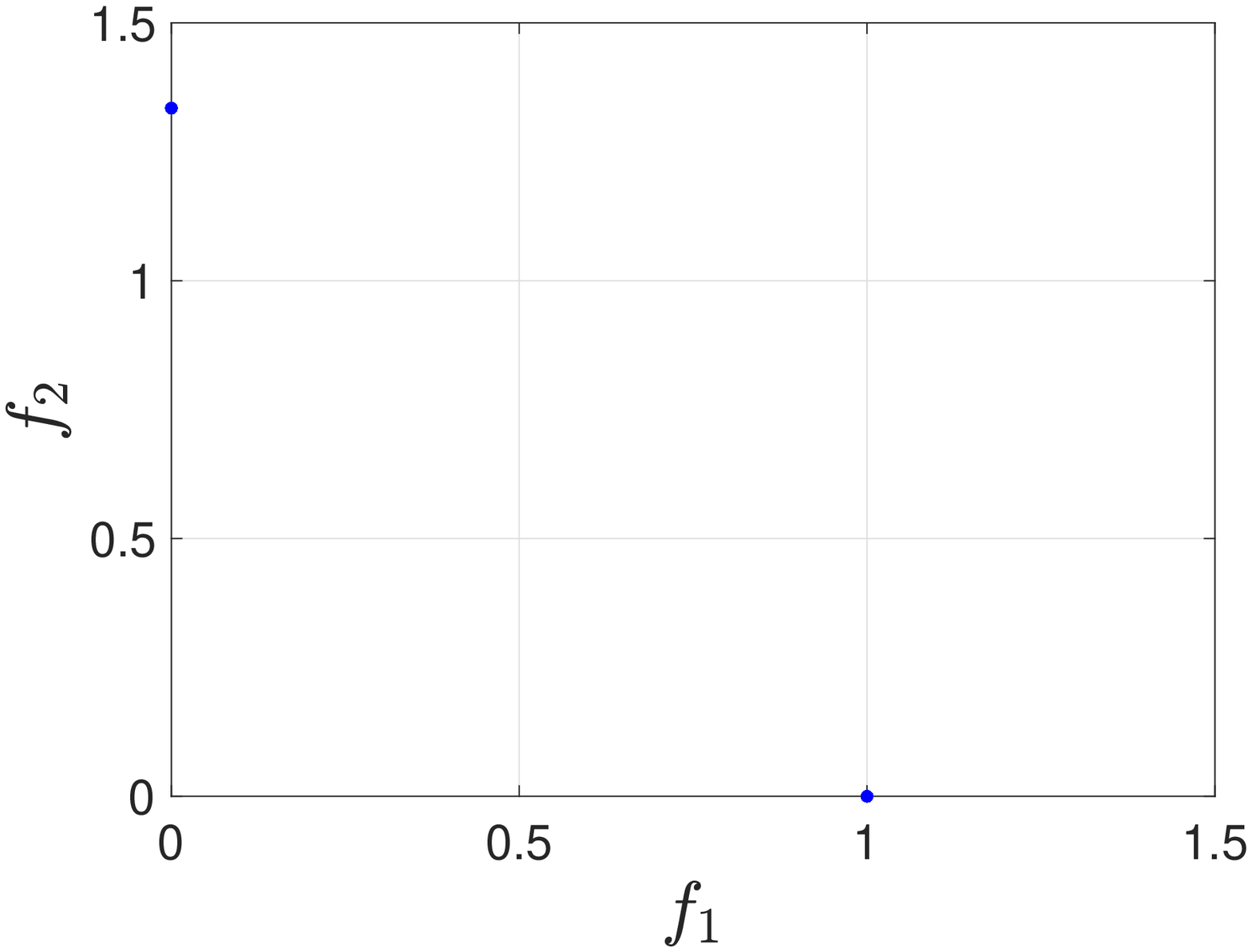}&
		\includegraphics[width=0.2\linewidth]{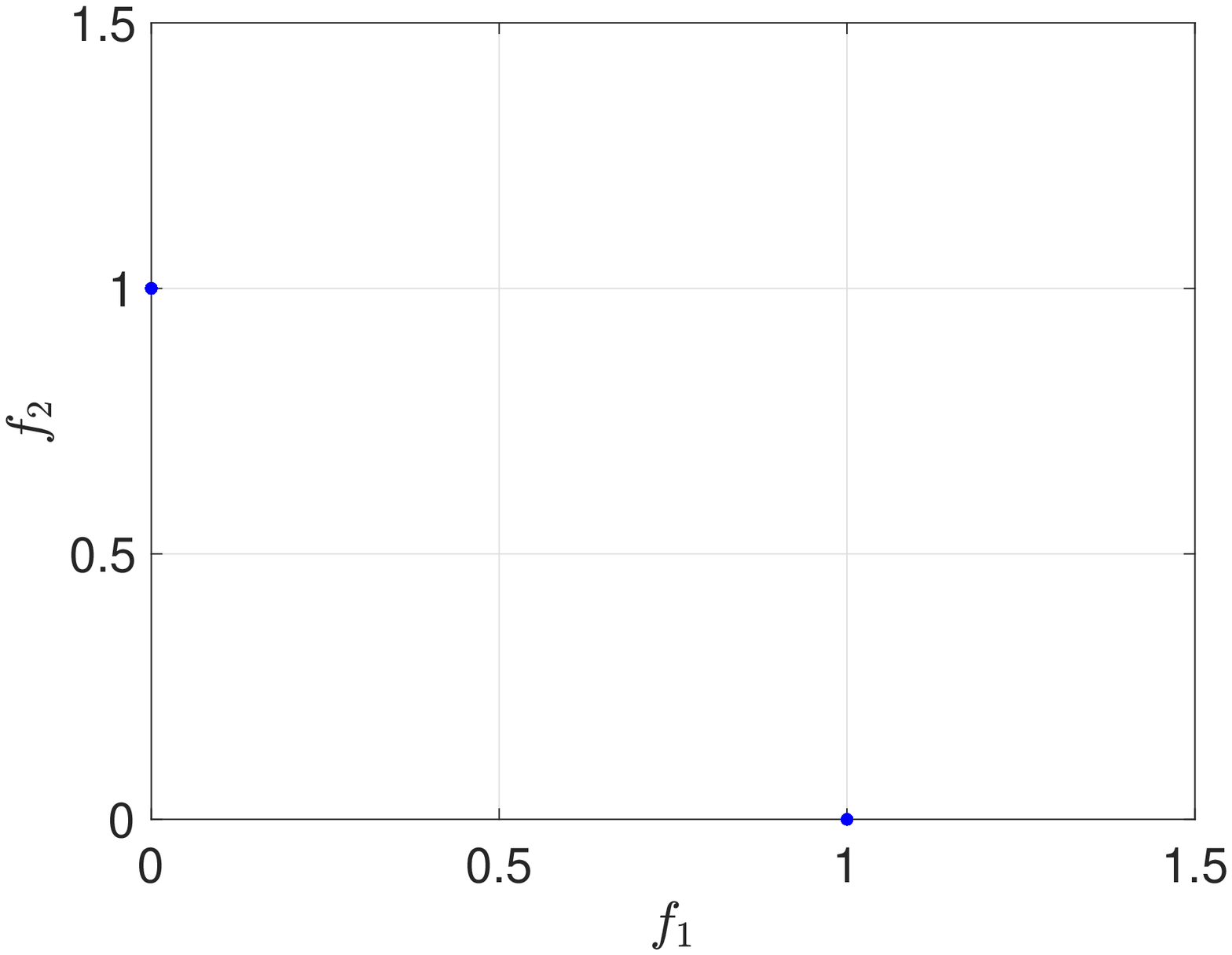}&
		\includegraphics[width=0.2\linewidth]{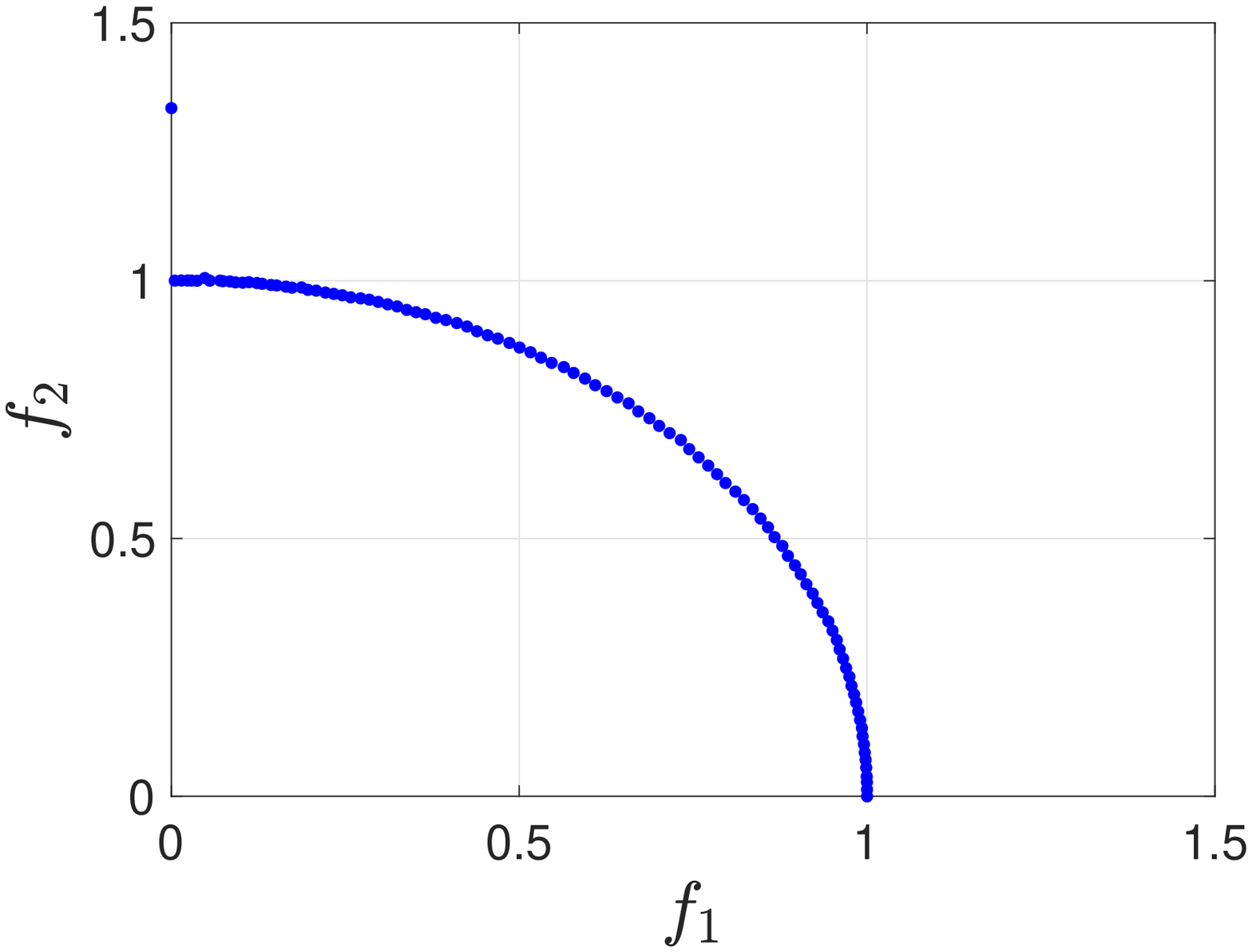}&
		\includegraphics[width=0.2\linewidth]{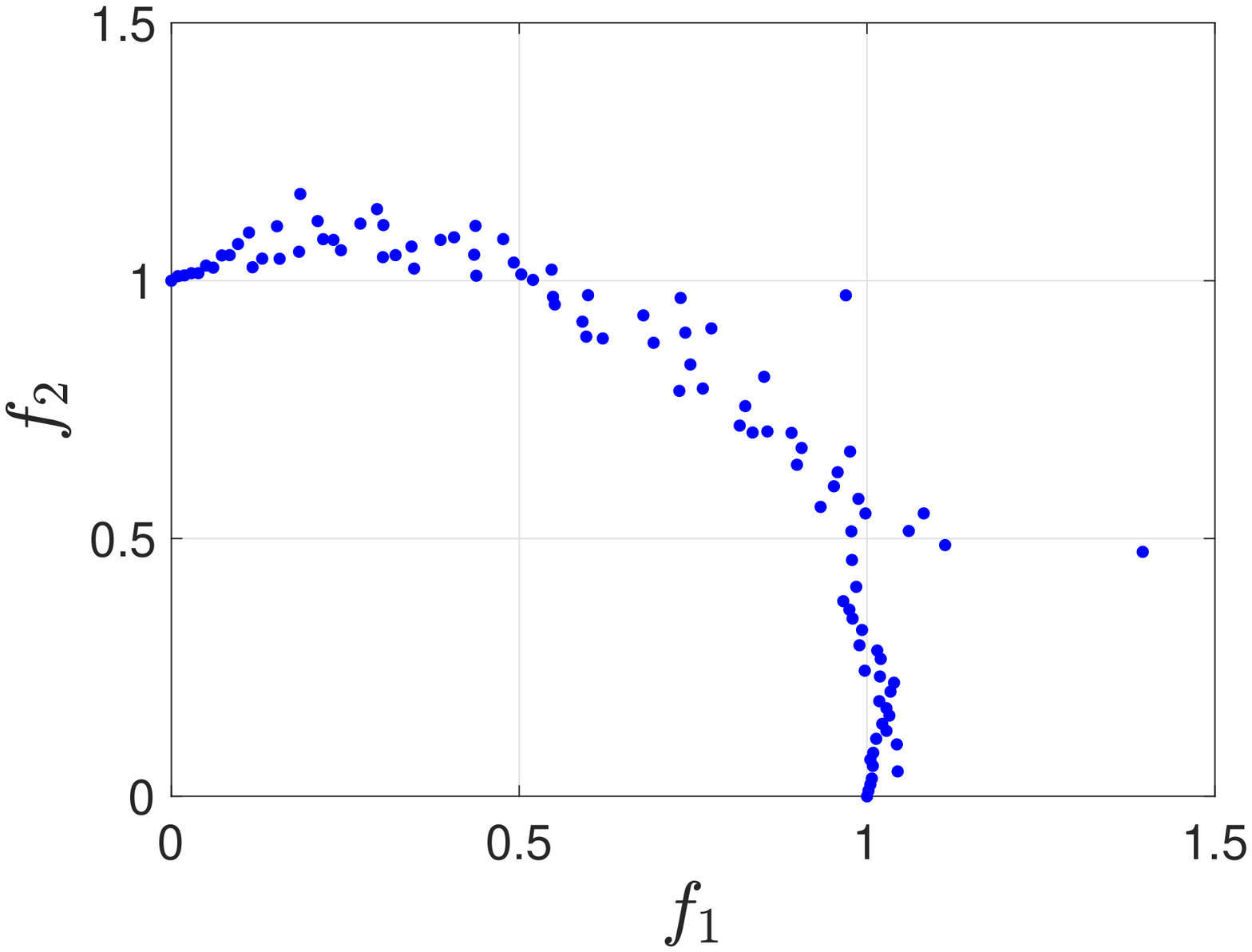}&
		\includegraphics[width=0.2\linewidth]{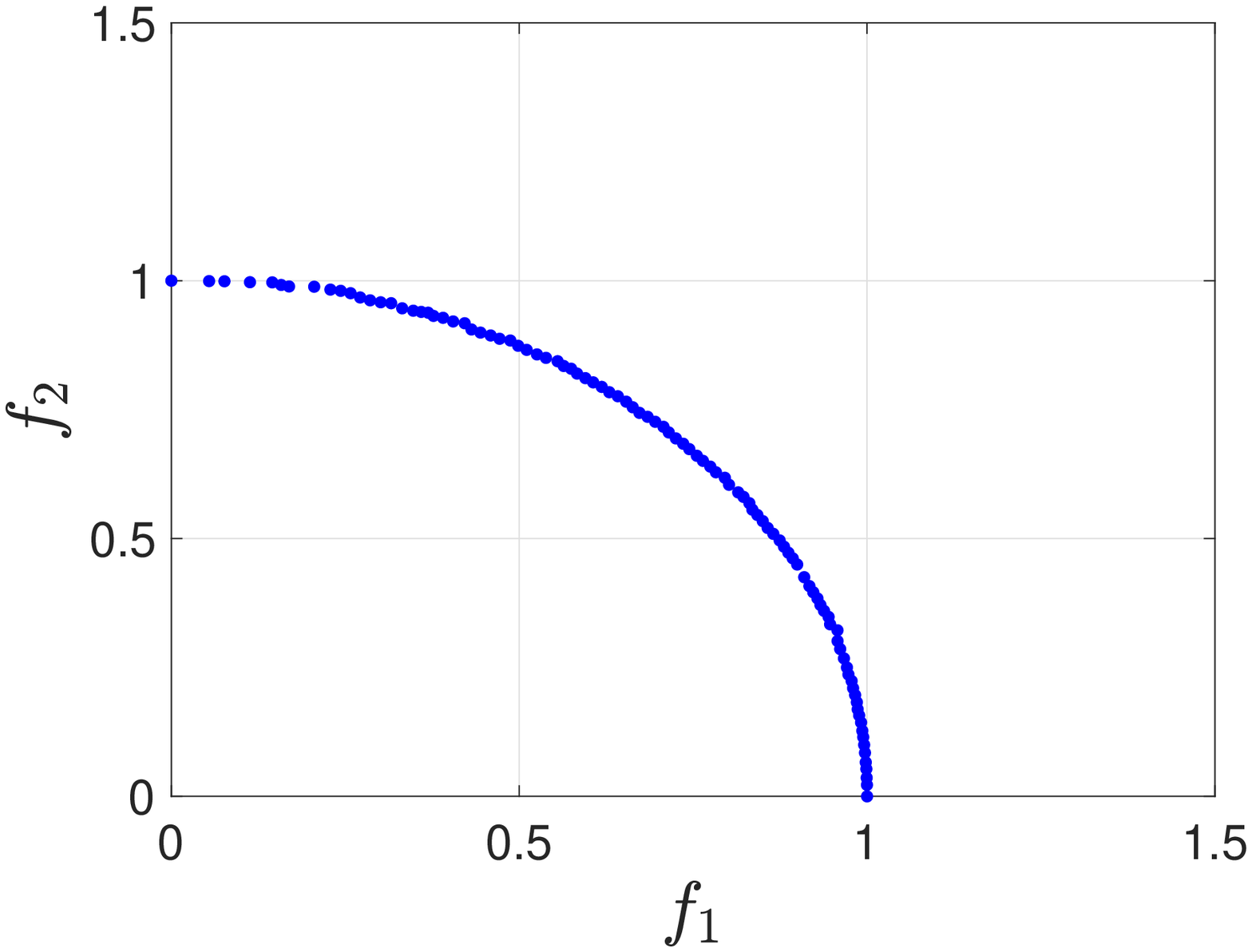}\\
		(a) MOEA/D & (b) ANSGA-III & (c) SPEA/R  & (d) RVEA & (e) AREA\\[-2mm]
	\end{tabular}
	\caption{PF approximation of MOP3 obtained by different algorithms.}
	\label{fig:mop3_pf}
	\vspace{-2mm}
\end{figure*}

Although AREA is not always the best algorithm on the MOP group, its robustness makes it on the top two list. Specifically, For the first five biobjective MOP instances, SPEA/R and AREA are the leading approaches: while SPEA/R wins on some problems, AREA wins on the others. For the two triobjective MOP instances, i.e. MOP6-7, RVEA shows the best performance, followed by AREA. The testing on this test suite, along with two visual examples shown in Figs.~\ref{fig:mop1_pf} and \ref{fig:mop3_pf}, therefore demonstrates that AREA is capable of dragging the population diversely toward the whole PF when some PF regions are significantly harder to approximate than others. This is actually one of the advantages over MOEA/D by placing the reference set in opposition to the first quadrant. RVEA seems to have slow convergence for the MOP group illustrated in Figs.~\ref{fig:mop1_pf} and \ref{fig:mop3_pf}. 

\begin{figure*}[!tp]
	\centering
	\begin{tabular}{@{}c@{}c@{}c@{}c@{}c}
		\includegraphics[width=0.2\linewidth]{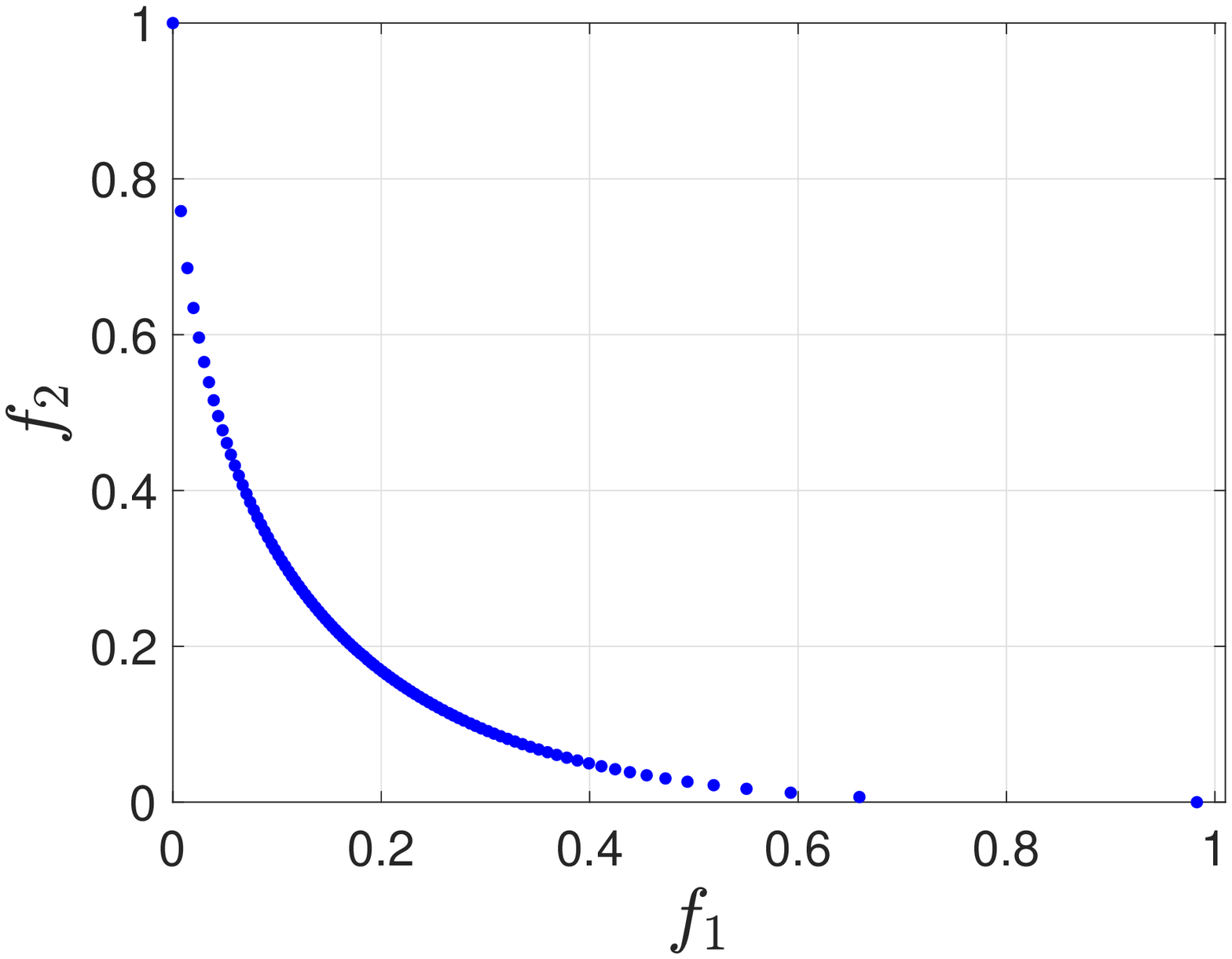}&
		\includegraphics[width=0.2\linewidth]{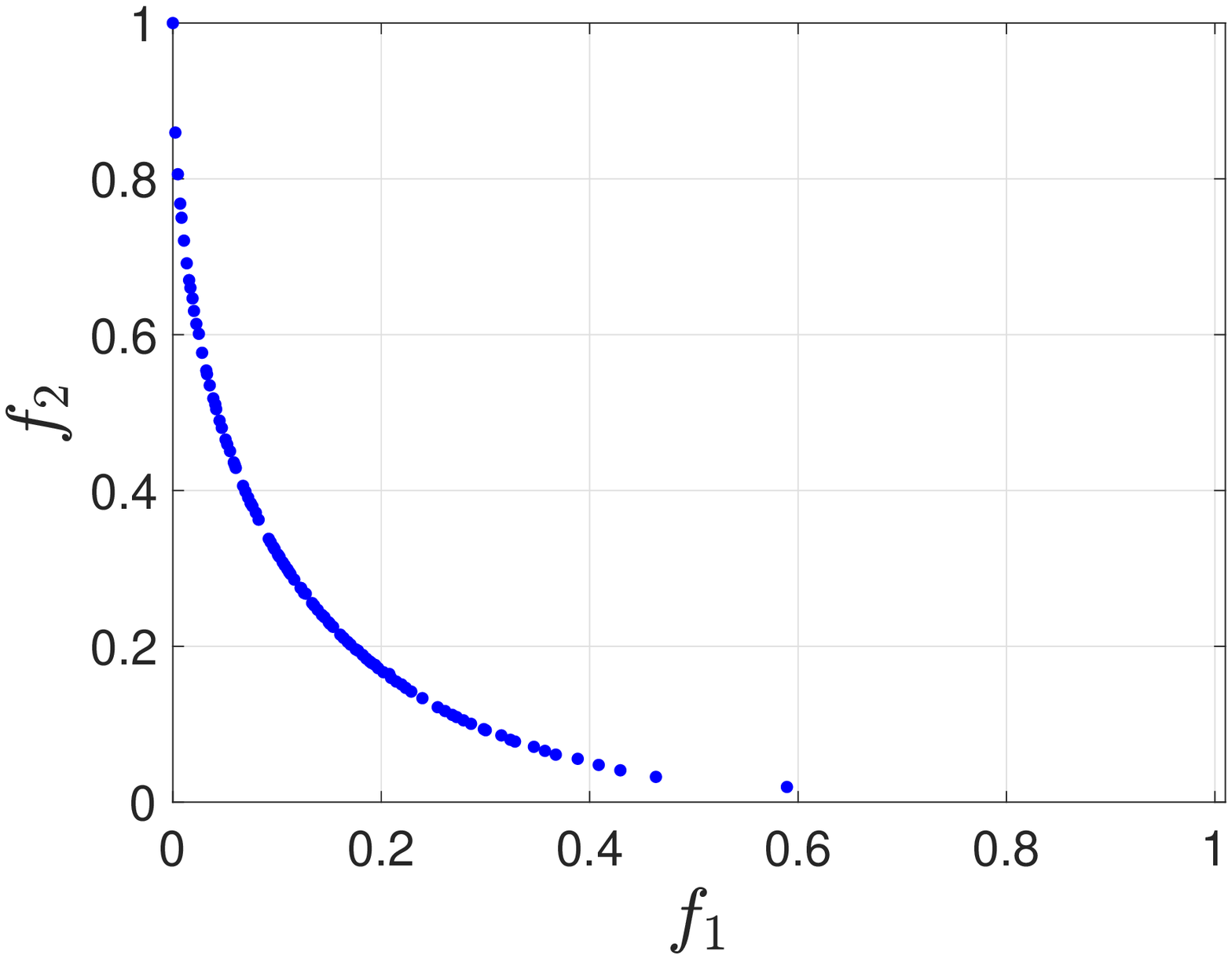}&
		\includegraphics[width=0.2\linewidth]{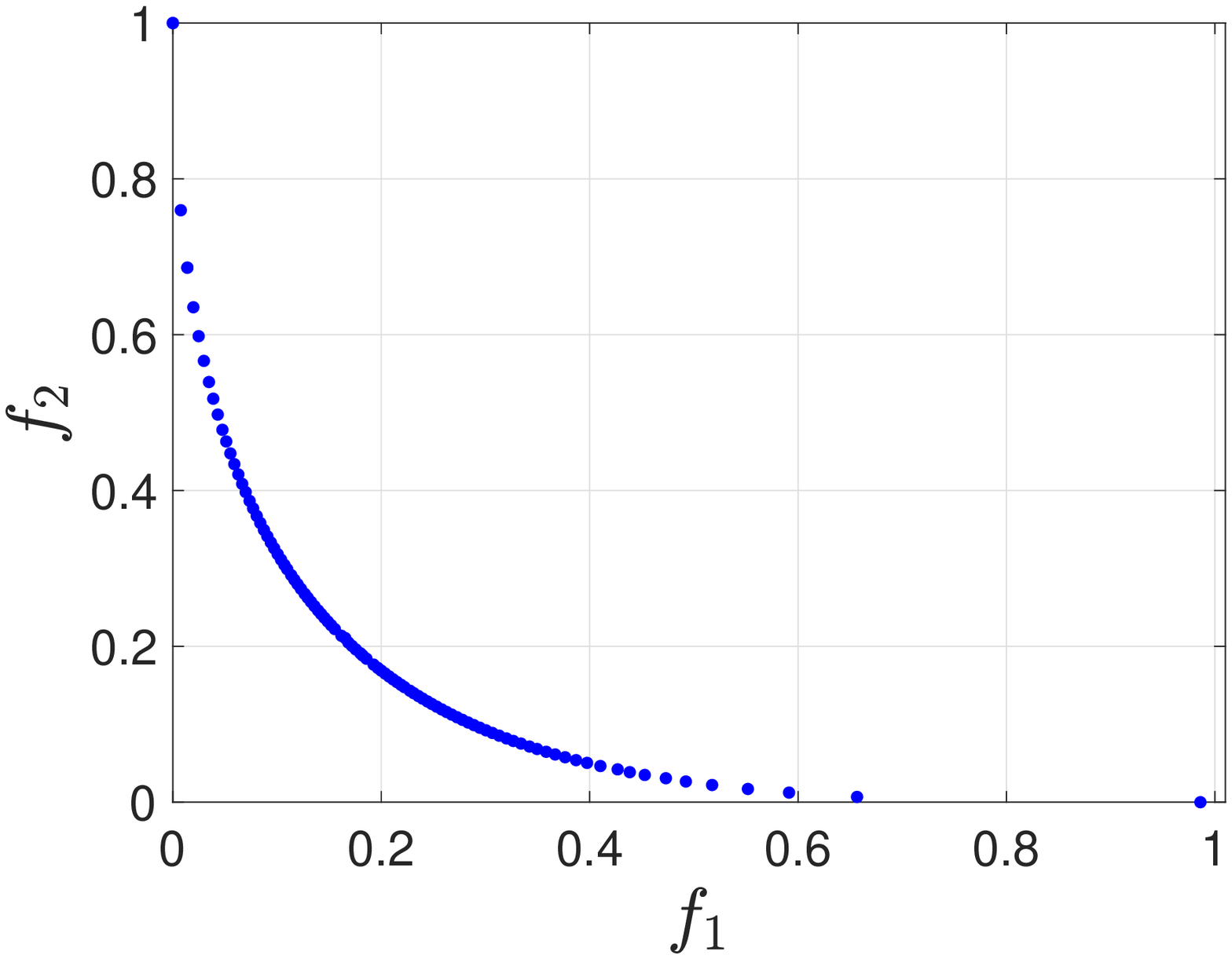}&
		\includegraphics[width=0.2\linewidth]{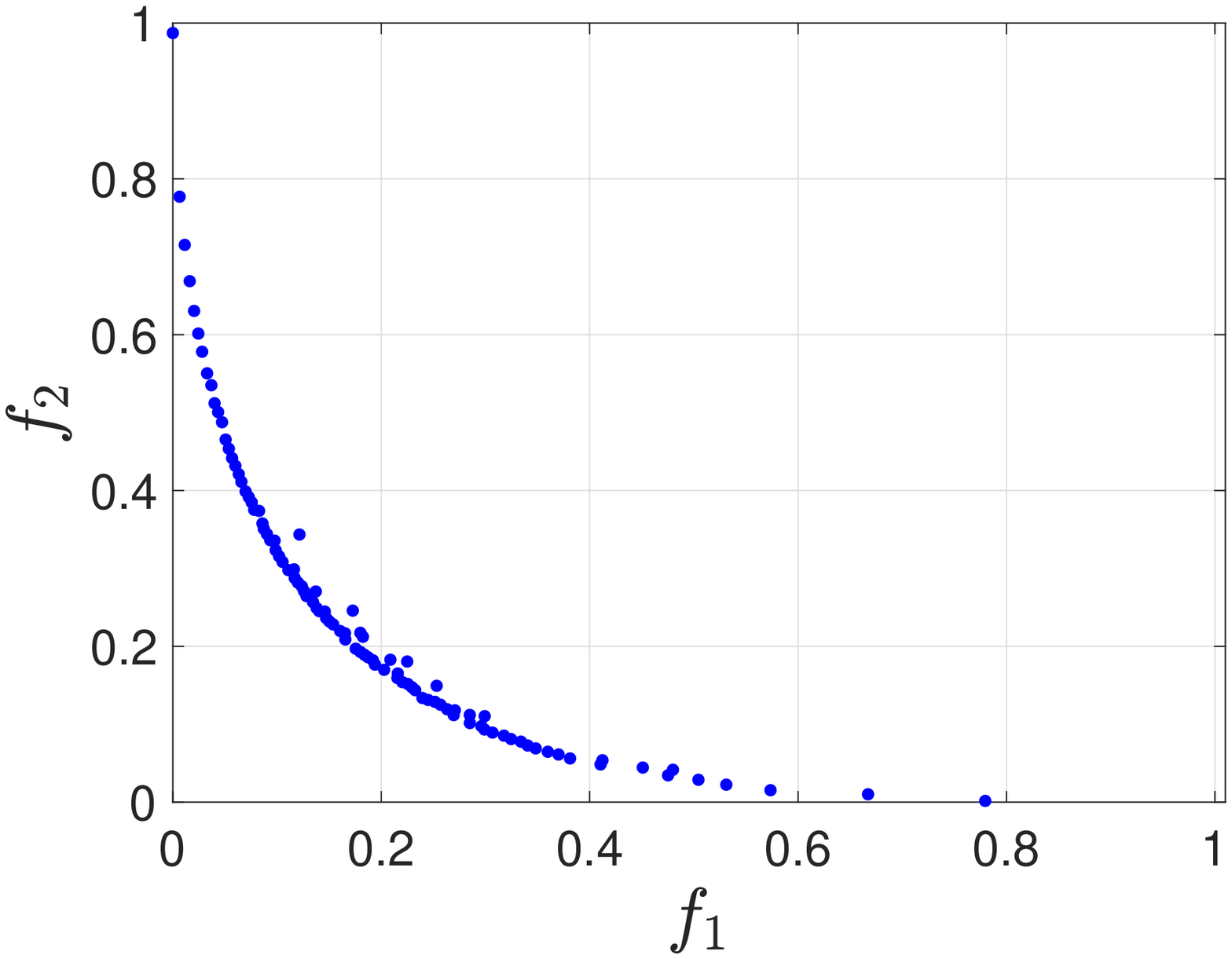}&
		\includegraphics[width=0.2\linewidth]{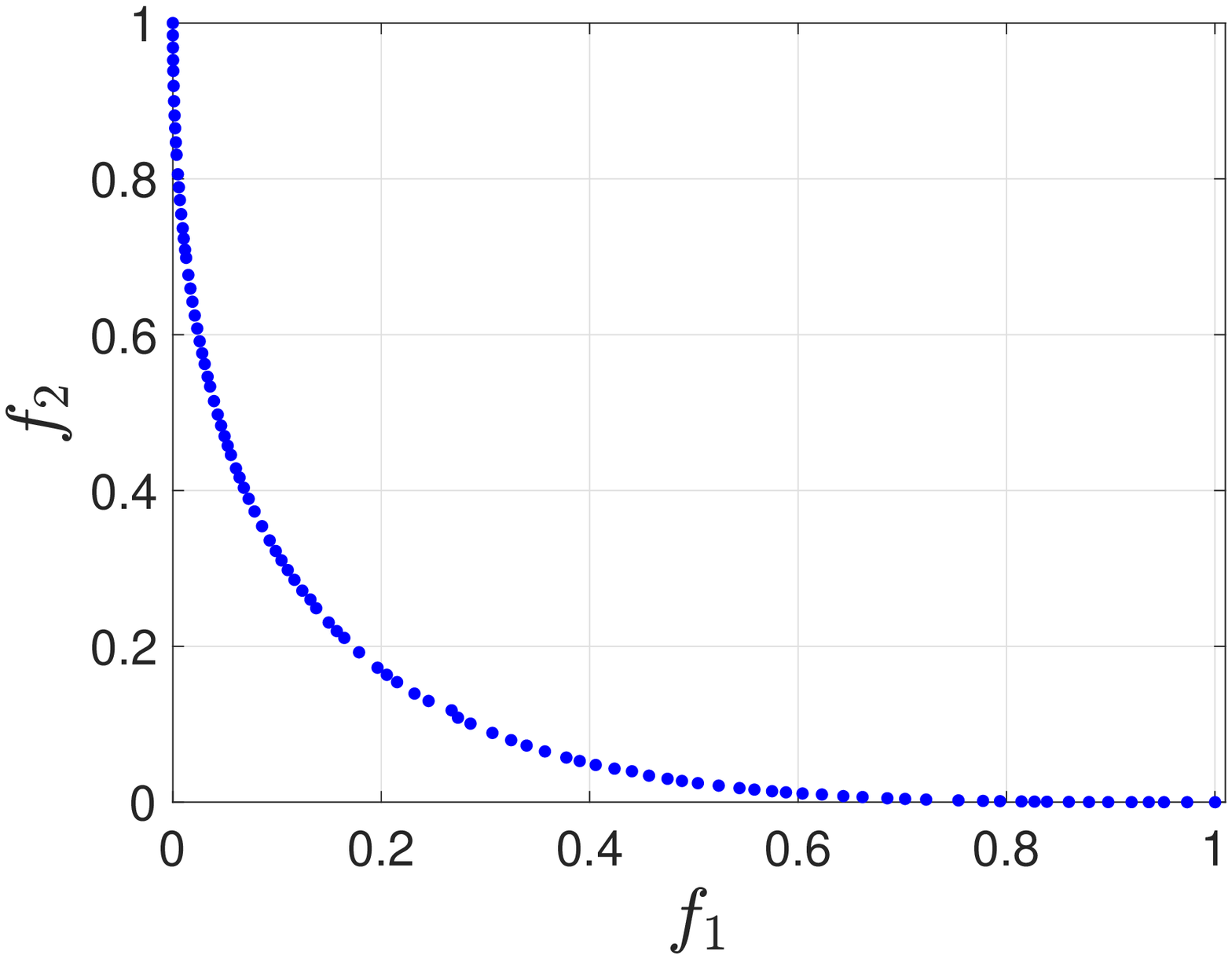}\\
		(a) MOEA/D & (b) ANSGA-III & (c) SPEA/R  & (d) RVEA & (e) AREA\\[-2mm]
	\end{tabular}
	\caption{PF approximation of F1 obtained by different algorithms.}
	\label{fig:f1_pf}
	\vspace{-2mm}
\end{figure*}
\begin{figure*}[!tp]
	\centering
	\begin{tabular}{@{}c@{}c@{}c@{}c@{}c}
		\includegraphics[width=0.2\linewidth]{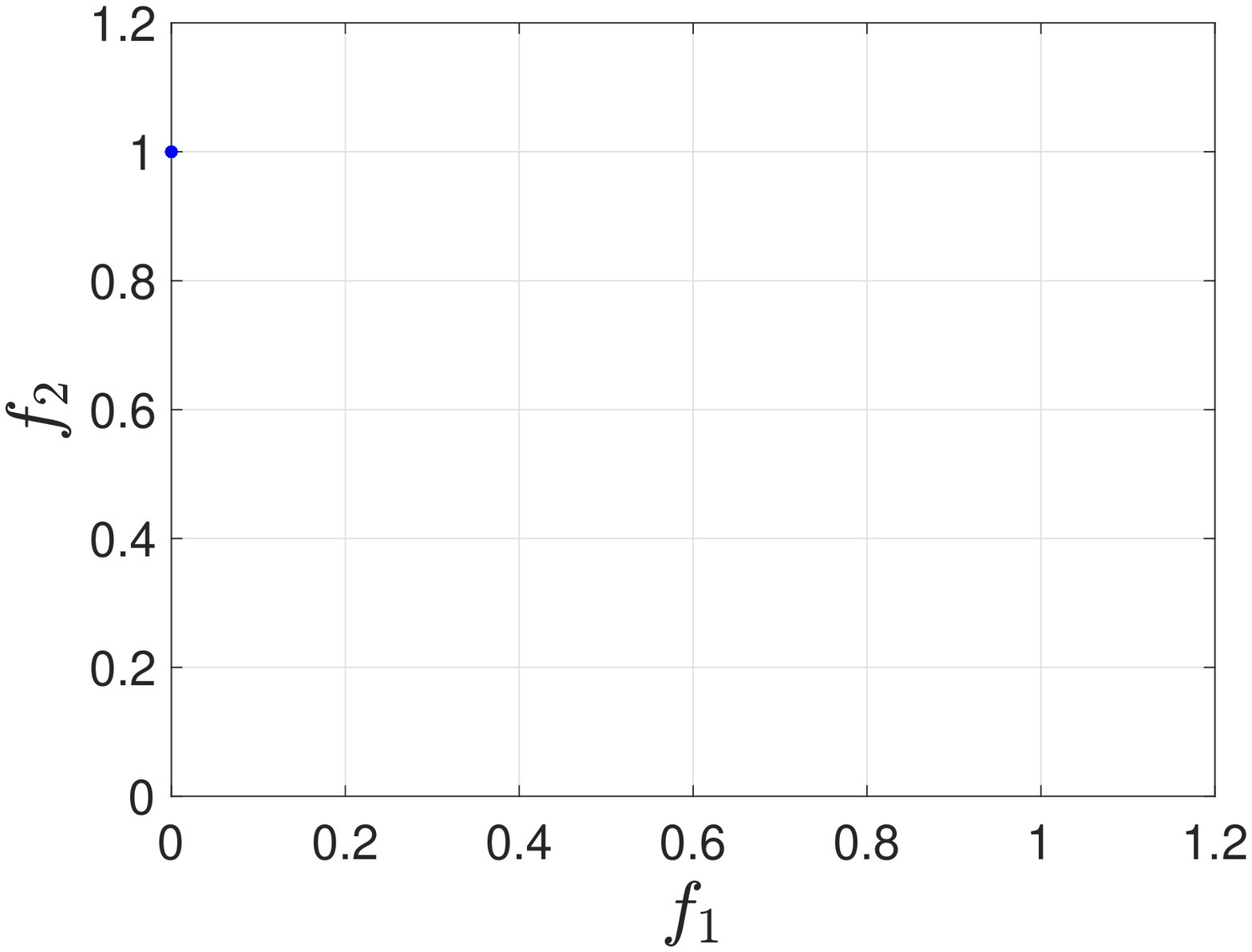}&
		\includegraphics[width=0.2\linewidth]{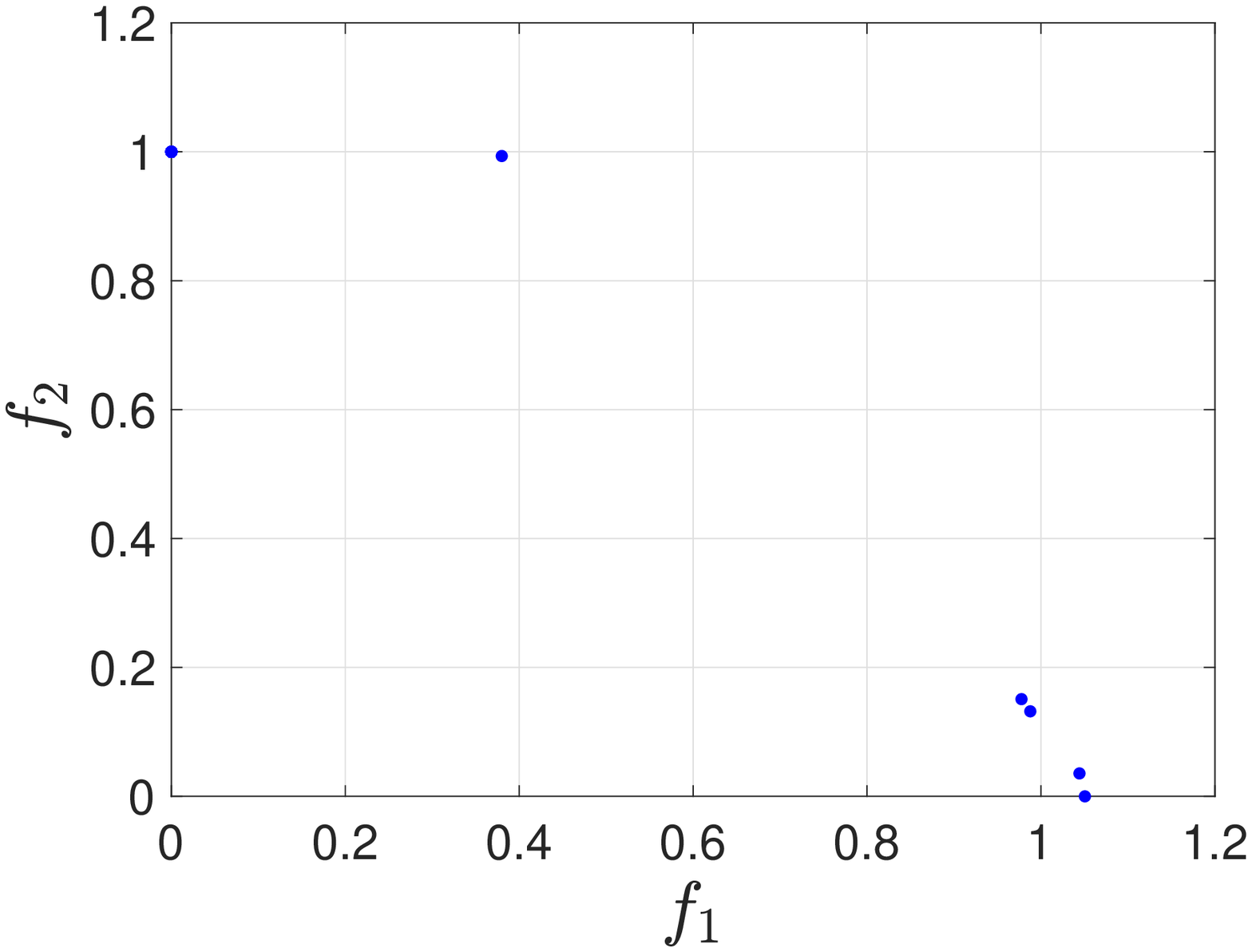}&
		\includegraphics[width=0.2\linewidth]{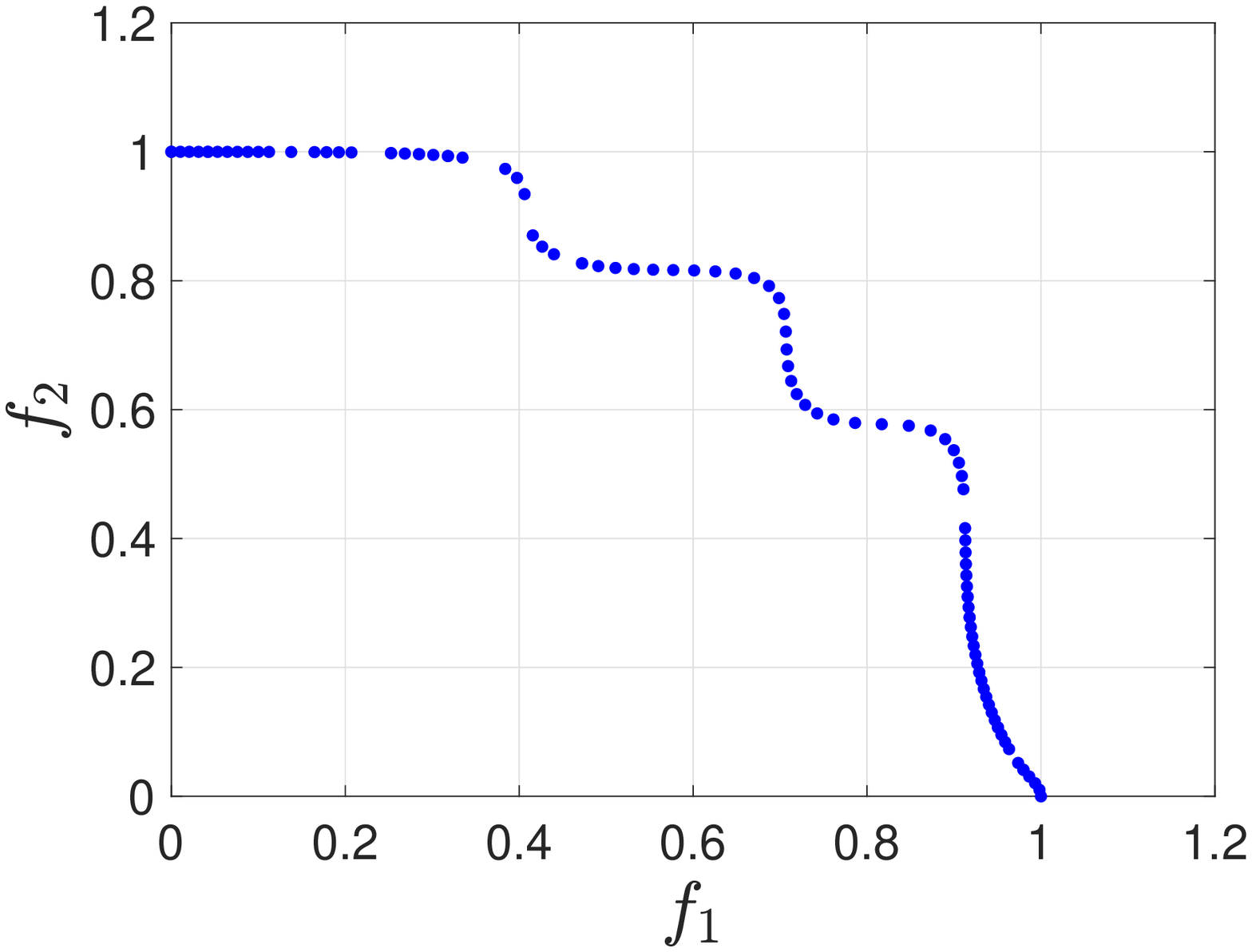}&
		\includegraphics[width=0.2\linewidth]{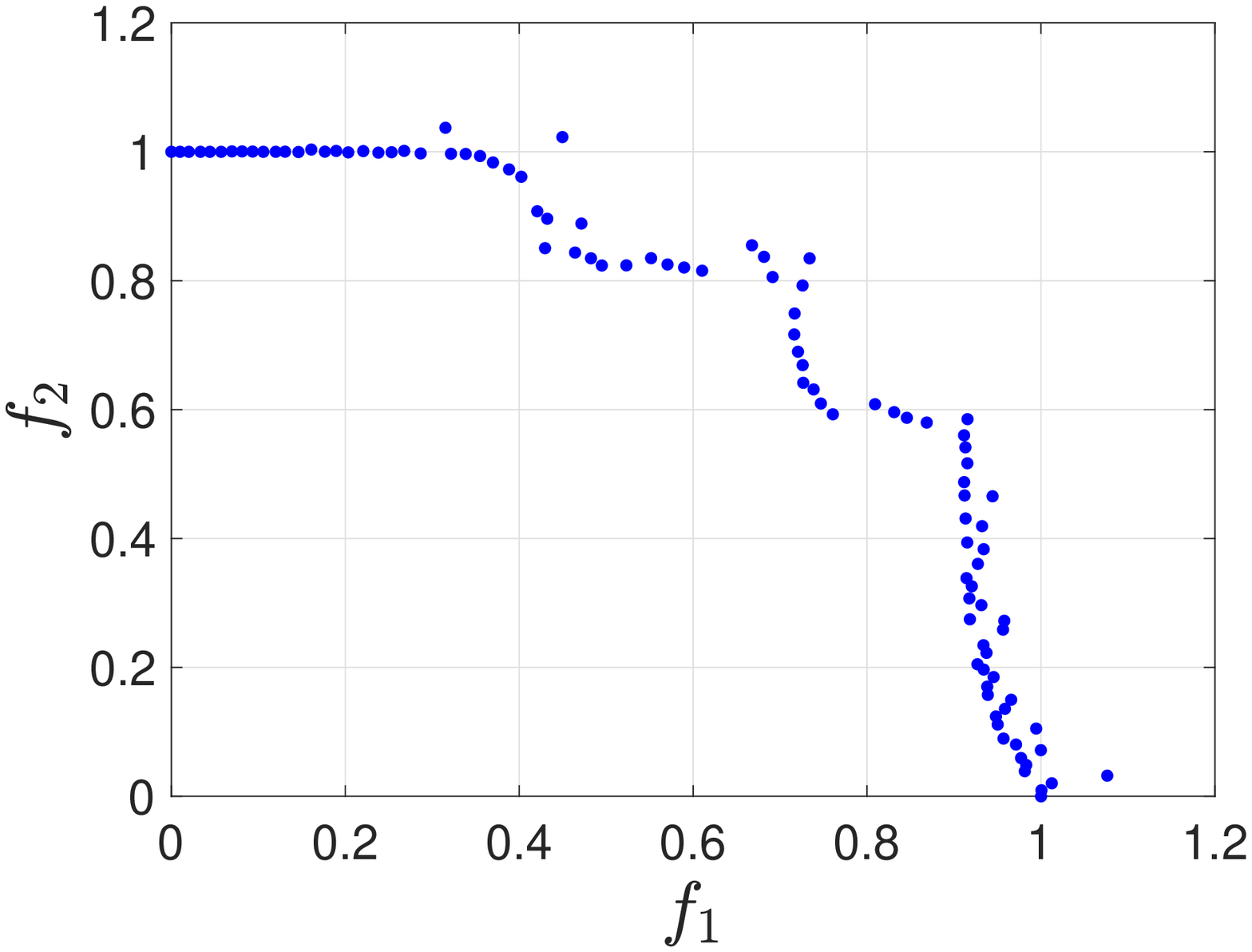}&
		\includegraphics[width=0.2\linewidth]{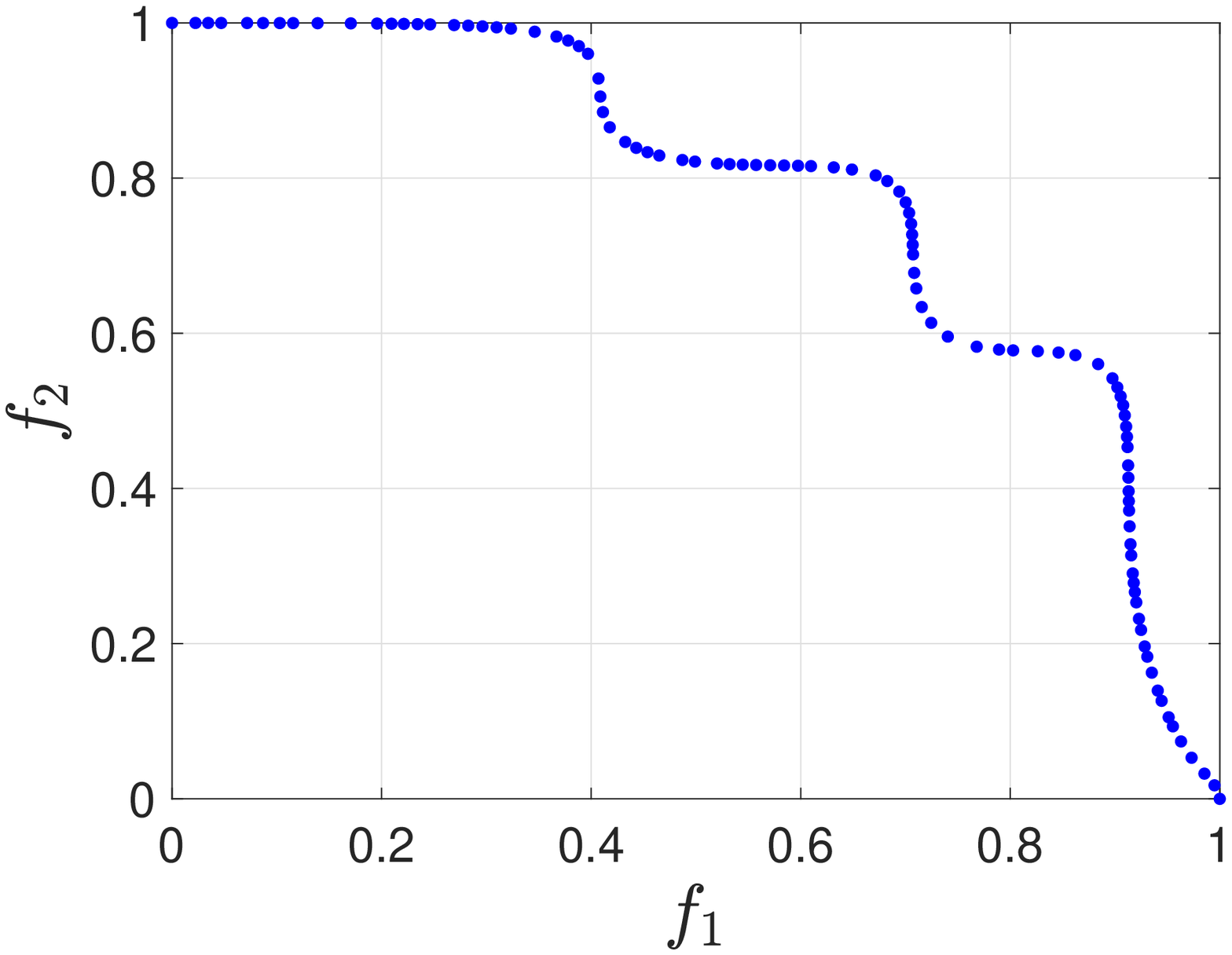}\\
		(a) MOEA/D & (b) ANSGA-III & (c) SPEA/R  & (d) RVEA & (e) AREA\\[-2mm]
	\end{tabular}
	\caption{PF approximation of F4 obtained by different algorithms.}
	\label{fig:f4_pf}
	\vspace{-2mm}
\end{figure*}
\begin{figure*}[!tp]
	\centering
	\begin{tabular}{@{}c@{}c@{}c@{}c@{}c}
		\includegraphics[width=0.2\linewidth]{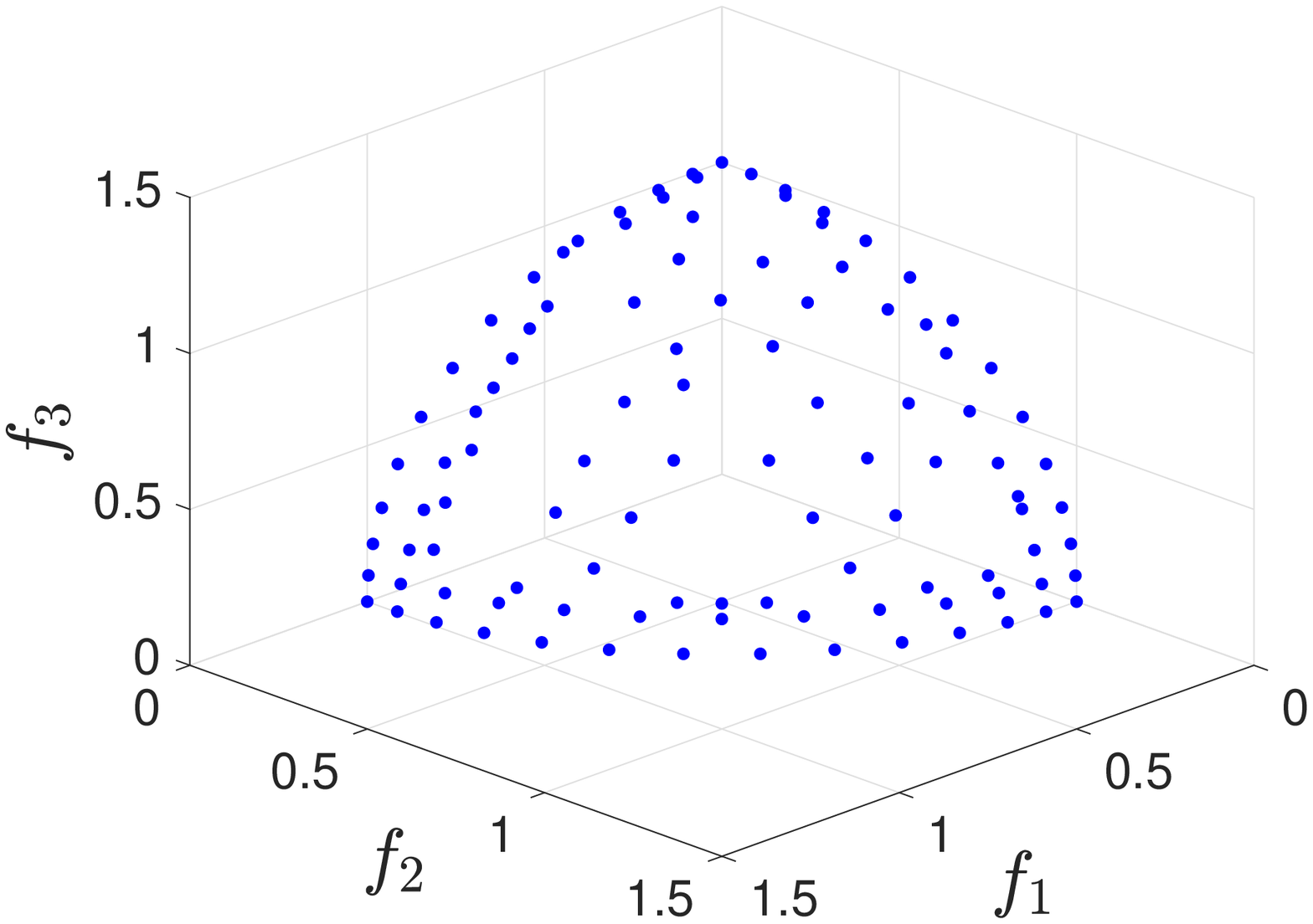}&
		\includegraphics[width=0.2\linewidth]{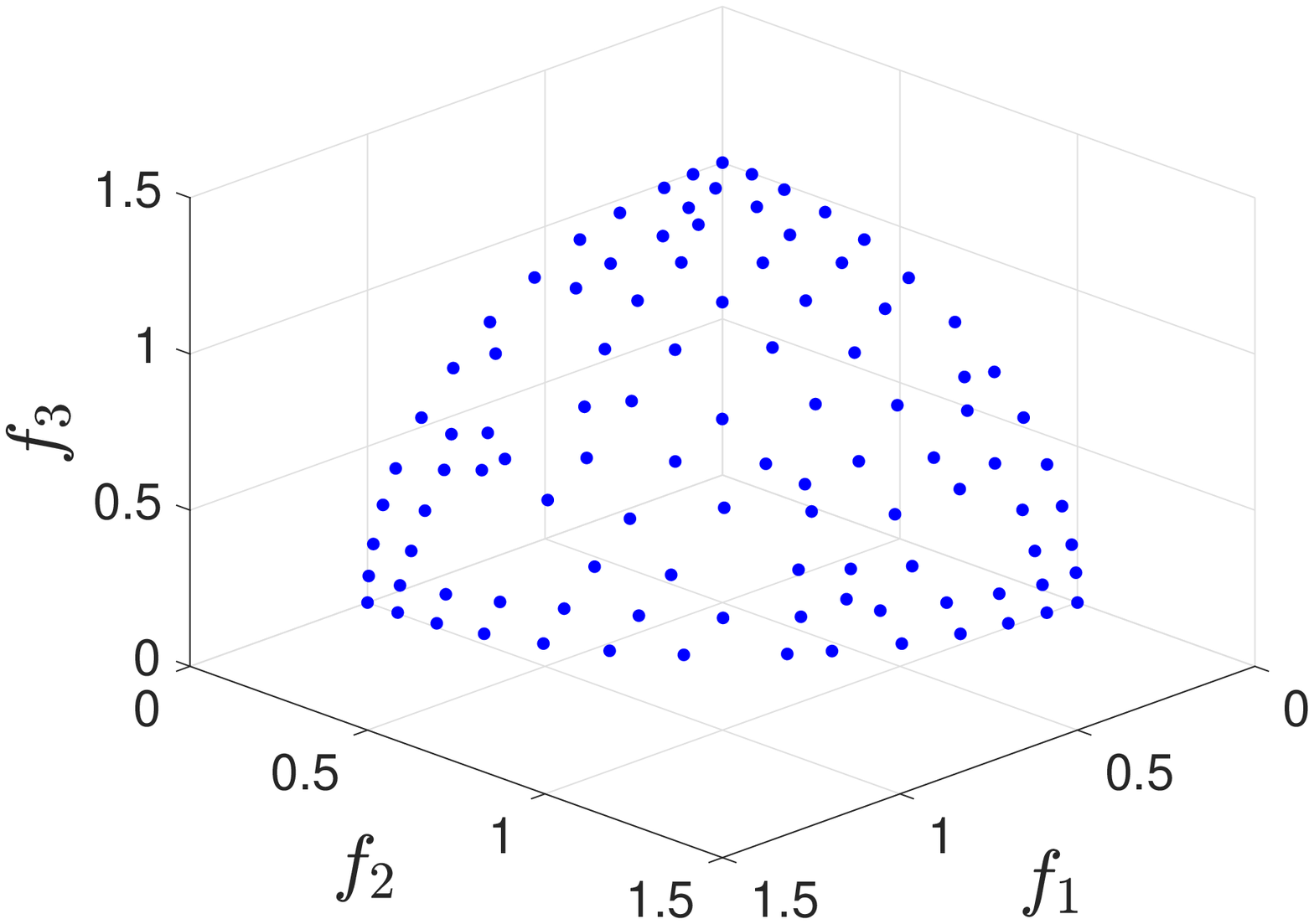}&
		\includegraphics[width=0.2\linewidth]{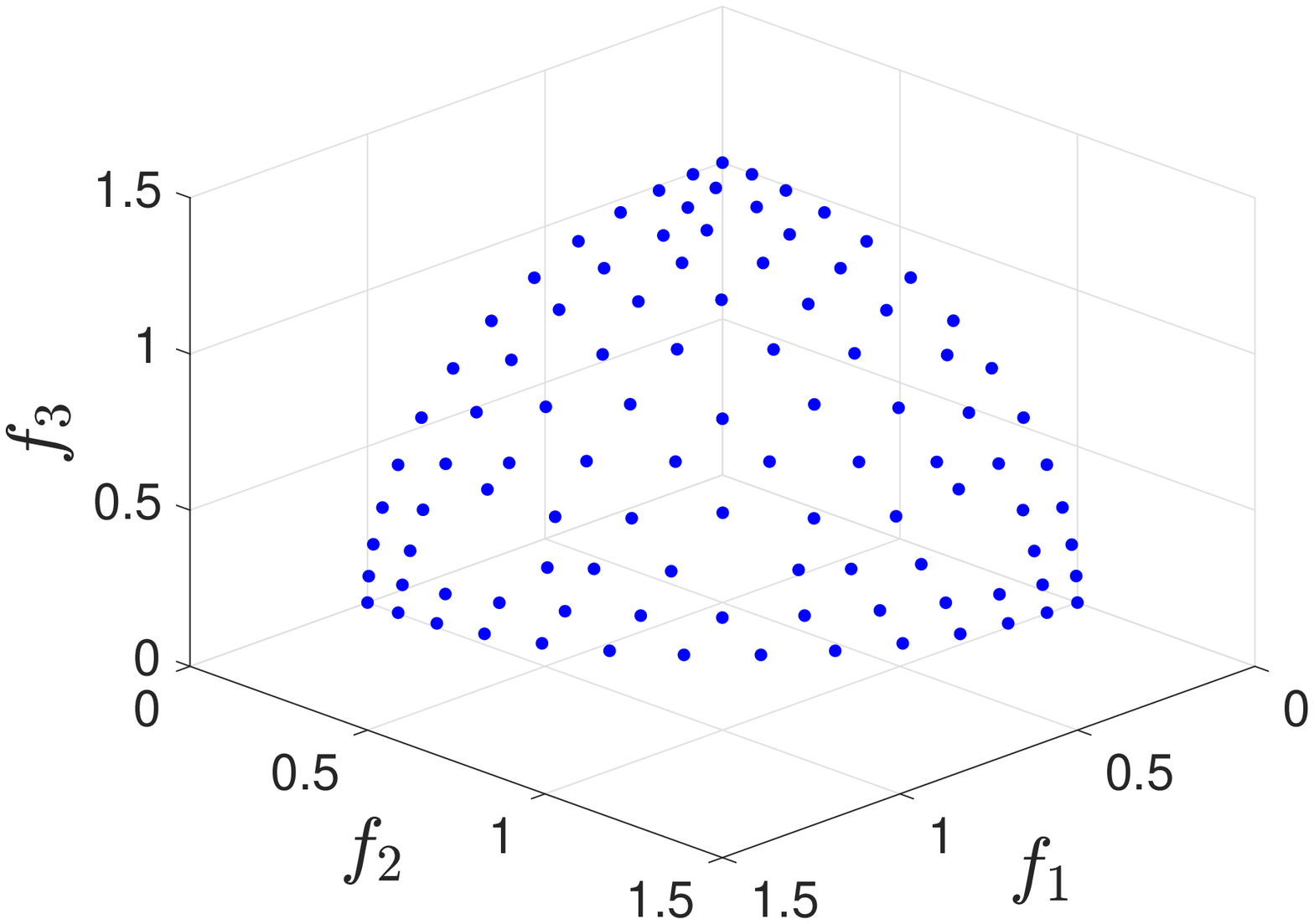}&
		\includegraphics[width=0.2\linewidth]{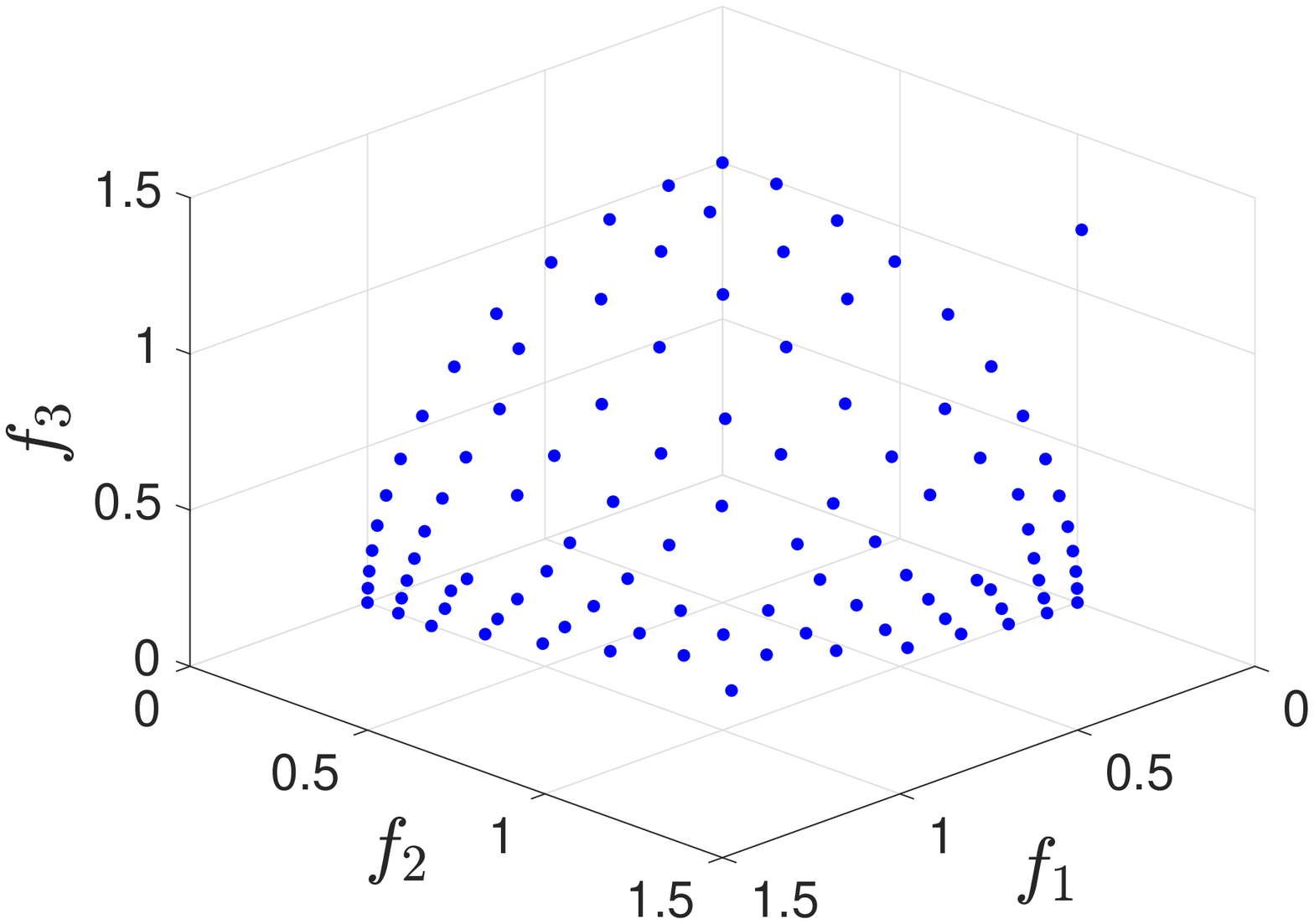}&
		\includegraphics[width=0.2\linewidth]{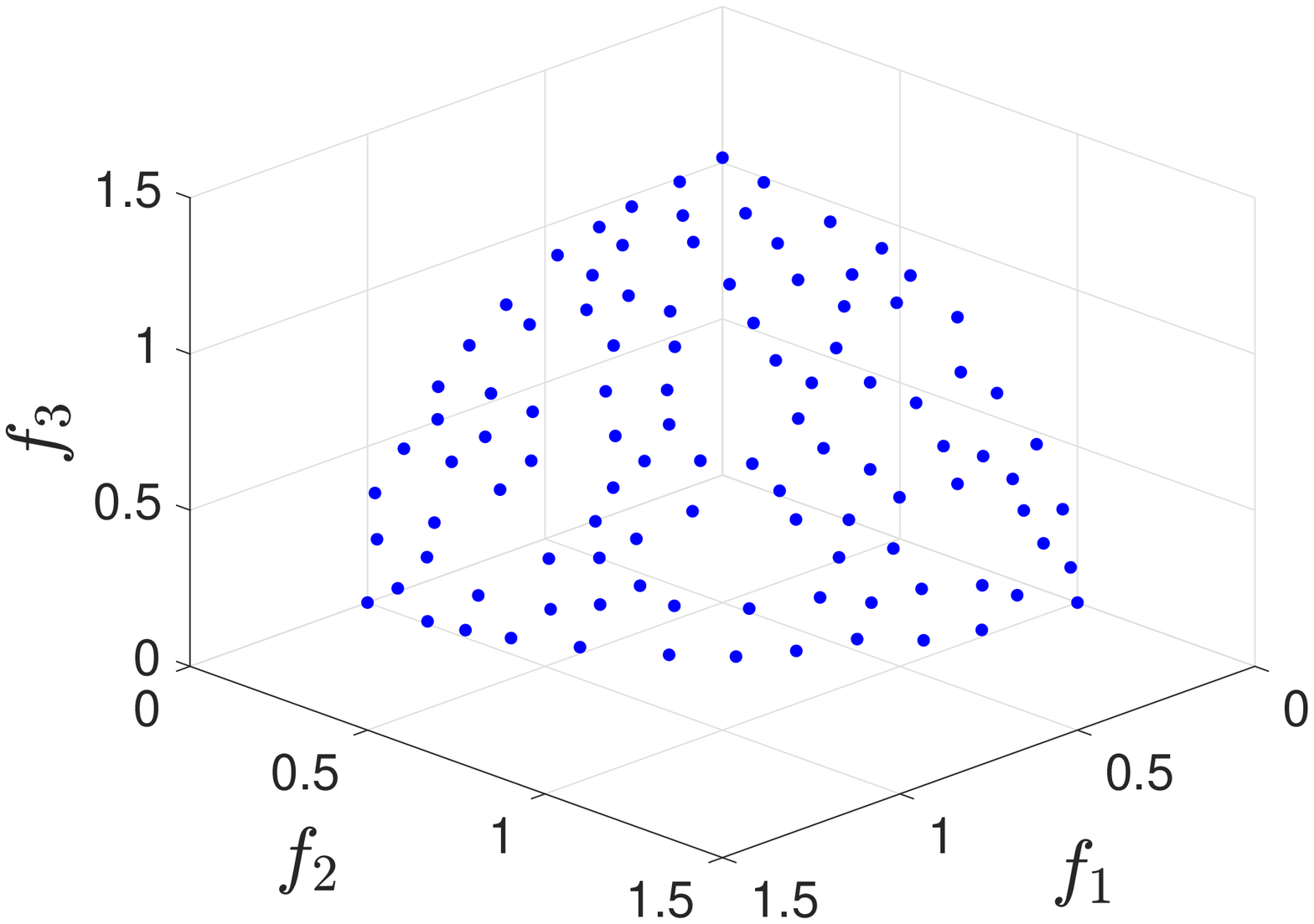}\\
		(a) MOEA/D & (b) ANSGA-III & (c) SPEA/R  & (d) RVEA & (e) AREA\\[-2mm]
	\end{tabular}
	\caption{PF approximation of F5 obtained by different algorithms.}
	\label{fig:f5_pf}
	\vspace{-2mm}
\end{figure*}
\begin{figure*}[!t]
	\centering
	\begin{tabular}{@{}c@{}c@{}c@{}c@{}c}
		\includegraphics[width=0.2\linewidth]{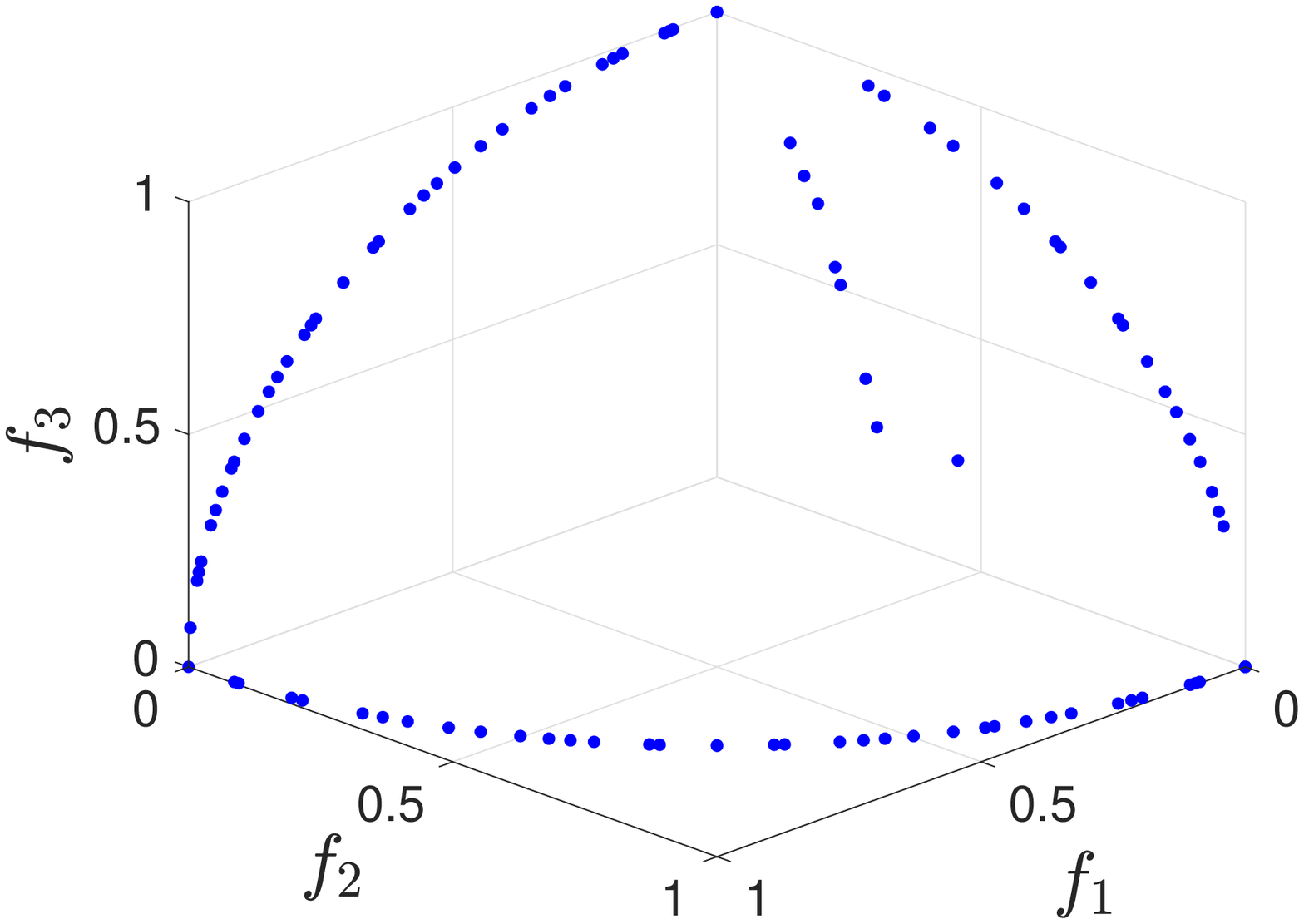}&
		\includegraphics[width=0.2\linewidth]{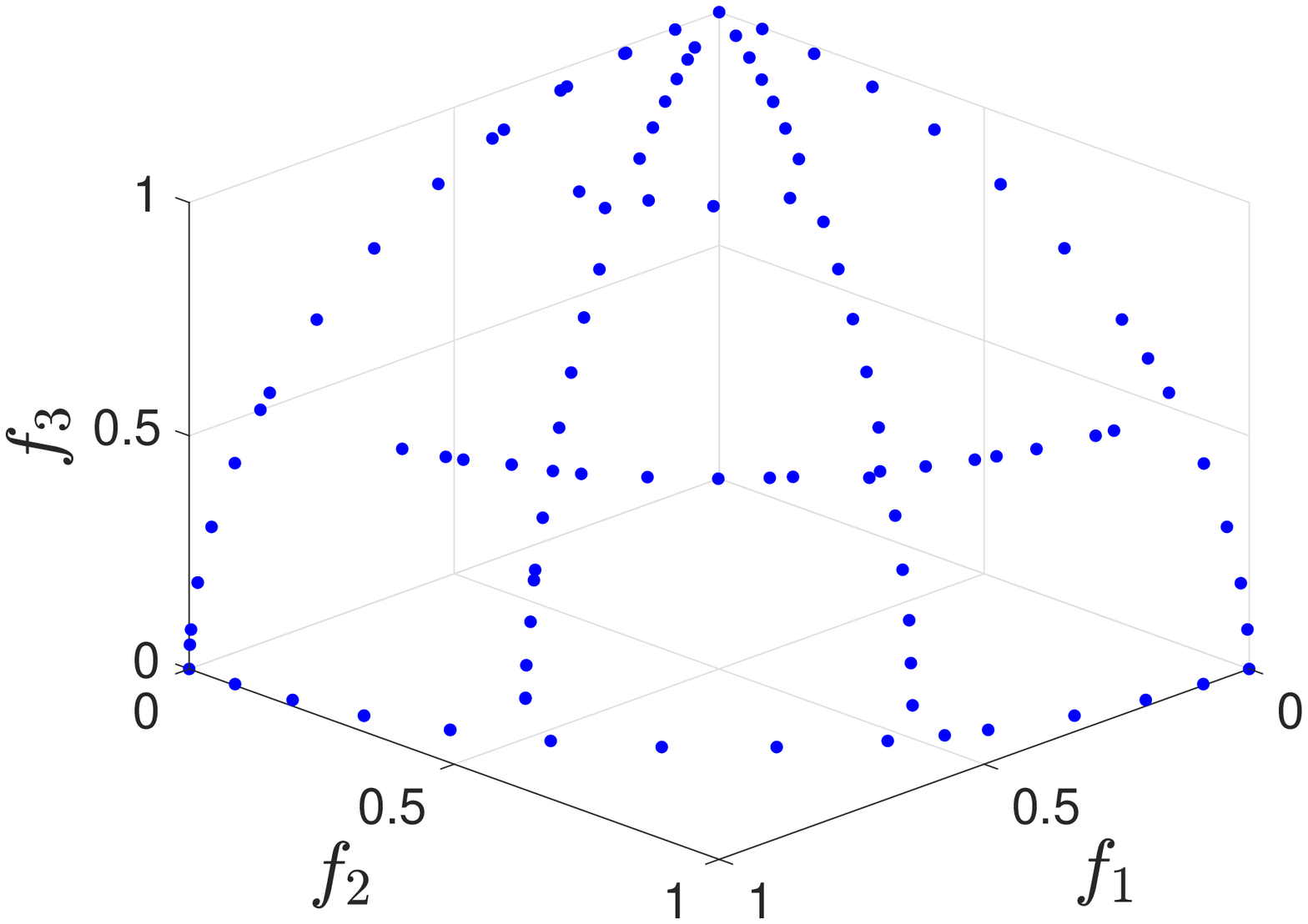}&
		\includegraphics[width=0.2\linewidth]{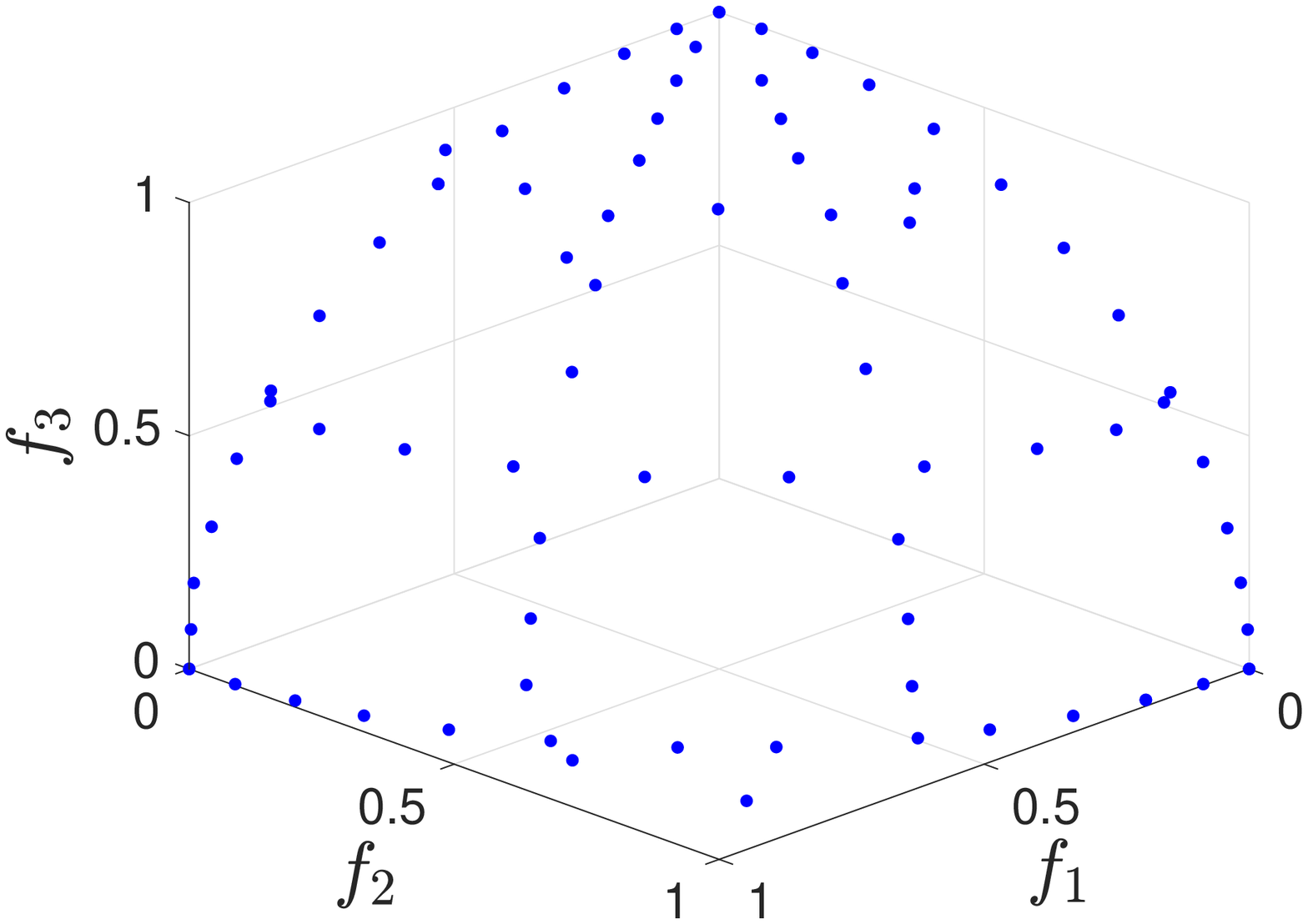}&
		\includegraphics[width=0.2\linewidth]{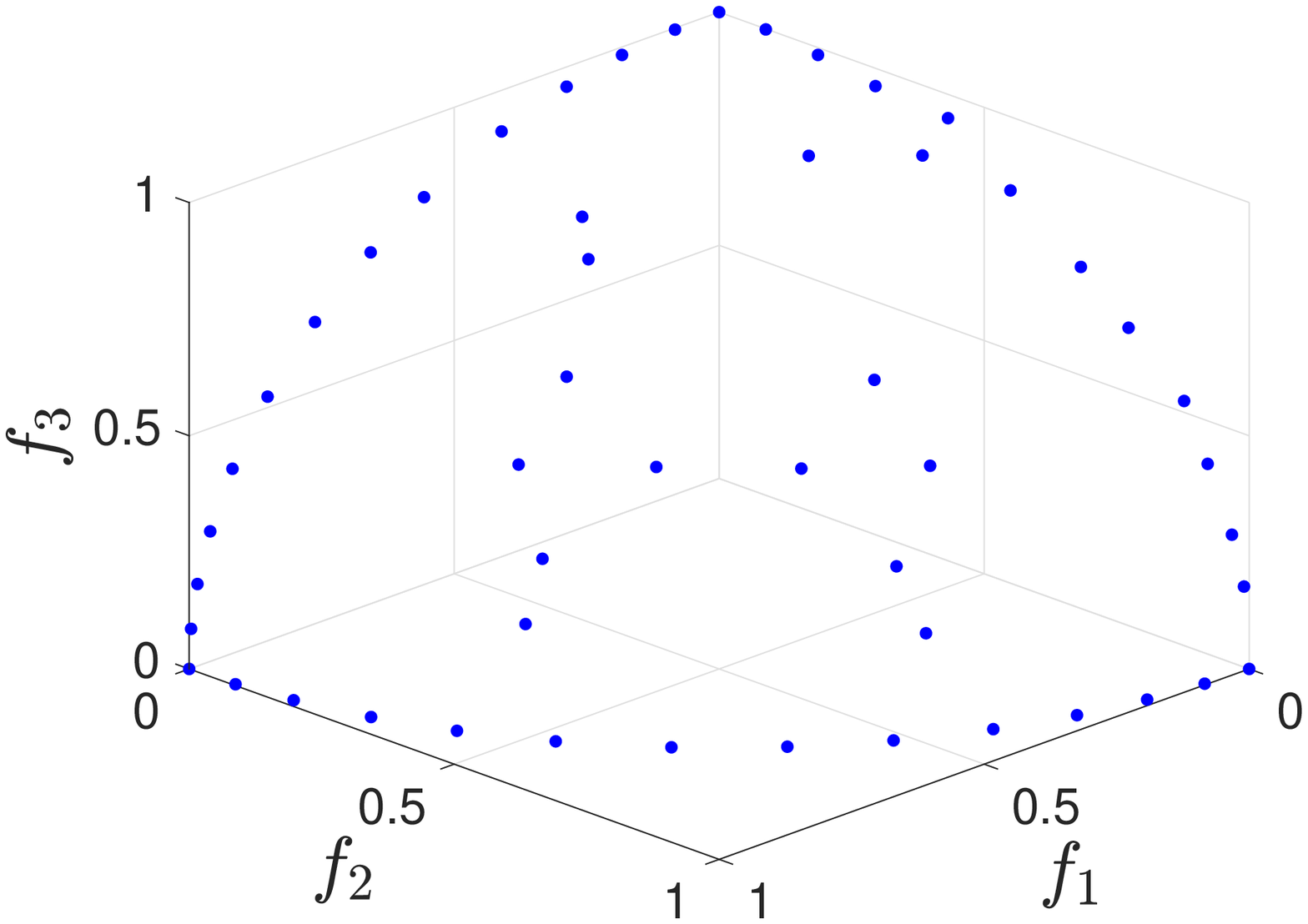}&
		\includegraphics[width=0.2\linewidth]{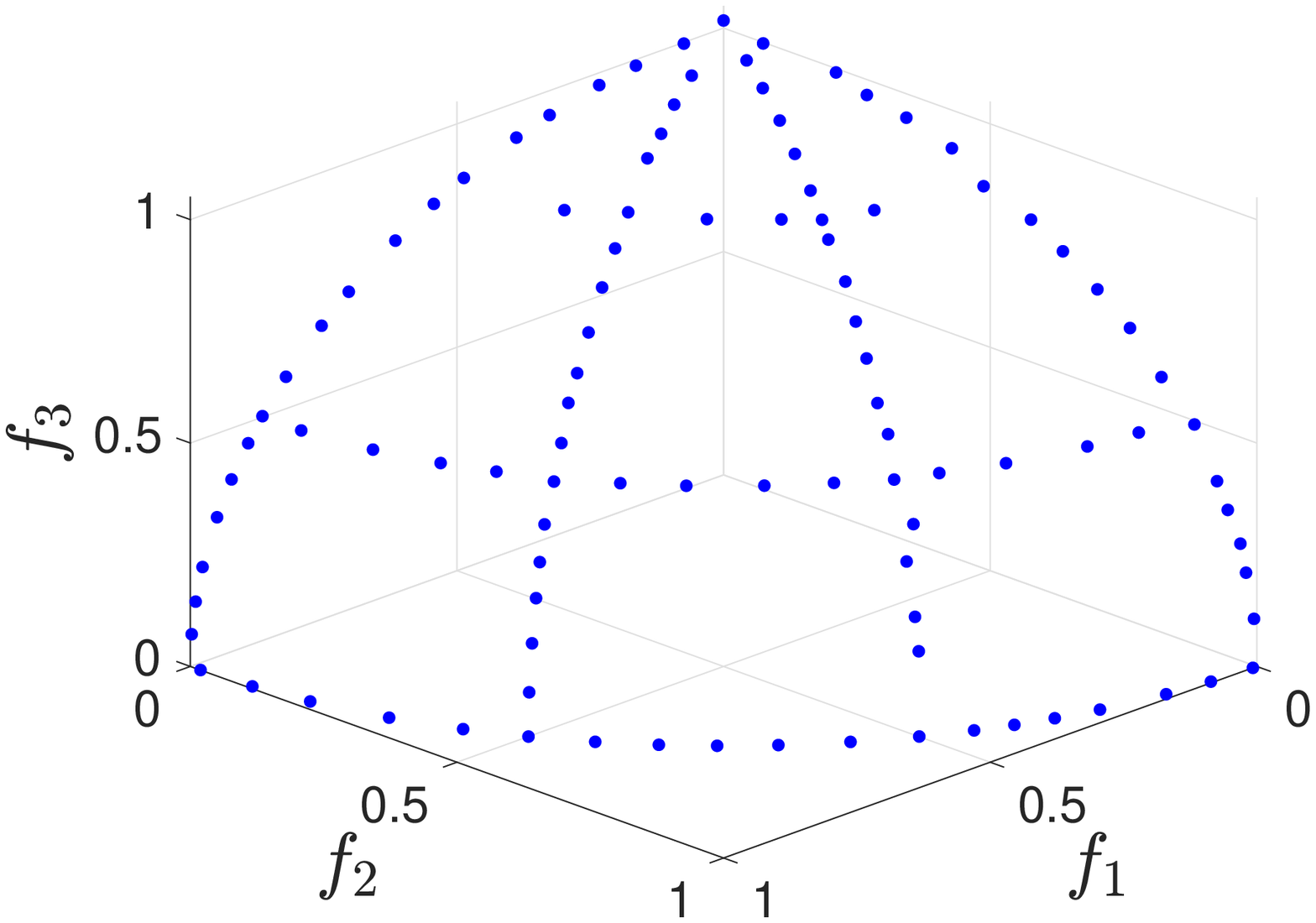}\\
		(a) MOEA/D & (b) ANSGA-III & (c) SPEA/R  & (d) RVEA & (e) AREA\\[-2mm]
	\end{tabular}
	\caption{PF approximation of F6 obtained by different algorithms.}
	\label{fig:f6_pf}
	\vspace{-2mm}
\end{figure*}

\begin{figure*}[!t]
	\centering
	\begin{tabular}{@{}c@{}c@{}c@{}c@{}c}
		\includegraphics[width=0.2\linewidth]{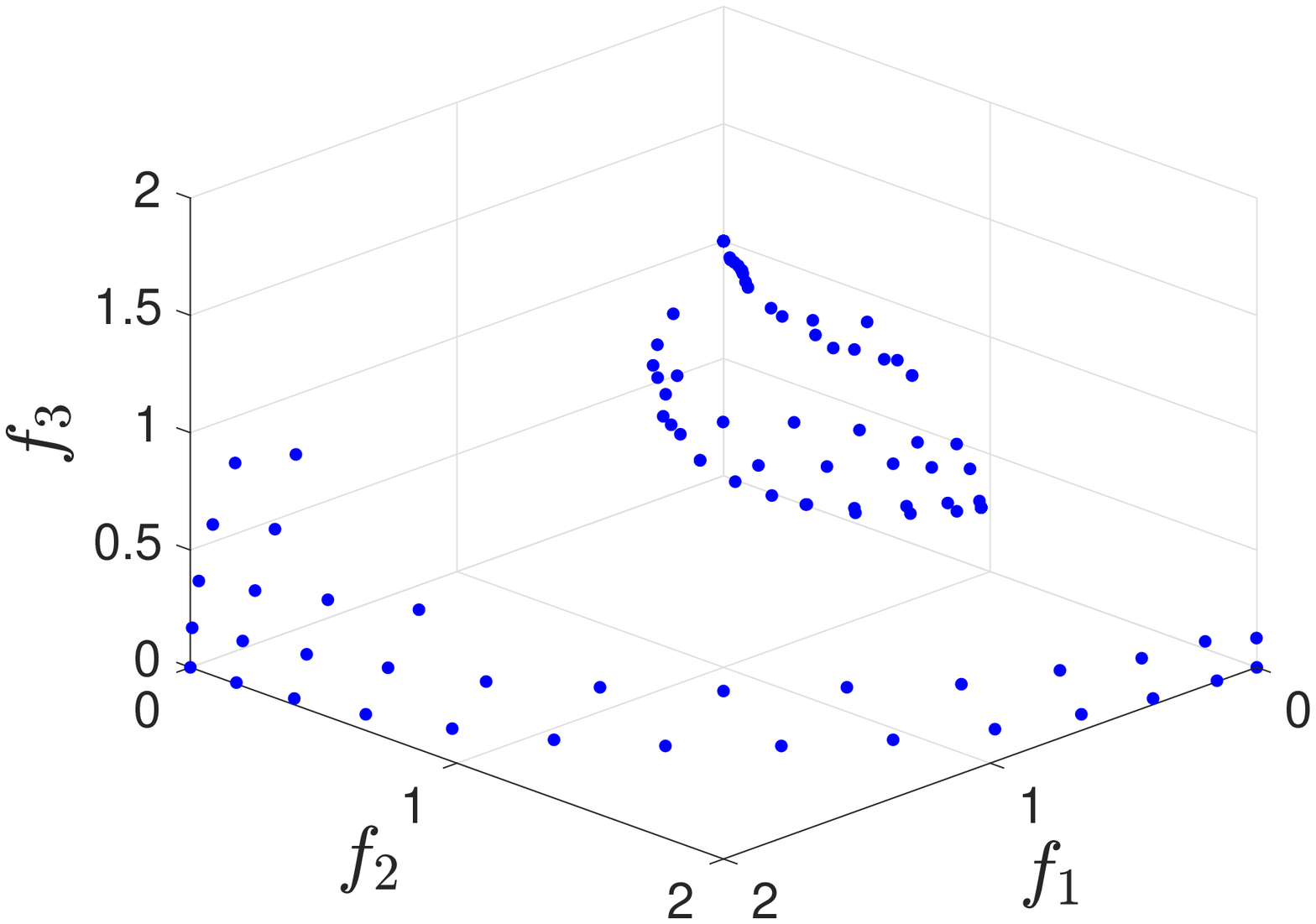}&
		\includegraphics[width=0.2\linewidth]{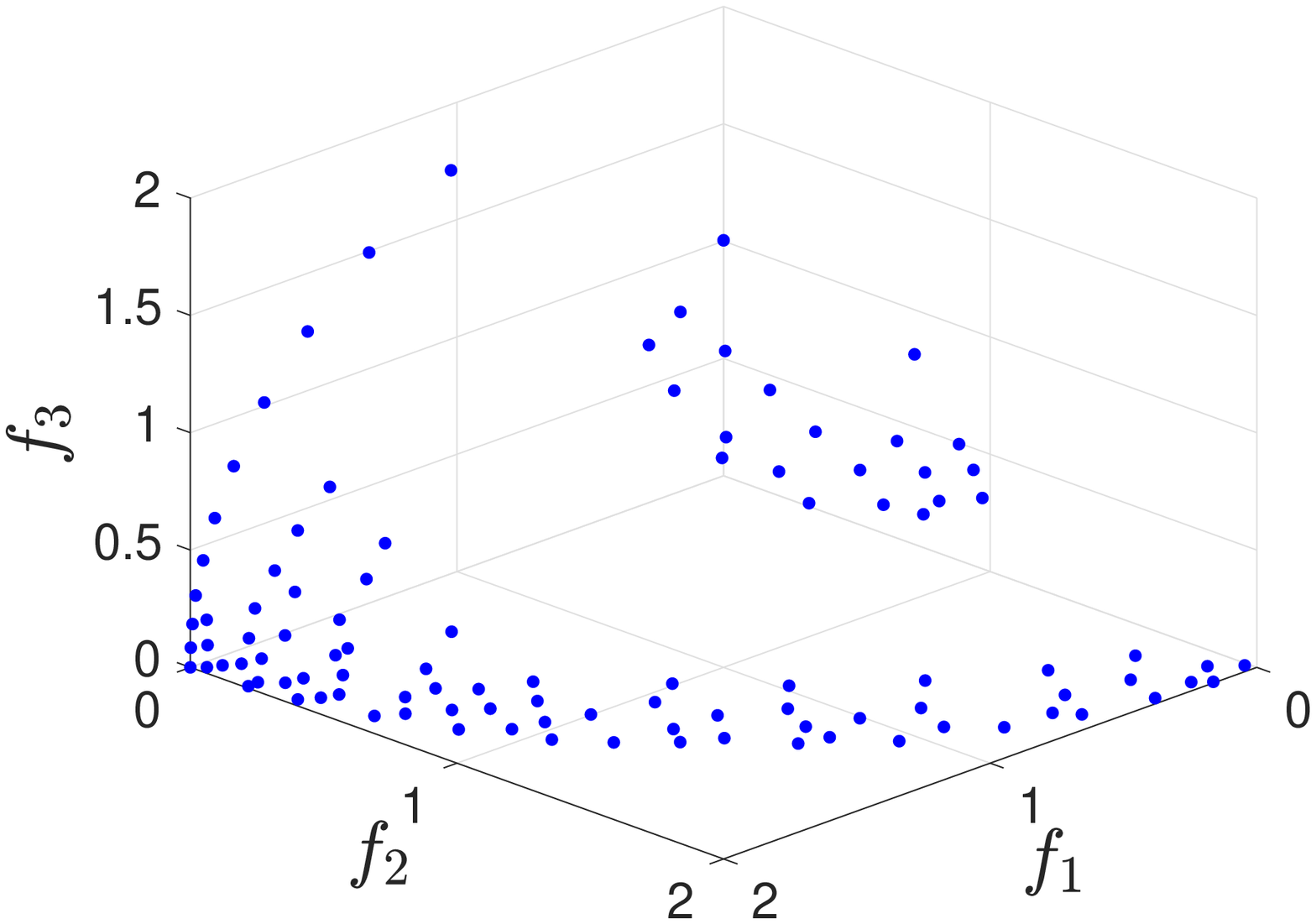}&
		\includegraphics[width=0.2\linewidth]{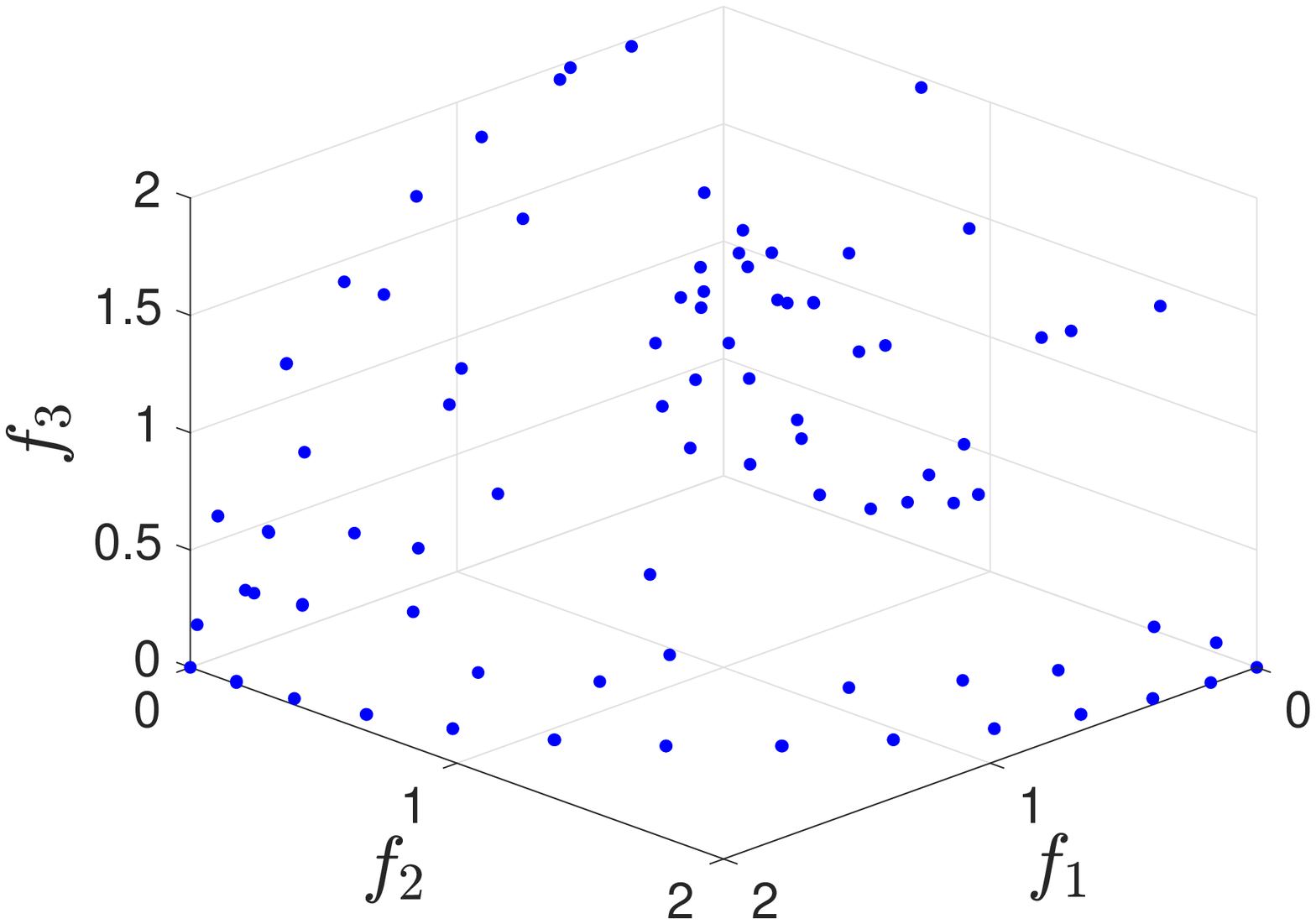}&
		\includegraphics[width=0.2\linewidth]{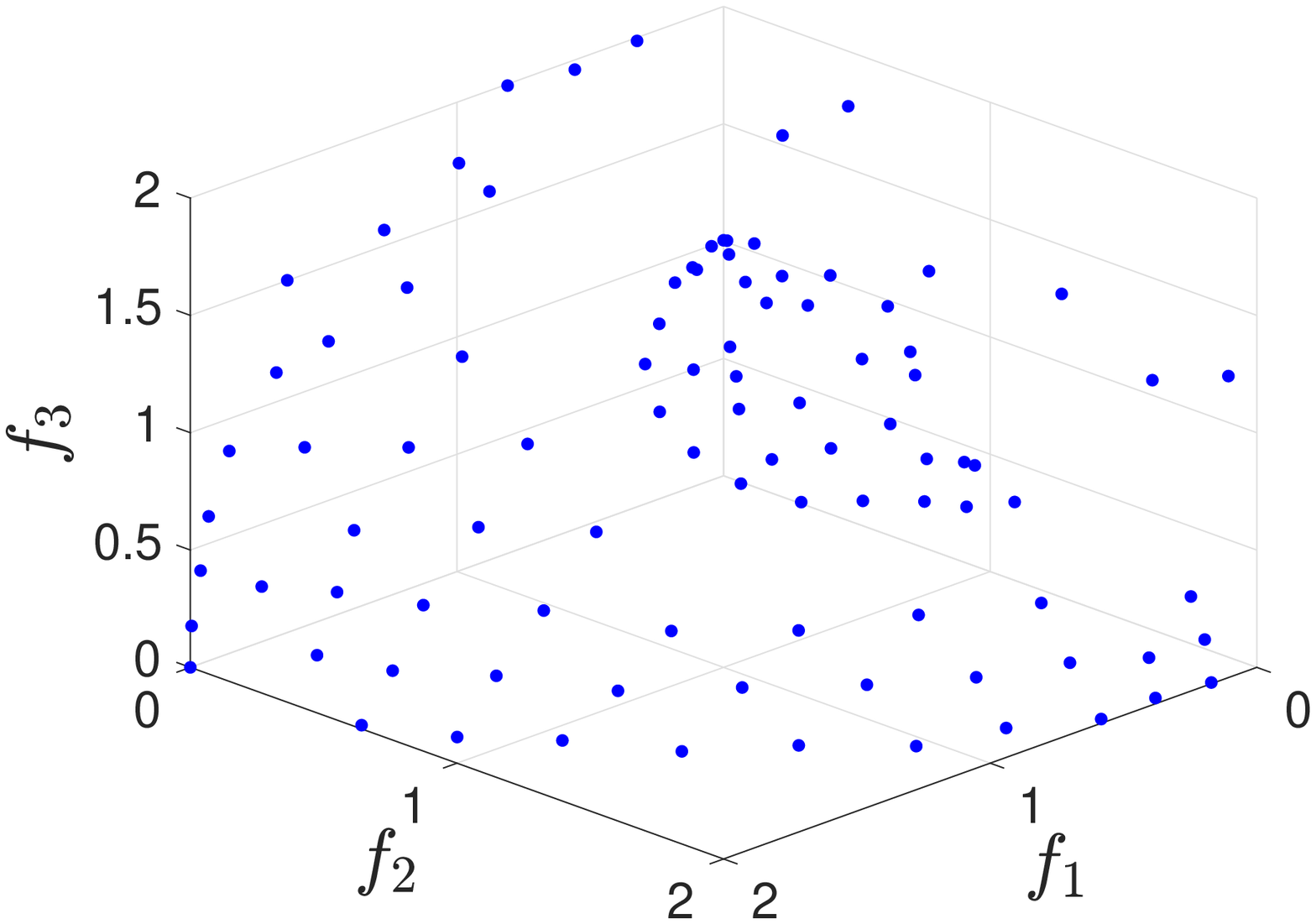}&
		\includegraphics[width=0.2\linewidth]{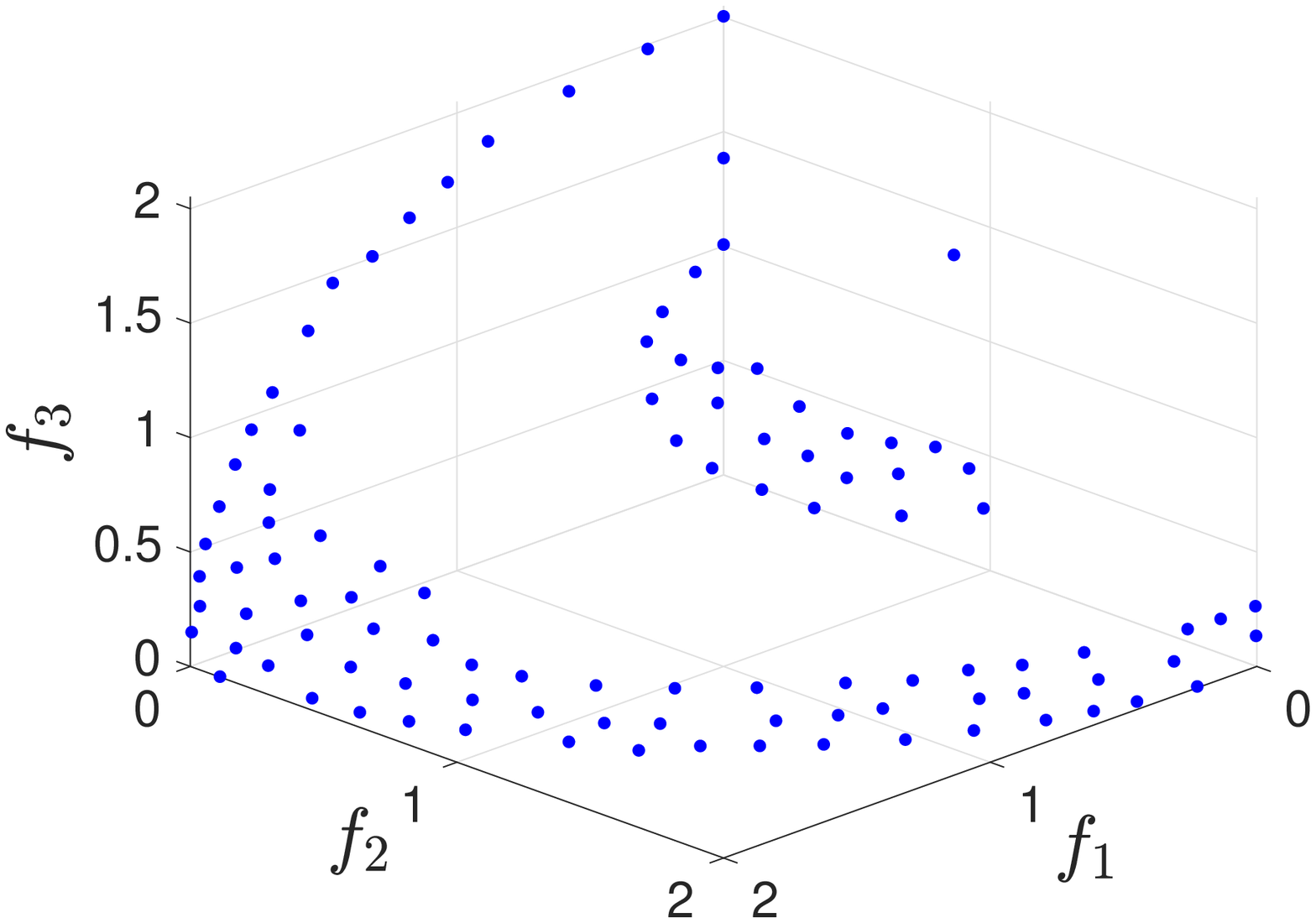}\\
		(a) MOEA/D & (b) ANSGA-III & (c) SPEA/R  & (d) RVEA & (e) AREA\\[-2mm]
	\end{tabular}
	\caption{PF approximation of F7 obtained by different algorithms.}
	\label{fig:f7_pf}
	\vspace{-2mm}
\end{figure*}

\begin{figure*}[!t]
	\centering
	\begin{tabular}{@{}c@{}c@{}c@{}c@{}c}
		\includegraphics[width=0.2\linewidth]{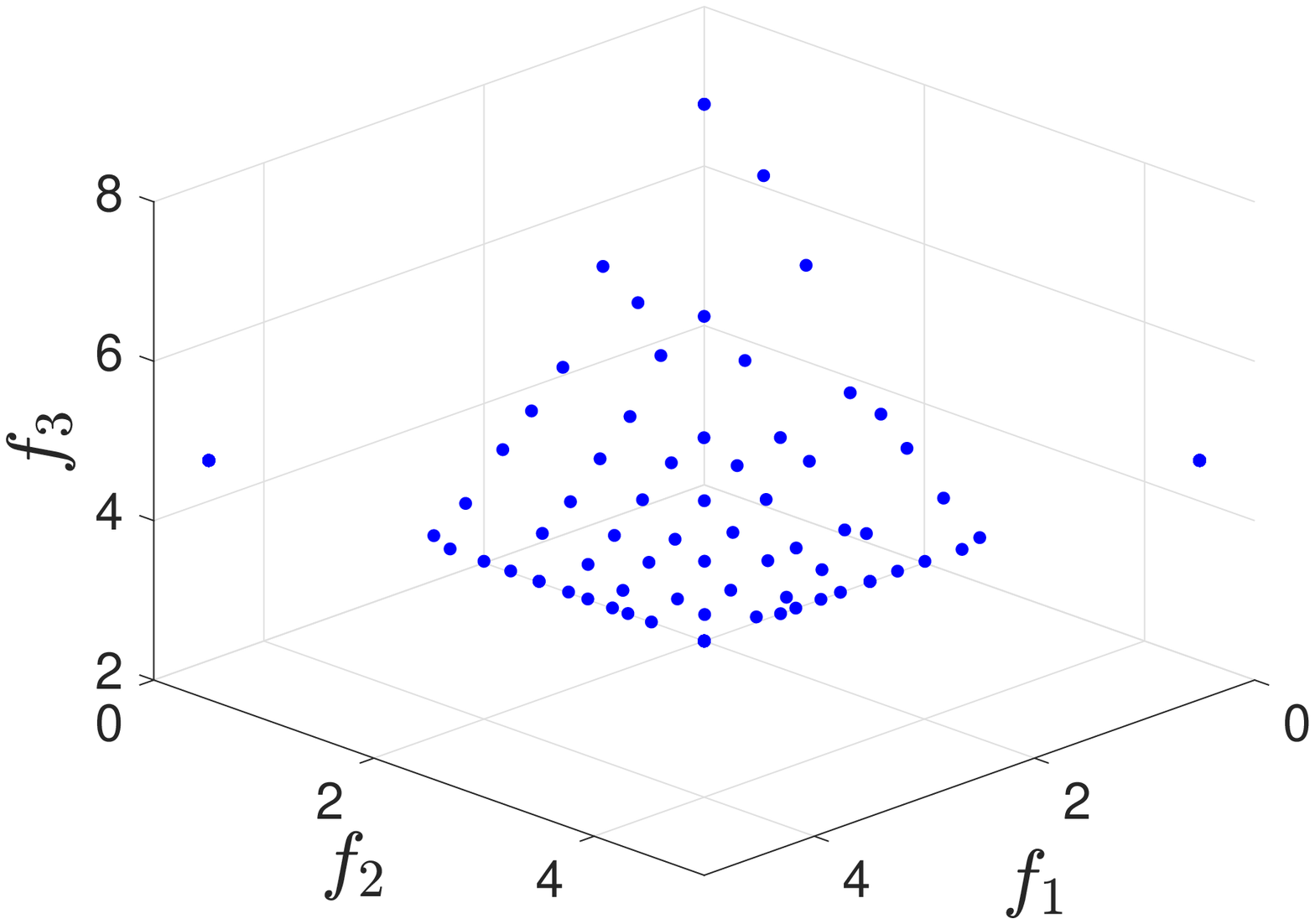}&
		\includegraphics[width=0.2\linewidth]{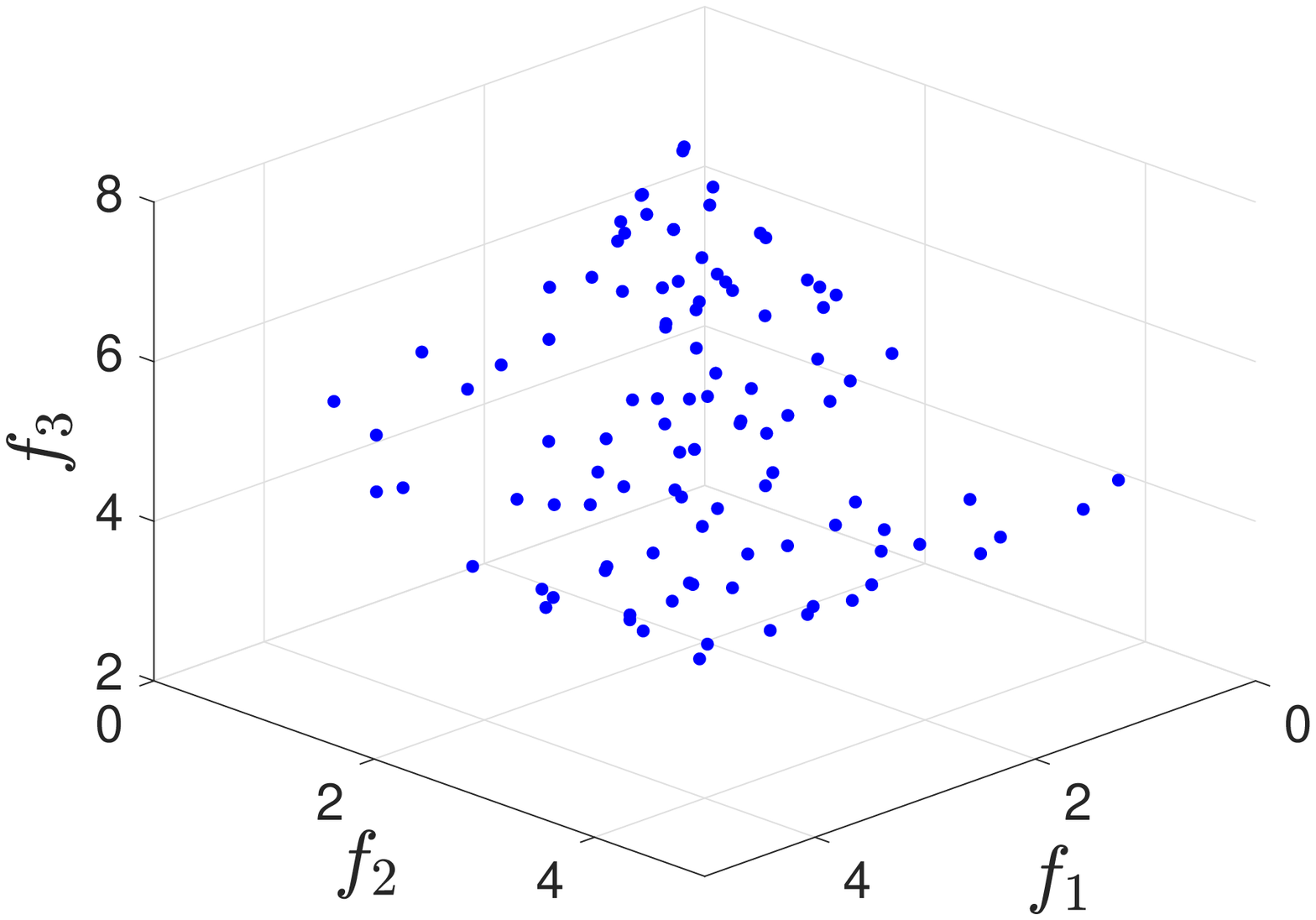}&
		\includegraphics[width=0.2\linewidth]{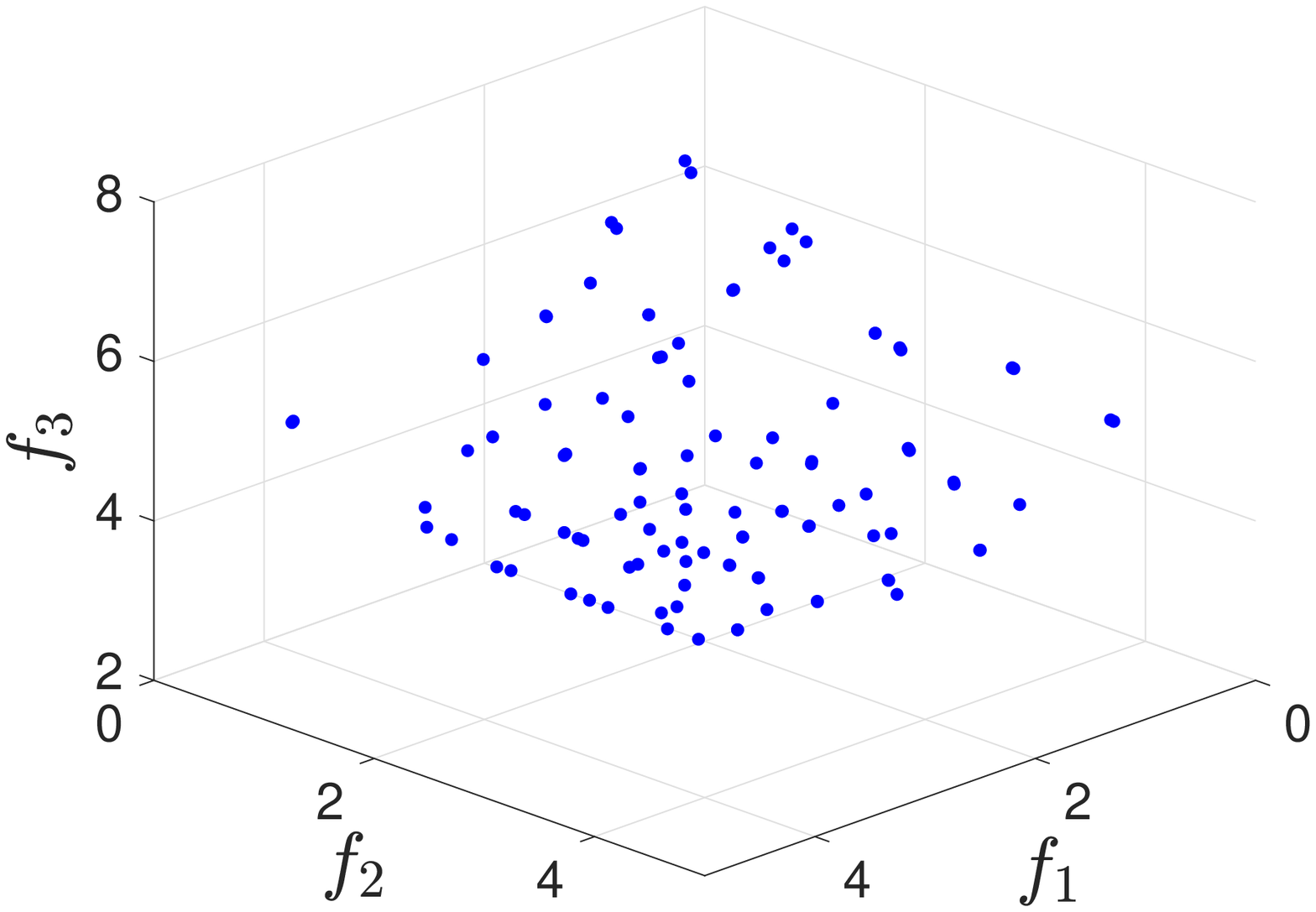}&
		\includegraphics[width=0.2\linewidth]{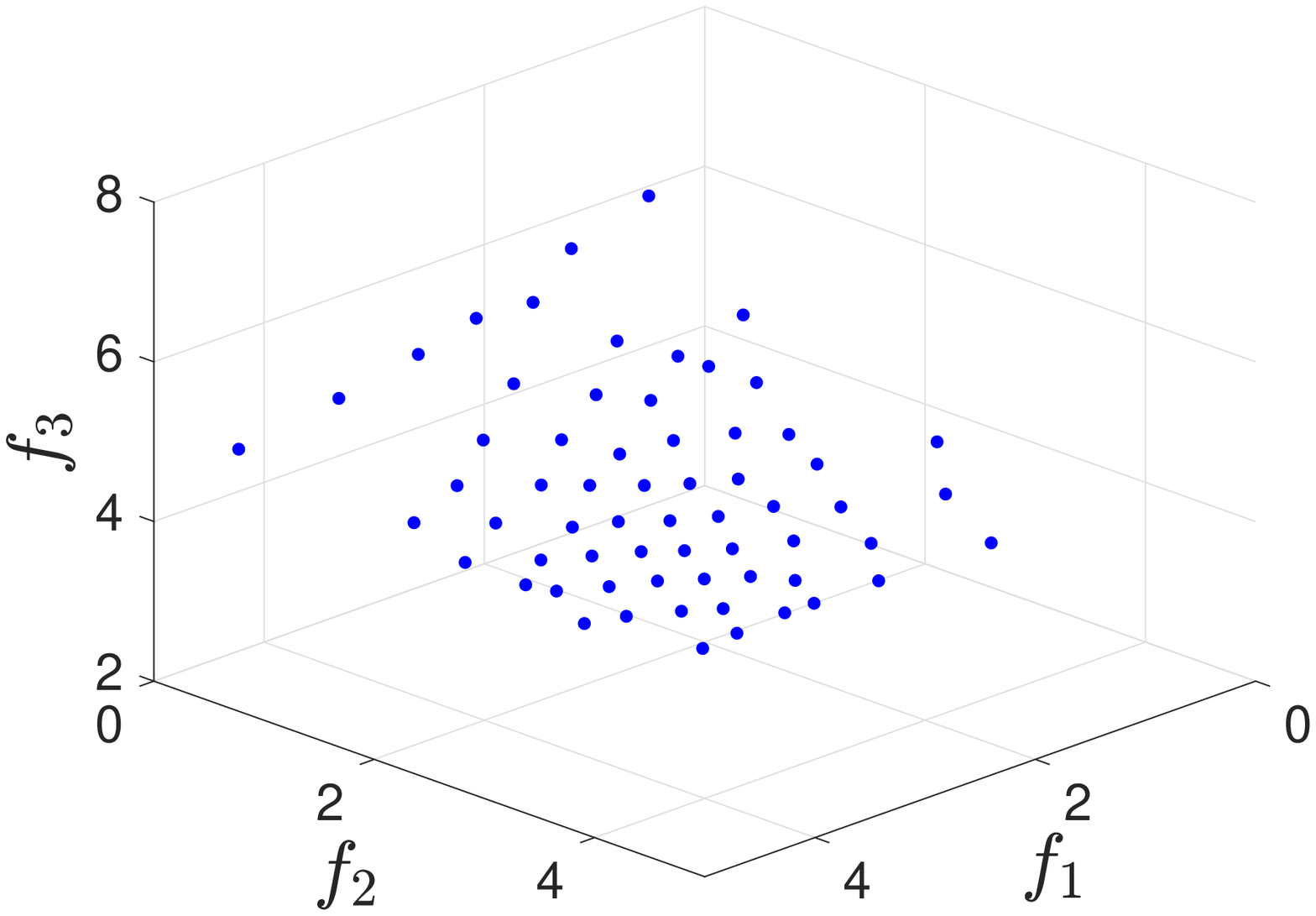}&
		\includegraphics[width=0.2\linewidth]{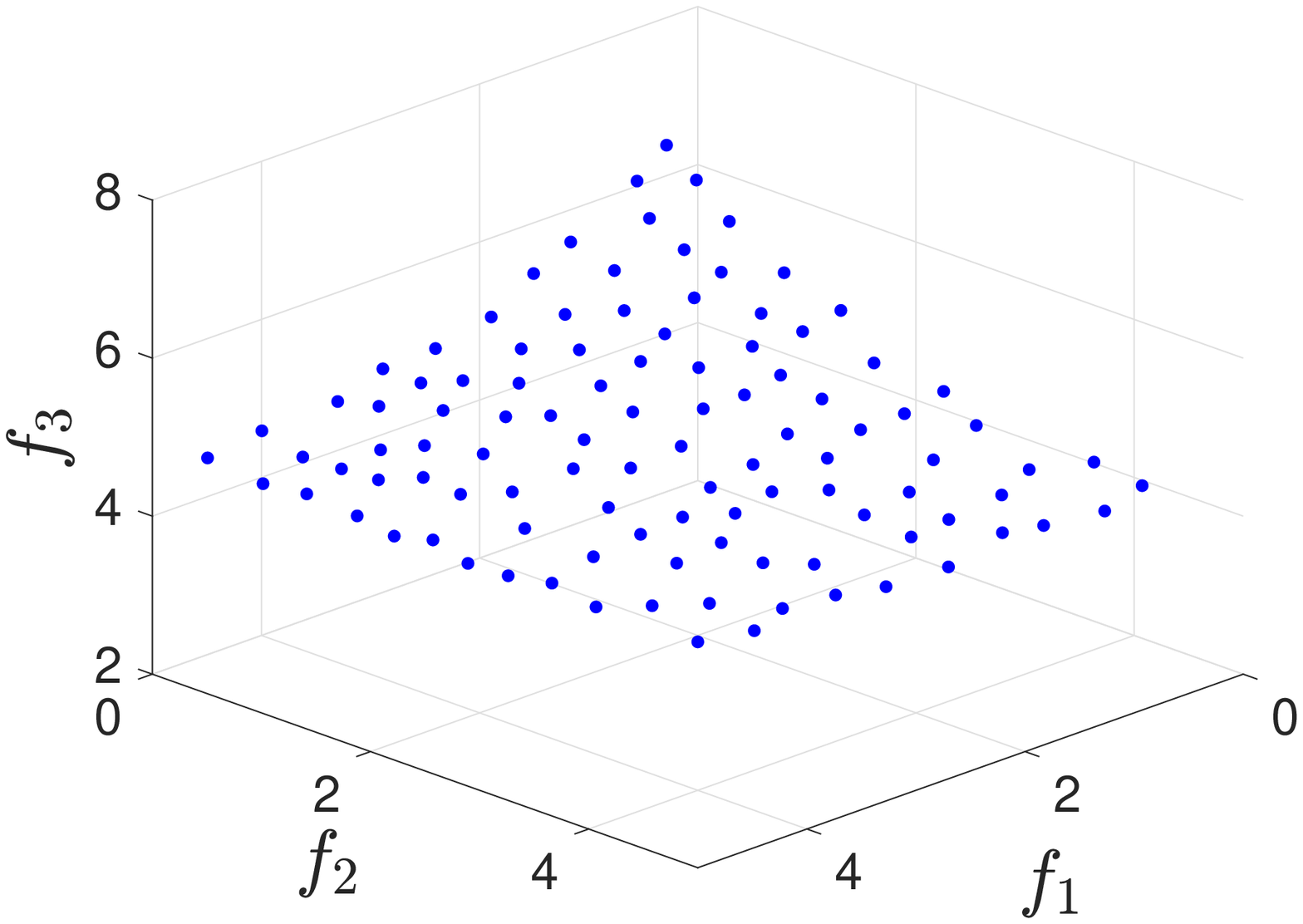}\\
		(a) MOEA/D & (b) ANSGA-III & (c) SPEA/R & (d) RVEA & (e) AREA\\[-2mm]
	\end{tabular}
	\caption{PF approximation of F8 obtained by different algorithms.}
	\label{fig:f8_pf}
	\vspace{-2mm}
\end{figure*}

For F1-F8, AREA outperforms the other algorithms on most of the problems except F5, demonstrating again the benefits of the proposed search strategies. To be more specific, F1--F4 are biobjective problems, and F5--F8 are triobjective ones. The PF approximations of F1 and F4, shown in Figs.~\ref{fig:f1_pf} and \ref{fig:f4_pf} respectively, are used to show the potential strengths and drawbacks of these algorithms in biobjective cases. It is seen that all the algorithms except AREA are not capable of finding boundary solutions on the PF of F1. AREA has a good coverage of the PF due to effective reference layout. For F4, MOEA/D and ANSGA-III find only two extreme points, just as what they do for MOP1. RVEA shows slow convergence again. SPEA/R obtains a good coverage of the PF, but the distribution is not as good as AREA. The PF approximations for the four triobjective problems are presented in Figs.~\ref{fig:f5_pf}--\ref{fig:f8_pf}.
For the non-simply connected F5, RVEA is less effective than the other algorithms in detecting the holes on the PF. The solution organisation can again explain the slightly worse IGD values of AREA than ANSGA-III and SPEA/R. For F6--F8, AREA clearly shows better performance than the others in terms of solution distribution and coverage of the PF. Thus, the proposed AREA is highly competitive and robust for a variety of problem characteristics.

\begin{figure*}[thb]
	\centering
	\begin{tabular}{@{}c@{}c@{}c}
		\includegraphics[width=0.32\linewidth, height=0.2\linewidth]{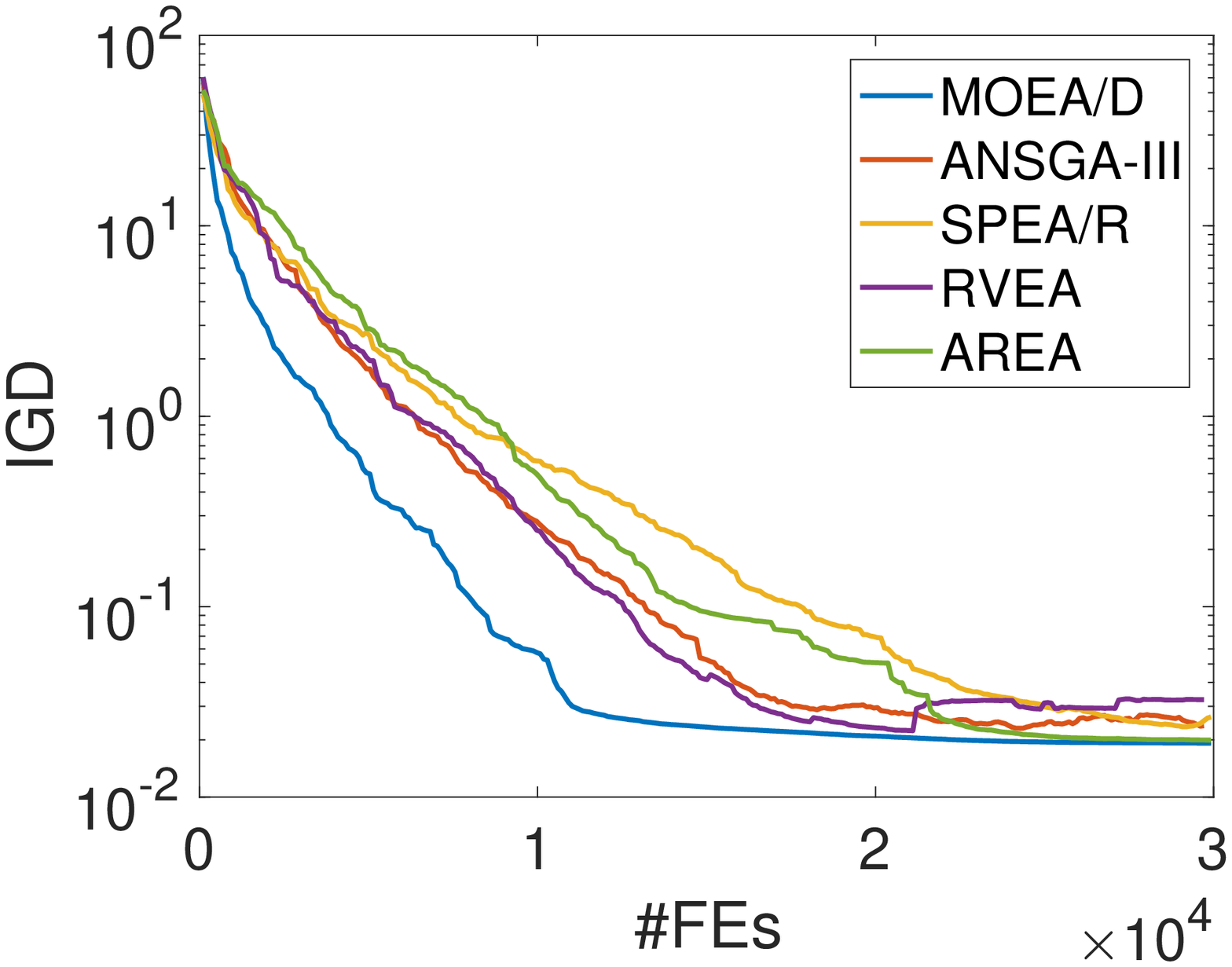}&
		\includegraphics[width=0.32\linewidth, height=0.2\linewidth]{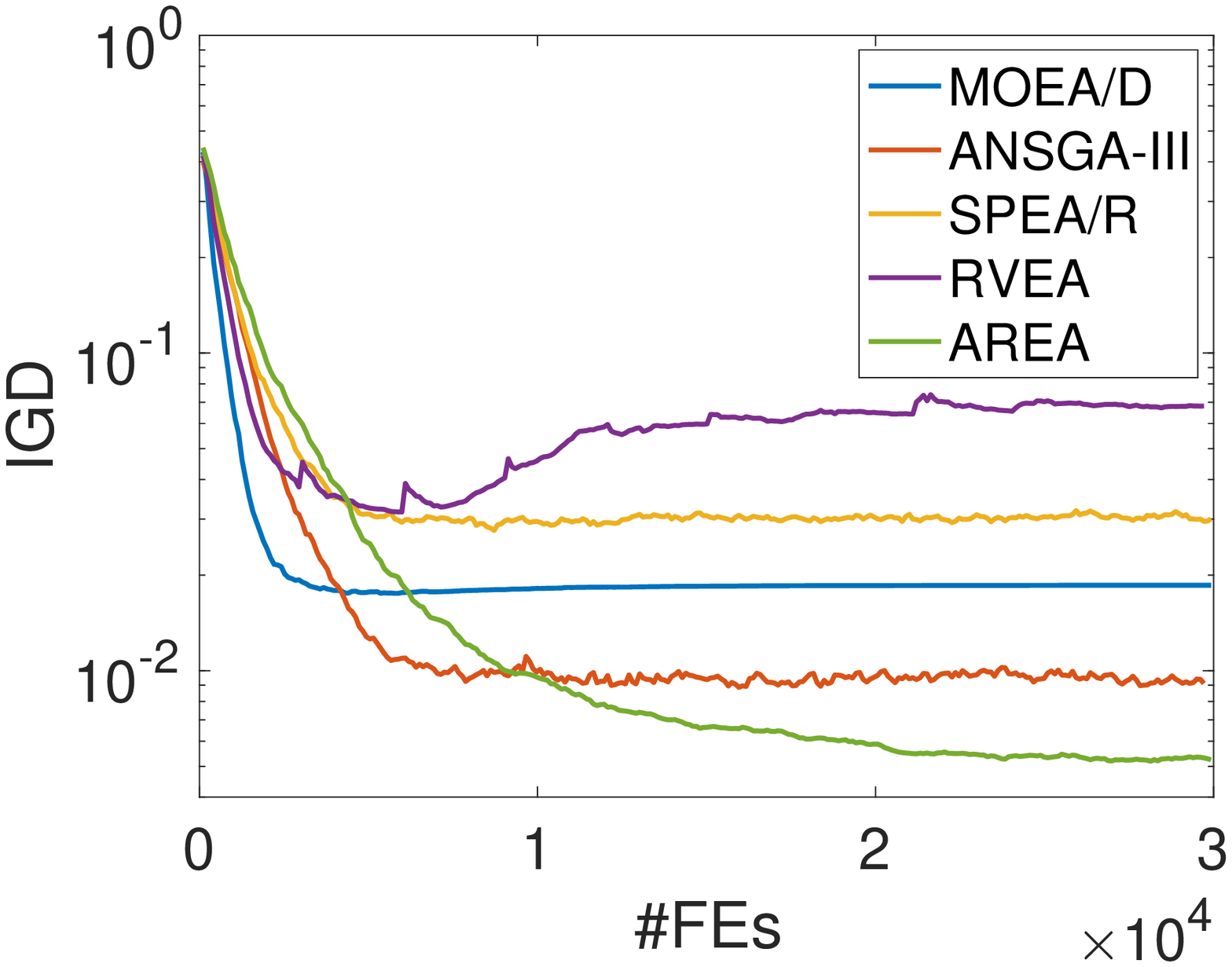}&
		\includegraphics[width=0.32\linewidth, height=0.2\linewidth]{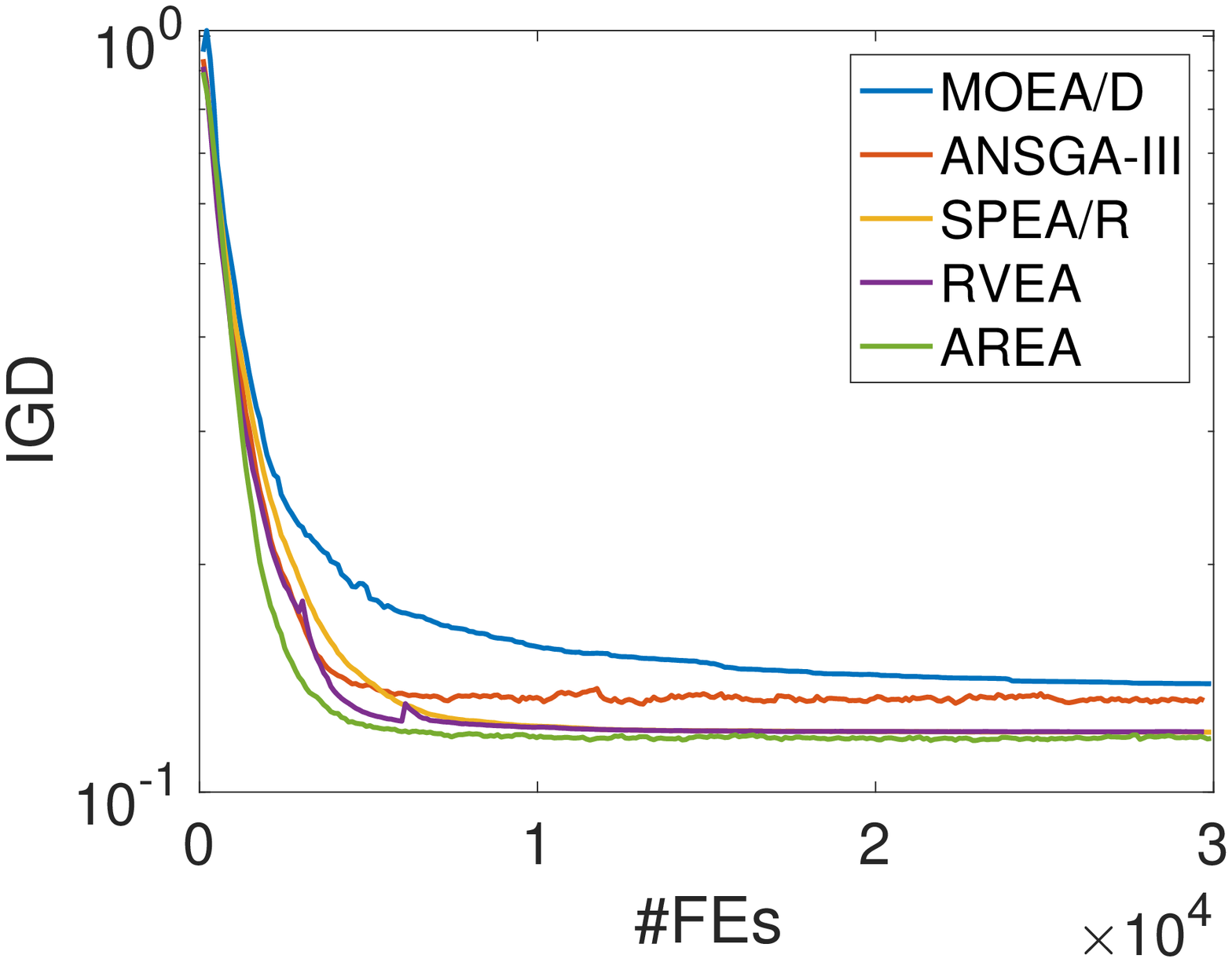}\\[-1mm]
		(a) DTLZ1 & (b) DTLZ5 & (c) SDTLZ2\\
		\includegraphics[width=0.32\linewidth, height=0.2\linewidth]{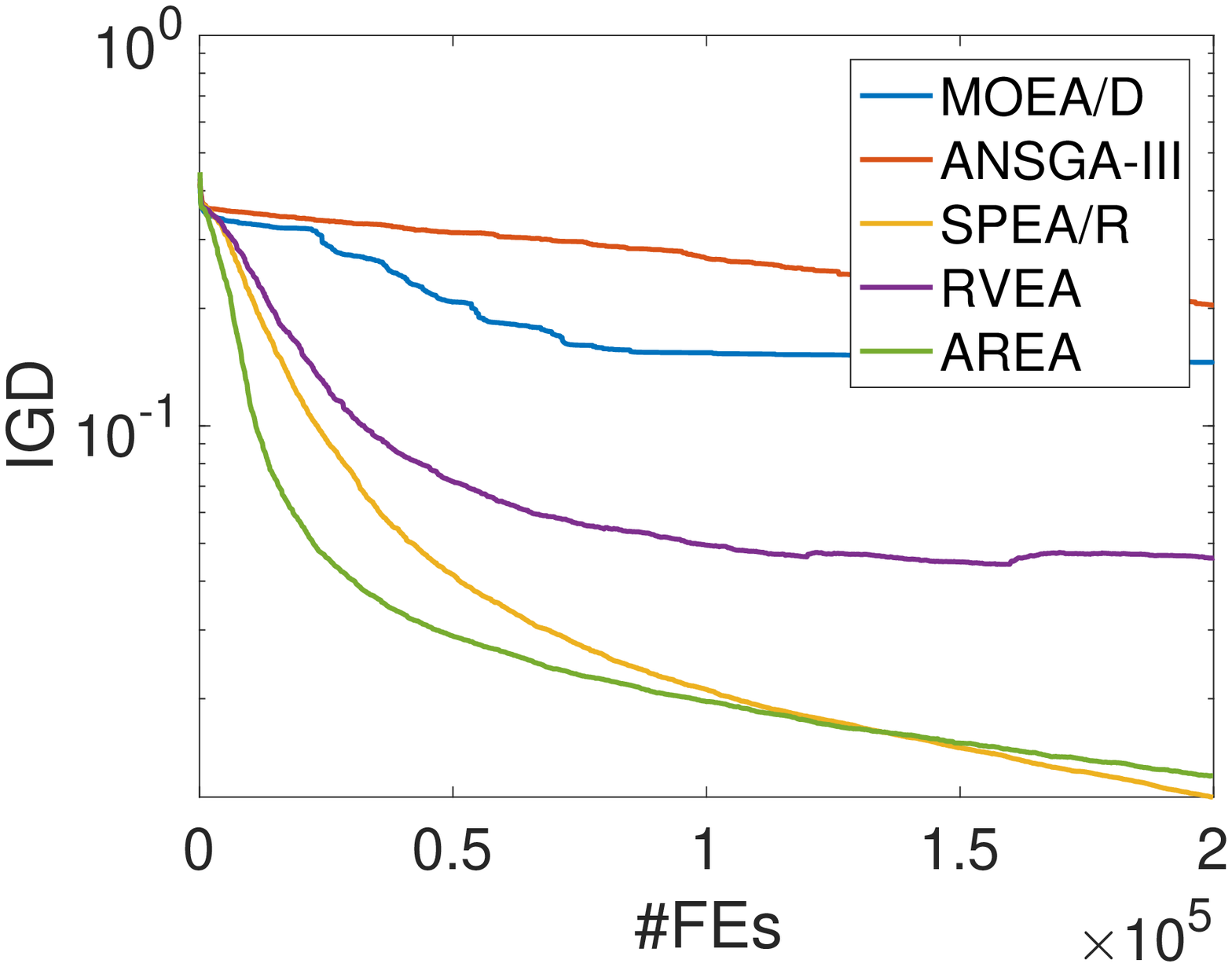}&
		\includegraphics[width=0.32\linewidth, height=0.2\linewidth]{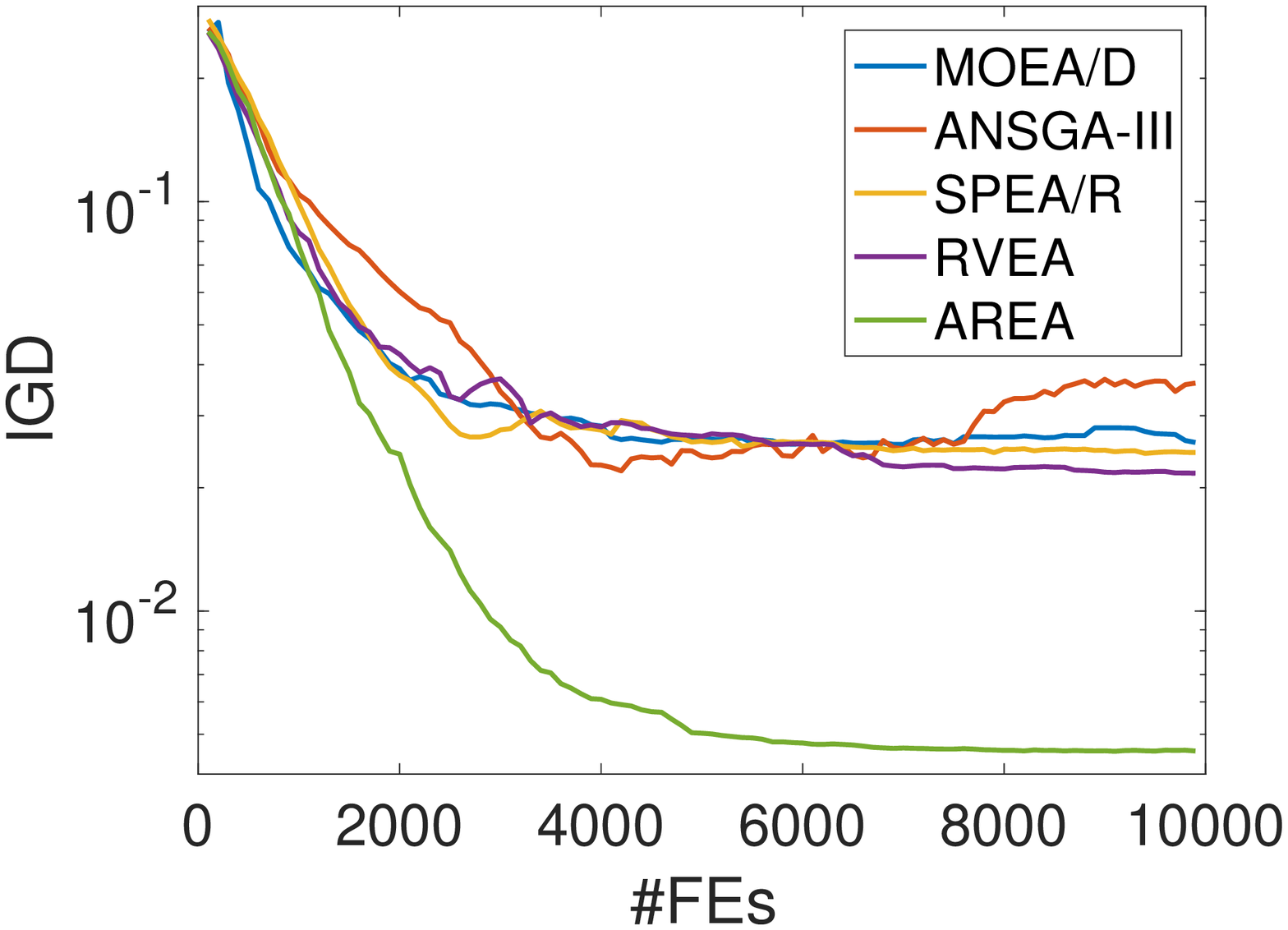}&
		\includegraphics[width=0.32\linewidth, height=0.2\linewidth]{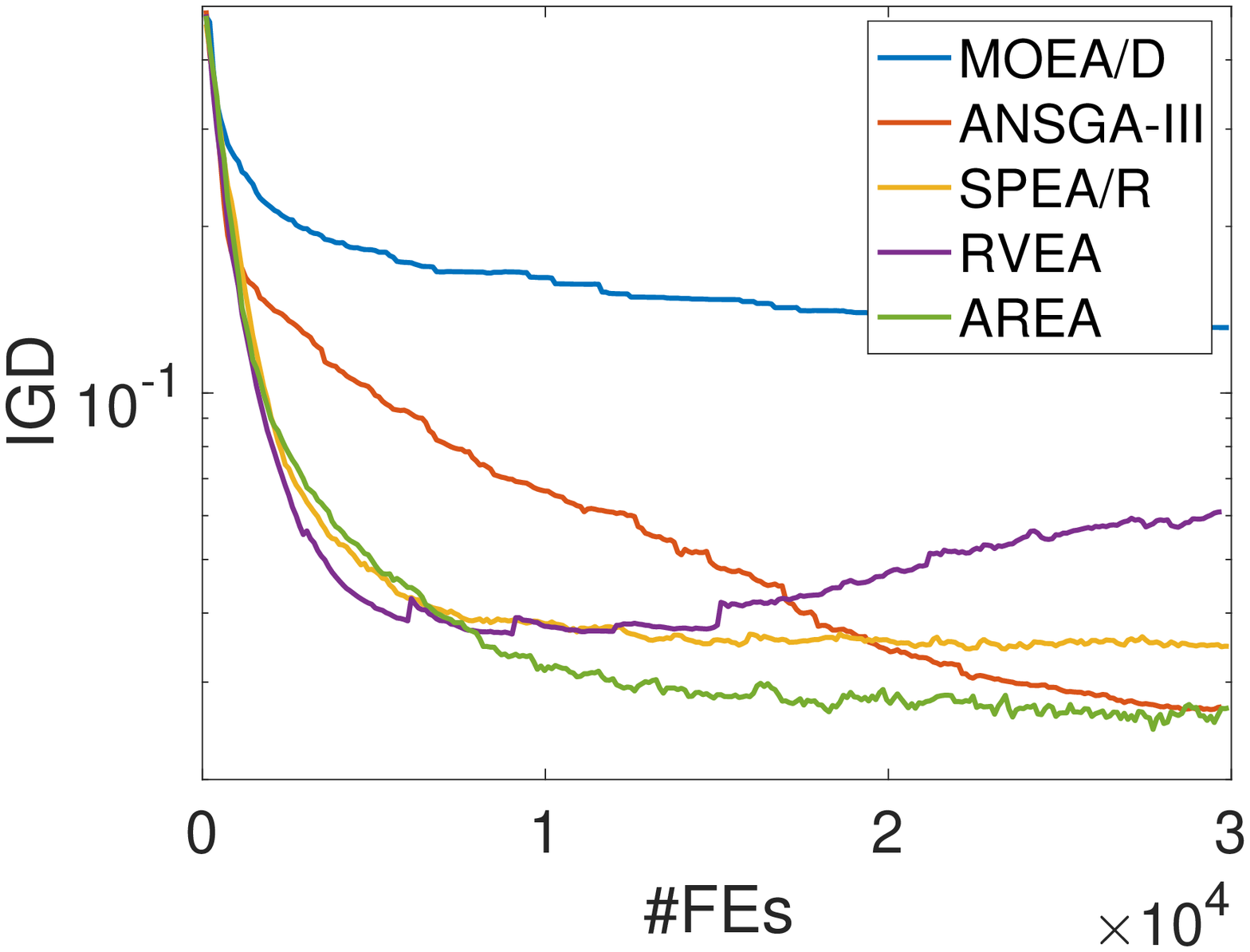}\\[-1mm]
		(d) MOP1 & (e) F1 & (f) F6\\[-2mm]
	\end{tabular}
	\caption{IGD values (averaging 30 runs) of algorithms against the number of FEs.}
	\label{fig:igd-curve}
	\vspace{-2mm}
\end{figure*}

Fig.~\ref{fig:igd-curve} plots the average IGD values of five algorithms against the number of function evaluations (FEs) for some selected problems. It is observed that the convergence speed differs significantly. Some algorithms, sucha as MOEA/D and RVEA, converge quickly toward the PF at the beginning of search but are not able to further reduce their IGD values as the number of FEs increase. In contrast, AREA is capable of lowering IGD values over function evaluations, therefore obtaining better IGD values at the end of search for the majority of the problems.

\subsection{Results on UF/WFG}
Here, we extend performance analysis to other test suites: UF \cite{Zhang2009} and WFG \cite{HHBW06}. UF contains a number of 2- and 3-objective difficult problems that have complicated PSs. Therefore, algorithms require a large number of iterations to approximate their PSs/PFs well. WFG is a set of scalable problems and frequently used for assessing algorithms' scalability in many-objective optimisation. For notational convenience, 'WFGK-M' represents the WFG$K$ instance with $M$ objectives. AREA is compared with MOEA/D \cite{LZ09} and RVEA \cite{CJOS-RVEA} on these two test suites, and the results are analysed separately in the following paragraphs.

\subsubsection{Results on UF}
The differential evolution (DE) helps to achieve better results on UF problems than simulated binary crossover (SBX) \cite{LZ09}. Therefore, we use DE in compared algorithms for this test suite. Population size is 600 and 595 for 2- and 3-objective problems \cite{Zhang-dra}, respectively. The maximum number of evaluations is 300,000 according to \cite{Zhang2009}.

\begin{figure*}[!t]
	\centering
	\begin{tabular}{@{}c@{}c@{}c}
		\includegraphics[width=0.33\linewidth]{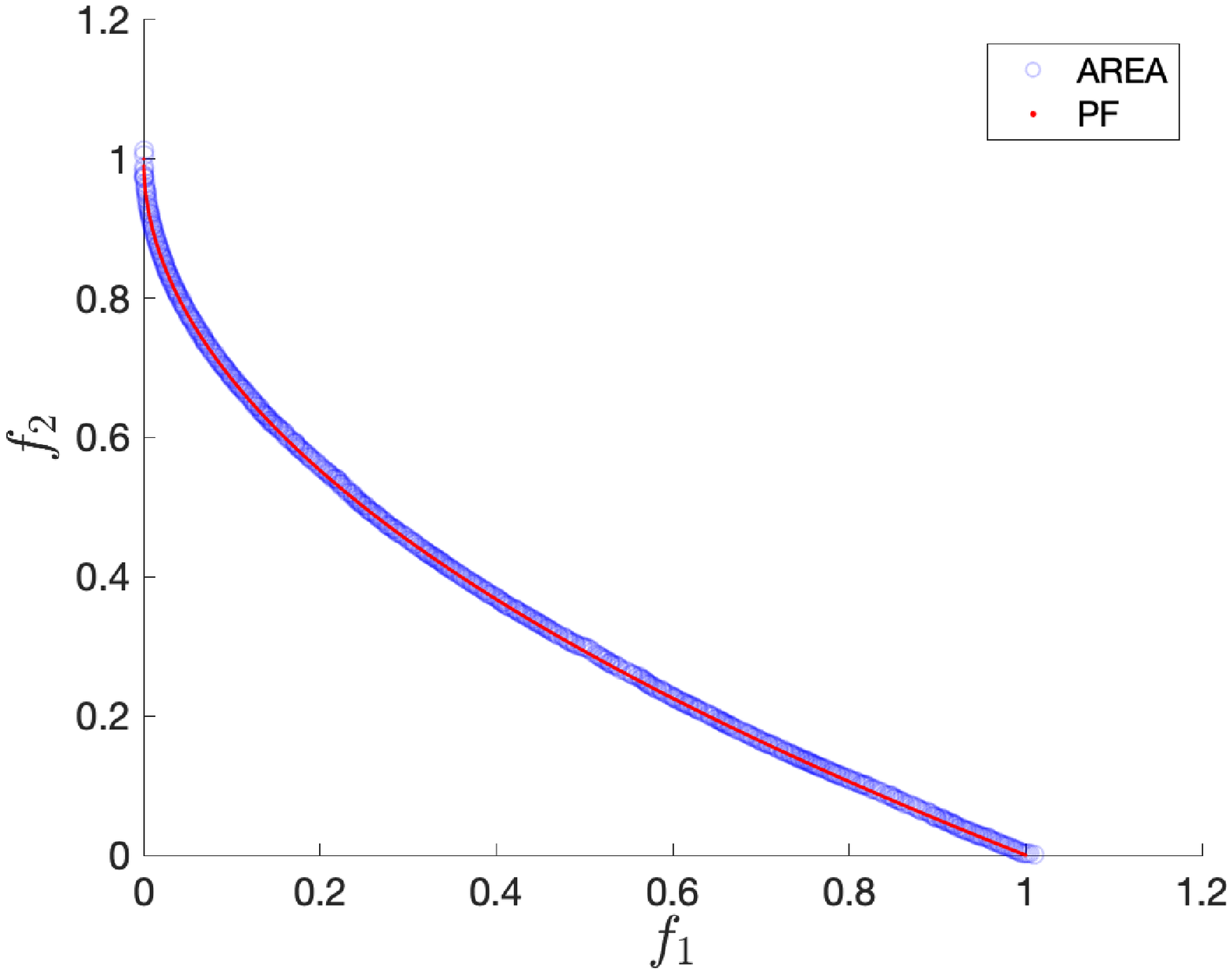}&
		\includegraphics[width=0.33\linewidth]{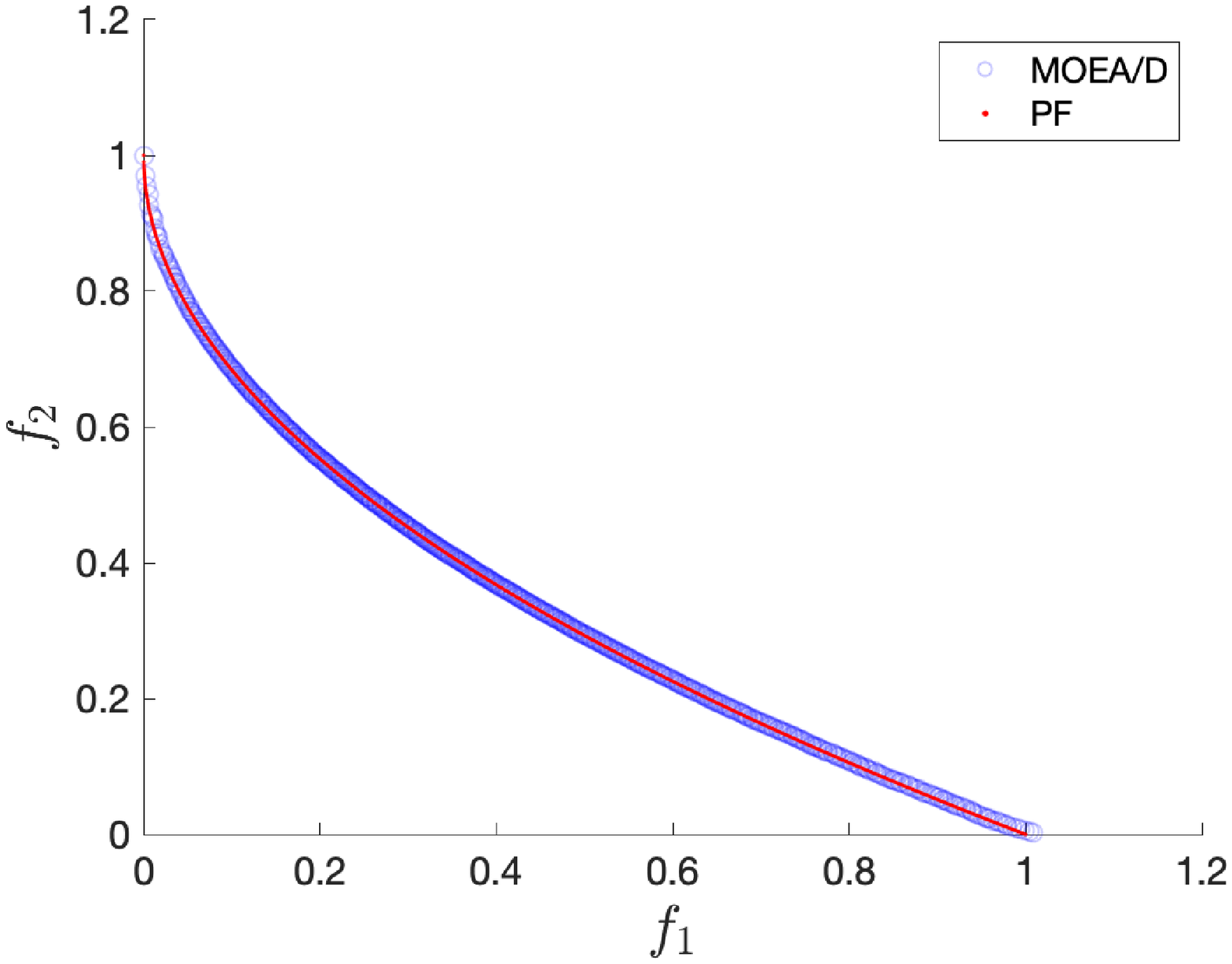}&
		\includegraphics[width=0.33\linewidth]{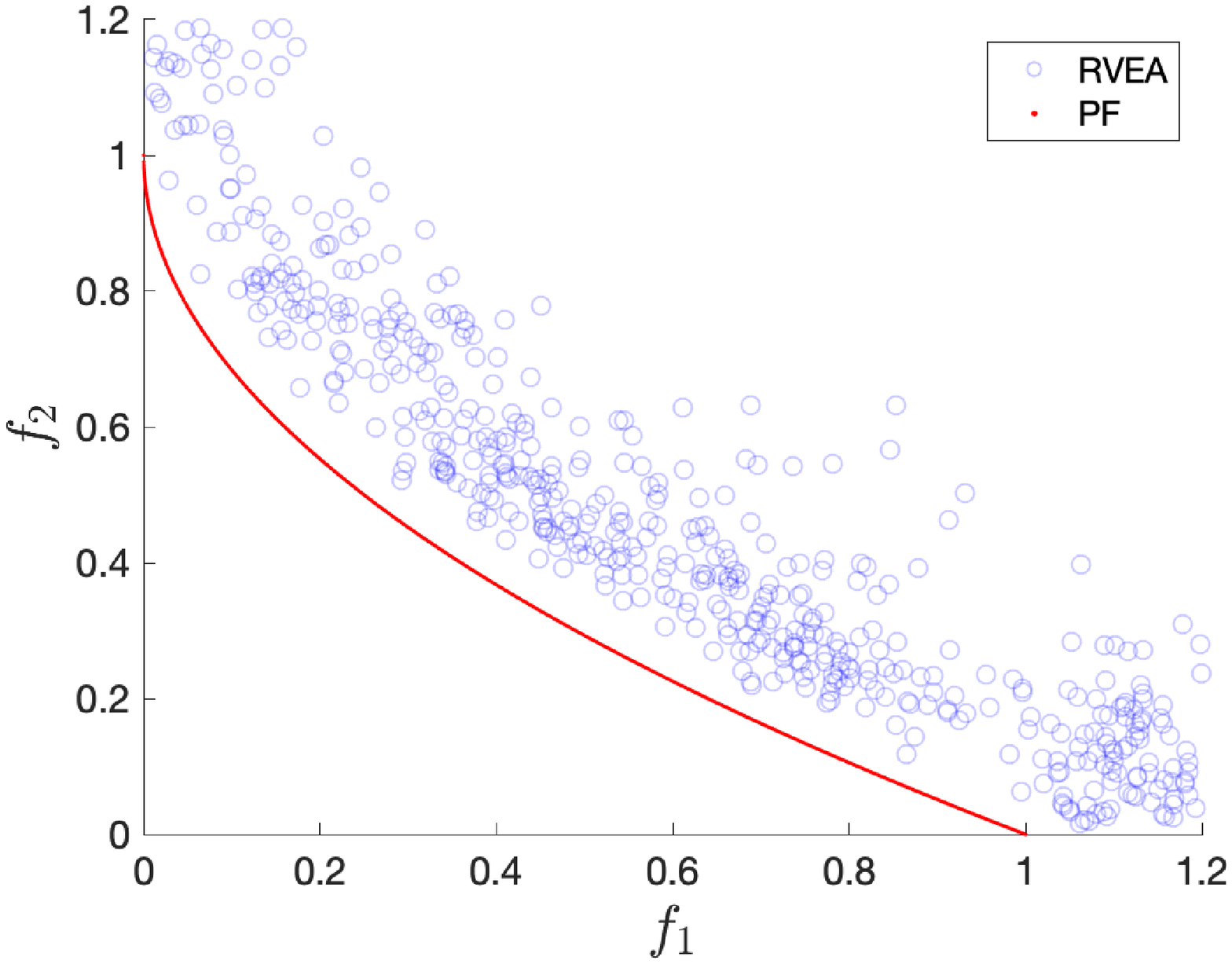}\\
		(a) AREA & (b) MOEA/D & (c)  RVEA\\[-2mm]
	\end{tabular}
	\caption{PF approximations obtained by three algorithms for UF1.}
	\label{fig:uf1}
	\vspace{-2mm}
\end{figure*}

\begin{figure*}[!t]
	\centering
	\begin{tabular}{@{}c@{}c@{}c}
		\includegraphics[width=0.33\linewidth]{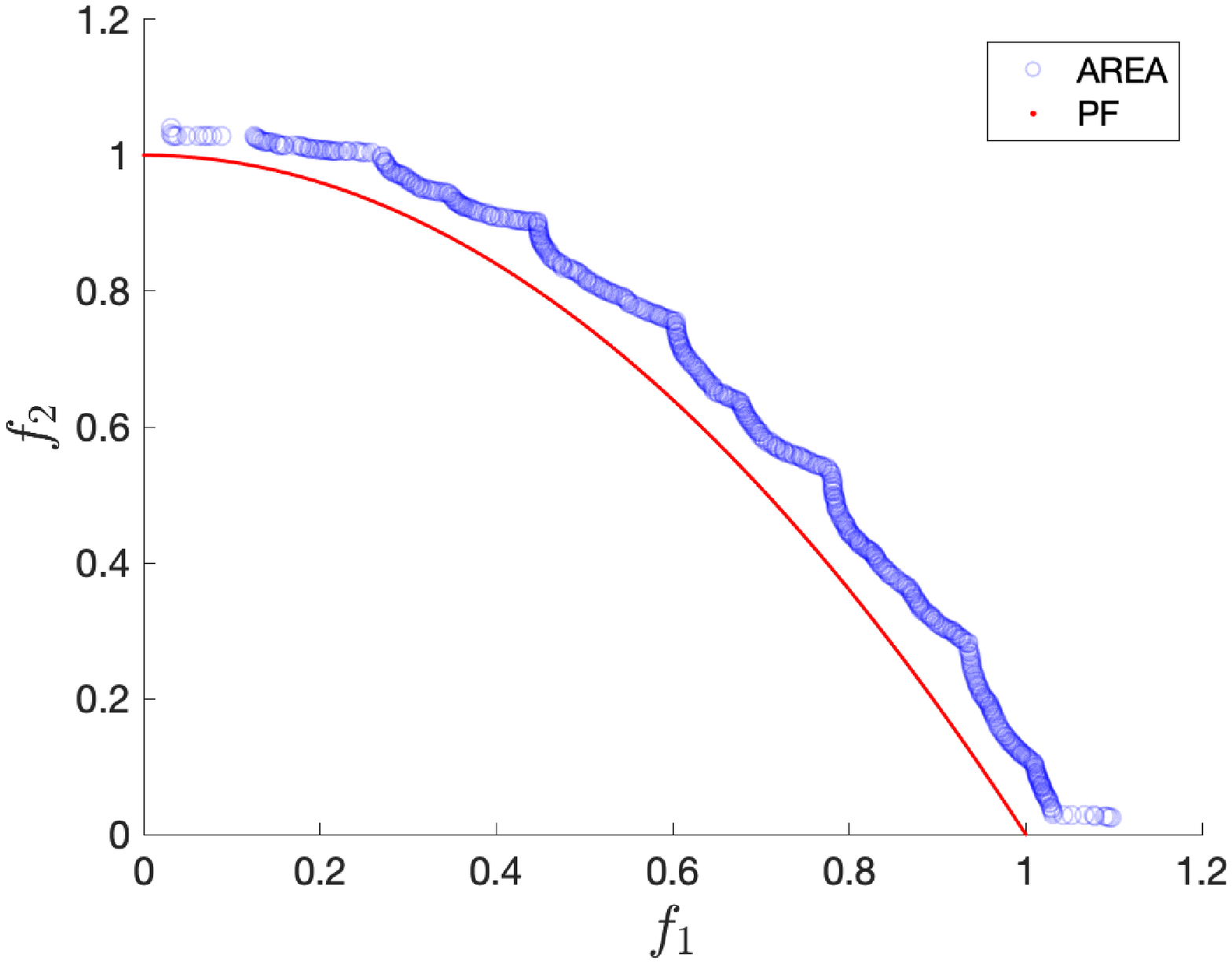}&
		\includegraphics[width=0.33\linewidth]{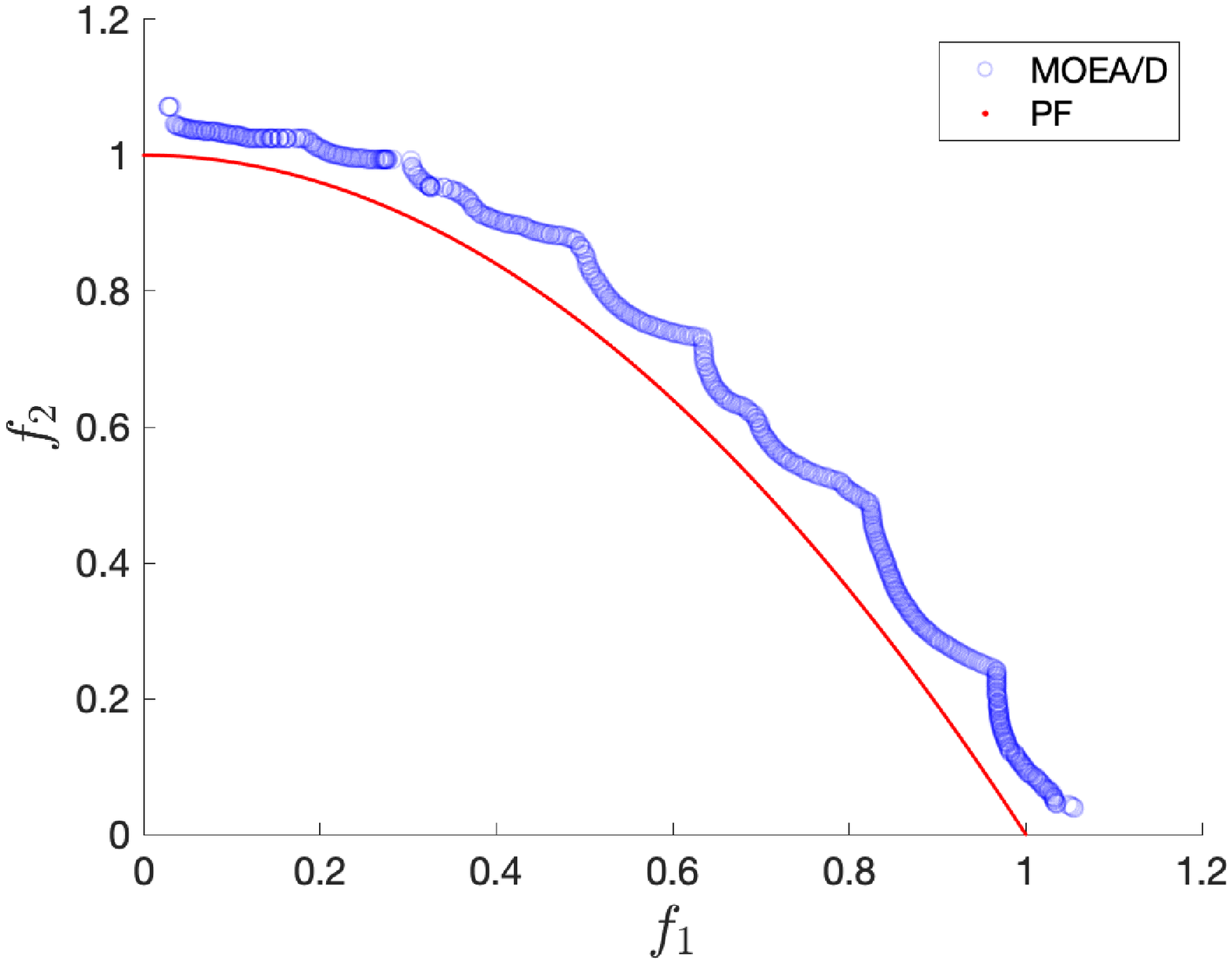}&
		\includegraphics[width=0.33\linewidth]{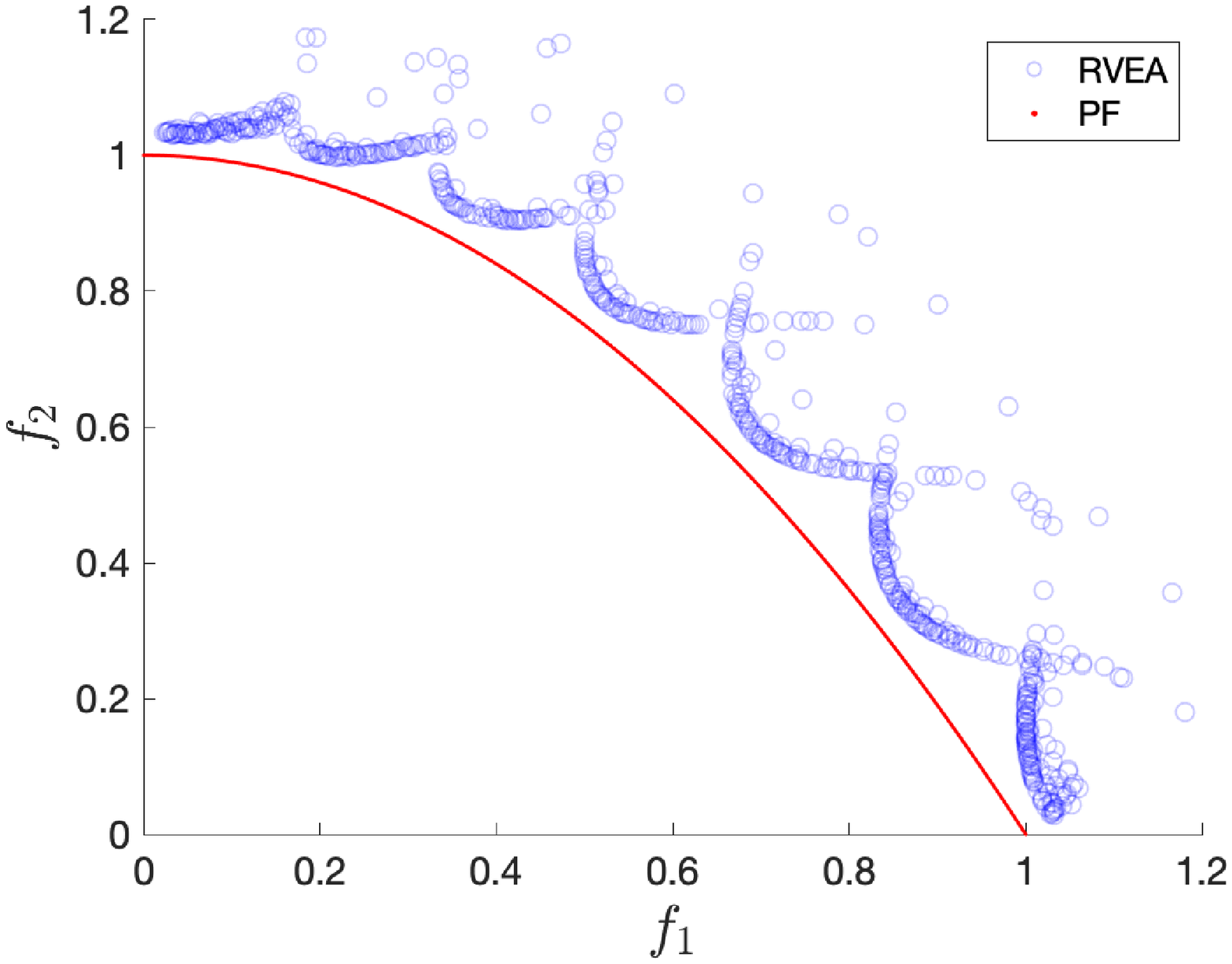}\\
		(a) AREA & (b) MOEA/D & (c)  RVEA\\[-2mm]
	\end{tabular}
	\caption{PF approximations obtained by three algorithms for UF4.}
	\label{fig:uf4}
	\vspace{-2mm}
\end{figure*}

Table \ref{tab:uf} presents the IGD values of the algorithms for nine UF problems. It shows that, while AREA performs the best on approximately half of the problems, MOEA/D wins the rest of comparison. A close inspection of their PF approximations in Figs. \ref{fig:uf1}--\ref{fig:uf4} suggests that both AREA and MOEA/D are effective for the UF problems, although one of them may have better IGD values. On the other hand, RVEA struggles to solve the UF problems, as indicated by both large IGD values and poor PF approximations. RVEA loses all the comparison due to slow convergence performance.

\subsubsection{Results on WFG} 
All the algorithms use SBX for crossover for the WFG problems, as recommended in \cite{DJ13,CJOS-RVEA}. Our experimental design has the same configurations as in \cite{DJ13}.

\begin{figure*}[!t]
	\centering
	\begin{tabular}{@{}c@{}c@{}c}
		\includegraphics[width=0.33\linewidth]{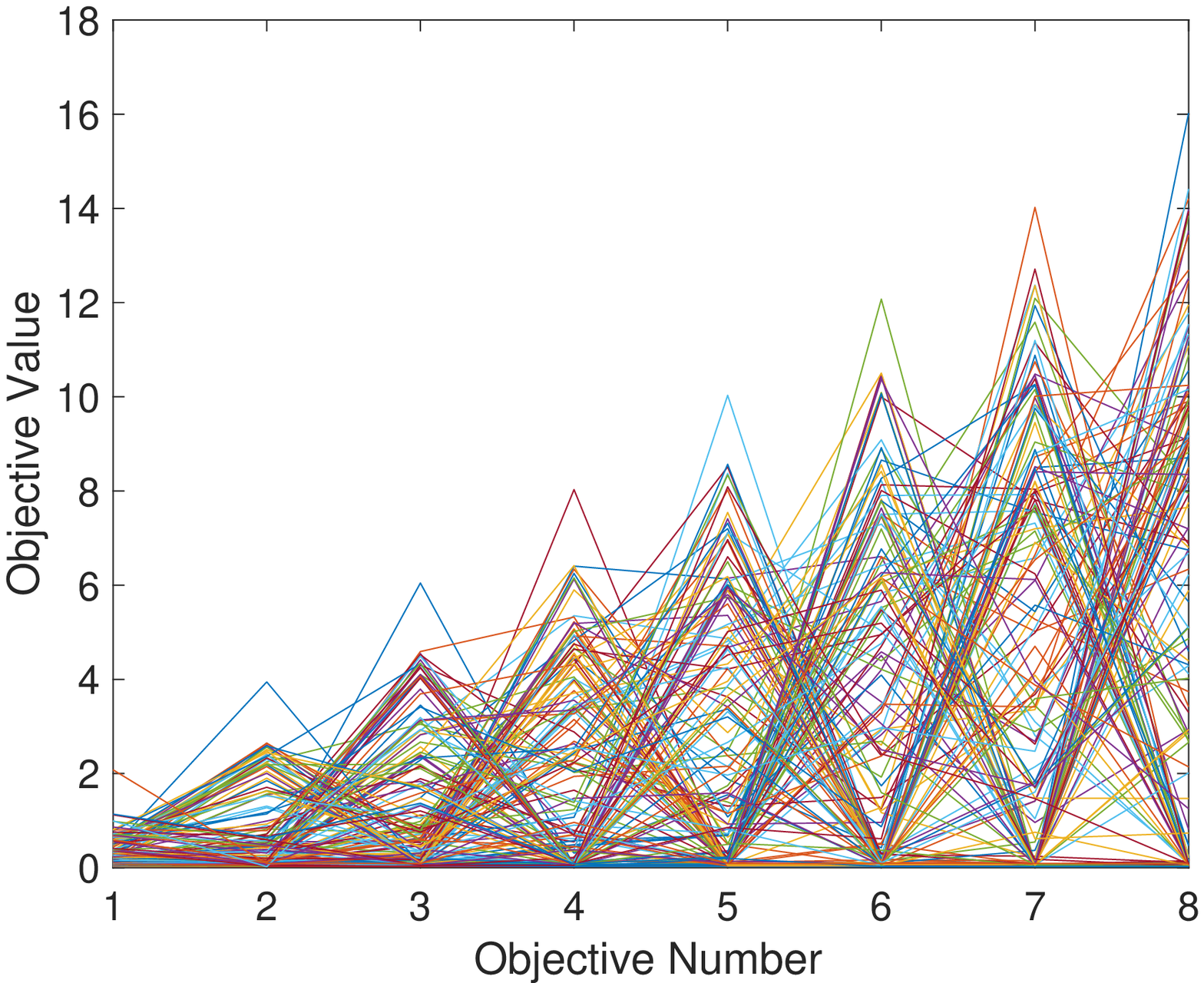}&
		\includegraphics[width=0.33\linewidth]{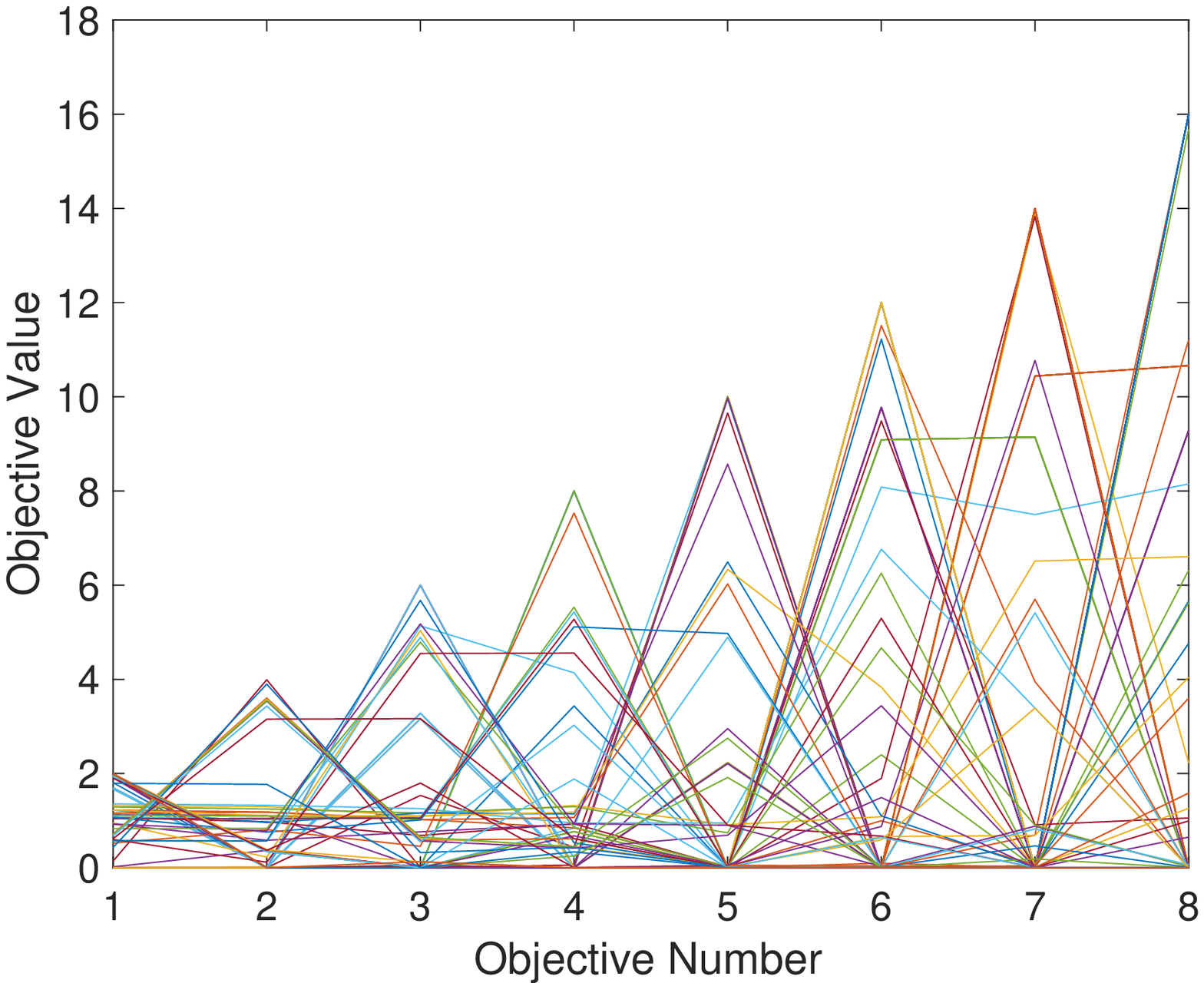}&
		\includegraphics[width=0.33\linewidth]{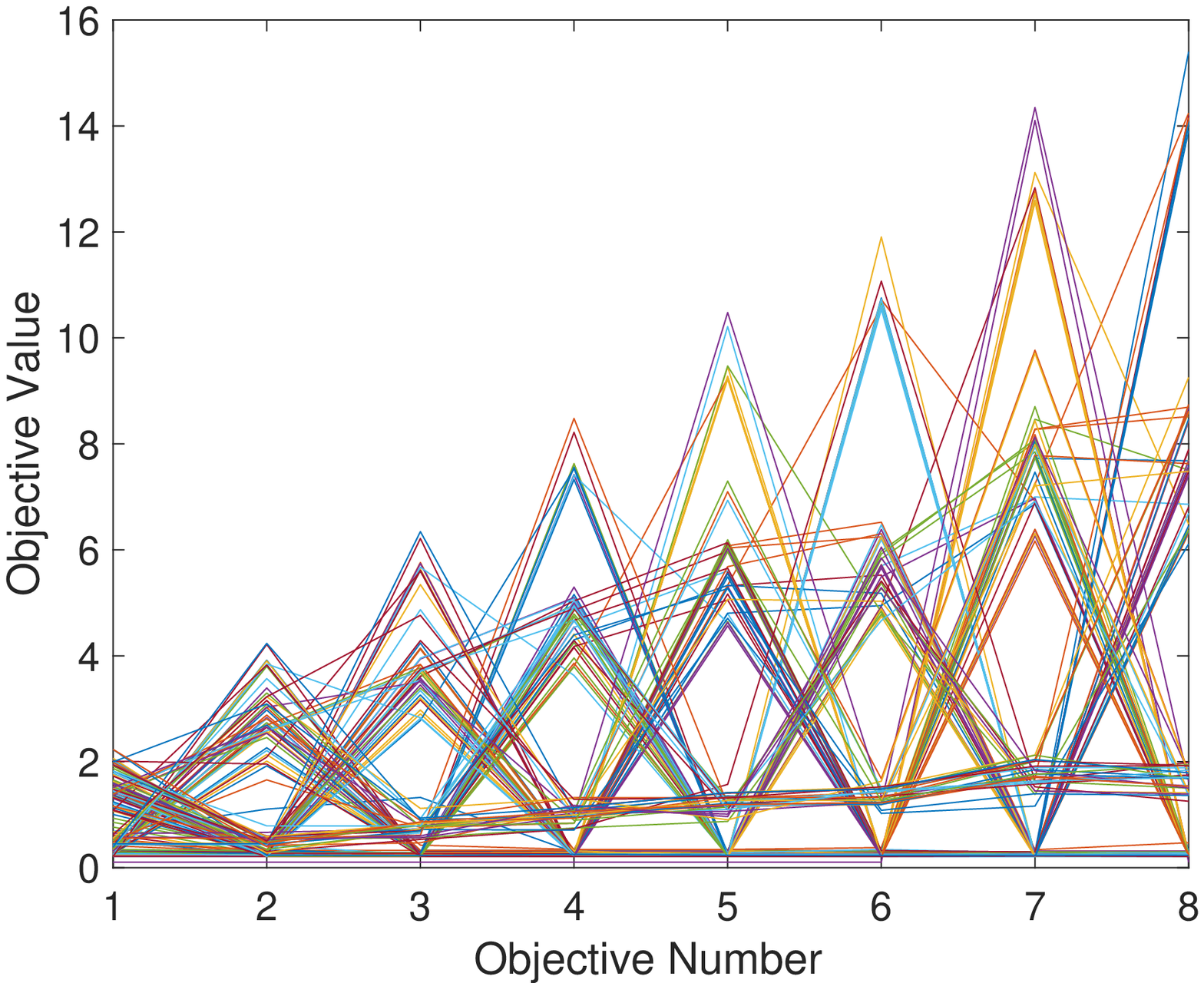}\\
		(a) AREA & (b) MOEA/D & (c)  RVEA\\[-2mm]
	\end{tabular}
	\caption{Plots of parallel coordinates of solutions obtained by three algorithms on WFG4 with 8 objectives.}
	\label{fig:wfg48}
	\vspace{-2mm}
\end{figure*}

Table \ref{tab:wfg} reports the IGD values of the three algorithms for selected WFG problems for different number of objectives. It is observed that AREA has the best performance on the 3-objective WFG problems, and MOEA/D is significantly worse than the other two. With an increase in the number of objectives, all the algorithms experience performance degradation since their IGD values become larger. For 8-objective WFG problems, RVEA performs better than the other two algorithms, and AREA is only slightly worse in term of IGD. When the number of objectives reaches 15, the IGD values of all the three algorithms are large, indicating they all suffer from the curse of dimensionality. Despite that, AREA shows better performance particularly for WFG4 and WFG6. We also show in Fig.~\ref{fig:wfg48} the parallel coordinate plots of the three algorithms for WFG4 with 8 objectives. This figure suggests that MOEA/D and RVEA tend to find numerous similar solutions as many parallel coordinate lines overlap each other. In contrast, the solutions obtained by AREA seem diversified as there are fewer overlapping lines. Despite that, we recognise that many-objective optimisation is challenging, and AREA can be improved to better handle many-objective problems.  
 
\begin{table*}[htbp]
\addtolength{\tabcolsep}{1.5pt}
\centering
\footnotesize
\caption{Mean (and standard deviation) values of IGD obtained by three algorithms for nine UF instances. Algorithms that are better than, worse than, or equivalent to AREA on ranksum test with a significance level of 0.05 are indicated by $+$, $-$, or $\approx$, respectively.}
\begin{tabular}{lcccc}
	\toprule
	Problem&$M$&AREA&MOEA/D&RVEA\\
	\midrule
	\multirow{1}{*}{UF1}&2&2.3897e-3 (4.80e-4)&\hl{1.5587e-3 (7.35e-5) $+$}&7.7344e-2 (8.70e-3) $-$\\
	\hline
	\multirow{1}{*}{UF2}&2&\hl{5.7683e-3 (8.01e-4)}&6.3921e-3 (1.68e-3) $-$&9.7725e-2 (4.91e-3)  $-$\\
	\hline
	\multirow{1}{*}{UF3}&2&1.3792e-2 (9.70e-3)&\hl{3.6351e-3 (2.04e-3) $+$}&2.3958e-1 (3.02e-2) $-$\\
	\hline
	\multirow{1}{*}{UF4}&2&\hl{4.8969e-2 (1.78e-3)}&5.6859e-2 (6.58e-3) $-$&6.5310e-2 (2.17e-2) $-$\\
	\hline
	\multirow{1}{*}{UF5}&2&\hl{2.3068e-1 (4.88e-2)}&2.8797e-1 (5.85e-2) $-$&1.1732e+0 (1.40e-1) $-$\\
	\hline
	\multirow{1}{*}{UF6}&2&2.7811e-1 (1.30e-1)&\hl{8.0531e-2 (1.19e-2) $+$}&5.1894e-1 (2.23e-2) $-$\\
	\hline
	\multirow{1}{*}{UF7}&2&\hl{2.0780e-3 (2.27e-4)}&{2.1632e-3 (1.85e-4) $\approx$}&7.0801e-2 (4.92e-3) $-$\\
	\hline
	\multirow{1}{*}{UF8}&3&\hl{6.6235e-2 (8.10e-3)}&7.4085e-2 (1.55e-2) $-$&2.3143e-1 (1.17e-2) $-$\\
	\hline
	\multirow{1}{*}{UF9}&3&2.0119e-1 (4.79e-2)&\hl{9.6465e-2 (7.76e-2) $+$}&2.0489e-1 (1.52e-2) $\approx$\\
	\hline
	\multicolumn{2}{c}{$+/-/\approx$}&&4/4/1&8/0/1\\
	\bottomrule
\end{tabular}
\label{tab:uf}
\end{table*}

\begin{table*}[htbp]
	\addtolength{\tabcolsep}{1.5pt}
	\centering
	\footnotesize
	\caption{Mean (and standard deviation) values of IGD obtained by three algorithms for selected WFG instances. Algorithms that are better than, worse than, or equivalent to AREA on ranksum test with a significance level of 0.05 are indicated by $+$, $-$, or $\approx$, respectively.}
	\begin{tabular}{lccc}
		\toprule
		Problem &AREA&MOEA/D &RVEA\\
		\midrule
        WFG2-3&\hl{1.5597e-1 (4.60e-3)}&4.1966e-1 (1.54e-2) $-$&1.9748e-1 (1.57e-2) $-$\\\hline
        
        WFG2-8&5.5505e+0 (7.46e-1)&\hl{2.9323e+0 (6.93e-1) $+$}&{4.3956e+0 (8.30e-1) $+$}\\\hline
        
        WFG2-15&1.9090e+1 (2.53e+0)&\hl{7.1471e-2 (3.49e-2) $+$}&1.5960+1(3.52e+0) $+$\\\hline
        
        WFG4-3 & \hl{2.2547e-1 (2.64e-3)}&2.7697e-1 (8.36e-3) $-$&2.2911e-1 (8.38e-3) $-$\\\hline
        
        WFG4-8&{3.1156e+0 (2.84e-2)}&3.7067e+0 (9.05e-2) $-$& \hl{3.081e+0 (1.07e-2) $\approx$}\\\hline
        
        WFG4-15&\hl{9.1102e+0 (4.67e-1)}&1.0388e+1 (4.70e-1) $-$&9.1849e+0 (1.25e-1) $-$\\\hline
        
        WFG6-3&\hl{2.4813e-1 (8.49e-3)}&2.9383e-1 (2.03e-2) $-$&2.6076e-1 (1.92e-2) $-$\\\hline
        
        WFG6-8&{3.1376e+0 (2.85e-2)}&3.3937e+0 (1.13e-1) $-$&\hl{3.0320e+0 (3.69e-1) $+$}\\\hline
        
        WFG6-15 &\hl{9.1245e+0 (6.63e-1)}&9.9702e+0 (1.04e+0) $-$&9.4863e+0 (3.99e-1) $-$\\\hline
		\hline
		{$+\!/\!-\!/\!\approx$}& &2/7/0& 4/4/1\\
		\bottomrule
	\end{tabular}
	\label{tab:wfg}
\end{table*}

\subsection{Comparison with Reference Adaptation Approaches}
Here, we would like to compare AREA with the algorithms that are equipped with specific reference adaptation mechanisms. RVEA$^{*}$ \cite{CJOS-RVEA} (which is another version of the RVEA algorithm) and MOEAD-AWA \cite{QMLJ14-AWS} have proven effective for various PF geometries, and therefore are used as comparison targets for AREA.

\begin{table*}[!t]
	\addtolength{\tabcolsep}{3.5pt}
	\centering
	\caption{Comparison with reference adaptation approaches. Mean (and standard deviation) values of IGD are reported. Algorithms that are better than, worse than, or equivalent to AREA on ranksum test with a significance level of 0.05 are indicated by $+$, $-$, or $\approx$, respectively.}
	\begin{tabular}{ccccc}
		\toprule
		Problem&$M$&RVEA$^{*}$ & MOEA/D-AWA &AREA\\
		\midrule
		\multirow{1}{*}{DTLZ1}&3 & 2.0124e-2 (5.17e-4) $\approx$&{1.9435e-2 (4.76e-4) $+$}         &2.0303e-2 (5.59e-4)\\
		\hline                                  
		\multirow{1}{*}{DTLZ2}&3 & 5.1207e-2 (3.43e-4) $+$&{5.0914e-2 (2.40e-4) $+$}               &5.2651e-2 (4.95e-4)\\
		\hline                                   
		\multirow{1}{*}{DTLZ5}&3 & 8.4923e-3 (4.08e-3) $-$&{4.1110e-3 (6.12e-5) $+$}          &{4.1568e-3 (9.44e-5)}\\
		\hline                                   
		\multirow{1}{*}{DTLZ7}&3 & 8.4604e-2 (9.12e-2) $-$&8.5103e-2 (8.90e-2) $-$               &{5.6225e-2 (1.47e-3)}\\
		\hline                                   
		\multirow{1}{*}{IDTLZ1}&3& 2.3392e-2 (2.25e-3) $-$&{2.0013e-2 (2.79e-4) $+$}          &{2.1485e-2 (2.05e-3)}\\
		\hline                                   
		\multirow{1}{*}{IDTLZ2}&3& 6.4158e-2 (1.70e-3) $-$&{5.1785e-2 (3.82e-4) $+$}          &{5.2069e-2 (5.01e-4)}\\
		\hline                                   
		\multirow{1}{*}{SDTLZ2}&3& 1.2813e-1 (2.66e-3) $-$&{1.1660e-1 (9.65e-4) $+$}          &{1.1792e-1 (1.25e-3)}\\
		\hline                                   
		\multirow{1}{*}{CDTLZ2}&3& 4.9651e-2 (2.62e-3) $-$&{3.2615e-2 (5.48e-4) $\approx$}          &{3.3358e-2 (7.83e-4)}\\
		\hline                                   
		\multirow{1}{*}{MOP1}  &2& 1.2414e-1 (9.37e-2) $-$&2.0304e-2 (3.01e-2) $-$               &{1.0964e-2 (2.65e-4)}\\
		\hline                                   
		\multirow{1}{*}{MOP2}  &2& 2.9402e-1 (5.93e-2) $-$&1.5528e-1 (9.17e-2) $-$         &{1.1691e-1 (8.64e-2)}\\
		\hline                                   
		\multirow{1}{*}{MOP3}  &2& 4.6595e-1 (2.92e-2) $-$&3.7836e-1 (1.33e-1) $-$               &{1.3660e-2 (3.33e-2)}\\
		\hline                                   
		\multirow{1}{*}{MOP4}  &2& 2.4221e-1 (2.31e-2) $-$&1.9705e-1 (4.64e-2) $-$               &{2.1709e-2 (3.37e-2)}\\
		\hline                                   
		\multirow{1}{*}{MOP5}  &2& 1.2436e-1 (1.21e-1) $-$&2.7638e-1 (3.88e-2) $-$               &{1.5858e-2 (7.78e-4)}\\
		\bottomrule
	\end{tabular}
	\label{tab:awa}
	\vspace{-2mm}
\end{table*}
Table \ref{tab:awa} presents the IGD values obtained by the three algorithms on a number of test problems. It is clear that MOEA/D-AWA is indeed very suitable for irregular PF shapes (from DTLZ5 to CDTLZ2), but its performance on the MOP problems with regular PF shapes is unexpectedly poor. RVEA$^{*}$ does not show much promise on the test problems. AREA achieves very similar results as MOEA/D-AWA on DLTZ-based irregular PF shapes. It also shows much better IGD values on the MOP problems than the other two algorithms. This demonstrates that AREA achieves a good balance between irregular PF shapes and regular ones. This highlights AREA is not designed to be the best for all types of problems, but should work well on a wide range of problems.

\subsection{Comparison with Preference Articulation Approaches}
We investigate whether the proposed algorithm can find a preferred region of interest (ROI) or not. We thus compare AREA with three specialist algorithms, which are g-NSGA-II \cite{Molina2009}, r-NSGA-II \cite{Said2010} and WV-MOEAP \cite{Zhangx2016}. Parameters in these algorithms are configured according to \cite{Rostami2017}. The ROI is specified through the definition of preferences, and here we focus on the ROI that dominates (is dominated by) a preference vector if the preference vector is inside (outside) the attainable objective space. The preference vector is set to (0.2,0.2,0.2), (0.5,0.5,0.5), (3.0,2.0,1.0), and (3.0,2.0,1.0) for DTLZ1, DTLZ2, WFG5, and WFG6, respectively. All the algorithms were run with a population size of 105 and a stopping criterion of 100,000 function evaluations.

Table \ref{tab:pref} presents the comparison results of normalised HV, which takes into account solutions in ROI only. It can be observed that AREA is neither the best nor the worst in finding the ROI of the four problems. g-NSGA-II performs the best for these problems except DTLZ1 for which r-NSGA-II seems more effective. WV-MOEAP obtains good results for some problems but fails for others. This is probably due to the small value for the extent of preference region parameter in its default setting. We believe a larger value for this parameter improves WV-MOEAP significantly in terms of the HV value. The comparison suggests in general specialist algorithms for ROI work better. Nevertheless, AREA can be improved if preferences defined by the decision maker are integrated into its search process. 

\begin{table}[htbp]
  \centering
  \caption{Comparison with preference articulation approaches. Mean (and standard deviation) values of normalised HV are reported. }
    \begin{tabular}{lcccc}
    \toprule
    Problem & g-NSGA-II & r-NSGA-II & WV-MOEAP & AREA \\
    \midrule
    DTLZ1 & 0.0000e-1 (0.00e-2) & 4.0613e-1 (1.64e-2) & 3.9252e-1 (3.93e-5) & 3.9067e-1 (3.15e-2) \\\hline
    DTLZ2 & 1.5951e-1 (1.98e-3)   & 1.2584e-1 (3.85e-3) & 1.2995e-1 (3.73e-5) & 1.2422e-1 (1.24e-2) \\\hline
    WFG5 & 1.5841e-1 (1.29e-4)   & 1.2174e-1 (7.45e-3)  & 1.0667e-1 (2.39e-4) & 1.2982e-1 (1.07e-2) \\\hline
    WFG6 & 1.5647e-1 (1.46e-3)   & 1.1599e-1 (7.03e-3) &  5.1171e-2 (4.48e-2) & 1.2613e-1 (8.17e-3) \\
    \bottomrule
    \end{tabular}%
  \label{tab:pref}%
\end{table}%

\section{Conclusion}
The use of references has great potential to direct population individuals toward the true PF. Weight set or its descendants, e.g. reference directions/vectors, is one of the most popular forms of references and has been widely adopted in a range of algorithms, particularly the well-known decomposition-based algorithms, for solving MOPs. While the weight-like form of reference set continues its success in the field of EMO, potential downsides are increasingly recognised. For example, one key downside is that a fixed weight set has difficulty in identifying the complete PF if the PF is shaped irregularly.

Instead of interpreting the reference set as weights, this paper refers to it as a set of targets, placed away from the attainable objective space, for population individuals to move toward. The reference set is adaptively adjusted by archived solutions. A couple of strategies are proposed to coordinate the evolution of the reference set and therefore optimise the reference-set guided search.
The proposed framework, called AREA, is examined on a number of problems with a wide variety of properties, along with extensive sensitivity analysis. A comparison with state of the arts shows that AREA is a powerful engine for solving the problems considered in this paper.

The time complexity of AREA is similar to that of MOEA/D, despite that additional computations are periodically required in the reference alternation process. Thus, AREA is computationally comparable to MOEA/D and has competitive runtime. AREA is also robust to problem types, as demonstrated by the experimental results. It works well particularly for problems with irregular PF shapes. It should be noted that AREA may have reduced performance when solving problems that do not need reference adjustment. However, it is often difficult to know in advance whether a fixed reference set is the best for the problem in question. Thus, it is wise to assume that the problem needs reference adjustment first, and then let AREA to decide how much changes needed in the reference set.      

Besides, there is still space to further improve AREA. AREA currently involves several parameters, although they have low sensitivity to problem characteristics studied in this paper. It is desirable to remove as many of these parameters as possible for the benefit of future application. Another improvement could be made to increase the capability to solve the problems that AREA currently works poorly. It will be also interesting to further study the interdependence between the reference set and archive management. All of these will be investigated in over future work. 

\section*{Acknowledgement}
SJ, MK, and NK acknowledge the Engineering and Physical Sciences Research Council (EPSRC) for funding project ``Synthetic Portabolomics: Leading the way at the crossroads of the Digital and the Bio Economies ({EP/N031962/1})''.

\end{document}